\documentclass[12pt]{article}
\usepackage{amsmath}
\usepackage{graphicx}
\usepackage[round]{natbib}
\usepackage{url} % not crucial - just used below for the UR
\usepackage[latin1]{inputenc} % for PCs
\newtheorem{theorem}{Theorem}

\newtheorem{lemma}{Lemma}

\newtheorem{remark}{Remark}
\newtheorem{proposition}{Proposition}

\newtheorem{definition}{Definition}
\newtheorem{assumption}{Assumption}

\newcommand\argmin{\operatorname{arg\,min}}
\newcommand\argmax{\operatorname{arg\,max}}
\usepackage{amssymb}
\usepackage{bm}
\usepackage{algorithmic}
\usepackage{booktabs}
\usepackage{amsmath}
\usepackage{algorithm}
\usepackage{algorithmic}
\renewcommand{\algorithmicrequire}{\textbf{Input:}}    \renewcommand{\algorithmicensure}{\textbf{Output:}}
\usepackage{color}
\usepackage{algorithm}
\usepackage{verbatim}%????
\usepackage{algorithmic}
\usepackage{makecell}
\usepackage{enumerate}
\usepackage{graphicx}
\usepackage{enumitem}
\usepackage{threeparttable}
\usepackage{multirow}
\usepackage{mathrsfs}
\usepackage{hyperref}
\usepackage{subfigure}
\usepackage{setspace}
\usepackage{fontawesome}
\usepackage[normalem]{ulem}

 \hypersetup{
     colorlinks=true,
     linkcolor=blue,
     filecolor=blue,
     citecolor = darkgreen,
     urlcolor=cyan,
     }
\newcommand{\T}{{\intercal}}
\newcommand{\tA}{{\tilde A}}

\newcommand{\tabincell}[2]{\begin{tabular}{@{}#1@{}}#2\end{tabular}}

\newcommand*{\QEDA}{\hfill\ensuremath{\blacksquare}}

\newcommand{\blind}{0}
\definecolor{darkgreen}{rgb}{0,0.6,0.2}

\begin{document}

\def\spacingset#1{\renewcommand{\baselinestretch}%
{#1}\small\normalsize} \spacingset{1}

%%%%%%%%%%%%%%%%%%%%%%%%%%%%%%%%%%%%%%%%%%%%%%%%%%%%%%%%%%%%%%%%%%%%%%%%%%%%%%

\if0\blind
{
  \title{\LARGE \bf Privacy-Preserving Transfer Learning for Community Detection using Locally Distributed Multiple Networks \thanks{Xiangyu Chang (xiangyuchang@xjtu.edu.cn) and Shujie Ma (shujie.ma@ucr.edu) are corresponding authors.}}
  \author{ Xiao Guo$^1$, Xuming He$^2$, Xiangyu Chang$^3$, Shujie Ma$^4$   \hspace{.2cm}  \\
    $1$ School of Mathematics, Northwest University, China\\
    $2$ Department of Statistics and Data Science,\\ Washington University in St. Louis, U.S.A.\\
    $3$  School of Management, Xi'an Jiaotong University, China\\
    $4$ Department of Statistics, University of California-Riverside, U.S.A.}
  \maketitle
} \fi

\if1\blind
{
 \bigskip
 \bigskip
 \bigskip
 \begin{center}
   {\LARGE\bf Privacy-Preserving Transfer Learning for Community Detection using Locally Distributed Multiple Networks}
\end{center}
 \medskip
} \fi

\bigskip
\begin{abstract}

Modern applications increasingly involve highly sensitive network data, where raw edges cannot be shared due to privacy constraints. We propose \texttt{TransNet}, a new spectral clustering-based transfer learning framework that improves community detection on a \emph{target network} by leveraging heterogeneous, locally stored, and privacy-preserved auxiliary \emph{source networks}. Our focus is the \textit{local differential privacy} regime, in which each local data provider perturbs edges via \textit{randomized response} before release, requiring no trusted third party. 
\texttt{TransNet} aggregates source eigenspaces through a novel adaptive weighting scheme that accounts for both privacy and heterogeneity, and then regularizes the weighted source eigenspace with the target eigenspace to optimally balance the two. Theoretically, we establish an error-bound-oracle property: the estimation error for the aggregated eigenspace depends only on \textit{informative sources}, ensuring robustness when some sources are highly heterogeneous or heavily privatized. We further show that the error bound of \texttt{TransNet} is no greater than that of estimators using only the target network or only (weighted) sources.
Empirically, \texttt{TransNet} delivers strong gains across a range of privacy levels and heterogeneity patterns. For completeness, we also present \texttt{TransNetX}, an extension based on Gaussian perturbation of projection matrices under the assumption that trusted local data curators are available.

\end{abstract}

\vfill

\newpage
\spacingset{1.5} % DON'T change the spacing!

\section{Introduction}

Network data are increasingly collected in a wide range of domains, including social media, neuroscience, and healthcare. A central problem in network analysis is community detection, which seeks to uncover latent node groupings that govern connectivity patterns. However, community detection may be unreliable when the \textit{target network} of interest has weak signal, a small number of nodes, or sparse edges. In many real-world applications, additional \textit{source networks} are available from other institutions or platforms. While these sources may differ in quality and structure, they often contain relevant information that can improve clustering on the target network. Leveraging such auxiliary networks falls under the umbrella of transfer learning \citep{bozinovski2020reminder}.

A motivating characteristic of these applications is that networks are often distributed across data providers such as hospitals or online platforms \citep{kairouz2021advances}, and raw edges cannot be shared due to sensitive content or privacy regulations \citep{dwork2014algorithmic}. These providers may also impose different privacy requirements \citep{xiao2006personalized}, and the similarity between source and target networks may vary significantly. Hence, an effective transfer-learning method must simultaneously handle distributional heterogeneity and privacy constraints.

Motivated by these practical considerations, we propose a new transfer-learning method for community detection on a \textit{target network} that leverages additional \textit{source networks}, under the setting where all networks are heterogeneous, locally stored, and privacy-preserved. To model this setting, we adopt the stochastic block model (SBM) \citep{holland1983stochastic}, a canonical model for community detection in networks, serving as a foundation for understanding and studying network structure and clustering \citep{holland1983stochastic,abbe2017community}. {\color{black}To protect the privacy of network edges, each data provider employs edge-wise randomized response (RR) \citep{warner1965randomized} and then sends the perturbed network to the local data curator. The RR mechanism ensures \textit{local differential privacy} (local DP) \citep{dwork2006calibrating,duchi2018minimax} without relying on any trusted third party and has been widely adopted in practice (e.g., Google RAPPOR \citep{erlingsson2014rappor}\footnote{https://research.google/blog/learning-statistics-with-privacy-aided-by-the-flip-of-a-coin/}).}
% To provide privacy without requiring trust among data providers, we employ edge-wise randomized response (RR) \citep{warner1965randomized}, which achieves \textit{local differential privacy} (local DP) \citep{dwork2006calibrating,duchi2018minimax} and is widely used in practice (e.g., Google RAPPOR \citep{erlingsson2014rappor}\footnote{https://research.google/blog/learning-statistics-with-privacy-aided-by-the-flip-of-a-coin/}). 
After each provider perturbs edges independently using network-specific probabilities, the data curator then applies bias correction to recover an unbiased estimator of the underlying network \citep{guo2023privacy,chang2024edge}.

We face two key challenges when leveraging distributed, heterogeneous, and privacy-preserved debiased source networks for transfer learning. First, how to aggregate sources effectively under varying heterogeneity and privacy. A simple average of source eigenspaces ignores differences in signal preservation and structural similarity. Second, how to integrate the aggregated source information with the target network's eigenspace to improve community detection; even strong sources may not fully represent the target, so the combination must be judicious.

We address these challenges with \texttt{TransNet}, a newly proposed three-step spectral method consisting of: (i) adaptive weighting: we aggregate source eigenspaces using data-driven weights that depend on privacy levels and heterogeneity, automatically adjusting each source's contribution to recovering the target's community structure. (ii) regularization: we integrate the weighted source eigenspace with the target eigenspace to optimally balance their information; {\color{black}the resulting estimator aims to achieve an error bound that is not larger than} that obtained by using only the sources or only the target network.
 (iii) clustering: we apply $k$-means to the regularized eigenspace to obtain the final community assignments.

The contributions of this paper are summarized as follows.  

$\bullet$ First, we develop a novel adaptive weighting strategy to aggregate the eigenspaces of the source networks. The weights are \textcolor{black}{data-adaptive}, and they match the practical intuition that the source network, which is more homogeneous to the target network and less privacy preserved is assigned with a greater weight. In the transfer learning for regression problems, the source datasets that have a similar underlying model to the target dataset are called  \emph{informative} data,  
%whose underlying model is `similar' to that of the target dataset is called \emph{informative},
see e.g., \citet{li2022transfer}. Similarly, we define the \emph{informative source networks} as the ones that {\color{black}are close to the target network within certain privacy-preserving and heterogeneity levels.} 
We derive two important theoretical properties for the adaptive weighting method. First, the adaptive weighting step satisfies the \emph{error-bound-oracle} property that under reasonable conditions, only the effects of {informative} source networks are involved in the error bound of the estimated eigenspace.  
Second, the eigenspace error bound achieved by the adaptive weighting strategy is of smaller order than the one obtained from the equal weighting strategy \textcolor{black}{under certain conditions}. These two properties provide theoretical evidence and support for the usefulness of our proposed adaptive weights.

$\bullet$ Second, we develop a regularization method that automatically balances between the eigenspace of the bias-corrected target network and the weighted average eigenspace of the source networks obtained from the first step. 
This step is a key component of the transfer learning process. Compared to transfer learning of the parameters in regression models (e.g., \citet{li2022transfer,tian2023transfer}), our proposed regularization method is for transfer learning of the eigenspaces, requiring a different technical treatment. We establish a useful error bound for the estimated eigenspace obtained by the {regularization method}. The bound reveals clearly how the heterogeneity and privacy-preserving levels affect the estimated eigenspace. {\color{black}We show that the regularized eigenspace generally yields a smaller error bound than that obtained using only the target network or the weighted average eigenspace of the source networks}, achieving the goal of transfer learning. %In particular, when both heterogeneity and privacy levels are \emph{low}, the regularization can provide a bound that is asymptotically smaller than that obtained using only the target network 
To the best of our knowledge, this is the first work that studies transfer learning for community detection of networks {\color{black}from a statistical perspective}.

$\bullet$ Third, the RR mechanism belongs to the \emph{input perturbation} methods for achieving DP. For completeness, we also develop a \emph{Gaussian mechanism} imposed on each projection matrix,  which is an \emph{output perturbation} method. The output perturbation actually assumes that the local
data providers trust the local data curators (local machines) such that the data curators can access the raw data and adds noise to the data or to summary statistics before any release \citep{dwork2006calibrating}. 
We extend our transfer-learning methodology and theory to this setting, leading to a new method called \texttt{TransNetX}. The privacy cost induced by the output perturbation-based Gaussian mechanism only affects the second-order statistical error, which is the benefit of having trusted local data curators. Formal privacy guarantees, error bounds, implementation details, and discussions of the two methods are all provided in the Supplement. 
{To the best of our knowledge, this is the first work to investigate the output perturbed DP method for spectral clustering-based community detection of SBMs.}

\subsection{Related work}

Transfer learning has become a major area in machine learning, with applications spanning natural language processing, speech recognition, and recommendation systems \citep{pan2009survey,Brown2020language,zhuang2020comprehensive}. In statistics, recent work has developed transfer-learning methods for high-dimensional regression \citep{bastani2021predicting,li2022transfer,tian2023transfer,li2024estimation,he2024transfusion,zhou2024doubly}, classification \citep{reeve2021adaptive}, nonparametric regression \citep{cai2024transfer,hu2023optimal}, graphical model estimation \citep{li2023transfer}, federated learning \citep{li2023targeting,li2024federated}, and semi-supervised learning \citep{cai2024semi}, and privacy-aware transfer learning for regression \citep{wei2020federated,wei2021user,shen2022federated,auddy2024minimax,cai2024federated,cai2024optimal2,li2024federated}. Most approaches in regression implicitly or explicitly include all source datasets in estimator construction, with informative-source selection typically handled using a two-step procedure: screening followed by estimation \citep[e.g.,][]{li2022transfer,tian2023transfer,li2024estimation}. In contrast, our method performs one-step adaptive weighting, simultaneously estimating the eigenspace and down-weighting non-informative sources, a desirable feature for spectral methods.

Community detection across multiple networks has also received increasing attention; see, for example, \citet{han2015consistent,paul2016consistent,paul2020spectral,arroyo2021inference,macdonald2022latent,zheng2022limit,agterberg2022joint,lei2023bias,huang2023spectral,guo2023privacy,zhang2024consistent}. Most existing works aim to estimate a shared or consensus community structure by pooling networks and typically assume strong similarity across layers. In contrast, we focus on improving a specific target network using heterogeneous source networks that remain locally stored and privacy-preserved.

Privacy-preserving community detection has largely focused on single-network settings \citep{zheleva2011privacy,wang2013learning,houghton2014privacy,borgs2015graphon,borgs2018revealing,ning2021benchmarking,laeuchli2022analysis,chakrabortty2024prime}. The use of RR for DP has been analyzed in \citet{chakrabortty2024prime,hehir2022consistent,imola2021locally}. Our contribution differs from the existing work by addressing multiple distributed networks with network-specific privacy levels and developing a transfer-learning framework with provable performance guarantees under two different DP regimes.

\subsection{Organization}
The remainder of this paper is organized as follows. Section \ref{sec:problem} presents the data-generating process and preliminaries of DP. 
Sections \ref{sec:adative weighting}, \ref{sec:regularization} and \ref{sec:clustering} provide the details of the three steps of \texttt{TransNet} and the corresponding error bounds, respectively.  
Sections \ref{sec:sim} and \ref{sec:real} include the simulation and real data experiments. Section \ref{sec:concl} concludes the paper. 
The proofs, additional experiments, and extension to output perturbation-based \texttt{TransNetX} method are all included in the Supplementary Materials.

\vspace{-0.5cm}

\section{Problem formulation}
\label{sec:problem}
In this section, we describe the data generating process for both source and target networks, including the statistical model, the RR mechanism, and the debiasing procedure. After that, we provide the preliminaries of DP and the RR mechanism. 
We introduce the notation first.

\subsection{Notation}
 {We use $[n]$ to denote the set $\{1,...,n\}$ for any integer $n$, whereas $\mathcal O_K$, $I_n$, and $\mathbf {1}_n$ denotes the set of $K \times K$ orthogonal matrices, the $n$-dimensional identity matrix and the $n$-dimensional vectors with all entries being 1, respectively. 
The projection distance, defined as $\mathrm{dist}(U, U') := \|UU^\T - U'U'^\T\|_2$, is used to quantify the subspace distance between any pair of orthogonal bases $U, U' \in \mathbb{R}^{n \times K}$.
We use $\lambda_{\min}(A)$ for the smallest non-zero eigenvalue of a given matrix $A$ and $\lambda_{j}(A)$ for the $j$th largest eigenvalue of matrix $A$. Note that the use of $\lambda_{\min}(A)$ here differs from its common usage as the smallest eigenvalue of a matrix.

The notations $\|A\|_2$ and $\|A\|_{  F}$ represent the spectral norm and the Frobenius norm of the matrix $A$, respectively. In addition, we adopt the following standard notation for $n\rightarrow \infty$ asymptotics. Let $f(n)\asymp g(n)$ denote $c g(n)\leq f(n)\leq C g(n)$ for some constants $0<c<C<\infty$. Denote $f(n)\lesssim g(n)$ or $f(n)=O(g(n))$ if $f(n)\leq Cg(n)$ for some constant $C<\infty$. Similarly, denote $f(n)\gtrsim g(n)$ or $f(n)=\Omega(g(n))$ if $f(n)\geq cg(n)$ for some constant $c>0$. Moreover, denote $f(n)=o(g(n))$ or $f(n)\ll g(n)$ if $f(n)/g(n)\rightarrow0$ as $n\rightarrow \infty$; $f(n)=\omega(g(n))$ or $f(n)\gg(g(n))$ if $g(n)/f(n)\rightarrow0$ as $n\rightarrow \infty$.
Finally, the constants $C$ and $c$ used here may be different from place to place.

\subsection{Data generating process and debiasing}

\noindent\textbf{Network model.} Suppose there is one target network and $L$ source networks with $n$ aligned nodes, where {$L$ may depend on $n$}. We denote the target network by $A_0$ and the source networks by $\{A_l\}_{l \in [L]}$, which are all symmetric and binary adjacency matrices. We consider a model-based data-generating process for the networks such that for $l\in \{0\}\cup[L]$,  each $A_l$ is independently generated from an SBM given as follows. We assume that each $A_l$ has $K$ communities, with each node belonging to exactly one community, where $K$ is assumed to be {known and fixed}, and denote its membership matrix by $\Theta_l\in \{0,1\}^{n\times K}$, where the $i$th row represents node $i$ with its $j$th column being 1 if node $i$ belongs to the community $j\in [K]$ and being 0 otherwise. Moreover, we denote its connectivity matrices by $B_l \in [0,1]^{K\times K}$. Then for given $\Theta_l$, each entry $A_{l,ij}$ $(i<j)$ of $A_l$ $(l\in \{0\}\cup[L])$ is independently generated according to 
\begin{equation}
\label{MOD:SBM}
A_{l,ij}\sim {\mathrm{Bernoulli}} (P_{l,ij})  \quad{\rm with} \quad P_l:= \Theta_l B_l\Theta_l^T,
\end{equation}
and $A_{l,ii}=0$.

The goal is to {better estimate} the membership matrix $\Theta_0$ of the target network using the $L$ source networks. We assume that the source networks are heterogeneous relative to the target network in both the membership matrix and connectivity matrix. 
It is worth mentioning that the model is different from the multi-layer SBMs studied by multiple network literature (see e.g., \citet{paul2020spectral,lei2023bias}), where the latter assumes homogeneous communities $\Theta_1=\Theta_2=...=\Theta_L$. To highlight the difference, we refer to the model for the $L$ source networks as \emph{Heterogeneous Multi-layer Stochastic Block Model} (HMSBM).

The heterogeneity in the community structures is inherent in the population eigenvectors. Denote $U_l$ as the $K$ leading eigenvectors of $P_l$, {\color{black}then we have that (see Lemma \ref{pro: eigendecom} in the Supplementary Materials)}
\begin{equation}
\label{eq:UandD}
U_l=\Theta_l D^{-1}_l L_l\quad {\rm with} \quad D_l:= {\rm diag} \{\sqrt{n_1^l},...,\sqrt{n_K^l}\}
\end{equation}
and $L_l$ being an orthonormal matrix, where $n_k^l$ is the number of nodes in community $k\in [K]$. Therefore, $\|U_lU^\T_l-U_mU^\T_m\|_2=0$ provided that $\Theta_l=\Theta_m$ for $l\neq m$.

% \textcolor{black}{\textbf{Privacy model.} We consider a federated setting in which each data provider (for example, a clinical unit within a hospital) generates its own adjacency matrix $A_\ell$, $\ell \in \{0\}\cup[L]$ (with $\ell=0$ denoting the target network). Providers send locally perturbed data to a local data curator (such as the hospital's data office) that coordinates computation and communication, where the  local data curator is treated as untrusted (semi-honest). A central server, the external data analyst, aggregates messages from the local curators. The server is modeled as honest-but-curious: it follows the protocol and never sees raw data, but may try to infer information from the messages and procedure. As a result, all transmissions must be privacy-preserved.}
\vspace{0.2cm}
\noindent\textbf{Privacy model.} 
We consider a federated learning framework involving the following actors:
{\color{black}
\begin{itemize}
[label=$\bullet$]
\item Local data providers: the data owners who generate their own adjacency matrices $A_l$, $l \in \{0\}\cup[L]$ (with $l=0$ denoting the target network), e.g., clinical units within a hospital;
\item Local data curators (local machines): parties that coordinate computation and communicate with local data providers and the external data analyst, e.g., a hospital's data office.
\item External data analyst (central machine): the party that receives messages from local data curators. The central machine is modeled as honest-but-curious: it follows the protocol and never sees the true information of data, but may try to infer information from the messages and procedure. As a result, all transmissions must be privacy-preserved before leaving the local machines.
\end{itemize}
}
% in which local machines act as local data curators (e.g., the hospital's data office), coordinating computation and communicating with local data providers (for example, a clinical unit within a hospital)  who generate their own adjacency matrix 
% $A_l$, $l \in \{0\}\cup[L]$ (with $l=0$ denoting the target network), and with the external data analyst who serves as the central machine. The central machine is modeled as honest-but-curious: it follows the protocol and never sees the true information of data, but may try to infer information from the messages and procedure. As a result, all transmissions must be privacy-preserved before leaving the local machines.

Depending on whether the local data curators (local machines) are trusted or not, there are two regimes. 
The first regime assumes that local data curators cannot be trusted by the local data providers, a setting typically referred to as local DP \citep{duchi2018minimax}. In this regime, the local data provider perturbs the raw data, and the local data curator simply forwards the locally perturbed data to the central server. The second regime assumes that the local data curators are trusted. This setting can be regarded as central DP \citep{dwork2006calibrating} to the local data curators. In this regime, the local data curator receives raw data from the local data provider, perturbs the data or its summary statistics, and then transmits the results to the central server. Notably, the first regime provides a stronger level of privacy protection, which inherently implies the guarantees of the second regime.

In the main paper, we consider the first DP regime. Extensions to the second DP regime are presented in the Supplementary Materials.

\vspace{0.2cm}
\noindent\textbf{Perturbation.} \textcolor{black}{Before any release, each local provider applies the RR mechanism to its network. RR protects the network data against any adversary, including the local machines and the central server. Beyond privacy, releasing perturbed adjacency data, rather than only summary statistics, supports a wider range of downstream analyses.}

Let $\tA_l$ be the perturbed network of $A_l$ and let $q_l,q'_l\in[0,1]$ be {edge-preserving} probabilities, that is, for any $i<j$, $\tA_{l,ij}$ is obtained by 
\begin{equation*}
\begin{aligned}
\mathbb P(\tilde{A}_{l,ij}=1|A_{l,ij}=1)=q_l \quad  {\rm and}\quad \mathbb P(\tilde{A}_{l,ij}=0|A_{l,ij}=0)=q'_l.
\end{aligned}
\end{equation*}
We allow for heterogeneous perturbation levels, that is, the $q_l$ can vary across different $l$, offering greater flexibility and generality compared to the homogeneous case studied in~\citet{guo2023privacy}.}

\vspace{0.2cm}
\noindent\textbf{Debiasing.} $\tA_l$ brings bias to the original network $A_l$. We adopt the following bias-adjustment procedure 
\begin{equation}\label{eq:db}
\hat{A}_l:=(\tA_l-(1-q'_l)\mathbf 1\mathbf 1^\intercal+{\rm diag}(1-q'_l))/(q_l+q'_l-1),
\end{equation}
which ensures that $\mathbb E(\hat{A}_l|A_l)=A_l$.
The bias-adjustment has been used and shown to be useful in \citet{chang2024edge,guo2023privacy}. We will illustrate the efficacy of bias-adjustment in Theorem \ref{theo: non-debiased} in Section \ref{app:add} of Supplementary Materials. 

The next subsection introduces the preliminaries of DP \citep{dwork2006calibrating} and provides a rigorous analysis on the RR mechanism. 
% The matrix $\hat{A}_l\ (l\in \{0\}\cup [L])$ in \eqref{eq:db} satisfies the rigorous 
% concept of DP \citep{dwork2006calibrating}; see the details and also the preliminaries of DP in the next subsection  \ref{app:dp}. 

\subsection{Differential privacy}
\label{sec:dp}
 In general, an algorithm is called DP if it has a similar output in probability when the input only varies by one data point. In the context of networks, the following notion of edge-DP \citep{rastogi2009relationship,karwa2017sharing,guo2023privacy} has been proposed to preserve the \textcolor{black}{edges of networks}. 

\begin{definition}[$\epsilon$-edge-DP]
\label{edp}
Let $A\in \{0,1\}^{n\times n}$ be the network of interest. For any range $\mathcal R$, let $\mathcal M(A)\in \mathcal R$ be a randomized mechanism that generates some synthetic information from $A$, indicated by a known parameter $\omega$. Given a privacy budget (also called privacy-preserving level in this paper) $\epsilon>0$, $\mathcal M(A)$ satisfies $\epsilon$-edge-DP if
\begin{equation*}
\underset{O\in \mathcal R}\max \,\underset{A',A,\, \Delta(A,A')=1}\max\; \log \frac{\mathbb P_{\omega}(\mathcal M(A)=O\,| \,A)}{\mathbb P_{\omega}(\mathcal M(A')=O \,|\,A')}\leq \epsilon,
\end{equation*}
where $\Delta(A,A')=1$ indicates that $A$ and $A'$ are neighboring networks, differing by only one edge.
\end{definition}

An edge-DP mechanism or algorithm indicates that an adversary can not easily infer whether there is an edge between one node pair from the algorithm's output and the edge information between all the rest node pairs.

\textcolor{black}{To understand the relationship between the RR mechanism and edge-DP, we first set up some notations. For each $l\in[L]\cup \{0\}$, denote the RR mechanism $\mathcal M_l:A_l\rightarrow \tilde{A}_l$. Denote $\mathcal A:= (A_1,...,A_L)$ and $\tilde{\mathcal{A}}:=(\tilde{A}_1,...,\tilde{A}_L)$. Define $\overline{\mathcal M}: \mathcal A\rightarrow \tilde{\mathcal A}$ as
\begin{equation}
\label{eq:mbar}
\overline{\mathcal M}(\mathcal A):= ({\mathcal M}_0(A_0),{\mathcal M}_1(A_1),...,{\mathcal M}_L(\mathcal A_L)).
\end{equation}
{\color{black}The following lemma establishes that both the individual RR mechanisms $\mathcal M_l$ and the joint mechanism $\overline{\mathcal M}$ satisfy edge-DP, with respect to a single network and multiple networks, respectively.}}

\begin{lemma}
\label{lem:dp}
\textcolor{black}{Suppose the edge-preserving probabilities satisfies
\begin{equation}
\label{eq:pribudg}
\log\max \left\{ \frac{q'_l}{1-q_l},\frac{1-q_l}{q'_l},\frac{1-q'_l}{q_l},\frac{q_l}{1-q'_l}\right\}\leq \epsilon_l,
\end{equation}
then the following properties hold:
\begin{enumerate}[label=(\arabic*), itemsep=0pt, topsep=2pt, parsep=0pt, partopsep=0pt]
\item For each $l \in [L] \cup \{0\}$, $\mathcal M_l(A_l)$ satisfies $\epsilon_l$-edge-DP.
\item (Parallel composition) $\overline{\mathcal{M}}(\mathcal{A})$ satisfies $\max_{l} \epsilon_l$-edge-DP.
\end{enumerate}}
\end{lemma}

\begin{remark}
{\color{black}
Property (1) has been studied in \citet{karwa2017sharing}. Property (1) considers the release of a single \(\mathcal M_l(A_l)\), where neighboring datasets \(A_l\) and \(A_l'\) differ in only one edge. Property (2) considers the joint release \(\overline{\mathcal M}(\mathcal A)\), where neighboring \(\mathcal A\) and \(\mathcal A'\) differ in one edge of a single network, while all other networks remain unchanged. Hence, a record corresponds to one edge within a single provider's network.
Since the original networks $A_l$'s are disjoint in edges, the parallel composition property \citep{McSherry2009PINQ,SmithAGMGT22} applies,  and the overall privacy budget remains bounded by $\max_l \epsilon_l$. The proof follows similar arguments as in Lemma \ref{lem:cdp} and is therefore omitted.
It is worth noting that if the privacy goal is instead to protect the same dyad jointly across layers, then the disjointness assumption no longer holds. In this case, a sequential composition argument (e.g., \(\sum_l \epsilon_l\)) should be used instead \citep{dwork2014algorithmic}.
}
\end{remark}

A typical choice of $q_l,q'_l$ satisfying (\ref{eq:pribudg}) is 
\begin{equation}
\label{eq: qe}
q_l=q'_l=\frac{\mathrm{e}^{\epsilon_l}}{1+\mathrm{e}^{\epsilon_l}},
\end{equation}
which has been proven to be optimal from different aspects \citep{ayed2020information,guo2023privacy}. Throughout the paper, we assume $q_l=q'_l$ relates to the privacy budget $\epsilon_l$ via (\ref{eq: qe}).   

\begin{remark}
\textcolor{black}{In practice, the central server (external data analyst) may propose a target privacy budget $\epsilon$. Each data provider can adopt this value or choose a different level, resulting in network-specific budgets $\epsilon_l$'s. The $\epsilon_l$ for each source is disclosed to the external analyst and the local curator, revealing
$\epsilon_l$ does not weaken privacy, since DP is defined against adversaries with arbitrary side information
\citep{dwork2014algorithmic}. The external analyst can enforce the desired privacy budget by omitting sources with
$\epsilon$ above the specified value.} %\citep{erlingsson2014rappor}. 
\end{remark}

The DP satisfies the post-processing property, which ensures that any post-processing of a DP  algorithm does not compromise its privacy guarantee, as long as the process does not use the \emph{original} data \citep{dwork2014algorithmic}.

\begin{lemma}[Post-processing property]
\label{lem:post-processing}
Let $\mathcal M(A):\{0,1\}^{n\times n}\rightarrow \mathcal R$ be a randomized mechanism that satisfies $\epsilon$-edge-DP in Definition \ref{edp}. Let $f:\mathcal{R}\to \mathcal{R}'$ be any (possibly randomized) mapping 
that is independent of the input data $A$. Define $\mathcal{M}':= f \circ \mathcal{M}$,
then $\mathcal{M}': \{0,1\}^{n\times n}\rightarrow \mathcal R'$ also satisfies $\epsilon$-edge-DP.
\end{lemma}

Therefore, the debiasing and further processing of the RR-perturbed adjacency matrices $\tilde{A}_l$'s do not degrade the privacy-preserving level of DP.

\section{Adaptive weighting}
\label{sec:adative weighting}

In this section, we propose the first step (i.e., the adaptive weighting step) of \texttt{TransNet}. Due to different levels of heterogeneity and privacy-preserving, the effect of each source network on the target network is different. Motivated by this, we propose an adaptive weighting strategy, which can lead to the error-bound-oracle property and a tighter {error} bound than the equal weighting strategy.  

The first step (i.e., the adaptive weighting step) of \texttt{TransNet} is based on the RR-perturbed and debiased $\hat{A}_l\ (l\in \{0\}\cup [L])$'s. To address the distributed nature of source networks and enhance communication efficiency, we aggregate the eigenspaces $\hat{U}_l$'s instead of the debiased networks $\hat{A}_l$'s. For $l\in \{0\}\cup [L]$, let $\hat{U}_l$ be the $K$ leading eigenvectors of $\hat{A}_l$, we aggregate $\hat{U}_l$'s weighted by $w_l$'s
to obtain
%\vspace{-0.5cm}
\begin{equation}
\label{eq:weiagg}
\hat{U}:=\sum\nolimits_{l=1}^L w_l\hat{U}_lZ_l\quad {\rm with}\quad Z_l:= \argmin _{Z\in \mathcal O_K} \|\hat{U}_lZ- \hat{U}_0\|_{F},
%\vspace{-0.5cm}
\end{equation}
where {$w_l~(1\leq w_l\leq 1 ;\sum_{l=1}^Lw_l=1)$} is to used to incorporate the heterogeneity, and 
$Z_l$ resolves the orthogonal ambiguity of eigenspaces  \citep{cape2020orthogonal,charisopoulos2021communication} and \textcolor{black}{has the closed form solution $S_lT_l^\intercal$ with $S_l$ and $T_l$ being the left and right eigenvectors of the $K\times K$ matrix $\hat{U}_l^\intercal \hat{U}_0$. Note that given $\hat{U}_l$'s, the construction of $\hat{U}$ is computationally efficient, which takes time complexity $O(LnK^2+LK^3)$. }
The matrix $\hat{U}$ is then orthogonalized using QR decomposition, yielding $\bar{U}$, i.e., {\color{black}$$\bar{U}, \bar{R}:= {\rm QR} (\hat{U}),\quad {\rm where} \quad \bar{U}^\intercal\bar{U} = I_K.$$}
The weights $w_i$ in \eqref{eq:weiagg} can be chosen appropriately, so it is of primary importance to understand how they affect the performance of $\bar{U}$. To this end, we impose the following assumptions. To fix ideas, recall that $q_l=q'_l=\frac{{\rm e}^{\epsilon_l}}{1+{\rm e}^{\epsilon_l}}$, and for notational convenience we define
\begin{equation}
\label{eq:nul}
\nu_l:= q_l+q'_l-1>0.
\end{equation}
Thus, $\nu_l$ represents the privacy-preserving level of the $l$-th network: smaller values of $\nu_l$ correspond to stricter privacy. Importantly, we also allow the high-privacy regime where both $q_l$ and $q'_l$ are close to $1/2$, i.e., $\nu_l$  close to $0$.

\begin{assumption}[Balanced sparsity]
\label{assu:sparse}
{The connectivity matrices $B_l =\rho {C}_{l,0}\ (l\in [L] \cup \{0\})$, where $\rho:=\rho_n=o(1)>0$ is the sparsity level which may depend on $n$, and $C_{l,0}$ is a $K\times K$ symmetric matrix with entries fixed in $[0,1]$.}
\end{assumption}

% {\color{black}Comment: Should we write $\rho = o(1)$ here, since no probability is involved?}
% \begin{assumption}[Heterogeneity]
% \label{assu:hetPo}
% The population matrix $P_l\ (l\in[L])$ of $L$ source networks are heterogeneous in that
% \begin{equation*}
% \label{eq:heterP}
% \|B_l-B_0\|_2:=\mathcal E_{b,l}\cdot \rho \quad {\rm and} \quad\|\Theta_l-\Theta_0\|_2:= \mathcal E_{\theta,l}\cdot \sqrt{n} \quad {\rm for\; }\; l \in [L],
% \end{equation*}
% {where $0\leq \mathcal E_{b,l}=O(1)$ and $0\leq \mathcal E_{\theta,l}=O(1)$, and both parameters may depend on $n$.} 
% %We assume $\mathcal E_{b,l}^2 \lesssim \mathcal E_{\theta,l}$ for all $l \in [L]$. 
% \end{assumption}

% \begin{assumption}[Privacy]
% \label{assu:hetPri}
% The edge-preserving parameters of $L$ source networks and the target network in RR are $(q_l,q'_l)_{l=0}^L$. \textcolor{black}{We assume $\min_{l\in [L] \cup \{0\}}(q_l+q'_l-1)\geq \nu>0$, where we abused the notation slightly in \eqref{eq:estim}.}
% \end{assumption}
% {\color{black}Comment: I feel that in Assumption 3, we probably don't need the condition that $\min_{l\in [L] \cup \{0\}}(q_l+q'_l-1)\geq \nu>0$. Instead, we can write $\nu_l=q_l+q'_l-1$, and replace $\nu$ by $\nu_l$ in our theories.Or we can further write $\nu=\min_{l\in [L] \cup \{0\}}\nu_l$ if  $\nu$ is involved in some conditions, so we can remove the aforementioned condition too.}

\begin{assumption}[Eigen-gap]
\label{assu:rank}
{For any $0\leq w_l\leq 1$ and $\sum_{l=1}^L w_l=1$}, we assume $B_0$ and $\sum_{l=1}^L w_l B_{l}$ are both of full rank $K$. In addition, $\lambda_{\rm   min}(P_0)\gtrsim n\rho$ and $\lambda_{\rm   min}(\bar{P})\gtrsim n\rho$ with 
\begin{equation}
\label{eq:barP}
\bar{P}:= \Theta_0 \left(\sum\nolimits_{l=1}^L w_l B_{l}\right) \Theta_0^\intercal. 
\end{equation}
\end{assumption}

\begin{assumption}[Order of $\rho$ and $L$]
\label{assu:conrhoL}
We assume
\begin{equation}
\label{eq:conrho1}
\rho \gtrsim \max\left\{\frac{\log n}{n\textcolor{black}{\nu_{\min}}},\; \frac{1}{n\textcolor{black}{\nu^2_{\min}}} \cdot\frac{1-\nu_{\min}}{\rho}\right\},
\end{equation}
\begin{equation}
\label{eq:conL}
\sum_{l\in[L]} \zeta_l \lesssim \max \left\{\frac{n\nu_{\min}^2}{\log n},\;\min_{l\in[L]\cup\{0\}} \zeta_l^2 n\rho^2\log n\right\},
\end{equation}
where 
$\zeta_l:=\nu_l^2/(\nu_l\rho+(1-\nu_l)/2)$ and $\nu_{\min}:=\min_{l\in[L]\cup\{0\}} \nu_l$.
\end{assumption}

% {\color{black}Comment: By checking the proofs in pages 1-4 of the supplment, probably we can relax the last condition in Assumption 3. The last condition seems to be very strong. For the second condition, I understand that because of the Bernstein inequality, $\nu_{max}$ is needed, but we can replace $\min_l\tilde{\rho}_l\nu_{\min}$ by the average of $\tilde{\rho}_l\nu_l$. For the last condition, I think we can use $\nu_l$ directly to find a condition. }

\begin{assumption}[Balanced communities]
\label{assu:balcom}
The community in the target network is balanced such that for any $k\in\{1,...,K\}$, {\color{black}$n_k^0\asymp n$}, where for $k\in[K]$, $n_k^0$ denotes the number of nodes in the $k$th community. 
\end{assumption}

{\color{black} For theoretical convenience, Assumption \ref{assu:sparse} assumes all networks to have sparsity of the same order, a condition also imposed in other SBM studies; see, e.g., \citet{lei2023bias}. Assumption \ref{assu:rank} imposes a lower bound on the minimum eigenvalues of $P_0$ and $\bar{P}$, a requirement that is typically satisfied in SBMs; see, e.g., \citet{lei2015consistency}.} 
For instance, when the $B_l$'s are identical and have a constant condition number, Assumptions \ref{assu:sparse} and \ref{assu:balcom} imply Assumption \ref{assu:rank}. Assumption \ref{assu:conrhoL} is required to derive and simplify the error bound of the estimated eigenspace after Step 1 under the general weights; see the discussions after Proposition \ref{theo: step1} for details. 
Assumption \ref{assu:balcom} is required for notational simplicity in deriving the misclassification rate of the proposed \texttt{TransNet} algorithm, but it can be relaxed.

Apart from the assumptions, we need the following notations. The heterogeneity among the $L$ source networks is quantified by the deviation of their population matrices $P_l$'s from the target, specified as:
\begin{equation}
\label{eq:heterP}
\|B_l-B_0\|_2:=\mathcal E_{b,l}\cdot \rho \quad {\rm and} \quad\|\Theta_l-\Theta_0\|_2:= \mathcal E_{\theta,l}\cdot \sqrt{n} \quad {\rm for\; }\; l \in [L],
\end{equation}
where both $\mathcal E_{b,l}$ and $\mathcal E_{\theta,l}$ may depend on $n$. Note that we naturally have $\|B_l-B_0\|\lesssim \rho$ and $\|\Theta_l-\Theta_0\| \lesssim \sqrt{n}$.  
Hence, $0\leq \mathcal E_{b,l}=O(1)$ and $0\leq \mathcal E_{\theta,l}=O(1)$ naturally. 
The edge-preserving parameters of $L$ source networks and the target network in RR are $\{(q_l,q'_l)\}_{l=0}^L$, which are possibly different across $l$.

We next present a unified result to examine the impact of weights on the performance of $\bar{U}$, providing guidance for selecting weights.  

\begin{proposition}
\label{theo: step1}
 Suppose Assumption \ref{assu:sparse}-\ref{assu:conrhoL} hold and {$\mathcal E_{b,l}^2 \lesssim \mathcal E_{\theta,l}$ for all $l \in [L]$}. 
Then with probability larger than $1-\frac{L}{n^{\kappa}}$ for some constant $\kappa>0$,  we have,
{\small
\begin{align}
\label{eq:step1err}
 {\rm dist}(\bar{U}, U_0)
&\lesssim \sqrt{\sum_{l=1}^L w_l^2\cdot \frac{(\nu_l\rho+ (1-\nu_l)/2)}{\rho}\cdot\frac{\log n}{n\rho\nu_l^2}}+\sum_{l=1}^L w_l \mathcal E_{\theta,l}\nonumber\\
&\lesssim  \sqrt{\sum_{l=1}^L w_l^2\cdot \frac{\log n}{n\rho\nu_l}}+\sqrt{\sum_{l=1}^L w_l^2\cdot \frac{1-\nu_l}{\rho}\cdot \frac{\log n}{n\rho\nu_l^2}}+\sum_{l=1}^L w_l \mathcal E_{\theta,l}
:= \mathcal I_1+ \mathcal I_2+ \mathcal I_3.
\end{align} 
}
\end{proposition}

% {\color{black}Comment: In the proof given in the line above (S7) in the supplement, why does $(\sum_l w_l \mathcal E_{b,l})^2\lesssim \sum_lw_l \mathcal E_{b,l}^2$ hold? }

\begin{remark}
For technical convenience, we require $\mathcal E_{b,l}^2 \lesssim \mathcal E_{\theta,l}$, which reduces the impact of $\mathcal{E}_{b,l}$ relative to $\mathcal{E}_{\theta,l}$ in the error bound. 
We next discuss the conditions \eqref{eq:conrho1} and \eqref{eq:conL} in Assumption \ref{assu:conrhoL}.
\eqref{eq:conrho1} is a technical condition that ensures the accuracy of the eigenspace computed using each debiased source network. When $q_l=q'_l=1$ for all $l\in[L]$, {i.e., without the privacy-preserving need}, \eqref{eq:conrho1} reduces to $\rho\gtrsim \log n/n$, which is also used in \citet{arroyo2021inference} when they studied the one-shot-based spectral embedding for multi-layer networks.  
Condition \eqref{eq:conL} is imposed to simplify the results and ensure that the distributed algorithm achieves a bound comparable to its centralized counterpart (i.e., the error bound for the eigenvector of $\sum_{l=1}^L w_l(\hat{A}_l-P_l)$). Note that $\zeta_l$ is an increasing function of $\nu_l$ for fixed $\rho$.
When $\zeta_l$ is of the same order for all $l\in[L]$, condition \eqref{eq:conL} actually requires that the number of source networks $L$ is upper bounded.
\end{remark}

\begin{remark}
% \iffalse
% {\color{black}The first two terms, $\mathcal{I}_1$ and $\mathcal{I}_2$, both involve the privacy parameter $\nu_l$. The third term $\mathcal I_3$ corresponds to the effect of heterogeneity; see a similar result in the regression setting \citep{li2022transfer}.
% When $\nu_l$'s are lower bounded by some positive constant, i.e., $q_l=q'_l>\frac{1+\tau}{2}$ for some small constant $\tau>0$, $\mathcal  I_1$ corresponds the standard statistical error of combining multiple networks (i.e., the error bound for the eigenvector of $\sum_lw_l A_l$) and $\mathcal I_2$ corresponds to the additional cost of RR mechanism. 
% % Smaller values of $1-q'_l$, i.e., the probability of flipping 0 to 1, are desirable for reducing the impact of RR mechanism.  
% Notably, when $1-\nu_l \asymp\rho$, the second term becomes comparable to the first, noting that $\rho=o_p(1)$ and $\nu_l$ is lower bounded by some positive constant in this case. 
% In addition, Theorem \ref{theo: non-debiased} in the Supplementary Materials indicates the necessity of debiasing in \eqref{eq:db}. Specifically, we show that when using the non-debiased $\tilde{A}_l$, the resulting error bound of estimated eigenspace would include extra bias. }
% \fi

{The first two terms in \eqref{eq:step1err}, $\mathcal I_1$ and $\mathcal I_2$, involve the privacy parameter $\nu_l$. 
\textcolor{black}{It is worth mentioning that the $\nu_l^{-1}$ effect on the error bound of estimated eigenspace has been observed in the literature, which studies the RR mechanism under the SBM and its variants, including \citet{chakrabortty2024prime,hehir2022consistent}. In particular, \citet{chakrabortty2024prime} provides the minimax optimality of RR mechanism under local DP within the framework of degree-corrected mixed membership SBMs. }  
The third term, $\mathcal I_3$, reflects the effect of heterogeneity; see also the analogous result in the regression setting \citep{li2022transfer}. When the privacy parameters are bounded away from zero, i.e., $q_l = q'_l > \tfrac{1}{2} + \tau$ for some small constant $\tau > 0$, the term $I_1$ corresponds to the standard statistical error from aggregating multiple networks (that is, the error bound for the eigenvectors of $\sum_l w_l A_l$), whereas $I_2$ captures the additional cost introduced by the RR mechanism. Notably, when $1 - \nu_l \asymp \rho$, the second term becomes comparable to the first, since $\rho = o(1)$ and $\nu_l$ remains bounded below by a positive constant in this case.

Furthermore, Theorem \ref{theo: non-debiased} in the Supplementary Materials highlights the necessity of debiasing in \eqref{eq:db}. Specifically, it shows that if one instead uses the non-debiased $\tilde{A}_l$, the resulting error bound for the estimated eigenspace would incur an additional bias term. {\color{black}In addition, we discuss the optimality of $\bar{U}$ in Section \ref{app:opt} in the Supplementary Materials.}
}
\end{remark}

% \begin{remark}
% \label{remark: db}
% We have used the eigenvectors of the debiased adjacency matrix $\hat{A}_l$ in \eqref{eq:weiagg}. In principle, we can also use the non-debiased matrix $\tA_l$, whose population matrix is $\tilde{P}_l:=\Theta \tilde{B}_l \Theta^\intercal$ with $\tilde{B}_l:= (q_l+q'_l-1)B_l + (1-q_l)\mathbf 1\mathbf 1^\intercal$. Theorem \ref{theo: non-debiased} in Section \ref{app:add} of Supplementary Materials provides theoretical evidence for the debiasing step under the generic weights. It turns out that using the non-debiased matrix $\tA_l$ would yield extra bias.  

% \end{remark}

\subsection{Adaptive weights and error-bound-oracle property}
For notational simplicity, for $l\in[L]$, we define
\begin{equation}
\label{eq:definitionofpl}
\mathcal P_l:= \sqrt{\frac{\nu_l\rho+(1-\nu_l)/2}{ \nu_l^2}\cdot \frac{\varsigma_n}{n\rho^2}}\quad {\rm with}  \quad\varsigma_n:= \max\{1,\frac{\log n}{L}\}.
\end{equation}
Here, $\mathcal P_l$ reflects the privacy-preserving level of the $l$-th source network. Specifically, noting that $\frac{\nu_l\rho+(1-\nu_l)/2}{ \nu_l^2}$ is a decreasing function of $\nu_l$ and recalling that $\nu_l=q_l+q'_l-1$, we have that the larger the value of $\mathcal P_l$, the higher the level of privacy preservation for the $l$-th source network.

The error bound in Proposition \ref{theo: step1}  motivates the selection of the weights $w_l(l\in[L])$. 
Indeed, we note that
\begin{align}
\label{eq:upperb}
 {\rm dist}(\bar{U}, U_0)\lesssim \left(\sum_{l=1}^L w_l^2 \left( \mathcal P_{l}^2 L +  \mathcal E_{\theta,l}^2 L\right)\right)^{1/2},
\end{align}  
{\color{black}where the optimal weights that minimize the right-hand side of \eqref{eq:upperb} can be obtained in closed form.} \textcolor{black}{Driven by this, we propose the following adaptive weighting strategy.} 
\begin{definition}[Adaptive weighting strategy]
\label{def:ada}
The adaptive weights are defined as
\begin{equation}
\label{eq: adapweight}
\hat{w}_l\varpropto \left(\hat{\mathcal P}_{l}^2+\hat{\mathcal E}_{\theta,l}^2 + \frac{{\varsigma_n}}{{n}{\hat{\rho}_l}^2}\right)^{-1} \;\; {\rm for}\; l\in[L],
\end{equation}
where $\hat{\mathcal P}_{l}$ is an estimator of $\mathcal P_l$ with $\rho$ replaced by $\hat{\rho}_l:= \frac{\sum_{ij} \hat{A}_{l,ij}}{n(n-1)}$, $\hat{\mathcal E}_{\theta,l}:= \|\hat{U}_l\hat{U}_l^\intercal - \hat{U}_0\hat{U}_0^\intercal\|_2$, and $\varsigma_n$ is defined in \eqref{eq:definitionofpl}.

% \begin{equation}
% \label{eq:estim}
% \hat{\mathcal P}_{l}^2:=[\nu_l\hat{\rho_l}+(1-\nu_l)/2]\cdot \frac{{\varsigma_n}}{{n}\hat{\rho}_l^2\textcolor{black}{\nu^2_l}}\; {\rm with}\; \hat{\rho}_l:= \frac{\sum_{ij} \hat{A}_{l,ij}}{n(n-1)},\; {\rm and} \; \hat{\mathcal E}_{\theta,l}:= \|\hat{U}_l\hat{U}_l^\intercal - \hat{U}_0\hat{U}_0^\intercal\|_2,\quad \varsigma_n:= \max\{1,\frac{\log n}{L}\}\; {\rm and\;} \nu_l\; {\rm is \; defined\; in \;} \eqref{eq:nul}.
% \end{equation}
\end{definition}

The first term and second term in \eqref{eq: adapweight} correspond to the estimated squared privacy level and estimated squared heterogeneity level, respectively. The third term is added to balance the weights for networks whose first and second terms are small (i.e., the \emph{informative networks} defined later), which prevents excessive weighting of a few networks and ensures that the information from other informative networks is fully utilized. 
% The first term $[\nu_l\hat{\rho_l}+(1-\nu_l)/2]\cdot \frac{{\varsigma_n}}{{n}\hat{\rho}_l^2\textcolor{black}{\nu^2_l}}$ relates to the level of privacy preservation, the second term $\hat{\mathcal E}_{\theta,l}^2 $ relates to the level of \textcolor{black}{estimated heterogeneity},  \textcolor{black}{and the third term $ \frac{\varsigma_n}{n{\rho^2}}$ is added for technical convenience, which controls the estimation error of $\hat{\mathcal E}_{\theta,l}^2$ and helps relate $\hat{\mathcal E}_{\theta,l}^2$ to its population counterpart ${\mathcal E}_{\theta,l}^2$.}
% which is actually the
% estimation error bound of $\hat{\mathcal E}_{\theta,l}^2 L$ with respective to its population counterpart  $\|U_l U_l^\T - U_0 U_0^\T\|_2^2L$. 
The expression (\ref{eq: adapweight}) therefore implies that when both the estimated privacy-preserving level and the estimated heterogeneity level are small, the weight $\hat{w_l}$ is large, which coincides with the intuition for good weighting.  It is worth noting that the adaptive weighting strategy is defined up to multiplicative constants; such constants do not affect the
theoretical results and the weights are normalized such that $\sum_{l=1}^L \hat{w}_l=1$.

An appealing property of the adaptive weights defined in \eqref{eq: adapweight} is that {the error bound of the estimated eigenspace, i.e., ${\rm dist}(\bar{U}, U_0)$, only depends on the \emph{informative networks}. More precisely, the influence of the {non-informative networks} on the error bound is dominated by that of the informative networks.} We refer to this as the \textbf{error-bound-oracle property}. 

{\color{black}Before presenting this result}, we first provide the formal definition of  \emph{informative networks}.

\begin{definition}[Informative networks]
\label{def:informative}
The informative networks set $S$ 
%with {\color{black}$|S|=m>0$} 
is given as:
\begin{equation*}
S=\left\{1\leq l\leq L : \;{\mathcal P}_{l}^2+ {\mathcal E}_{\theta,l}^2\leq \eta_n\right\}, 
\end{equation*}
where $\eta_n\asymp \frac{\varsigma_n}{n\rho^2}$. Correspondingly, the source networks in $S^c$ are called non-informative networks. 
\end{definition}

\begin{remark}
\textcolor{black}{
The informative set $S$ collects source networks whose privacy level (via $\nu_l$) and heterogeneity $\mathcal E_{\theta,l}$ are jointly controlled. Indeed, we require that both $\mathcal P_l^2\lesssim \varsigma_n/(n\rho^2)$ and $\mathcal E_{\theta,l}^2\lesssim \varsigma_n/(n\rho^2)$ hold simultaneously. 
In particular, $\mathcal P_l^2\lesssim \varsigma_n/(n\rho^2)$ effectively enforces $\nu_l\geq \tau$ for some constant $0<\tau\leq 1$ among informative networks, while $\mathcal E_{\theta,l}^2\lesssim \varsigma_n/(n\rho^2)$ bounds the squared heterogeneity by its estimation error (see Lemma  \ref{lem: accuracyofe} in the Supplement). 
% Throughout, we adopt  Definition \ref{def:informative}. One may relax the order of $\eta_n$ in Definition \ref{def:informative} to $\eta_n\asymp\varsigma_n/({n\rho^2\nu_{*}})$, in which $\nu_{*}\in(0,1]$ denotes the acceptable privacy level for informative sources and it is allowed to approach $0$; the adaptive weights and their theoretical guarantees extend accordingly. 
}
\end{remark}
\iffalse
\textcolor{black}{The informative networks set $S$ consists of source networks with privacy-preserving levels and heterogeneity jointly bounded.
Specifically, $\mathcal P_l^2\lesssim \frac{\varsigma_n}{n\rho^2}$ essentially requires that $\nu_l\geq \tau$ for some constant $0<\tau\leq 1$; $\mathcal E_{\theta,l}^2\lesssim \frac{\varsigma_n}{n\rho^2}$ requires that the squared heterogeneity level is bounded above by its estimation error bound, $\frac{\varsigma_n}{n\rho^2}$ (see Lemma \ref{lem: accuracyofe} in the Supplementary Materials). 
}

\begin{remark}
In this paper, we focus on Definition \ref{def:informative} for the informative networks. It is also worth mentioning that, depending on the specific application, the order of $\eta_n$ can be relaxed to $\frac{\varsigma_n}{n\rho^2\nu_{*}}$, where $0 < \nu_{*} \leq 1$ represents the expected privacy-preserving level of the informative networks that is acceptable to data analysts. The adaptive weights and its theoretical properties can be similarly defined and analyzed. 
\end{remark}
\fi

\vspace{-1cm}
\begin{remark}
\textcolor{black}{
In practice, the informative set $S$ is unknown and need not be estimated explicitly; it is introduced solely to \textcolor{black}{study} the error-bound-oracle property. The adaptive weighting scheme provides a data-driven mechanism that automatically up-weights networks with controlled heterogeneity and privacy, while down-weighting highly privatized or heterogeneous ones. Consequently, the aggregated eigenspace estimator attains the oracle-like error bound in the theorem below, as if $S$ were known \textit{a priori}.} 
% \textcolor{black}{Note that the theorem below assumes that the cardinality $m$ of the set of informative networks is greater than zero. When $m=0$, i.e., there is no informative source network, the error-bound-oracle property does not apply. Nevertheless, even in this case, the aggregated eigenspace estimator with adaptive weights still achieves an error bound that is no larger than that of the estimator using equal weights. 
%one can directly apply Proposition \ref{theo: step1} to obtain the error bound.

\end{remark}
\iffalse
\textcolor{black}{The informative networks need not to be estimated. Instead, the adaptive weights automatically upweight informative networks and downweight non-informative ones.} The next theorem demonstrates the effectiveness of the adaptive weights, which can cancel out the effect of non-informative source networks on the error bound automatically. 
\fi
% Recall the definition of $\nu_l$ in \eqref{eq:nul}. By the definition of informative networks, we have
% $$\nu_l\gtrsim \nu_{*}\quad {\rm for}\quad l\in S, $$ and we further assume that $$\nu_l \lesssim \nu^{*} \quad {\rm for}\quad l\in S.$$ 

% {\color{black}Comment: In condition (19), can we replace $\min_{l\in S^c} \left(\mathcal E_{\theta,l}^2 + {\mathcal P_{l}^2}\right)$ by $\eta_n$, which seems to be a sufficient condition for (19) and it makes the inequality looks simpler?}

% {\color{black}Comment: In condition (19), I think that $\varsigma_n\cdot \min\{{\frac{\log n}{L}},\;\frac{L}{\log n}\}=\min\{{\frac{\log n}{L}},\;1\}$. By replacing  $\min_{l\in S^c} \left(\mathcal E_{\theta,l}^2 + {\mathcal P_{l}^2}\right)$ with $\eta_n$, the condition in (19) becomes $\frac{L-m}{m}\lesssim \min\{1,\frac{\nu^2_{*}}{(\nu^{*})^2}\}\cdot\min\{{\frac{\log n}{L}},\;1\}$ which leads to $\frac{L-m}{m}\lesssim 1$.

% Without the replacement, the condition can be written $$\frac{L-m}{m}\lesssim \eta_n^{-1}\min\{1,\frac{\nu^2_{*}}{(\nu^{*})^2}\}\min\{{\frac{\log n}{L}},\;1\}\min_{l\in S^c} \left(\mathcal E_{\theta,l}^2 + {\mathcal P_{l}^2}\right)$$.}

\begin{theorem}[Error-bound-oracle]
\label{theo: effectofadaptive}
Suppose the assumptions in Proposition \ref{theo: step1} all hold, $\lambda_{\min} (P_l) \gtrsim n\rho$ for all $l \in [L]$, and $m=|S|>0$.
If 
\begin{equation}
\label{eq:conditionfororacle}
\frac{L-m}{m}\lesssim \frac{ \min_{l\in S^c} ({\mathcal E}_{\theta,l}^2+ {\mathcal P}_{l}^2) }{\eta_n} \cdot \min\{1,\;\frac{\log n}{L}\},
\end{equation}
then with probability larger than $1-\frac{L}{n^{\kappa}}$ for some constant $\kappa>0$, 
the adaptive weights $\hat{w}_l$'s defined in (\ref{eq: adapweight}) satisfy
\begin{equation}
\label{eq:order}
{w}_l \asymp \frac{1}{m}  \quad {\rm for }\quad l \in S;
\end{equation}
{\small
\begin{align}
\label{eq:orcalepro}
{\rm dist} (\bar{U}, U_0) &\lesssim   \sqrt{\frac{1}{m^2}\sum_{l\in S} \frac{\log n}{n\rho\nu_l}}+\sqrt{\frac{1}{m^2}\sum_{l\in S} \frac{1-\nu_l}{\rho}\cdot \frac{\log n}{n\rho\nu_l^2}}+\frac{1}{m}\sum_{l\in S}\mathcal E_{\theta,l}
:=\mathcal M+ \mathcal P + \mathcal H.
\end{align}
}
\end{theorem}

\begin{remark}
{\color{black}The error-bound-oracle property in Theorem \ref{theo: effectofadaptive} naturally requires that $S$ be nonempty. When it is empty, this property no longer applies; however, the aggregated eigenspace estimator still obeys the error bound in Proposition \ref{theo: step1}, ensuring that the subsequent transfer-learning step remains effective.}
%The condition $\lambda_{\min} (P_l) \gtrsim n\rho$ is required to establish the accuracy of $\hat{\mathcal E}_{\theta,l}$ in estimating its population analogue.
Condition \eqref{eq:conditionfororacle} requires that the ratio of
non-informative to informative networks be upper bounded in terms of $\min_{l\in S^c} ({\mathcal E}_{\theta,l}^2+{\mathcal P}_{l}^2)$. 
This requirement is mild. For a simple illustration, suppose $L \lesssim \log n$, then \eqref{eq:conditionfororacle} simplifies to $\frac{L-m}{m}\lesssim \frac{ \min_{l\in S^c}({\mathcal E}_{\theta,l}^2+{\mathcal P}_{l}^2) }{\eta_n}$. By the definition of $S$ (informative networks), we have
$\min_{l\in S^c}({\mathcal E}_{\theta,l}^{2}+{\mathcal P}_{l}^2)> \eta_n$; hence
$(L-m)/m\lesssim 1$ is a sufficient condition for
\eqref{eq:conditionfororacle} (i.e., a non-vanishing fraction of networks are
informative).
\end{remark}

\begin{remark}
{The error bound \eqref{eq:orcalepro} depends only on the indices of the informative networks; its three components have the same interpretation as in Proposition \ref{theo: step1}. {\color{black}Note that by Definition \ref{def:informative}, for $l\in S$, $\nu_l\geq \tau$ for some constant $0<\tau\leq 1$. When the privacy protection for a source network is strong (i.e., $\nu_l$ approaches zero), the adaptive weighting scheme assigns it a near-zero weight. Consequently, the error bound in Theorem \ref{theo: effectofadaptive} depends only on networks with $\nu_l\geq \tau$, thereby mitigating potential blow-up under heterogeneous privacy levels.} 
When $\mathcal E_{\theta,l}$'s are small and $\nu_l$'s are large, the adaptive weighting step alone ensures good performance in eigenspace estimation. By contrast, if $\mathcal E_{\theta,l}$'s are large and $\nu_l$'s are small, the second and third terms can dominate the first term.  
Therefore, performing the regularization step introduced in Section \ref{sec:regularization} is beneficial for mitigating the adverse effects of privacy noise and heterogeneity on the estimated eigenspace.}
\end{remark}

By Proposition \ref{theo: step1}, the error bound under the equal weighting strategy, i.e., $w_l=1/L$ for all $l\in[L]$, turns out to be $$ {\rm dist} (\bar{U}, U_0) \lesssim\sqrt{\frac{1}{L^2}\sum_{l\in [L]}\frac{\log n}{n\rho{\nu_l}}}+\sqrt{\frac{1}{L^2}\sum_{l\in [L]}\frac{1-\nu_l}{\rho}\cdot\frac{\log n}{n\rho\nu_l^2}}+\frac{1}{L}\sum_{l\in [L]} \mathcal E_{\theta,l}.$$ 
The next theorem shows that the adaptive weighting strategy leads to an order-wise smaller bound than the equal weighting strategy.  

\begin{theorem}
\label{theo: effectofadaptivesecond}
Suppose all the conditions in Theorem \ref{theo: effectofadaptive} hold. If we further have 
\begin{equation}
\label{eq:genequal}
 \frac{ \min_{l\in S^c} {(\mathcal E}_{\theta,l}^2+\mathcal P_l^2) }{\eta_n}\cdot (L-m)^2=w\left(\frac{L^3}{m}\cdot  \max\{1,\frac{L}{\log n}\}\right),
\end{equation}
then, with the same probability as in Theorem \ref{theo: effectofadaptive}, the error bound of ${\rm dist}(\bar{U}, U_0)$ under the adaptive weights $\hat{w}_l$'s is of smaller order than the {\color{black}error bound of ${\rm dist}(\bar{U}, U_0)$} under the equal weights.
\end{theorem}

\begin{remark}
Condition \eqref{eq:genequal} actually requires that 
the privacy-corrected heterogeneity gap between informative and non-informative networks should be large. Indeed, by the definition of informative networks, we have  $\frac{ \min_{l\in S^c} ({\mathcal E}_{\theta,l}^2 + {\mathcal P}_{l}^2)}{\eta_n}=\Omega(1)$. 
Suppose 
$L \lesssim \log n$ and $m< L$, then \eqref{eq:genequal} reduces to $\frac{ \min_{l\in S^c} {(\mathcal E}_{\theta,l}^2+\mathcal P_l^2) }{\eta_n}=w\left(\frac{L^3}{m (L-m)^2}\right)$ which 
enforces a larger $\min_{l\in S^c} ({\mathcal E}_{\theta,l}^2+ {\mathcal P}_{l}^2)$ when $m$ is small. 

\end{remark}

\section{Regularization}
\label{sec:regularization}
The adaptive weighting step in Section \ref{sec:adative weighting} computes a good estimator of the population target eigenspace $U_0$ by weighting each source network differently according to their heterogeneity and privacy-preserving levels.
In this section, we introduce the second step (i.e., the regularization step) of \texttt{TransNet}, where we propose a regularization method to further enhance the accuracy of estimating the target eigenspace using the target network.  
The theoretical error bound for the estimated eigenspace is also provided.   

Specifically, we propose the following ridge-type regularization:  
\begin{equation}
\label{eq:finetune}
{\hat{U}_0^{  RE}}(\lambda) \in \argmax_{V^\intercal V=I} \left\{{\rm tr}(V^\intercal {\hat{U}_0\hat{U}_0^\intercal}  V)-\frac{\lambda}{2} \|VV^\intercal- \bar{U}  \bar{U}^\intercal \|_F^2\right\},
\end{equation}
where the first term represents the spectral clustering objective function on ${\hat{U}_0\hat{U}_0^\intercal}$, whose solution is $\hat{U}_0$ \textcolor{black}{if no penalty (i.e., $\lambda=0$) is used}, and the second term is a penalty function aimed at minimizing the difference between the estimated eigenspace and $\bar{U}$.

\begin{remark}
Problem \eqref{eq:finetune} can also be written as 
\begin{equation*}
{\min} \;\|{\hat{U}_0\hat{U}_0^\intercal} -(\bar{U}\bar{U}^\intercal+\delta)\|_F^2 + \lambda \|\delta\|_F^2, \quad  s.t., \quad (\bar{U}\bar{U}^\intercal+\delta)^\intercal (\bar{U}\bar{U}^\intercal+\delta)=I,
\end{equation*}
which shares a similar spirit with the transfer learning method in the high-dimensional regression context \citep{li2022transfer}.
\end{remark}

The next theorem provides the accuracy of ${\hat{U}_0^{  RE}}(\lambda)$ and shows how the regularization step can further improve the accuracy in terms of the target eigenspace estimation.

\begin{theorem}
\label{theo: step2}
Suppose the conditions in Theorem \ref{theo: effectofadaptive} all hold. Then with the same probability as in Theorem \ref{theo: effectofadaptive}, the eigenspace $\hat{U}_0^{  RE}(\lambda^*)$  satisfies that,
\begin{align}
\label{eq:step2err}
 {\rm dist}(\hat{U}_0^{  RE}(\lambda^*), U_0)
&\lesssim \mathcal M+ \frac{1}{\sqrt{n\rho\nu_0}}\max\left\{1, \sqrt{\frac{1-q'_0}{\rho\nu_0}}\right\} \wedge \left(\mathcal P\vee \mathcal H\right),
\end{align} 
where $\mathcal M,\mathcal P$ and $\mathcal H$ are defined in \eqref{eq:orcalepro} and
$\lambda^*\asymp \frac{1}{\sqrt{n\rho\nu_0}}\max\left\{1, \sqrt{\frac{1-q'_0}{\rho\nu_0}}\right\} /(\mathcal P\vee \mathcal H)$.
\end{theorem}

\begin{remark}
Note that the error bound when using only the eigenspace $\hat{U}_0$ of the target network is given by $\mathcal T:=\frac{1}{\sqrt{n\rho\nu_0}}\max\left\{1, \sqrt{\frac{1-q'_0}{\rho\nu_0}}\right\}$. Thus, 
bound \eqref{eq:step2err} shows the regularization estimator $\hat{U}_0^{RE}(\lambda^*)$ improves both the bounds of $\bar{U}$ (\textcolor{black}{see \eqref{eq:orcalepro}}) and $\hat{U}_0$. In particular, when both the effects of privacy and heterogeneity are small, such as when $\mathcal P\vee \mathcal H\lesssim \mathcal T$, the accuracy of $\hat{U}_0^{  RE}(\lambda^*)$ is enhanced as if the knowledge within the source networks are transferred to the target network. On the other hand, when the effect of privacy or heterogeneity is large, meaning the auxiliary information in the source networks is noisy, such as when
$\mathcal P\vee \mathcal H \gtrsim \mathcal T$, the performance of $\hat{U}_0^{  RE}(\lambda^*)$ is at least as good as that achieved using only the target network. \textcolor{black}{We establish the error bound \eqref{eq:step2err} under the condition that the set of informative source networks $S$ is nonempty (i.e., $m>0$). In the case $m=0$, we can similarly obtain an error bound for $\hat{U}_0^{RE}(\lambda^*)$  by using the bound for $\bar{U}$ in Proposition \ref{theo: step1} in place of Theorem \ref{theo: effectofadaptive}.} 
\end{remark}

\begin{remark}
Although we used the adaptive weighting strategy when deriving the bound for the regularization step. Similar results hold when using arbitrary weights $w_l$'s. Indeed, $\mathcal M$, $\mathcal P$ and $\mathcal H$ in \eqref{eq:step2err} can be replaced by their corresponding terms under the chosen weights.    
\end{remark}
%\vspace{-1cm}

\section{Algorithm}
\label{sec:clustering}
\textcolor{black}{
In this section, we present the complete \texttt{TransNet} algorithm and provide a theoretical analysis of its misclassification rate.}

With the estimated eigenspace obtained in the regularization step, we only need to perform the $k$-means on the rows of the eigenspace to obtain the final clusters. The whole algorithm is then summarized in Algorithm \ref{alg:transnet}. {\color{black}Note that, for communication efficiency, the debiased matrices $\hat{A}_l$'s
 are retained locally by the data curators and are not transmitted to the central server or released to the outside. Instead, the central server only receives the local eigenvectors $\hat{U}_l$'s
 for aggregation. The final estimated eigenspace and cluster labels are released to outside. By the post-processing property of DP, all computations based on $\hat{A}_l$'s without using raw data remain DP with the same privacy guarantee.}

\begin{algorithm}[!htbp]
\small

\renewcommand{\algorithmicrequire}{\textbf{Input:}}

\renewcommand\algorithmicensure {\textbf{Output:} }

\caption{\textbf{Trans}fer learning of \textbf{net}works (\texttt{TransNet})}

\label{alg:transnet}

\begin{algorithmic}[1]

\STATE \textbf{Input:} Bias-corrected source networks $\{\hat{A}_l\}_{l=1}^L$, bias-corrected target network $\hat{A}_0$, the number of clusters $K$, and the tuning parameter $\lambda>0$. \\
\STATE \,\underline{\textbf{Step 1 (Adaptive weighting):}}
\STATE \, \texttt{Eigen-decomposition:} Compute the $K$ leading eigenvectors $\hat{U}_l$'s of $\hat{A}_l$'s for $l\in \{0\}\cup [L]$;  \\
\STATE \, \texttt{Procrustes fixing:} Obtain $\hat{U}_lZ_l$ with $Z_l:= \argmin _{Z\in \mathcal O_K} \|\hat{U}_lZ- \hat{U}_0\|_{F}$ for $l\in [L]$;  \\
\STATE\, \texttt{Weighted aggregation:} Compute the weights $\hat{w}_l$'s according to \eqref{eq: adapweight} and aggregate \\
\,\,\,$\hat{U}:=\sum_{l=1}^L \hat{w}_l\hat{U}_lZ_l$ and orthogonalize $\bar{U}, \bar{R}:= {\rm QR} (\hat{U})$.  \\

\STATE \underline{\textbf{Step 2 (Regularization):}}
\STATE \, Optimize and obtain ${\hat{U}_0^{  RE}}(\lambda) \in \argmax_{V^\intercal V=I} {\rm tr}(V^\intercal {\hat{U}_0\hat{U}_0^\intercal}  V)-\frac{\lambda}{2} \|VV^\intercal- \bar{U}  \bar{U}^\intercal \|_F^2$.

\STATE \underline{\textbf{Step 3 (Clustering):}}
\STATE \, Conduct $k$-means clustering on the \textcolor{black}{rows} of ${\hat{U}_0^{  RE}}(\lambda)$ to obtain $K$ clusters. \\
\STATE \textbf{Output:} The estimated eigenspace ${\hat{U}_0^{  RE}}(\lambda)$ and the estimated $K$ clusters. 

\end{algorithmic}
\end{algorithm}

With the error bound of the estimated eigenspace, it is straightforward to obtain the error bound on the misclassification rate after the clustering step, given by the next theorem.

\begin{theorem}
\label{theo: step3}
Suppose Assumption \ref{assu:balcom} and the conditions in Theorem  \ref{theo: step2} all hold. Denote the set of misclassified nodes by $\mathbb M$.
Then, with probability larger than $1-\frac{L}{n^{\kappa}}$ for the constant $\kappa>0$ appeared in Theorem \ref{theo: step1}, Algorithm \ref{alg:transnet} with the adaptive weighting defined in \eqref{eq: adapweight} has misclassification rate upper bounded by
$\frac{|\mathbb M|}{n}\lesssim Err^2,$
where $Err$ denotes the RHS of \eqref{eq:step2err}.
\end{theorem}

Theorem \ref{theo: step3} shows that the misclassification rate is {\color{black}bounded by} the squared error of the estimated eigenspace derived in Theorem \ref{theo: step2}. As a result, the effect of transfer learning on clustering can be discussed similarly as in Theorem \ref{theo: step2}. Therefore, we omit the detailed discussions. Additionally, the misclassification rate for arbitrary weights can be similarly derived by replacing $\mathcal E$ with the corresponding error bound of the estimated eigenspace obtained after the regularization step.

\section{Simulations}
\label{sec:sim}
In this section, we test the finite sample performance of the proposed algorithm and compare it with several competitors. 

\vspace{0.2cm}
\noindent\textbf{Methods compared.} 
We consider two transfer learning-based methods. The first is the proposed algorithm, denoted by \texttt{TransNet-AdaW}, where we use the adaptive weights defined in Definition \ref{def:ada}. The second is the counterpart of the proposed algorithm that uses equal weights, denoted by \texttt{TransNet-EW}, where $w_l = 1/L$. The tuning parameter $\lambda$ in both algorithms is selected via cross-validation, following a procedure similar to that in \citet{chen2018network}; see the Supplementary Materials for the details.
We compare the transfer learning-based methods with the distributed counterpart of \texttt{TransNet-EW} using only the source networks, denoted by \texttt{Distributed SC}, which corresponds to the \textcolor{black}{method of} \citet{charisopoulos2021communication}; and the spectral clustering algorithm using only the target network, denoted by \texttt{Single SC}.

\begin{itemize}
    \item \texttt{Distributed SC}: obtained from the Step 1 of Algorithm \ref{alg:transnet} under the equal weighting strategy and the follow-up $k$-means;
    \item \texttt{Single SC}: obtained from the spectral clustering on the debiased matrix $\hat{A}_0$ of the target network. 
\end{itemize}
\vspace{0.2cm}
\noindent\textbf{Performance measures.} \textcolor{black}{Consistent with} the theoretical analysis, we use the \emph{projection distance} to measure the subspace distance between the estimated eigenspaces and the true eigenspaces and use the \emph{misclassification rate} to measure the proportion of misclassified (up to permutations) nodes.

% \subsection{Algorithm evaluation}
% \label{sub:alge}
\vspace{0.2cm}
\noindent\textbf{Network generation.} The networks are generated from the HMSBM in (\ref{MOD:SBM}). We fix $n=120$, $K=3$, and let the number of source networks $L$ vary from 8 to 24. Next, we provide the setup of connectivity matrices, the community assignments, and the privacy-preserving levels for the target network and source networks, respectively. 

For the target network, we set
%\begin{equation*}
$\scriptsize
B_0=\begin{bmatrix}
0.3 & 0.1 & 0  \\
0.1 & 0.3 & 0.06\\
0 & 0.06 & 0.3
\end{bmatrix},
$
%\end{equation*}
and the communities are of equal size, with the first community consisting of the initial 40 nodes, and so on. The probability parameters $q_0$ and $q'_0$ in RR are both fixed to be 0.95. 

We classify the source networks into four equal-sized groups. In the same group, the connectivity matrices, the community assignments, and the privacy-preserving levels are identical. 
We set $B_l=B^{(1)}$ for $l\in \{1,...,[L/4]\}$, $B_l=B^{(2)}$ for $l\in \{[L/4]+1,...,[L/2]\}$, $B_l=B^{(3)}$ for $l\in \{[L/2]+1,...,[3L/4]\}$ and $B_l=B^{(4)}$ for $l\in \{[3L/4]+1,...,L\}$ with 
\begin{equation*}
\scriptsize
B^{(1)}=\begin{bmatrix}
0.3 & 0.1 & 0.1  \\
0.1 & 0.3 & 0.06\\
0.1 & 0.06 & 0.3
\end{bmatrix}, 
 B^{(2)}=\begin{bmatrix}
0.3 & 0.1 & 0  \\
0.1 & 0.2 & 0.06\\
0 & 0.06 & 0.2
\end{bmatrix}, 
B^{(3)}=\begin{bmatrix}
0.3 & 0.1 & 0  \\
0.1 & 0.3 & 0.1\\
0 & 0.1 & 0.3
\end{bmatrix}, 
B^{(4)}=\begin{bmatrix}
0.3 & 0.15 & 0  \\
0.15 & 0.3 & 0.06\\
0 & 0.06 & 0.3
\end{bmatrix}.
\end{equation*}
For the community assignments of source networks within each group, a certain number of nodes in each target community are randomly selected to change their community randomly to the other two remaining communities. For source networks in group $i(i\in[4])$, the proportion of nodes that need to change their community are denoted by $\mu^{(i)}$. We also denote $\mu = (\mu^{(1)},\mu^{(2)},\mu^{(3)},\mu^{(4)})$. \textcolor{black}{Larger values in $\mu$ correspond to higher heterogeneity levels.} 
For the privacy-preserving levels, the edge-preserving parameters in RR (including 0 to 0 and 1 to 1) in group $i(i\in[4])$ is fixed to be  $q^{(i)}$, and we denote 
$q=(q^{(1)},q^{(2)},q^{(3)},q^{(4)})$. \textcolor{black}{Lower values of $q$ correspond to higher privacy-preserving levels.}  

We consider three different set-ups, depending on whether the source networks have heterogeneous community assignments (compared to the target network) and the privacy-preserving demand. To fix ideas, in the following, we refer to the source networks as \emph{heterogeneous} if they differ from the target network in community assignments. The three set-ups correspond to the following Experiments I--III, respectively.  

\begin{table}[!htb]
\centering
\small
\caption{The parameter set-ups of Experiments I--III. For each experiment, we consider three cases. In all cases, the source networks are classified into 4 groups. We denote $\mu = (\mu^{(1)},\mu^{(2)},\mu^{(3)},\mu^{(4)})$, where $\mu^{(i)}$ denotes the proportion of nodes in each of the $i$th group of source networks that randomly alters their communities in the target network. We denote $q=(q^{(1)},q^{(2)},q^{(3)},q^{(4)})$, where $q^{(i)}$ denotes the edge-preserving parameters in RR (including 0 to 0 and 1 to 1) in each of the $i$th group of source networks.  }\vspace{0.5cm}
\def\arraystretch{1.3}
\begin{tabular}{c|c|c|c}
\hline
\textbf{Cases} &\tabincell{c}{ \textbf{Experiment I}} & \tabincell{c}{\textbf{Experiment II}} & \tabincell{c}{\textbf{Experiment III}} \\
\hline
{Case I}&$q=(0.95,0.95,0.7,0.7)$&$\mu=(0.02,0.02,0.5,0.5)$& \tabincell{c}{$\mu=c(0.02,0.02,0.5,0.5)$\\$q=c(0.95,0.95,0.7,0.7)$ }\\\hline
{Case II}&$q=(0.95,0.95,0.8,0.8)$&$\mu=(0.02,0.02,0.3,0.3)$&\tabincell{c}{$\mu=(0.1,0.1,0.5,0.5)$\\$q=c(0.95,0.95,0.7,0.7)$ } \\\hline
{Case III}&$q=(0.8,0.8,0.8,0.8)$&$\mu=(0.3,0.3,0.3,0.3)$& \tabincell{c}{$\mu=c(0.02,0.02,0.5,0.5)$\\$q=c(0.8,0.8,0.95,0.95)$ }\\

\hline
\end{tabular}
\label{table:experi}
\end{table}

\vspace{0.2cm}
\noindent\textbf{Experiment I: Private but non-heterogeneous.} We consider the setup where the community assignments of the source networks are the same as the target network, while the source networks are private. Specifically, we test three cases of the privacy-preserving level $q$ for the source networks; see the first column of Table \ref{table:experi}. The average results over 10 replications are displayed in Figure \ref{sim_pri} (Cases I and II) and Figure \ref{sim_pri_case3} in Section \ref{add:num} of Supplementary Materials (Case III).

We have the following observations. First, the transfer learning-based methods \texttt{TransNet-EW} and \texttt{TransNet-AdaW} perform better than the distributed learning-based method \texttt{Distributed SC} and \texttt{Single SC}, which is based on the single target network. As expected, when $L$ is large and the privacy-preserving levels of the source networks are highly different (Case I and Case II), the advantage of \texttt{TransNet-AdaW} over \texttt{TransNet-EW} is apparent. Second, when the source networks are highly privacy-preserved (Case III; see Figure \ref{sim_pri_case3} in Section \ref{add:num} of Supplementary Materials) and hence of low quality, \texttt{Distributed SC} performs poorly and the transfer learning-based methods \texttt{TransNet-AdaW} and \texttt{TransNet-EW} become necessary.

\begin{figure*}[t]{}
%6.80 6.43
\centering
 \subfigure[Case I]%{\includegraphics[height=4.5cm,width=4.5cm,angle=0]{pri_dis_case1.pdf}}
{\includegraphics[height=4.2cm,width=4.2cm,angle=0]{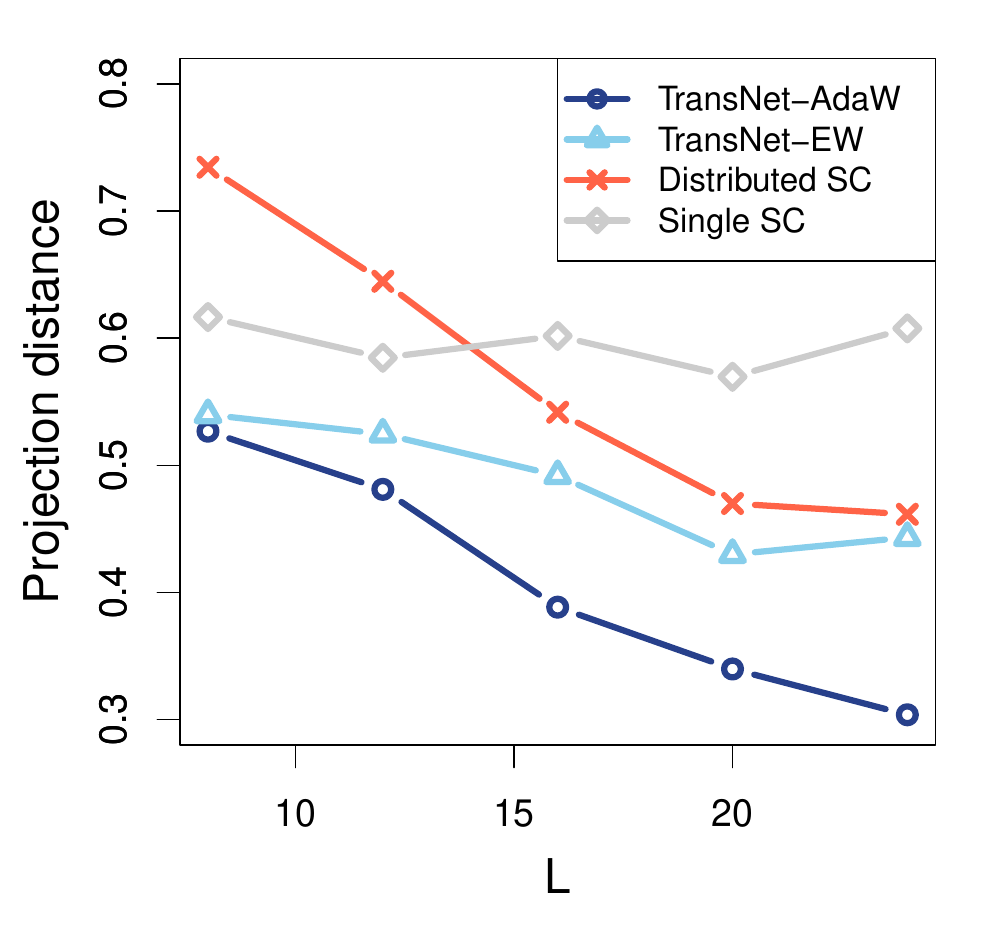}}
\subfigure[Case II]{\includegraphics[height=4.2cm,width=4.2cm,angle=0]{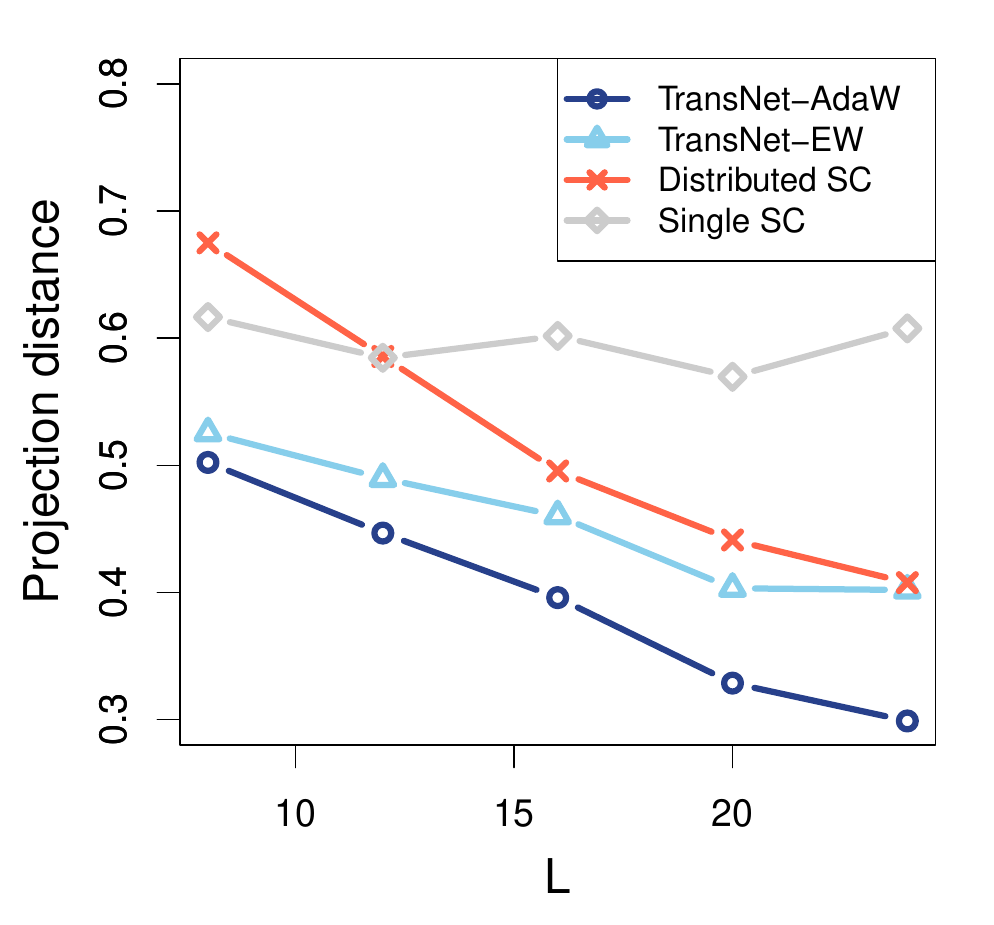}}\newpage
\subfigure[Case I]{\includegraphics[height=4.2cm,width=4.2cm,angle=0]{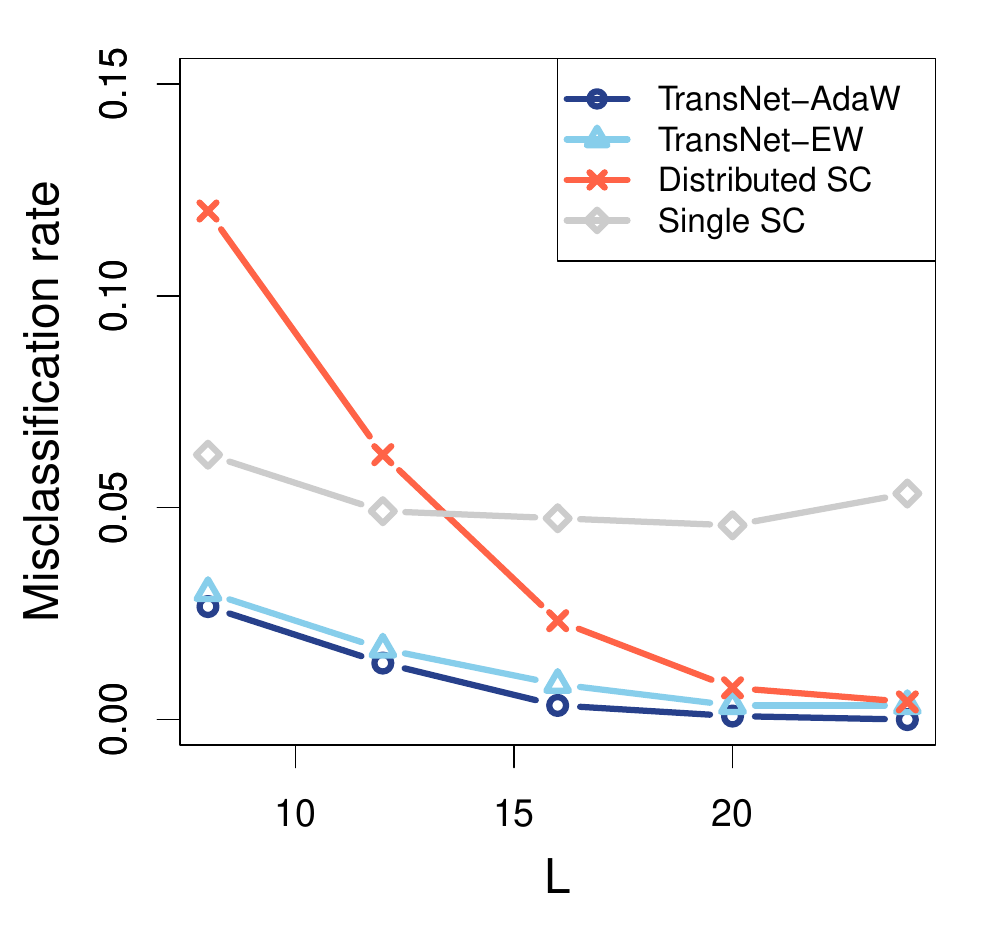}}
\subfigure[Case II]{\includegraphics[height=4.2cm,width=4.2cm,angle=0]{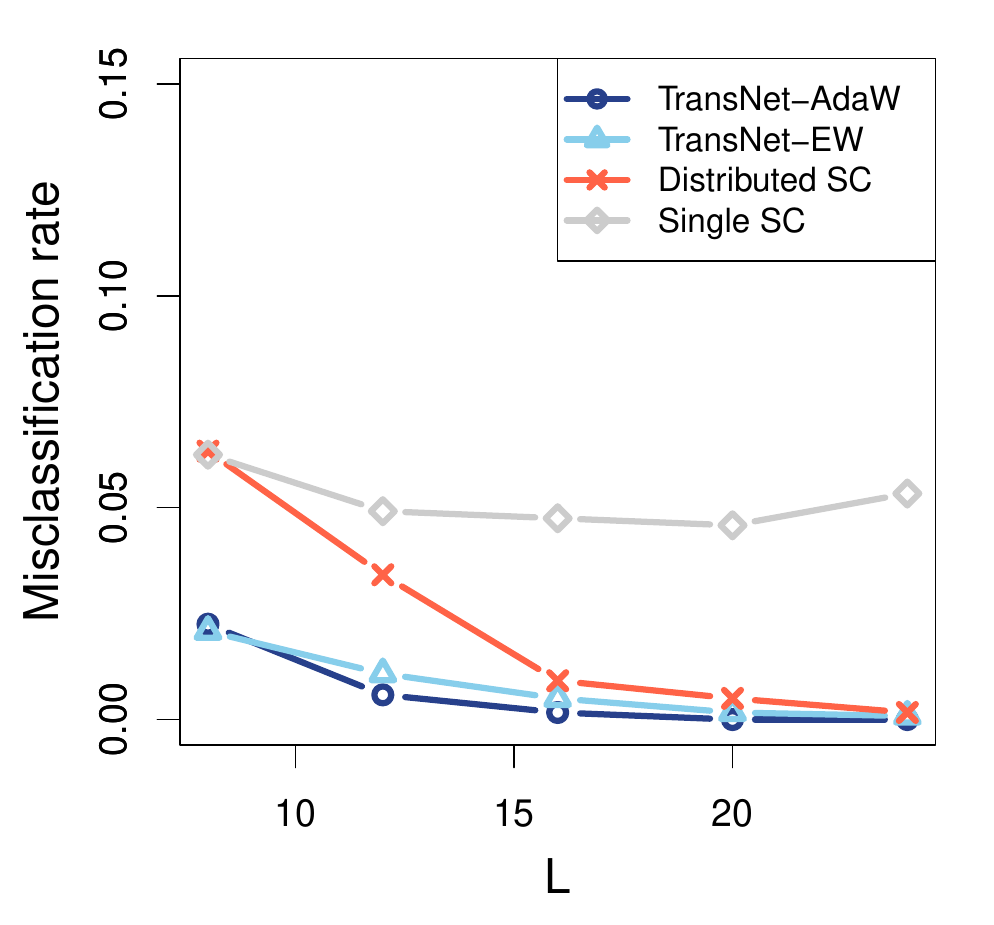}}
\caption{The projection distance (the first row) and misclassification rate (the second row) of each method under Cases I and II of Experiment I (Private but non-heterogeneous). }\label{sim_pri}
\end{figure*}

\noindent\textbf{Experiment II: Heterogeneous but non-private.}
We consider the setup where the community assignments of the source networks are different from those of the target network, and the source networks and target network are public, which means that we use the original network adjacency matrices in all methods. Specifically, we test three cases of $\mu$ for the source networks; see the second column of Table \ref{table:experi}. The average results over 10 replications are displayed in Figure \ref{sim_het} (Cases I and II) and Figure \ref{sim_het_case3} in Section \ref{add:num} of Supplementary Materials (Case III).

We have the following observations. First, \texttt{TransNet-AdaW} outperforms \texttt{TransNet-EW}, especially when the source networks differ in their levels of heterogeneity. Second, both \texttt{TransNet-EW} and \texttt{TransNet-AdaW} are superior to the distributed method \texttt{Distributed SC}, as the latter fails to effectively capture information from the target network when the source networks are heterogeneous.  
Second, when the heterogeneity is high (Case III), the performance of the transfer learning-based methods is close to the single spectral clustering based on only the target network, which is consistent with the theory.

\begin{figure*}[!h]{}
\centering
\subfigure[Case I]{\includegraphics[height=4.2cm,width=4.2cm,angle=0]{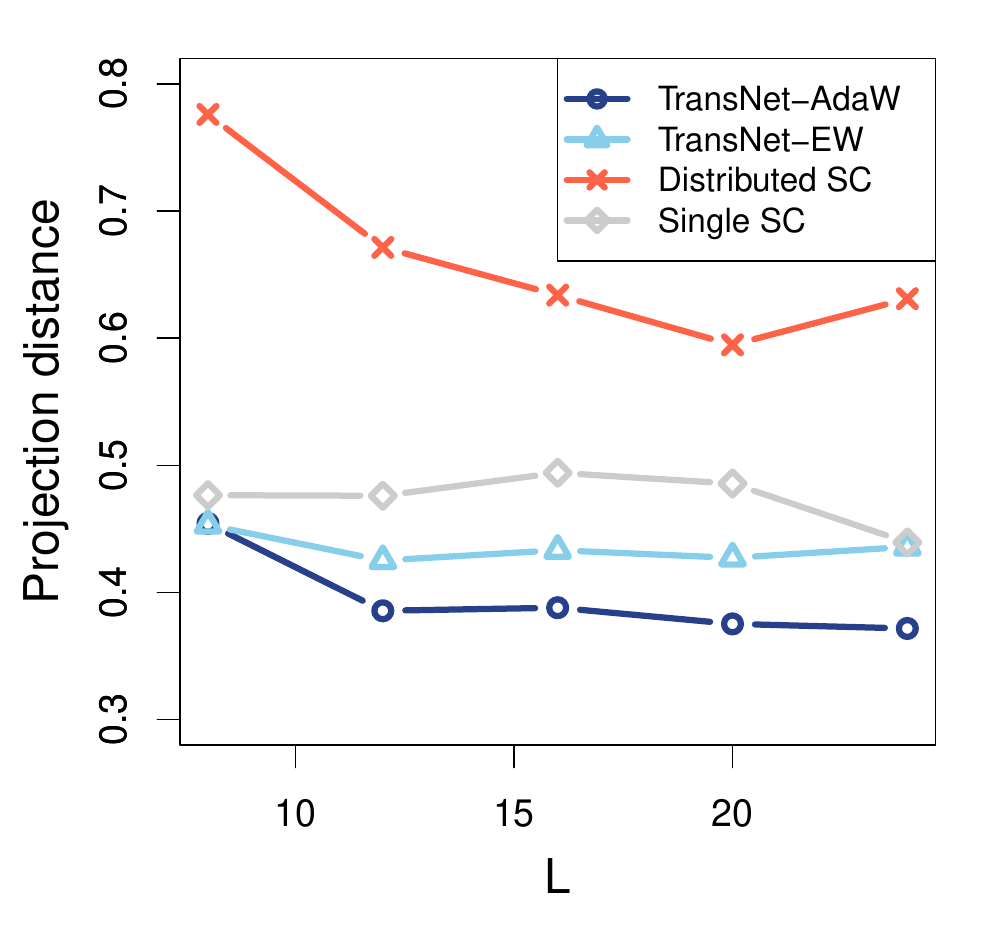}}
\subfigure[Case II]{\includegraphics[height=4.2cm,width=4.2cm,angle=0]{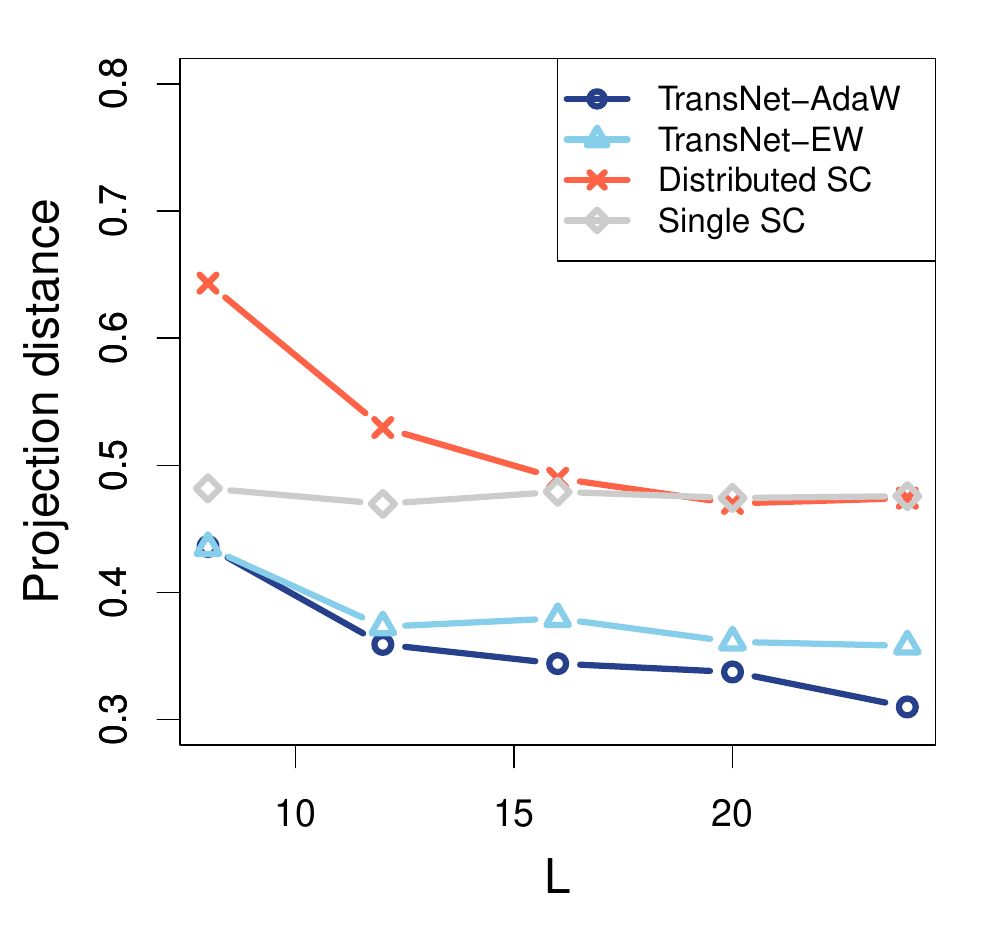}}\newpage
\subfigure[Case I]{\includegraphics[height=4.2cm,width=4.2cm,angle=0]{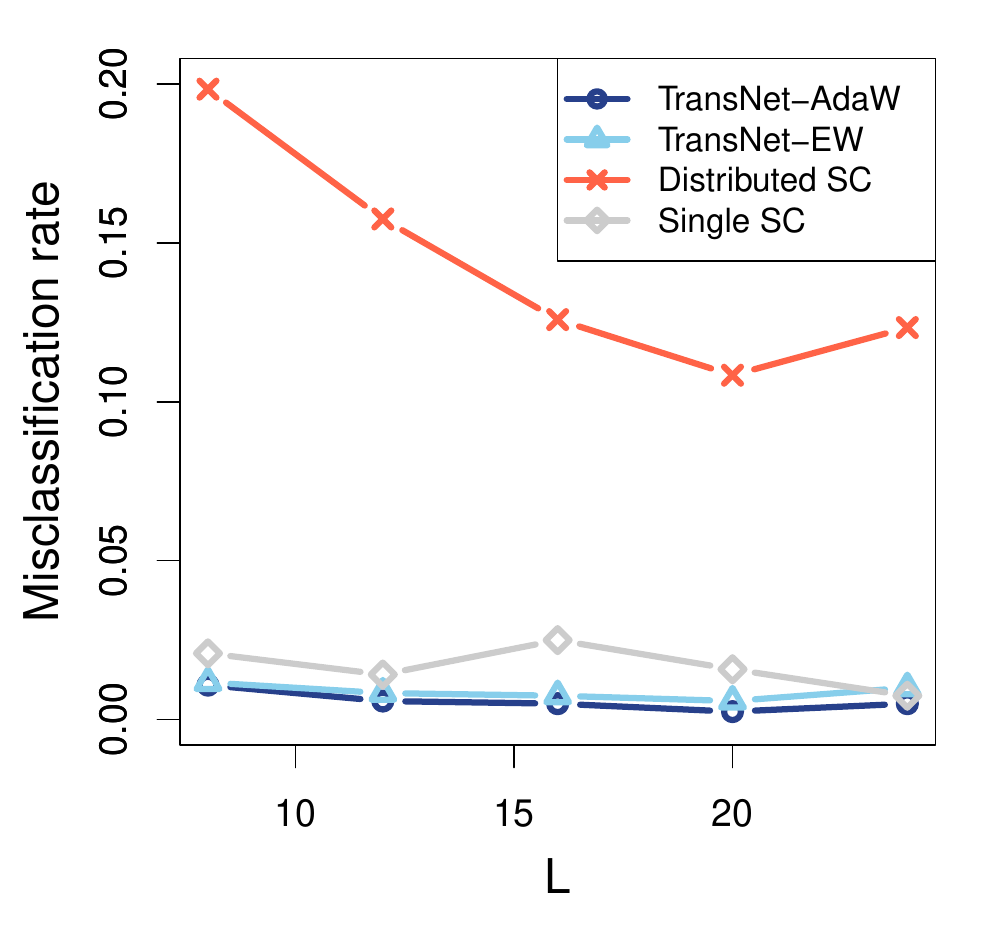}}
\subfigure[Case II]{\includegraphics[height=4.2cm,width=4.2cm,angle=0]{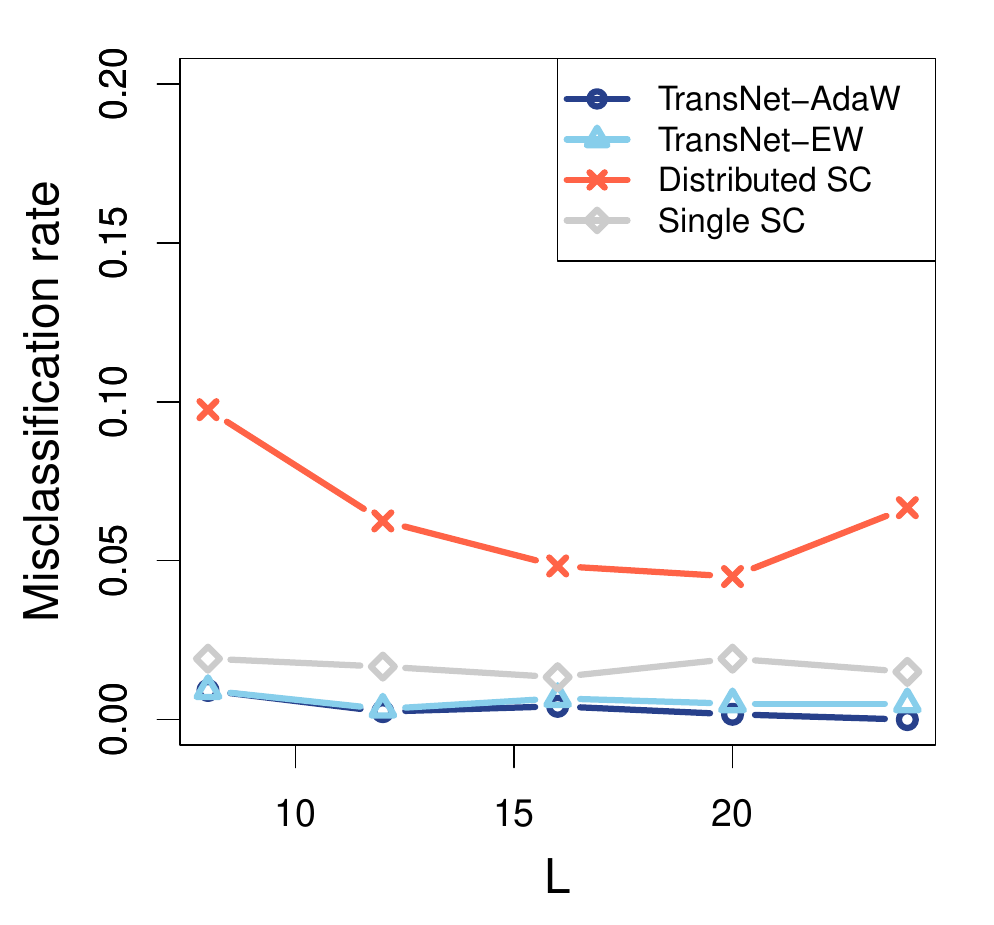}}
\caption{The projection distance (the first row) and misclassification rate (the second row) of each method under Cases I and II of Experiment II (Heterogeneous but non-private). }\label{sim_het}
\end{figure*}

\vspace{0.2cm}
\noindent\textbf{Experiment III: Heterogeneous and private}
We consider the setup where the community assignments of the source networks are different from those of the target network, and the source networks need to be privacy-preserved. Specifically, we test three cases of the couple $(\mu,q)$ for the source networks; see the third column of Table \ref{table:experi}. The average results over 10 replications are displayed in Figure \ref{sim_het_pri} (Cases I and II) and Figure \ref{sim_het_pri_case3} in Supplementary Materials (Case III).

We have the following observations. The transfer learning-based methods are generally better than the distributed method and the single spectral clustering. In particular, when the privacy-preserving level and the heterogeneity level of source networks are positively correlated (Cases I and II), the adaptive weighting strategy is better than the equal weighting strategy. By contrast, when the privacy-preserving level and the heterogeneity level of source networks are negatively correlated (Case III; see Figure \ref{sim_het_pri_case3} in the Supplementary Materials), the performance of the transfer learning-based methods is \textcolor{black}{not much better than} that of the single spectral clustering method. At this time, the distributed method performs poorly because of the low quality of the source data.

{\color{black}
In Section \ref{add:num} of the Supplementary Materials, we include additional simulations (Experiments IV-VI) to assess the sensitivity of the proposed method. Experiments IV and V study the performance of the proposed method as the number of informative networks varies, while Experiment VI examines varying privacy parameters, including a stringent regime where the $ q_l$ values for non-informative networks are close to 0.5. The numerical results show that the proposed method has satisfactory performance even under tight privacy budgets. We refer to Section \ref{add:num} for detailed results and discussion.
\iffalse
In the Supplementary Materials, we conduct additional simulations to test the sensitivity of the proposed method via Experiments IV-VI. In particular, Experiments IV and V evaluate the performance of the proposed method as the number of informative networks varies, where either the parameter configurations of the informative networks vary or those of the non-informative networks vary. Experiment VI  evaluates the performance of the proposed method as the privacy parameters vary, including a stringent privacy regime in which $q_l$'s corresponding to non-informative networks are close to 0.5.
\fi
}

\begin{figure*}[!h]{}
\centering
\subfigure[Case I]{\includegraphics[height=4.2cm,width=4.2cm,angle=0]{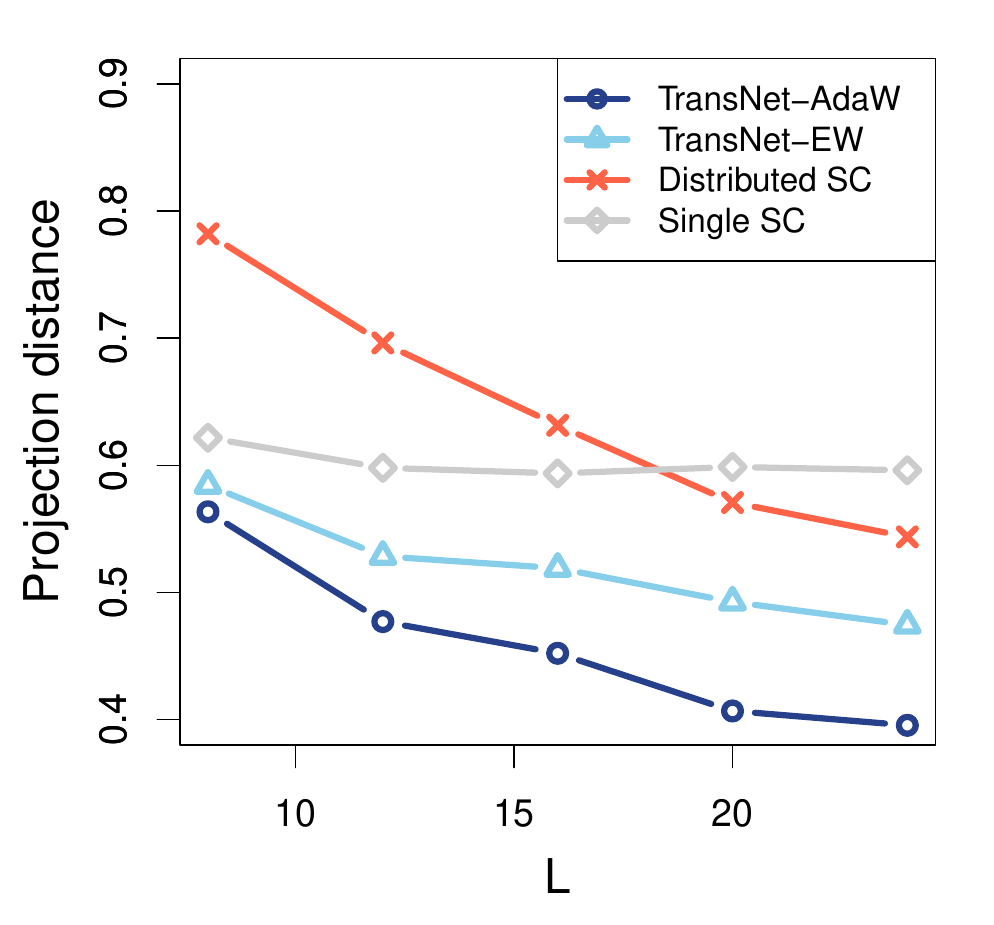}}
\subfigure[Case II]{\includegraphics[height=4.2cm,width=4.2cm,angle=0]{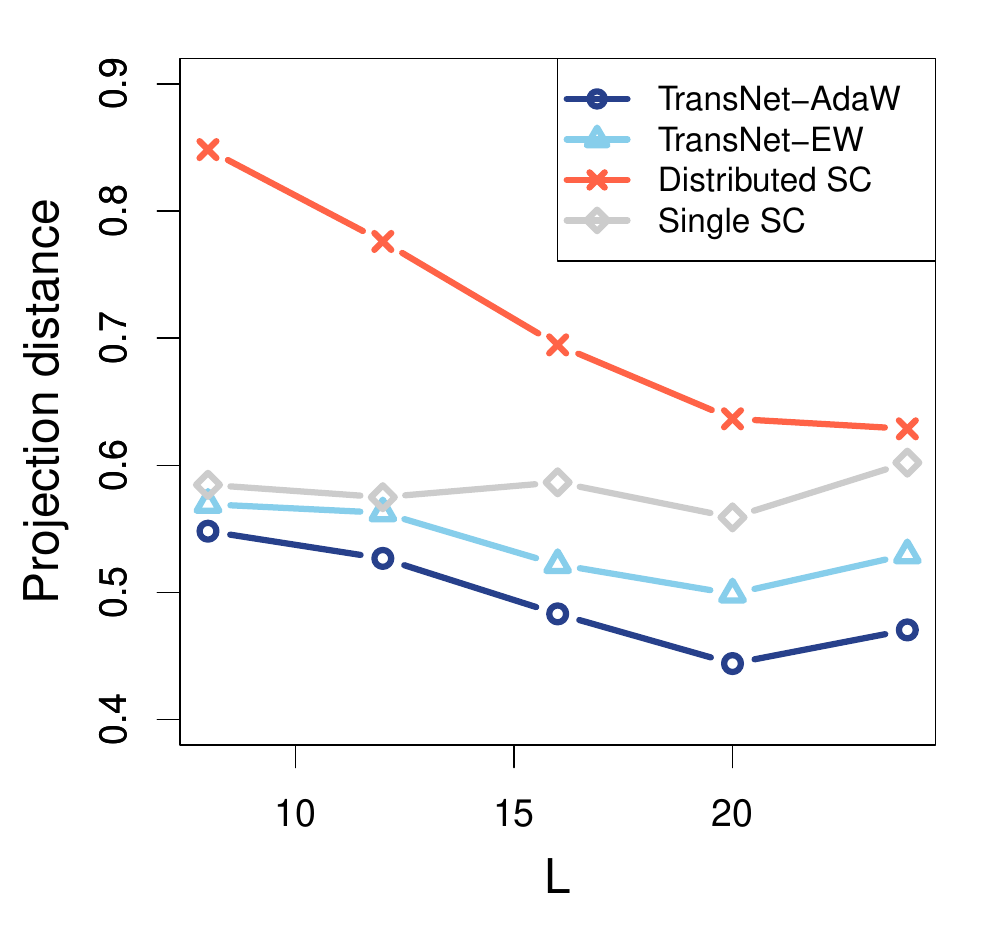}}\newpage
\subfigure[Case I]{\includegraphics[height=4.2cm,width=4.2cm,angle=0]{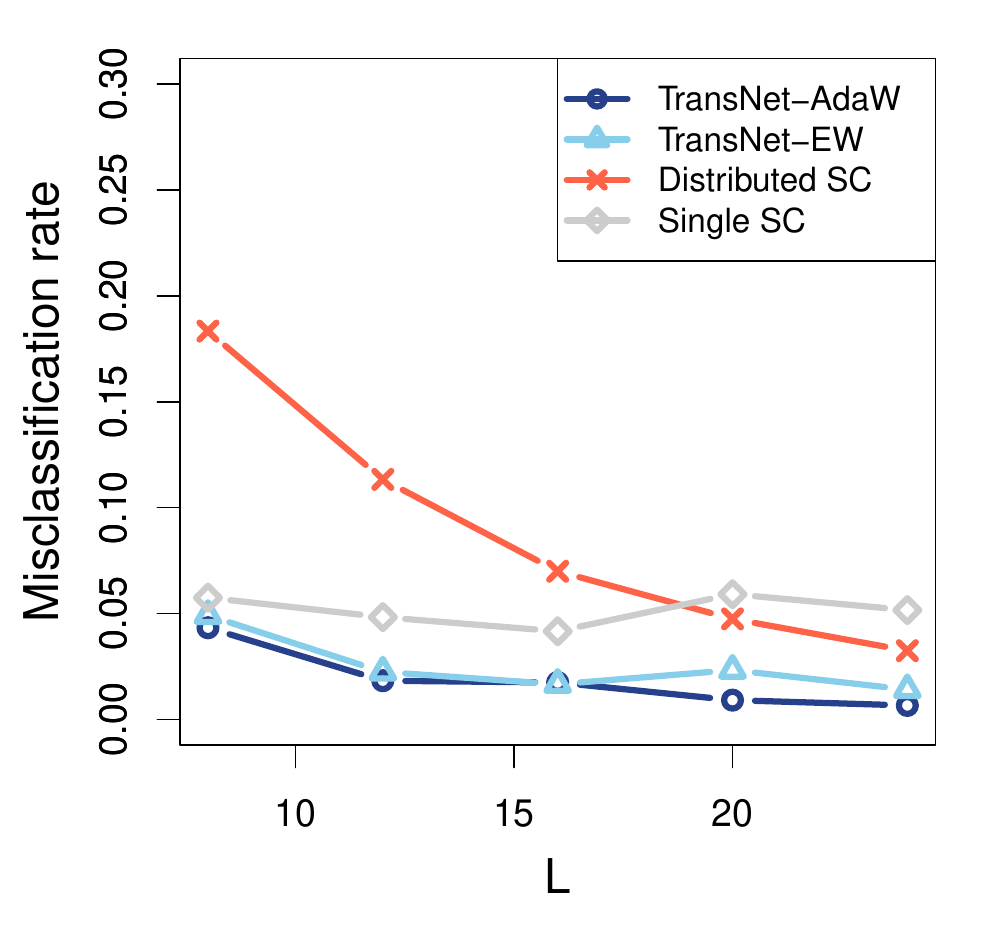}}
\subfigure[Case II]{\includegraphics[height=4.2cm,width=4.2cm,angle=0]{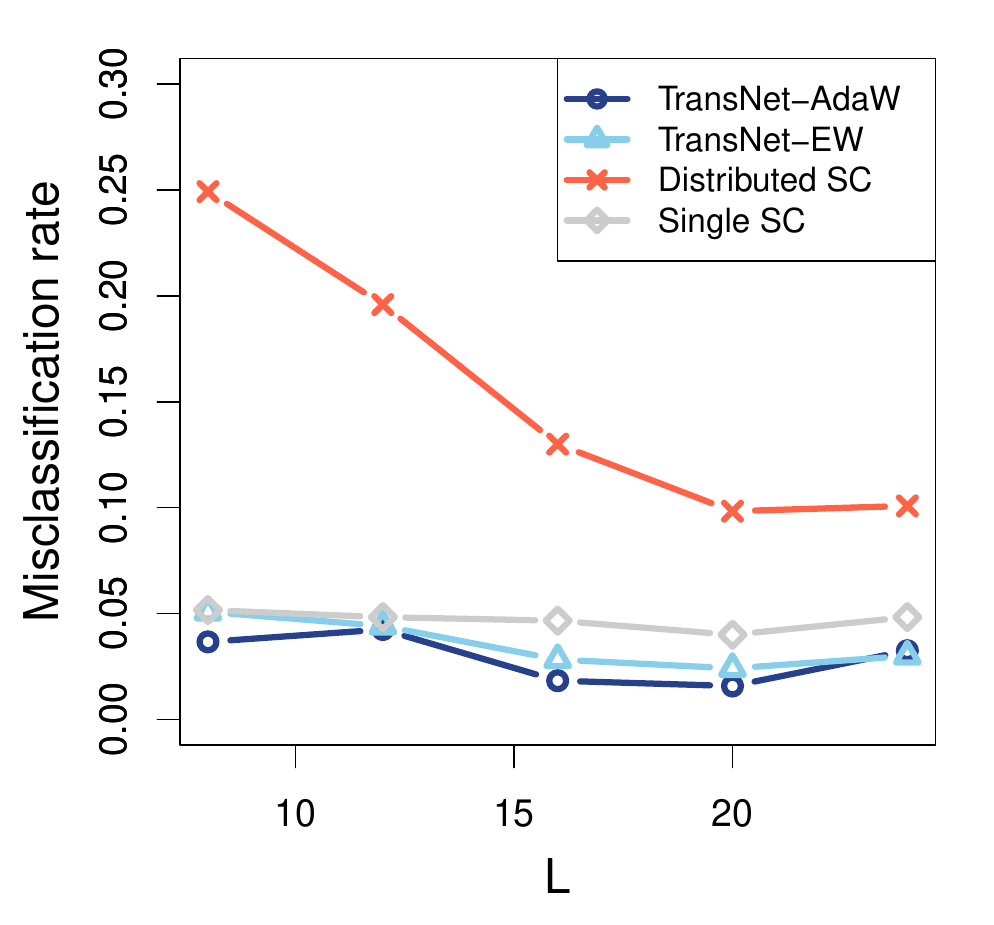}}
\caption{The projection distance (the first row) and misclassification rate (the second row) of each method under Cases I and II of Experiment III (Heterogeneous and private). }\label{sim_het_pri}
\end{figure*}

\section{Real data analysis}
\label{sec:real}
\textcolor{black}{In this section, we analyze two real-world datasets, the \texttt{AUCS} dataset \citep{rossi2015towards} and the \texttt{Politics} dataset \citep{greene2013producing}. Both datasets consist of multiple networks and are publicly known. We mimic the privacy-aware transfer learning set-up to demonstrate the efficacy of the proposed algorithm. Due to space limitations, the description and analysis of the \texttt{Politics} dataset are provided in Section~\ref{sec:politics} of the Supplementary Materials.}

The \texttt{AUCS} dataset includes five types of relationships among the 61 employees at the Department of Computer Science of Aarhus University, including current working relationships (Work), friendship on Facebook (Facebook), repeated leisure activities (Leisure), regularly eating lunch together (Lunch), and co-authorship of a publication (Coauthor).
The eight research groups of 55 out of 61 employees are available. A visualization of the five networks can be found in Section \ref{sec:politics} of the Supplementary Materials.

% The nodes in the same research group are ordered next to each other. 
% It is obvious that the community structures of each network are heterogeneous. For example, for the lunch network, the nodes within the same research group are densely connected compared with those in distinct research groups. For the coauthor network, all the nodes are loosely connected, and in some research groups (say the third one), there are no connections among nodes.  

% \begin{figure*}[!h]{}
% \centering
% \subfigure[Work]{\includegraphics[height=4cm,width=4cm,angle=0]{work.pdf}}
% \subfigure[Facebook]{\includegraphics[height=4cm,width=4cm,angle=0]{facebook.pdf}}
% \subfigure[Leisure]{\includegraphics[height=4cm,width=4cm,angle=0]{leisure.pdf}}
% \subfigure[Lunch]{\includegraphics[height=4cm,width=4cm,angle=0]{lunch.pdf}}
% \subfigure[Coauthor]{\includegraphics[height=4cm,width=4cm,angle=0]{coauthor.pdf}}
% \caption{A visualization of the \texttt{AUCS} network data. Each sub-figure corresponds to one of the five relationships. The nodes 
% are ordered according to underlying research groups.}\label{AUCS_vis}
% \end{figure*}

We regard one of these five networks as the target network and the other four as source networks. Each network is alternately used as the target network. The research groups are regarded as the true clustering of the target network. To mimic the privacy-preserving set-up, we perturb each network using the RR. We let the privacy parameters $q,q'(q=q')$ of the target network vary, and the privacy parameters of source networks are kept fixed. We test the following privacy parameter settings: 0.95 (Work), 0.8 (Facebook), 0.8 (Leisure), 0.95 (Lunch), 0.8 (Coauthor). Other set-ups can be similarly tested.  

\begin{figure*}[!h]{}
\centering
\subfigure[Work]{\includegraphics[height=3.9cm,width=4.1cm,angle=0]{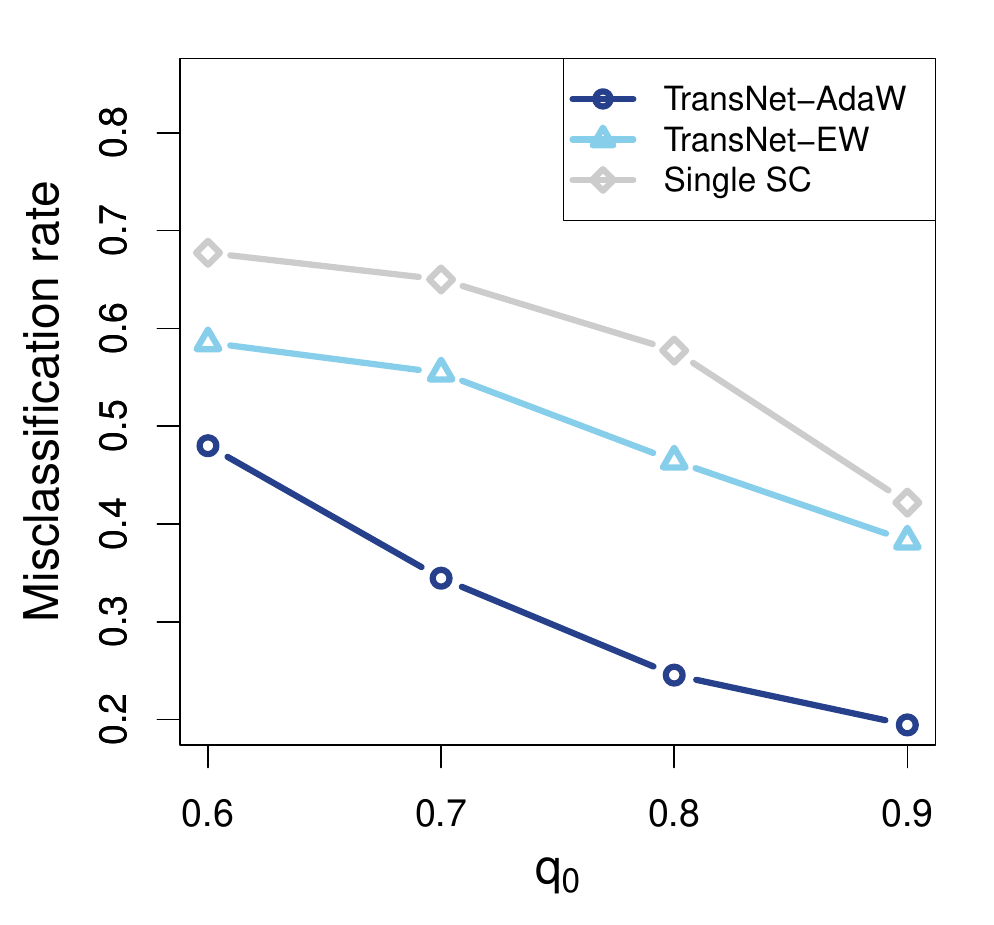}}
\subfigure[Facebook]{\includegraphics[height=3.9cm,width=4.1cm,angle=0]{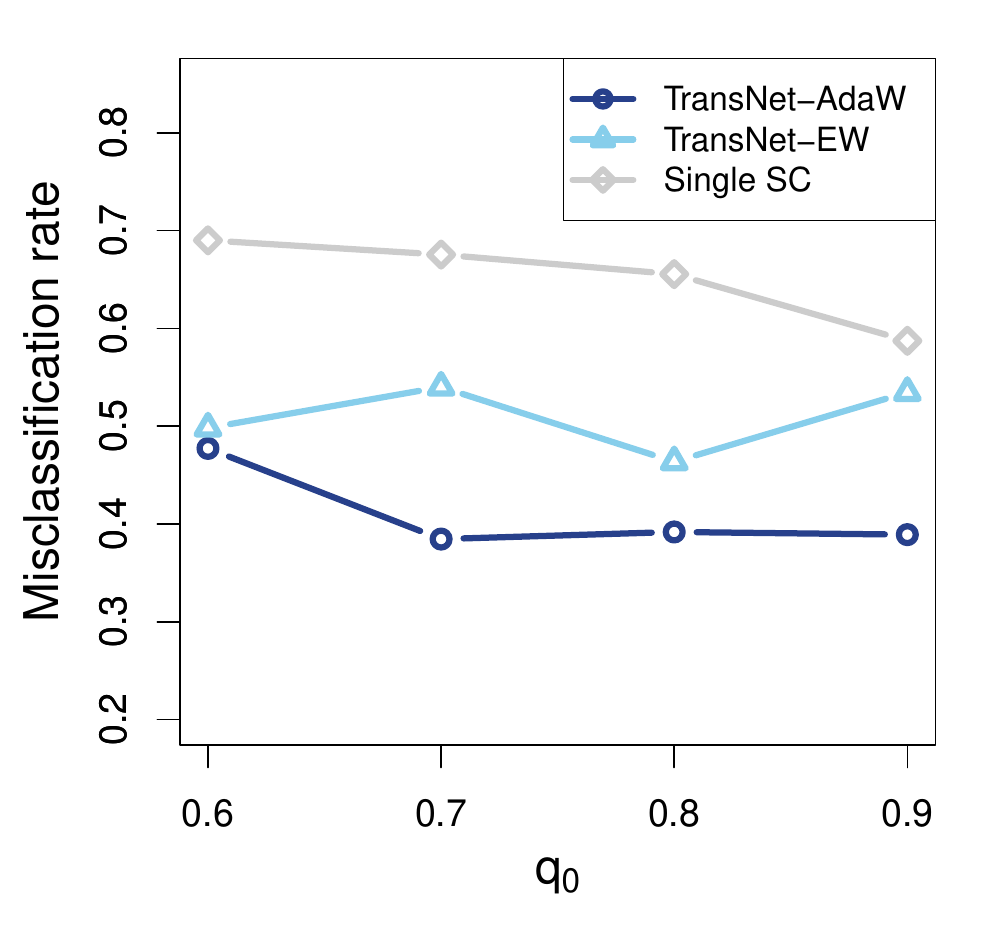}}
\subfigure[Leisure]{\includegraphics[height=3.9cm,width=4.1cm,angle=0]{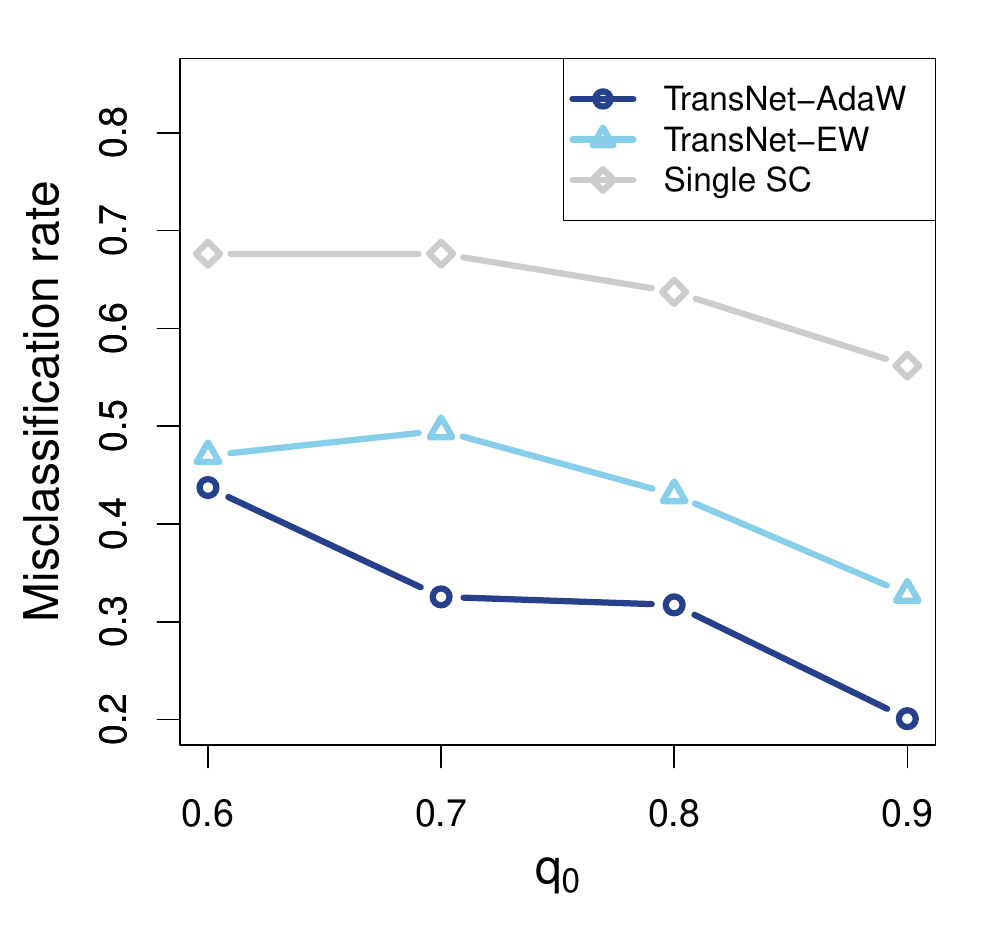}}
\subfigure[Lunch]{\includegraphics[height=3.9cm,width=4.1cm,angle=0]{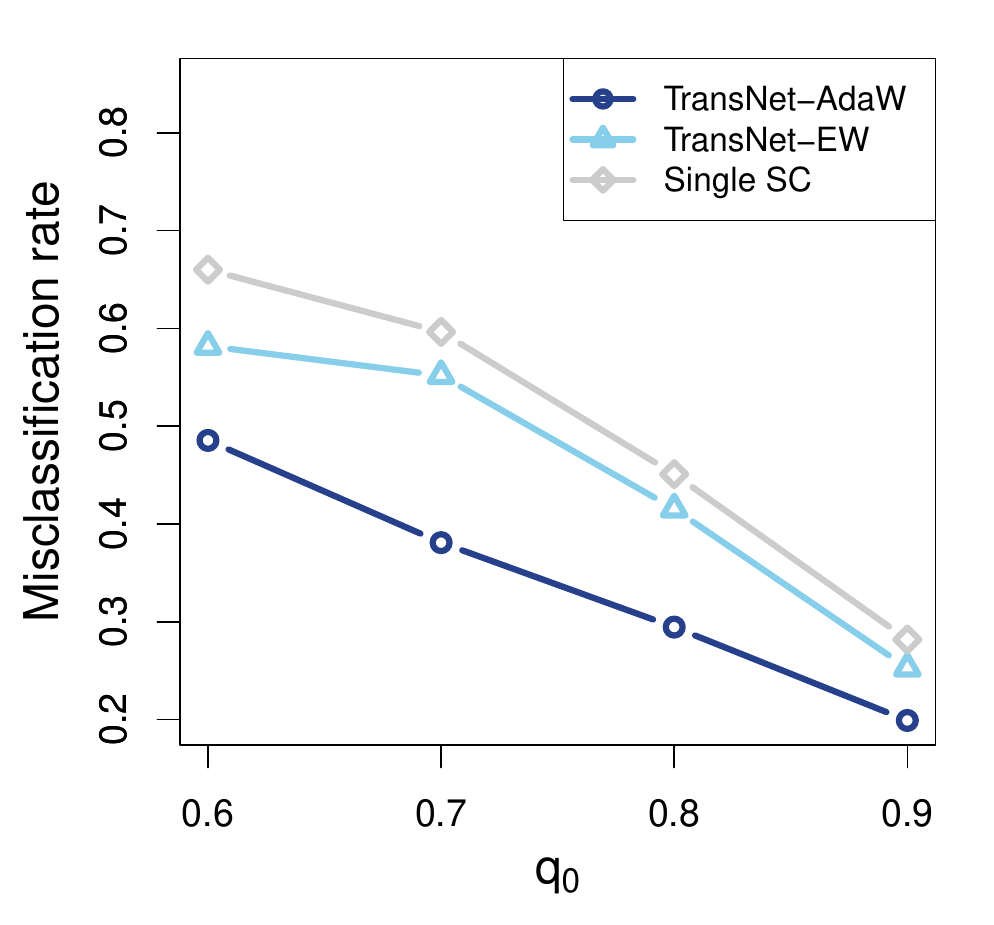}}
\subfigure[Coauthor]{\includegraphics[height=3.9cm,width=4.1cm,angle=0]{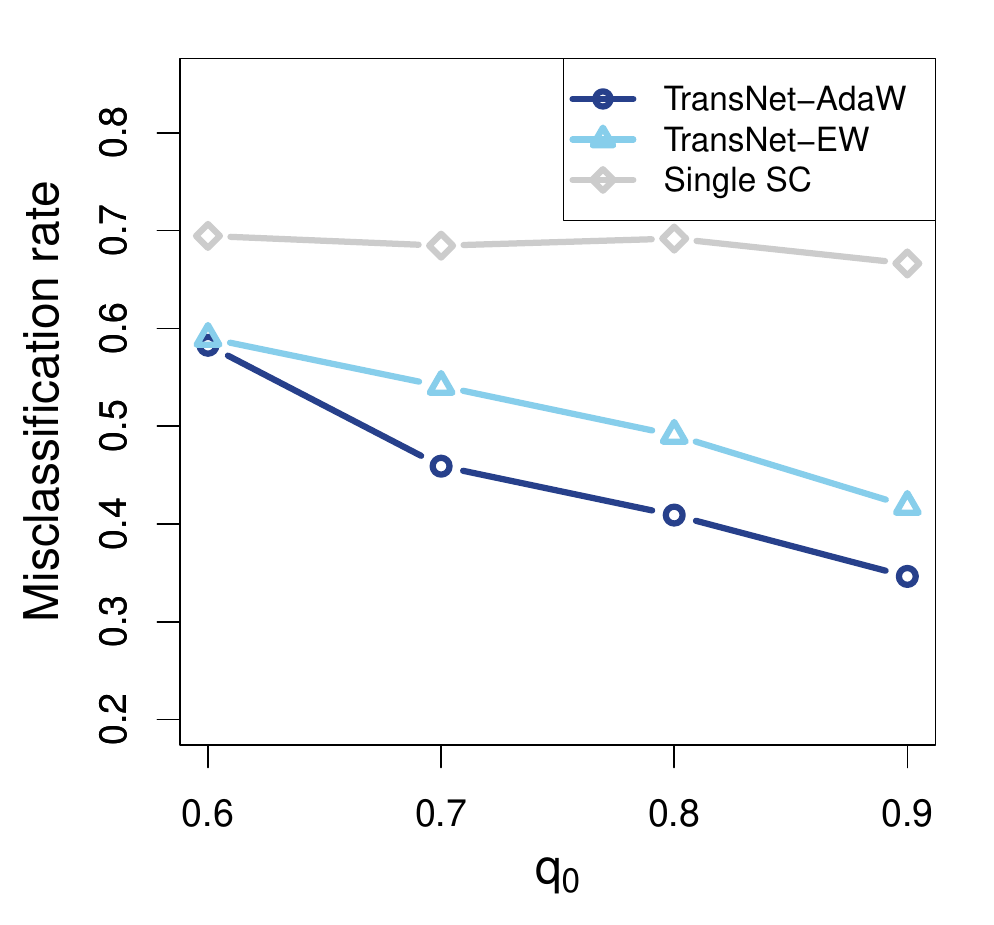}}
\caption{ The misclassification rate of each method for the \texttt{AUCS} network. (a)-(e) correspond to the scenarios where Work, Facebook, Leisure, Lunch, and Coauthor are considered as the target network, with the remaining networks serving as source networks.}\label{AUCS_results}
\end{figure*}

We focus on evaluating how the proposed method \texttt{TransNet-AdaW} can help improve the accuracy of the single spectral clustering on the target network (\texttt{Single SC}).For comparison, we also include the equal-weight counterpart, \texttt{TransNet-EW}. The details of three methods can be found in Section \ref{sec:sim}. Figure \ref{AUCS_results} provides the averaged misclassification rates of three methods over 20 replications for five scenarios, where Work, Facebook, Leisure, Lunch, or Coauthor are considered as the target network, respectively, with the remaining networks serving as source networks. We have the following observations. 
First, the proposed transfer learning methods \texttt{TransNet-AdaW} and \texttt{TransNet-EW} improve the accuracy of \texttt{Single SC} under all five of the considered scenarios,    
indicating the benefit of combining source networks with the target network. The transfer learning methods show superiority, particularly when the Leisure and Coauthor networks are served as the target network, respectively, because these two networks are relatively sparse, containing limited community structures on their own; see Figure \ref{AUCS_vis}. Second, the \texttt{TransNet-AdaW} performs better than  \texttt{TransNet-EW}, showing the effectiveness of the weighting strategy. 
Third, the performance of the three methods \textcolor{black}{improves as the privacy parameter $q_0$ increases}, except when the Facebook network serves as the target network. In the latter case, the  \emph{original} Facebook network is \textcolor{black}{not so} informative (see Figure \ref{AUCS_vis}), let alone the RR-perturbed networks.

%\vspace{-1cm}

\section{Conclusion}
\label{sec:concl}
{\color{black}

This paper introduces a novel spectral clustering-based method, \texttt{TransNet}, designed to learn the community memberships of a target network using multiple source networks that are locally stored, private, and heterogeneous. Edges in each network are perturbed via the randomized response (RR) mechanism, with the key consideration that different source networks may adhere to distinct privacy levels and exhibit varying degrees of heterogeneity. To effectively utilize information from the source networks, we propose an adaptive weighting scheme that aggregates their eigenspaces with data-driven weights, considering both privacy and heterogeneity. Additionally, we introduce a regularization step that merges the weighted average eigenspace of the source networks with the eigenspace of the target network to achieve an optimal balance.
Theoretically, we establish an error-bound-oracle property, demonstrating that the estimation error bound for the aggregated eigenspace depends solely on information from informative source networks. Furthermore, \texttt{TransNet} achieves an error bound that is never larger than that of estimators based exclusively on the target network or the source networks. Simulation studies and real data analyses support the effectiveness of \texttt{TransNet}. Note that
\texttt{TransNet} is specifically designed for environments where local machines are not necessarily trusted for data storage and handling. For a comprehensive approach, we extend both the methodology and theory to settings where local machines are trusted, introducing \texttt{TransNetX}. Additionally, the methodology may be extended to directed networks or weighted networks with mixed memberships, offering promising directions for future research.}

% \textcolor{black}{Finally, we note that there are two typical regimes for DP. The first is {local DP} \citep{duchi2018minimax}, which we considered in this work. In local DP, the data provider or users do not have to trust the data curator and thus the noise is added to the input data points before sharing it to the data curator or other external parties. 
% The second is \emph{central DP} \citep{dwork2006calibrating}, where the data provider or users trust the data curator enough to share data with it. The data curator adds noise to the output of a query of the data. The data curator protects the individuals' privacy from people who are querying the data. The local DP has advantage in that the data provider or user does not have to trust the data curator. The noise is added directly to the raw data, which facilitates downstream analyses and the perturbed data can be used repeatedly without privacy loss accumulation. However, since noise is added directly to the raw data, it may compromise the algorithmic utility (i.e., accuracy).
% On the other hand, the central DP has advantage in the algorithmic utility as the data curator can access the raw data and only perturb the summary data. However, each data provider has to trust the data curator enough to share data with it, which might be difficult. In addition, repeated queries on the raw data leads to privacy loss accumulation. It is of independent interest to generalize the content to the central DP regime and explore other mechanism tailored for central DP, such as the Gaussian mechanism \citep{dwork2014algorithmic}. }

\section{Acknowledgments}

The authors thank the Editor, the Associate Editor, and anonymous reviewers for their constructive comments that have helped us improve the article substantially.

\bigskip
 \bigskip
 \bigskip
\begin{center}
{\large\bf SUPPLEMENTARY MATERIALS}
\end{center}
 \medskip

\renewcommand*{\thetheorem}{S\arabic{theorem}}
\renewcommand*{\thelemma}{S\arabic{lemma}}
\renewcommand*{\theproposition}{S\arabic{proposition}}
\renewcommand*{\thedefinition}{S\arabic{definition}}
\renewcommand*{\theequation}{S\arabic{equation}}
\renewcommand*{\thefigure}{S\arabic{figure}}
\renewcommand*{\thealgorithm}{S\arabic{algorithm}}
\renewcommand*{\theremark}{S\arabic{remark}}

\renewcommand*{\thesubsection}{\Alph{subsection}}

Section \ref{app:tech} includes the proof of Proposition \ref{theo: step1} and lemmas that are useful in the proof of main theorems. 
Section \ref{app:main proof} provides the proofs for Theorems \ref{theo: effectofadaptive}-\ref{theo: step3}. 
Section \ref{app:add} presents the additional results on the effect of the bias-adjustment.
Section \ref{app:auxiliary} includes the auxiliary lemmas. Section \ref{app:opt} discusses the optimality of \texttt{TransNet}. Section \ref{add:num} presents the details of the tuning parameter selection and the additional numerical results.
Section \ref{app:ext} provides the extension to \texttt{TransNetX}, including the method, theory and numerical experiments. Section \ref{app:cdp} contains all the preliminaries, lemmas and proofs involved in Section \ref{app:ext}.

\subsection{Proofs of propositions and technical lemmas}
\label{app:tech}

\subsubsection*{Proof of Proposition \ref{theo: step1}}

At the beginning, it is worth mentioning that the true eigenspace $U_0$ is only identical up to orthogonal transformation. For simplicity, we assume that the sample version eigenspace $\hat{U}_0$ is already aligned with $U_0$ in the sense that $\arg \min_{Z\in\mathcal O_K} \|\hat{U}_0 Z-U_0\|_F=I_K$. Alternatively, we could work with $U_0Z$ for some $Z\in \mathcal O_K$ instead.

We will use the following results, obtained by modifying Lemmas 1-3 from \citet{charisopoulos2021communication}, to prove Proposition \ref{theo: step1}.

\begin{proposition}
\label{prop: cha}
  Suppose
  \begin{equation}
  \label{eq:conofcha}
\sum_{l=1}^L w_l\|\hat{A}_l-\bar{P}\|_2^2\lesssim  {\lambda^2_{\min}(\bar{P})}.
  \end{equation}
 Then, the $\bar{U}$ obtained by Step 1 of Algorithm \ref{alg:transnet} satisfies that
\begin{align}
\label{eq:decompo}
{\rm dist}(\bar{U}, U_0)&\lesssim{\sum_{l=1}^L\frac{w_l\|\hat{A}_l-\bar{P}\|_2^2}{\lambda^2_{  \rm min}(\bar{P})} }+{\frac{\|\sum_{l=1}^L w_l\hat{A}_l-\bar{P}\|_2}{\lambda_{  \rm min}(\bar{P})}}+{\frac{\|\hat{A}_0-\bar{P}\|_2^2}{\lambda^2_{  \rm min}(\bar{P})} }.
\end{align}
\end{proposition}

To facilitate the analysis, we further decompose \eqref{eq:decompo} as follows, 
{\small
\begin{align}
\label{eq: decomposition}
{\rm dist}(\bar{U}, U_0)&
 \lesssim{\sum_l\frac{w_l\|\hat{A}_l-P_l\|_2^2}{\lambda^2_{  \rm min}(\bar{P})} } +{\sum_l\frac{w_l\|P_l-\bar{P}\|_2^2}{\lambda^2_{  \rm min}(\bar{P})}} \nonumber\\
&\quad\quad\quad +{\frac{\|\sum_l (w_l\hat{A}_l-w_lP_l)\|_2}{\lambda_{  \rm min}(\bar{P})}}+{\frac{\|\sum_l w_lP_l-\bar{P}\|_2}{\lambda_{  \rm min}(\bar{P})}}+\frac{\|\hat{A}_0-\bar{P}\|_2^2}{\lambda^2_{  \rm min}(\bar{P})} \nonumber\\
&:=\mathcal T_1+\mathcal T_2+\mathcal T_3+\mathcal T_4+\mathcal T_5.
\end{align}
}
In the following proof, we bound $\mathcal T_1$, $\mathcal T_2$, $\mathcal T_3$, $\mathcal T_4$ and $\mathcal T_5$, respectively. 

For $\mathcal T_1$, we have with probability larger than $1-\frac{1}{n^\kappa}$ for some constant $\kappa>0$ that 
\begin{align*}
\|\hat{A}_l-P_l\|_2^2&= \frac{1}{(q_l+q'_l-1)^2}\cdot \left\|
\tA_l-\left[(q_l+q'_l-1)P_l+(1-q'_l)\mathbf 1\mathbf 1^\intercal-{\rm diag}(1-q'_l)\right]\right\|_2^2\\
&\lesssim {n((q_l+q'_l-1)\rho+1-q'_l)/(q_l+q'_l-1)^2},
\end{align*}
where the first equality follows from (\ref{eq:db}) and the second inequality follows from the fact that $(q_l+q'_l-1)P_l+(1-q'_l)\mathbf 1\mathbf 1^\intercal-{\rm diag}(1-q'_l)$ is the population matrix of the binary matrix $\tA_l$, the concentration inequality of 
adjacency matrix (see \citet{lei2015consistency} for example), and Assumption \ref{assu:sparse} and the condition of $\rho$ in Assumption \ref{assu:conrhoL}. Taking use of Assumption \ref{assu:rank} and the union bound, we further have
with probability larger than $1-\frac{L}{n^\kappa}$ for some constant $\kappa>0$ that
\begin{equation}
\label{eq:t1}
\mathcal T_1 \lesssim \sum_l\frac{ w_l n((q_l+q'_l-1)\rho+1-q'_l)}{n^2\rho^2(q_l+q'_l-1)^2}\lesssim \sum_l \frac{w_l}{n\rho\nu_l} + \sum_l\frac{ w_l (1-q'_l)}{n\rho^2\nu_l^2}.
\end{equation}

For $\mathcal T_2$, we first note that
{\small
\begin{align}
\label{eq:diff}
P_l-\bar{P}&=\Theta_l B_l\Theta_l^\intercal -\Theta_0 \sum_l (w_l B_l) \Theta_0^\intercal \nonumber\\
&=\left(\Theta_l B_l\Theta_l^\intercal-\Theta_l B_l\Theta_0^\intercal\right)+\left(\Theta_l B_l\Theta_0^\intercal-\Theta_0 B_l\Theta_0^\intercal\right)\nonumber\\
&\quad\quad\quad\quad+\left(\Theta_0 B_l\Theta_0^\intercal-\Theta_0 B_0\Theta_0^\intercal\right)+\Big(\Theta_0 B_0\Theta_0^\intercal-\Theta_0 \sum_l(w_lB_l)\Theta_0^\intercal\Big).
\end{align}
}
Noting \eqref{eq:heterP}, the following results hold,
{\small
\begin{align*}
&\|P_l-\bar{P}\|_2\lesssim n\rho \mathcal E_{\theta,l}+n\rho \mathcal E_{b,l}+n\rho \sum_l (w_l \mathcal E_{b,l}),\nonumber\\
&\|P_l-\bar{P}\|_2^2\lesssim (n\rho)^2\Big(\mathcal E_{\theta,l}^2+\mathcal E_{b,l}^2+(\sum_lw_l \mathcal E_{b,l})^2+\mathcal E_{\theta,l}\mathcal E_{b,l}+\mathcal E_{\theta,l}\sum_l(w_l\mathcal E_{b,l})+\mathcal E_{b,l}\sum_l(w_l\mathcal E_{b,l})\Big).
\end{align*} 
}
We then have
{\small
\begin{align}
\label{eq:heterop}
\sum_lw_l\|P_l-\bar{P}\|_2^2&\lesssim (n\rho)^2\Big(\sum_lw_l\mathcal E_{\theta,l}^2+{\sum_lw_l\mathcal E_{b,l}^2}\nonumber\\
&\quad\quad\quad\quad\quad+{(\sum_lw_l \mathcal E_{b,l})^2}+\sum_lw_l \mathcal E_{\theta,l}\mathcal E_{b,l}+(\sum_lw_l\mathcal E_{b,l})(\sum_lw_l\mathcal E_{\theta,l})\Big).
\end{align}
}All the terms in the RHS of \eqref{eq:heterop} can be bounded by $(n\rho)^2\sum_l w_l\mathcal E_{\theta,l}$ by noting that $\mathcal E_{b,l}$'s and $\mathcal E_{\theta,l}$'s are constant, 
and $(\sum_l w_l \mathcal E_{b,l})^2\lesssim \sum_lw_l \mathcal E_{b,l}^2\lesssim \sum_l w_l \mathcal E_{\theta,l}$ implied by our condition.
Using Assumption \ref{assu:rank}, we then have
\begin{equation}
\label{eq:t2}
\mathcal T_2 \lesssim \sum_l w_l\mathcal E_{\theta,l}.
\end{equation}

For $\mathcal T_3$, we will use the matrix Bernstein equality (\citet{chung2011spectra}; see also Lemma \ref{pro:matrixbern}) bound. 
For $i<j$, let $E^{(ij)}$ be a matrix with 1 in the $(i,j)$th and $(j,i)$th position and 0 elsewhere. Let
\begin{equation*}
X^{(ijl)} =(w_l \hat{A}_{l,ij}-w_l P_{l,ij})E^{(ij)}= \frac{1}{q_l+q'_l-1}\cdot \left( w_l \tA_{l,ij}-w_l \tilde{P}_{l,ij}\right),
\end{equation*}
where $\tilde{P}_{l,ij}= (q_l+q'_l-1)P_{l,ij}+1-q'_l$. 
It is easy to see that $X^{(ijl)}$'s are $n\times n$ symmetric matrices with mean $0$ and they are mutually independent. Note that
\begin{equation*}
\|X^{(ijl)}\|_2\leq \frac{w_l}{q_l+q'_l-1}\leq \frac{1}{\nu_{\min
}}\quad {\rm for}\; {\rm any}\; i,j,l, 
\end{equation*}
and
\begin{align*}
\|\sum_l\sum_{ij}{\rm var}(X^{(ijl)})\|_2&=\|\sum_l\sum_{ij}{\mathbb E}(X^{(ijl)})^2\|_2\nonumber\\
&=\|\sum_l \sum_{ij} \frac{(w_l^2\tilde{P}_{l,ij}-w_l^2\tilde{P}_{l,ij}^2)}{(q_l+q'_l-1)^2}\cdot(E^{(ii)}+E^{(jj)})\|_2\\
&=\max_i \left(\sum_l\sum_j \frac{w_l^2 (\tilde{P}_{l,ij}-\tilde{P}_{l,ij}^2)}{(q_l+q'_l-1)^2}\right)\nonumber\\
&\lesssim n \sum_l \frac{w_l^2[(q_l+q'_l-1)\rho+1-q'_l]}{(q_l+q'_l-1)^2}:= n \sum_l \frac{w_l^2\tilde{\rho}_l}{\nu_l^2}.
\end{align*}
Then, we apply Lemma \ref{pro:matrixbern} on $X^{(ijl)}$'s. Specifically, 
taking $a= \sqrt{4\sum_l(w_l^2\tilde{\rho}_ln \log (2n^2)/\nu^2_l)}$, Lemma \ref{pro:matrixbern} implies that
\begin{align*}
\mathbb P\left(\|\sum_l w_l \hat{A}_l-w_l P_l\|_2\geq a\right)&\leq 2n\,\exp\left(-\frac{4\sum_l (w_l^2\tilde{\rho}_ln \log(2n^2)/\nu^2_l)}{2\sum_l(w_l^2\tilde{\rho}_ln/\nu^2_l)+2a/(3\nu_{\min})}\right)\\
&\leq 2n\,\exp\left(-\frac{4\sum_l (w_l^2\tilde{\rho}_ln \log(2n^2)/\nu_l^2)}{4\sum_l(w_l^2\tilde{\rho}_ln/\nu_l^2)}\right)\leq \frac{1}{n},
\end{align*}
where the second inequality follows by 
\begin{equation*}
\frac{2a}{3\nu_{\min}}:= \frac{4\sqrt{\sum_l(w_l^2\tilde{\rho}_ln \log (2n^2)/\nu^2_l)}}{3\nu_{\min}}\lesssim 2\sum_l\frac{w_l^2\tilde{\rho}_ln}{\nu_l^2},
\end{equation*}
which holds naturally if we note condition (\ref{eq:conL}) that $\sum_{l\in[L]} \frac{\nu_l^2}{\tilde{\rho}_l } \lesssim \frac{n\nu_{\min}^2}{\log n}$ and  $$\sum_l\frac{w_l^2\tilde{\rho}_l}{\nu_l^2}\gtrsim (\sum_{l\in[L]}\frac{\nu_l^2}{\tilde{\rho}_l })^{-1}.$$ Therefore, we have with probability larger than $1-\frac{1}{n}$ that
\begin{equation}
\label{eq:t3}
\mathcal T_3 \lesssim \sqrt{\sum_{l=1}^L w_l^2\cdot \frac{\log n}{n\rho\nu_l}}+\sqrt{\sum_{l=1}^L w_l^2\cdot \frac{1-q'_l}{\rho}\cdot \frac{\log n}{n\rho\nu_l^2}}.
\end{equation}

For $\mathcal T_4$, we have 
\begin{align*}
\sum_lw_lP_l-\bar{P}=\sum_l\left((\Theta_l-\Theta_0)w_lB_l\Theta_l^\intercal\right) +\Theta_0\left(w_lB_l (\Theta_l-\Theta_0)^\intercal \right)\lesssim n\rho\sum_l w_l \mathcal E_{\theta,l},
\end{align*}
where the last inequality follows from \eqref{eq:heterP} and the triangle inequality.
By Assumption \ref{assu:rank}, we obtain
\begin{equation}
\label{eq:t4}
\mathcal T_4 \lesssim \sum_l w_l \mathcal E_{\theta,l}. 
\end{equation}

For $\mathcal T_5$, we have
\begin{equation*}
\frac{\|\hat{A}_0-\bar{P}\|_2^2}{\lambda^2_{  \rm min}(\bar{P})}\lesssim \frac{\|\hat{A}_0-{P}_0\|_2^2} {\lambda^2_{  \rm min}(\bar{P})}+ \frac{\|P_0-\bar{P}\|_2^2}{\lambda^2_{  \rm min}(\bar{P})}.
\end{equation*}
For the first term, similar to the derivation of $\mathcal T_1$, it is easy to see that 
\begin{equation}
\label{eq:t5deri}
\frac{\|\hat{A}_0-{P}_0\|_2^2} {\lambda^2_{  \rm min}(\bar{P})}\lesssim \frac{n[(q_0+q'_0-1)\rho+1-q'_0]}{n^2\rho^(q_0+q'_0-1)^2}\lesssim \frac{1}{n\rho\nu_0} + \frac{1}{n\rho\nu^2_0} \cdot \frac{1-q'_0}{\rho},
\end{equation}
with probability larger than $1-{n^{-\kappa}}$ for some constant $\kappa$. For the second term, by \eqref{eq:diff} and Assumption \ref{assu:rank}, we have 
\begin{align*}
\frac{\|P_0-\bar{P}\|_2^2}{\lambda^2_{  \rm min}(\bar{P})}&\lesssim (\sum_lw_l \mathcal E_{b,l})^2\lesssim \sum_l w_l \mathcal E_{b,l}^2\lesssim \sum_l w_l\mathcal E_{\theta,l},
\end{align*}
where the last inequality follows from \eqref{eq:heterP}. Therefore, we have with probability larger than $1-\frac{1}{n^{\kappa}}$ for some constant $\kappa>0$ that
\begin{equation}
\label{eq:t5}
\mathcal T_5 \lesssim \frac{1}{n\rho\nu_0} + \frac{1}{n\rho\nu^2_0} \cdot \frac{1-q'_0}{\rho}+\sum_l w_l \mathcal E_{\theta,l}. 
\end{equation}

With the bound of $\mathcal T_1$-$\mathcal T_5$, {we now simplify the results by showing that $\mathcal T_1$ is dominated by $\mathcal T_3$ and the first two terms in $\mathcal T_5$ is dominated by $\mathcal T_3$.} To that end, we first observe that both $\mathcal T_1$ and the first two terms in $\mathcal T_5$ is upper bounded by $\frac{1}{n\rho^2 }\max_{l\in[L]\cup\{0\}}\frac{\tilde{\rho}_l}{\nu_l^2}$, where $\tilde{\rho}_l:=(q_l+q'_l-1)\rho+1-q'_l$. Therefore, it suffices to show 
\begin{equation}
\label{eq:sufficon}
\frac{1}{n\rho^2 }\max_{l\in[L]\cup\{0\}}\frac{\tilde{\rho}_l}{\nu_l^2}\lesssim \sqrt{\sum_l\frac{w_l^2\tilde{\rho}_l \log n}{n\rho^2 \nu^2_l}}.
\end{equation}
Note that by taking the optima of $w_l$'s, we have
\begin{equation*}
\sum_lw_l^2 \frac{\tilde{\rho}_l}{\nu_l^2}\geq \left(\sum_{l}\frac{\nu_l^2}{\tilde{\rho}_l}\right)^{-1}.
\end{equation*}
As a result, by condition \eqref{eq:conL} in Assumption \ref{assu:conrhoL}, \eqref{eq:sufficon} holds.

It remains to show the condition (\ref{eq:conofcha}) of Proposition \ref{prop: cha} holds, which suffices if $\mathcal T_1$ and $\mathcal T_2$ are both smaller than a constant. Note that $\mathcal T_2\lesssim 1$ naturally and $\mathcal T_1\lesssim 1$ by condition \eqref{eq:conrho1} in Assumption \ref{assu:conrhoL}. Consequently, we arrive the results of Proposition \ref{theo: step1} that
{\small
\begin{align*}
 {\rm dist}(\bar{U}, U_0)
&\lesssim \sqrt{\sum_{l=1}^L w_l^2\cdot \frac{\log n}{n\rho\nu_l}}+\sqrt{\sum_{l=1}^L w_l^2\cdot \frac{1-q'_l}{\rho}\cdot \frac{\log n}{n\rho\nu_l^2}}+\sum_{l=1}^L w_l \mathcal E_{\theta,l}.
\end{align*} 
}
\QEDA

We next provide two technical lemmas which are useful for studying the property of the adaptive weights. Lemma \ref{lem: accuracyofe} and \ref{lem: accuracyofrho} establish the accuracy of $\hat{\mathcal E}_{\theta,l}$ and $\hat{\rho}_l$ (see Definition \ref{def:ada}) in estimating their population analogues, respectively.

\begin{lemma}
\label{lem: accuracyofe}
Suppose Assumptions \ref{assu:rank}-\ref{assu:conrhoL} hold. Also suppose $\lambda_{\min} (P_l) \gtrsim n\rho$ for $l\in[L]$. Then for any $l\in[L]$, we have 
\begin{equation}
\label{eq: forme}
\mathcal E_{\theta,l} \asymp \|{U}_l{U}_l^\intercal -{U}_0{U}_0^\intercal\|_2,
\end{equation}
and with probability larger than $1-\frac{L}{n^{\kappa}}$ for some constant $\kappa>0$ that,
\begin{equation}
\label{eq:acce}
\hat{\mathcal E}_{\theta,l}\asymp \mathcal E_{\theta,l} +  
O_p\left(\sqrt{\frac{{(q_l+q'_l-1)\rho+1-q'_l}}{{n\rho^2}(q_l+q'_l-1)^2}}\right).
\end{equation}
\end{lemma}

\noindent \emph{Proof.} First, we prove \eqref{eq: forme}. 
Recall that we have defined in \eqref{eq:heterP} that $$\|\Theta_l-\Theta_0\|\asymp \mathcal E_{\theta,l}\sqrt{n}\quad l\in [L].$$ We now show that 
$$\mathcal E_{\theta,l} \asymp \|{U}_l{U}_l^\intercal -{U}_0{U}_0^\intercal\|_2. $$
On the one hand, by the Davis-Kahan theorem (see Lemma \ref{pro:DK}), we have
\begin{equation}
\label{eq:uupper}
\|{U}_l{U}_l^\intercal -{U}_0{U}_0^\intercal\|_2 \lesssim \frac{\|P'_l-P_0\|_2}{\lambda_{\min }(P_0)}\lesssim \mathcal E_{\theta,l},
\end{equation}
where $P'_l:=\Theta_l B_0 \Theta_l^\intercal$ and we used \eqref{eq:heterP} and Assumption \ref{assu:rank}, and the fact that $U_l$ is the eigenvector of $P'_l$ up to orthogonal transformation (see Lemma \ref{pro: eigendecom}). On the other hand, by the standard analysis of the misclassification rate of spectral clustering (see \citet{lei2015consistency} for example), we have
\begin{equation}
\label{eq:ulower}
\|{U}_l{U}_l^\intercal -{U}_0{U}_0^\intercal\|_2 \gtrsim \frac{\|\Theta_l-\Theta_0\|_0^{1/2}}{\sqrt{n}}\asymp\frac{\|\Theta_l-\Theta_0\|_F}{\sqrt{n}}\gtrsim \frac{\|\Theta_l-\Theta_0\|_2}{\sqrt{n}}=\mathcal E_{\theta,l},
\end{equation}
where we regarded $\Theta_l$ as the estimated membership matrix obtained by the $k$-means on the $U_l$ because $U_l$ has only $K$ distinct rows (see Lemma \ref{pro: eigendecom}), and the misclassification rate can be written as $\frac{\|\Theta_l-\Theta_0\|_0^{1/2}}{\sqrt{n}}$.
Combining \eqref{eq:uupper} and \eqref{eq:ulower}, we have shown (\ref{eq: forme}) holds.

Next, we show (\ref{eq:acce}). Note that
\begin{equation}
\hat{U}_l\hat{U}_l^\intercal -\hat{U}_0\hat{U}_0^\intercal={U}_l{U}_l^\intercal -{U}_0{U}_0^\intercal+\left(\hat{U}_l\hat{U}_l^\intercal-{U}_l{U}_l^\intercal\right)+\left({U}_0{U}_0^\intercal-\hat{U}_0\hat{U}_0^\intercal\right),
\end{equation}
where for last two terms, by the Davis-Kahan theorem, the concentration inequality of the adjacency matrix (see \citet{lei2015consistency} for example) and the condition of $\rho$ in Assumption \ref{assu:conrhoL}, we have with probability larger than $1-\frac{L}{n^{\kappa}}$ for some constant $\kappa>0$ that, for all $l\in [L]$,
\begin{equation*}
\|\hat{U}_l\hat{U}_l^\intercal-{U}_l{U}_l^\intercal\|_2 \lesssim \frac{\|\hat{A}_l-P_l\|_2}{\lambda_{\min} (P_l)} \lesssim \frac{\sqrt{n((q_l+q'_l-1)\rho+1-q'_l)}}{{n\rho}(q_l+q'_l-1)}.
\end{equation*}
Therefore, we obtain 
the results.
\QEDA

\begin{lemma}
\label{lem: accuracyofrho}
Suppose Assumptions \ref{assu:sparse} and \ref{assu:conrhoL} hold. Then we have for any $l \in [L]$, $\rho \asymp \sum_{i\neq j}\frac{P_{l,ij}}{n(n-1)}$, and with probability larger than $1-\frac{L}{n^n}$, 
\begin{equation}
\label{eq:accrho}
\hat{\rho}_l \asymp \rho+ O_p(\rho).
\end{equation}
\end{lemma}

\noindent\emph{Proof.} By Assumption \ref{assu:sparse}, $\rho \asymp \sum_{i\neq j}\frac{P_{l,ij}}{n(n-1)}$ holds naturally. To show $\hat{\rho}_l \asymp \rho+ O_p(\rho)$, we use the Hoeffding bound for the sum of sub-Gaussian variables; see Proposition 2.5 in \citet{wainwright2019high} for example. Specifically, we have that each $\frac{\hat{A}_{l,ij}}{n(n-1)}$ is bounded that
\begin{equation*}
\frac{\hat{A}_{l,ij}}{n(n-1)}\in\left[ \frac{q'_l-1}{(q_l+q_l'-1)n(n-1)}, \frac{q'_l}{(q_l+q_l'-1)n(n-1)}\right],
\end{equation*}
and thus it is sub-Gaussian with parameter $\frac{1}{2n(n-1)(q_l+q'_l-1)}$. Applying the Hoeffding bound, we have
\begin{align*}
\mathbb P(\left|\sum_{i\neq j}\frac{(\hat{A}_{l,ij}-P_{l,ij})}{n(n-1)}\right| \geq t) &\leq 2{\rm exp}\left(-\frac{t^2}{{(2n(n-1)(q_l+q'_l-1)^2)^{-1}} }\right).
\end{align*}
Taking $t\asymp \sum_{i\neq j}\frac{P_{l,ij}}{n(n-1)}\asymp \rho$ in the last inequality, we have
that $t^22n(n-1)(q_l+q'_l-1)^2\gtrsim \log n$ by noting the condition $\rho\gtrsim \log n/(n\nu_{\min})$  in Assumption \ref{assu:conrhoL}. Therefore, we have with probability larger than than $1-\frac{L}{n^{\alpha}}$ that, $\hat{\rho}_l \asymp \rho+ O_p(\rho)$ for all $l\in[L]$.

\QEDA

\subsection{Main proofs}
\label{app:main proof}

\subsubsection*{Proof of Theorem \ref{theo: effectofadaptive}}

Recall the definition of $\mathcal P_l$ in \eqref{eq:definitionofpl} and define 
$$\iota_n:= \frac{n\rho^2}{\varsigma_n},$$
then by Lemmas \ref{lem: accuracyofe} and \ref{lem: accuracyofrho}, for $l\in[L]$, we can write $\hat{w}_l$ as
\begin{equation*}
\hat{w}_l\varpropto \left({{\mathcal P}_{l}^2 }+O_p({{\mathcal P}_{l}^2 })+{\mathcal E}_{\theta,l}^2 + O_p(\frac{(\nu_l \rho+(1-\nu_l)/2)}{n\rho^2\nu^2_l})+\iota_n^{-1}+ O_p(\iota_n^{-1})\right)^{-1}.
\end{equation*}
By our condition $\eta_n \asymp \frac{\varsigma_n}{n\rho^2}$, we then have $\hat{w}_l\varpropto \iota_n$ for $l\in S$ and $\hat{w}_l\varpropto \frac{1}{(\mathcal E_{\theta,l}^2+\mathcal P_{l}^2)}$ for $l\in S^c$. In particular, we have
\begin{equation}
\label{eq:orderre}
\hat{w}_l\asymp \frac{\iota_n}{m\iota_n+\sum_{l\in S^c}1/((\mathcal E_{\theta,l}^2+\mathcal P_{l}^2)) }\asymp\frac{1}{m}\quad {\rm for}\quad l\in S,
\end{equation}
where we used the fact that 
\begin{equation*}
\sum_{l\in S^c}1/((\mathcal E_{\theta,l}^2+{\mathcal P_{l}^2}))\lesssim \frac{L-m}{\min _{l\in S^c} (\mathcal E_{\theta,l}^2+\mathcal P_{l}^2) } \lesssim  m\iota_n \asymp \frac{mn\rho^2}{\varsigma_n},
\end{equation*}
which holds by condition \eqref{eq:conditionfororacle}.

To prove \eqref{eq:orcalepro}, we need to show that the error corresponds to the non-informative networks are dominated by that corresponds to the informative networks. Define 
\begin{equation}
\label{eq:tilP}
\tilde{\mathcal P}_l:= \sqrt{\frac{(\nu_l\rho+(1-\nu_l)/2)\log n}{n\rho^2\nu_l^2L}}.
\end{equation}
By Proposition \ref{theo: step1}, we have 
\begin{equation*}
{\rm dist}(\bar{U},U_0)\lesssim \left(\sum_{l\in [L]}  {\hat{w}_l^2{\tilde{\mathcal P}}_{l}^2L}\right)^{1/2}+\sum_{l\in [L]}\hat{w}_l\mathcal E_{\theta,l}.
\end{equation*}
Hence, we need to show
\begin{equation}
\label{eq:inequa}
\left(\sum_{l\in S^c}  {\hat{w}_l^2{\tilde{\mathcal P}}_{l}^2L}\right)^{1/2}+\sum_{l\in S^c}\hat{w}_l\mathcal E_{\theta,l} \lesssim \left(\sum_{l\in S} {\hat{w}_l^2{\tilde{\mathcal P}}_{l}^2L}\right)^{1/2}+\sum_{l\in S}\hat{w}_l\mathcal E_{\theta,l}.
\end{equation}
Note that 
\begin{equation*}
\sum_{l\in S^c}\hat{w}_l\mathcal E_{\theta,l} \lesssim \left(\sum_{l\in S^c} \hat{w}_l^2 \mathcal E_{\theta,l}^2L\right)^{1/2}\quad {\rm and}\quad \tilde{\mathcal P}_{l}\leq \mathcal P_l. 
\end{equation*}
To establish \eqref{eq:inequa}, 
it is sufficient to show 
\begin{equation}
\label{eq:suff}
\sum_{l\in S^c} \hat{w}_l^2 \left({{\mathcal P}_{l}^2L}+\mathcal E_{\theta,l}^2L\right)\lesssim \sum_{l\in S} \hat{w}_l^2 \frac{\log n((1-\nu_l)/2+\nu_l\rho)}{n\rho^2\nu_l^2}\lesssim  \frac{\log n}{mn\rho^2},
\end{equation}
where we used \eqref{eq:orderre} and $\nu_l \lesssim 1$ for $l\in S$ in the last equality. 
Noting 
\begin{equation}
\label{eq:orderunre}
\hat{w}_l\asymp \frac{1/{(\mathcal E_{\theta,l}^2+\mathcal P_{l}^2)}}{m\iota_n+\sum_{l\in S^c}1/(\mathcal E_{\theta,l}^2+\mathcal P_{l}^2) }\asymp\frac{1}{m\iota_n (\mathcal E_{\theta,l}^2+\mathcal P_{l}^2)}\quad {\rm for}\quad l\in S^c,
\end{equation}
we have
\begin{equation*}
  \sum_{l\in S^c} \hat{w}_l^2 \left( \mathcal E_{\theta,l}^2L+{\mathcal P_{l}^2L}\right)\asymp\sum_{l\in S^c} \frac{L}{ m^2\iota_n^2(\mathcal E_{\theta,l}^2+\mathcal P_{l}^2) }.
\end{equation*}
Therefore, the following condition suffices for achieving \eqref{eq:suff}, 
\begin{equation*}
\frac{(L-m)L}{m^2\iota^2_n  \min_{l\in S^c}(\mathcal E_{\theta,l}^2+\mathcal P_{l}^2)} \lesssim \frac{\log n}{mn\rho^2},
\end{equation*}
which is implied by condition (\ref{eq:conditionfororacle}). As a result, we have demonstrated \eqref{eq:inequa} holds. 

Finally, plugging in $\hat{w}_l$ for $l\in S$, we obtain the oracle error bound (\ref{eq:orcalepro}).
\QEDA

\subsubsection*{Proof of Theorem \ref{theo: effectofadaptivesecond}}
Recall the definition of $\tilde{\mathcal P}_l$ in \eqref{eq:tilP}. By Proposition \ref{theo: step1}, the error bound of ${\rm dist}(\bar{U}, U_0)$ under the equal weights can be written as 
\begin{equation*}
\left(\frac{1}{L^2}\sum_{l=1}^L{ \tilde{\mathcal P}_{l}^2 L}\right)^{1/2}+\frac{1}{L}\sum_{l=1}^L \mathcal E_{\theta,l}.
\end{equation*}
The goal is to show the bound under adaptive weights are of smaller order than that under the equal weights. By the result of Theorem \ref{theo: effectofadaptive}, it is thus sufficient to show 
\begin{equation}
\label{eq:reducedrelation}
\sum_{l\in S} \hat{w}_l^2\left(\mathcal E_{\theta,l}^2L+{\tilde{\mathcal P}_{l}^2 L}\right) = o\left(\sum_{l\in S^c} \frac{\tilde{\mathcal P}_{l}^2 }{L} + \frac{1}{L^2} \left(\sum_{l\in S^c} \mathcal E_{\theta,l}\right)^2\right).
\end{equation}
By the condition $\eta_n \asymp \frac{\varsigma_n}{n\rho^2}$, we have
\begin{equation*}
\sum_{l\in S} \hat{w}_l^2\left(\mathcal E_{\theta,l}^2L+{\tilde{\mathcal P}_{l}^2 L}\right) \lesssim \frac{L\varsigma_n}{mn\rho^2}. 
\end{equation*}
On the other hand, we have
\begin{equation*}
\sum_{l\in S^c} \frac{\tilde{\mathcal P}_{l}^2 }{L} + \frac{1}{L^2} \left(\sum_{l\in S^c} \mathcal E_{\theta,l}\right)^2 \gtrsim \frac{L-m}{L} \min_{l\in S^c} {\tilde{\mathcal P}_{l}^2} + \frac{(L-m)^2}{L^2} \min_{l\in S^c} \mathcal E_{\theta,l}^2.
\end{equation*}
Hence, the following condition suffices for (\ref{eq:reducedrelation}),
\begin{equation*}
\frac{L\varsigma_n}{mn\rho^2} =o \left(  \frac{(L-m)}{L} \cdot\left(\min_{l\in S^c} \tilde{\mathcal P_{l}^2} + \frac{L-m}{L}\cdot\min_{l\in S^c} \mathcal E_{\theta,l}^2\right)\right),
\end{equation*}
which actually can be implied by our condition \eqref{eq:genequal} and the relationship between $\tilde{\mathcal P}_l$ and ${\mathcal P}_l$. The result then follows. 

\QEDA

\subsubsection*{Proof of Theorem \ref{theo: step2}}
As beginning, we note that $\hat{U}_0^{  RE}(\lambda)$ is the top-$K$ eigenvector of 
$\hat{U}_0\hat{U}_0^\intercal+\lambda \bar{U}\bar{U}^\intercal.$ For convenience, we  denote $\lambda:= \frac{\gamma}{1-\gamma}$ for $0<\gamma <1$. Also note that the population matrix $P_0$'s eigenvector $U_0$ is also the top-$K$ eigenvector of $U_0 U_0^\intercal$ and the minimum non-zero eigenvalue of $U_0 U_0^\intercal$ is 1. Then by 
the Davis-Kahan theorem (see Lemma \ref{pro:DK}),
we have 
\begin{align*}
{\rm dist}(\hat{U}_0^{FT}, U_0) &\leq \|(1-\gamma)\hat{U}_0\hat{U}_0^\intercal+\gamma \bar{U}\bar{U}^\intercal-U_0 U_0^\intercal\|_2 \nonumber\\
&\leq (1-\gamma) \|\hat{U}_0\hat{U}_0^\intercal-U_0 U_0^\intercal\|_2 + \gamma \|\bar{U}\bar{U}^\intercal-U_0 U_0^\intercal\|_2:= \mathcal R_1 + \mathcal R_2.
\end{align*}
For $\mathcal R_1$, similar to \eqref{eq:t5deri}, we have with probability larger than $1-\frac{1}{n^\kappa}$ for some constant $\kappa>0$ that 
\begin{equation*}
\mathcal R_1=(1-\gamma) \|\hat{U}_0\hat{U}_0^\intercal-U_0 U_0^\intercal\|_2\lesssim (1-\gamma)\cdot \frac{\|\hat{A}_0-P_0\|_2}{\lambda_{  \rm min}({P_0})}\lesssim \frac{1-\gamma}{\sqrt{n\rho\nu_0}}\cdot \max\{1,\,\sqrt{\frac{1-q'_0}{\rho\nu_0}}\}:=(1-\gamma) \mathcal T.
\end{equation*}
For $\mathcal R_2$, we have by Theorem  \ref{theo: effectofadaptive} that
\begin{align*}
\mathcal I_2 &\lesssim \gamma \cdot\left(\sqrt{\frac{\log n}{mn\rho\nu_{*}}}+\sqrt{\frac{\log n}{mn\rho\nu_{*}}}\cdot\sqrt{\frac{1-\nu_{*}}{\rho\nu_{*}}}+\frac{1}{m}\sum_{l\in S}\mathcal E_{\theta,l}\right)\\
&:=\gamma (\mathcal M+ \mathcal P + \mathcal H). 
\end{align*}

{Now we analyze the final bound of $\mathcal R_1+\mathcal R_2$, which can be written as the following function of $\gamma$},
\begin{align}
\label{boundfunction}
f(\gamma) &:=  (1-\gamma) \mathcal T+\gamma \left(\mathcal M+\mathcal P + \mathcal H\right)\nonumber\\
&\lesssim (1-\gamma )\mathcal T + \gamma (\mathcal M+ \mathcal P \vee \mathcal H)\nonumber\\
&\lesssim \max\{(1-\gamma )\mathcal T,\; \gamma (\mathcal P \vee \mathcal H)\} + \gamma\mathcal M.
\end{align}
If $(1-\gamma)\mathcal T \lesssim \gamma (\mathcal P \vee \mathcal H)$, i.e., $\gamma \gtrsim \frac{\mathcal T}{\mathcal T+\mathcal P \vee \mathcal H}$,
then (\ref{boundfunction}) reduces to
\begin{equation*}
\gamma (\mathcal P \vee \mathcal H) + \gamma \mathcal M,
\end{equation*}
which attains the minimum at $\gamma^*\asymp \frac{\mathcal T}{\mathcal T+\mathcal P \vee \mathcal H}$, and we then have
\begin{align*}
f(\gamma^*)\lesssim\frac{\mathcal T}{\mathcal T+\mathcal P \vee \mathcal H} (\mathcal P \vee \mathcal H) + \gamma \mathcal M\lesssim \mathcal T\wedge
(\mathcal P \vee \mathcal H)+ \mathcal M,
\end{align*}
where we used the property $\frac{a\cdot b}{a+b} \leq \min\{a,b\}$ in the last inequality. On the other hand, $(1-\gamma)\mathcal T \gtrsim\gamma (\mathcal P \vee \mathcal H)$, i.e., $\gamma \lesssim \frac{\mathcal T}{\mathcal T+\mathcal P \vee \mathcal H}$,
then (\ref{boundfunction}) reduces to
\begin{equation*}
(1-\gamma) \mathcal T+ \gamma \mathcal M,
\end{equation*}
which also attains the minimum at $\gamma^*\asymp \frac{\mathcal T}{\mathcal T+\mathcal P \vee \mathcal H}$, and 
we then have
\begin{align*}
f(\gamma^*)\lesssim\frac{\mathcal P \vee \mathcal H}{\mathcal T+\mathcal P \vee \mathcal H} \mathcal T + \gamma \mathcal M\lesssim \mathcal T\wedge
(\mathcal P \vee \mathcal H)+ \mathcal M.
\end{align*}

Combining the two cases, we have for $\gamma^*\asymp \frac{\mathcal T}{\mathcal T+\mathcal P \vee \mathcal H}$, i.e., $\lambda^*\asymp \frac{\mathcal T}{\mathcal P \vee \mathcal H}$ that
\begin{align*}
 {\rm dist}(\hat{U}_0^{  RE}(\lambda^*), U_0)
&\lesssim \mathcal M+ \frac{1}{\sqrt{n\rho\nu_0}}\max\{1, \sqrt{\frac{1-q'_0}{\rho\nu_0}}\} \wedge \left(\mathcal P\vee \mathcal H\right).
\end{align*} 

\QEDA

\subsubsection*{Proof of Theorem \ref{theo: step3}}
The derivation the misclassification rate based on the bound of eigenspace is a standard technique in the SBM literature; see \citet{lei2015consistency,chen2021spectral} for example. Hence, we omit the details.

\QEDA

\subsection{Theoretical results on effect of bias-adjustment}
\label{app:add}

In Algorithm \ref{alg:transnet}, we used the debiased matrix $\hat{A}_l$ ($l\in \{0\}\cup [L]$) as the input. In the following, we provide the bound of Step 1 (i.e., adaptive weighting step), provided that the input of Algorithm \ref{alg:transnet} is replaced by the non-debiased perturbed matrix $\tA_l$ ($l\in \{0\}\cup [L]$). For clarity, we denote the estimated eigenspace by $\tilde{U}$. 
We need the following Assumption \ref{assu:rankprime} that is the modification of Assumption \ref{assu:rank}.

\renewcommand{\theassumption}{2'}
\begin{assumption}[Eigen-gap]
\label{assu:rankprime}
We assume $\tilde{B}_0$ and $\sum_{l=1}^L w_l \tilde{B}_{l}$ are both of full rank $K$. In addition, $\lambda_{\rm   min}(\tilde{P}_0)\gtrsim n\rho$ and $\lambda_{\rm   min}(\tilde{P})\gtrsim n\rho$.
\end{assumption}
Assumption \ref{assu:rankprime} is natural. When the source networks and target network all follow the two-parameter SBMs (i.e., each connectivity matrix takes two values on the diagonal and non-diagonal, respectively), it can be shown that the minimum eigenvalue of $\tilde{P}$ and $\bar{P}$ are both of order $n\rho$ \citep{rohe2011spectral}. 

In what follows, \textcolor{black}{we assume for simplicity that $\nu_{\min}$ is bounded below by a positive constant, and the results can be easily extended to the general case.}
\begin{theorem}
\label{theo: non-debiased}
Suppose the Assumptions \ref{assu:sparse},  \ref{assu:rankprime} and \ref{assu:conrhoL} hold. Then with probability larger than $1-\frac{L}{n^{\kappa}}$ for some constant $\kappa>0$, the eigenspace $\tilde{U}$ satisfies that
{\small
\begin{align*}
\label{eq:step11err}
 {\rm dist}(\tilde{U}, U_0)
&\lesssim \sqrt{\frac{\log n}{n\rho}}\cdot \sqrt{\sum_{l=1}^L w_l^2}+\sqrt{\frac{\log n}{n\rho}}\cdot \sqrt{\sum_{l=1}^L w_l^2\cdot \frac{1-q'_l}{\rho}}+\sum_{l=1}^L w_l \mathcal E_{\theta,l}
\nonumber\\
&\quad\quad+\sum_l w_l \mathcal E_{\theta,l} (1-q'_l)/\rho+\sum_l w_l \mathcal E_{\theta,l}^2 (1-q'_l)^2/\rho^2  +\sum_l w_l |q'_0-q'_l|^2/\rho^2,
\end{align*} 
}
provided that the last two terms being smaller than a small constant. 
\end{theorem}

\subsubsection*{Proof of Theorem \ref{theo: non-debiased}} The proof follows closely with that of Proposition \ref{theo: step1}, hence we only highlight the difference. 

Define the population matrix of $\tA_l$ by $\tilde{P}_l:=\Theta \tilde{B}_l \Theta^\intercal$ with $\tilde{B}_l:= (q_l+q'_l-1)B_l + (1-q_l)\mathbf 1\mathbf 1^\intercal$ and denote $\tilde{P}:=\sum_l w_l \tilde{P}_l$, then similar to \eqref{eq: decomposition}, we have
\begin{align*}
{\rm dist}(\tilde{U}, U_0)&
 \lesssim{\sum_l\frac{w_l\|{\tA}_l-\tilde{P}_l\|_2^2}{\lambda^2_{  \rm min}(\tilde{P})} } +{\sum_l\frac{w_l\|\tilde{P}_l-\tilde{P}\|_2^2}{\lambda^2_{  \rm min}(\tilde{P})}} \nonumber\\
&\quad\quad\quad +{\frac{\|\sum_l (w_l\tilde{A}_l-w_l\tilde{P}_l)\|_2}{\lambda_{  \rm min}(\tilde{P})}}+{\frac{\|\sum_l w_l\tilde{P}_l-\tilde{P}\|_2}{\lambda_{  \rm min}(\tilde{P})}}+\frac{\|\tilde{A}_0-\tilde{P}\|_2^2}{\lambda^2_{  \rm min}(\tilde{P})} \nonumber\\
&:=\mathcal R_1+\mathcal R_2+\mathcal R_3+\mathcal R_4+\mathcal R_5.
\end{align*}

For $\mathcal R_1$, similar to the derivation of (\ref{eq:t1}), by Assumption \ref{assu:rankprime} and Lemma \ref{pro:DK}, we have with high probability that 
\begin{equation*}
\label{eq:r1}
\mathcal R_1 \lesssim \frac{1}{n\rho} + \frac{\sum_l w_l (1-q'_l)}{n\rho^2}.
\end{equation*}

For $\mathcal R_2$, we note that 
{\small
\begin{align*}
\tilde{P}_l-\tilde{P}&=\Theta_l \tilde{B}_l\Theta_l^\intercal -\Theta_0 \sum_l (w_l \tilde{B_l}) \Theta_0^\intercal \nonumber\\
&=\left(\Theta_l \tilde{B}_l\Theta_l^\intercal-\Theta_l \tilde{B}_l\Theta_0^\intercal\right)+\left(\Theta_l \tilde{B}_l\Theta_0^\intercal-\Theta_0 \tilde{B}_l\Theta_0^\intercal\right)\nonumber\\
&\quad\quad\quad\quad+\left(\Theta_0 \tilde{B}_l\Theta_0^\intercal-\Theta_0 \tilde{B}_0\Theta_0^\intercal\right)+\Big(\Theta_0 \tilde{B}_0\Theta_0^\intercal-\Theta_0 \sum_l(w_l\tilde{B}_l)\Theta_0^\intercal\Big).
\end{align*}
}
Also note that 
\begin{align*}
\|\tilde{B}_l-\tilde{B}_0\|_2 &=\|(q_l+q'_l-1)B_l + (1-q_l)\mathbf 1\mathbf 1^\intercal-[(q_0+q'_0-1)B_0 + (1-q_0)\mathbf 1\mathbf 1^\intercal]\|_2\nonumber\\
&\lesssim \|B_l-B_0\|_2 + \|(q'_0-q'_l)\mathbf 1\mathbf 1^\intercal\|_2\lesssim \rho \mathcal E_{b,l} + |q'_0-q'_l|. 
\end{align*}
Hence, by defining $\tilde{\rho}_l:= (q_l+q'_l-1)\rho + 1-q'_l$, we have
{\small
\begin{align*}
\|\tilde{P}_l-\tilde{P}\|_2&\lesssim n\tilde{\rho}_l \mathcal E_{\theta,l}+n\rho \mathcal E_{b,l}+n\rho \sum_l (w_l \mathcal E_{b,l})+ n|q'_0-q'_l|+n\sum_l w_l |q'_0-q'_l|,
\end{align*}
}
and
{\small
\begin{align*}
\sum_l w_l\|\tilde{P}_l-\tilde{P}\|_2^2&\lesssim n^2 \sum_l w_l\mathcal E_{\theta,l}^2 \tilde{\rho}_l^2 + n^2\rho^2 \sum _l w_l \mathcal E_{b,l}^2 \nonumber\\
& \quad \quad \quad\quad \quad+ n^2\rho^2 (\sum_l w_l \mathcal E_{b,l})^2 +n^2 \sum_l w_l|q'_0-q'_l|^2 + n^2 (\sum_l w_l|q'_0-q'_l|)^2\nonumber \\
&\lesssim  n^2 {\rho}^2\sum_l w_l\mathcal E_{\theta,l}^2  + n^2 \sum_l w_l\mathcal E_{\theta,l}^2 (1-q'_l)^2 + n^2\rho^2 \sum _l w_l \mathcal E_{b,l}^2 +n^2 \sum_l w_l|q'_0-q'_l|^2 \nonumber\\
&\lesssim n^2 {\rho}^2\sum_l w_l\mathcal E_{\theta,l}  + n^2 \sum_l w_l\mathcal E_{\theta,l}^2 (1-q'_l)^2 +n^2 \sum_l w_l|q'_0-q'_l|^2,
\end{align*} 
}

\noindent where we used \eqref{eq:heterP} and the inequality that $(\sum_l w_l a_l)^2\leq \sum_l w_l a_l^2$ for any $a_l$'s. Then, by Assumption \ref{assu:rankprime}, we have
\begin{equation*}
\label{eq:r2}
\mathcal R_2 \lesssim \sum_l w_l\mathcal E_{\theta,l} + \sum_l w_l \mathcal E_{\theta,l}^2\frac{ (1-q'_l)^2}{\rho^2}  +\sum_l w_l \frac{|q'_0-q'_l|^2}{\rho^2}.
\end{equation*}

For $\mathcal R_3$, similar to the derivation of (\ref{eq:t3}), we have with high probability that
\begin{equation*}
\label{eq:r3}
\mathcal R_3 \lesssim \sqrt{\frac{\log n}{n\rho}}\cdot \sqrt{\sum_{l=1}^L w_l^2}+\sqrt{\frac{\log n}{n\rho}}\cdot \sqrt{\sum_{l=1}^L w_l^2\cdot \frac{1-q'_l}{\rho}}.
\end{equation*}

For $\mathcal R_4$, similar to the derivation of (\ref{eq:t4}) and noting the definition of $\tilde{P}_l$ and $\tilde{P}$, we can easily derive that 
\begin{equation*}
\label{eq:r4}
\mathcal R_4 \lesssim \sum_l w_l\mathcal E_{\theta,l} + \sum_l w_l\mathcal E_{\theta,l} \frac{ (1-q'_l)}{\rho}.
\end{equation*}

For $\mathcal R_5$, similar to the derivation of (\ref{eq:t5}) and using the decomposition of $\tilde{P}_0-\tilde{P}$, we have
\begin{equation*}
\label{eq:r5}
\mathcal R_5 \lesssim \frac{1}{n\rho} + \frac{1}{n\rho}\cdot \frac{1-q'_0}{\rho}+\sum_l w_l\mathcal E_{\theta,l} +\sum_l w_l \frac{|q'_0-q'_l|^2}{\rho^2}.
\end{equation*}

By condition (\ref{eq:conL}) in Assumption \ref{assu:conrhoL}, we can show that $\mathcal R_1$ is dominated by $\mathcal R_3$ and the first two terms in $\mathcal R_5$ is dominated by $\mathcal R_3$. In addition, by condition (\ref{eq:conrho1}) and the condition of Theorem \ref{theo: non-debiased}, $\mathcal R_1$ and $\mathcal R_2$ are both smaller than a small constant, which met the condition (\ref{eq:conofcha}) of Proposition \ref{prop: cha}. 

Consequently, the conclusion of Theorem \ref{theo: non-debiased} is achieved. 

\QEDA

\subsection{Auxiliary lemmas}
\label{app:auxiliary}
\begin{lemma}[\citep{chung2011spectra}]
\label{pro:matrixbern}
Let $X_1,...,X_p$ be independent random $n\times n$ Hermitian matrices. Assume that $\|X_i-\mathbb E(X_i)\|_2\leq M$ for all $i$, and $\|\sum_i{\rm var}(X_i)\|_2\leq v^2$. Then for any $a>0$,
\begin{equation*}
\mathbb P\left(\|\sum_iX_i-\mathbb E(\sum_iX_i)\|_2>a\right)\leq 2n\,\exp\left(-\frac{a^2}{2v^2+2Ma/3}\right).
\end{equation*}
\end{lemma}
\QEDA
\begin{lemma}[Davis-Kahan theorem \citep{davis1970rotation}]
	\label{pro:DK}
Let $A\in\mathbb R^{n\times n}$ and $B$ be two symmetric matrix with ${U}_A,{U}_B\in \mathbb R^{n\times K}$ being their corresponding  $K$ leading eigenvectors, respectively. Let $\delta=\min_{1\leq i \leq K, K+1\leq j \leq n} |\lambda_i(A)-\lambda_j(B)|$ be the eigen-gap. Then, for any unitarily invariant norm $\|\cdot\|$,
    \begin{equation*}
\|{U}_A{U}_A^\intercal - {U}_B{U}_B^\intercal\| \asymp\|{\rm sin}\Theta ({U}_A,{U}_B)\|\leq \frac{\|{U}_B^\intercal (A-B){U}_A\|}{\delta}\leq \frac{\|A-B\|}{\delta}.
\end{equation*}	
In particular, if $B$ is of rank $K$, then \begin{equation*}
\|{U}_A{U}_A^\intercal - {U}_B{U}_B^\intercal\|\asymp\|{\rm sin}\Theta ({U}_A,{U}_B)\| \leq \frac{\|A-B\|}{\lambda_K(A)}.
\end{equation*}	
\end{lemma}
\QEDA

\begin{lemma}
\label{pro: eigendecom}
Given an SBM parameterized by $(\Theta_{n\times K}, B_{K \times K})$ with ${\rm rank}(B)=K$, define the population matrix $P=\Theta {B}\Theta^\intercal$ and define $D = {\rm diag}(\sqrt{n_1},...,\sqrt{n_K})$, where $n_k$ denotes the number of nodes in community $k(k\in[K])$. Denote the eigen-decomposition $P={V}{\Sigma}{V}^\intercal$ and $D BD^\intercal=L\Sigma L^\intercal$, respectively. Then, $V=\Theta D^{-1}L$. Specifically, for $\Theta_{i\ast}=\Theta_{j\ast}$, we have ${V}_{i\ast}={V}_{j\ast}$; while for $\Theta_{i\ast}\neq\Theta_{j\ast}$, we have $\|{V}_{i\ast}-{V}_{j\ast}\|_2=\sqrt{(n_{g_i})^{-1}+(n_{g_j})^{-1}}$, where $g_i$ denotes the community index of node $i$ ($i\in[n]$). 
\end{lemma}
\noindent\emph{Proof.} See \citet{lei2015consistency,zhang2022randomized}, among others.
\QEDA

\subsection{{\color{black}Discussions on the optimality}}
\label{app:opt}

We analyze the optimality of \texttt{TransNet} in terms of eigenspace estimation in the homogeneous setting, where the population eigenspace and the privacy parameter $\epsilon$ are shared across source and target networks. In this setting, Step 1 of \texttt{TransNet} with equal weights is sufficient, and the Step 2 is not needed. Recall that $$\nu=q+q'-1 \quad {\rm and}\quad q=q'=\frac{{\rm e}^\epsilon}{1+{\rm e}^\epsilon},$$ then Proposition \ref{theo: step1} provides that the estimator obtained after Step 1 satisfies the following upper bound on the projection distance,
\begin{equation}
\label{eq:upp}
\sqrt{\frac{(1-\nu)\log n}{Ln\rho^2\nu^2}}\asymp \sqrt{\frac{{\rm e}^\epsilon\log n}{Ln\rho^2({\rm e}^\epsilon-1)^2}},
\end{equation}
provided that $\epsilon\in [0,\epsilon_0]$ for some constant $\epsilon_0>0$, which implies that $\nu<c<1$ for come constant $0<c<1$; and the last equality follows from 
$$\frac{{\rm e}^\epsilon} {({\rm e}^\epsilon -1)^2}=\frac{(1-q)q}{(q+q'-1)^2}\asymp \frac{1-\nu}{\nu^2}.$$

For notational simplicity, we first focus on directed networks and then extend the analysis to undirected networks. We consider the following parameter space:
\begin{equation}
\label{eq: paramspace}
{\mathcal Y}(n, L, \rho, K, \sigma_{*})=\{l\in [L], P_l\in [0, \rho]^{n\times n}: {\rm rank}(P_l)=K, \Pi({P_l})= \Pi, \sigma_K(P_l)\geq \sigma_{*}\},
\end{equation}
where $\Pi({P_l})$ denotes the projection matrix of the left singular space of $P_l$, $\sigma_K(P_l)$ denotes the singular value of $P_l$. Suppose that given $l\in[L]$, the entries of adjacency matrix $A_l$ are generated independently from 
\begin{equation*}
A_{l,ij}\sim {\rm Bernoulli}(P_{l,ij})\quad {\rm for}\quad i,j\in[n]\times [n],  
\end{equation*}
where the membership information is not incorporated, since it is not used in Proposition \ref{theo: step1}. Moreover, the matrices $A_1,...,A_L$
are mutually independent.
Let us consider $\mathcal X _{{\epsilon}}$, the class of all estimators of the common projection matrix $\Pi$ that are constructed from $L$ datasets released by the $L$ data providers after perturbing the original adjacency matrices using any randomized mechanism to enforce non-interactive $\epsilon$-edge-local-DP  for the $l$th dataset, where each privatized entry of adjacency matrix only depends on the corresponding entry of the original adjacency matrix \citep{duchi2013local}. The following theorem provides the lower bound for the minimax risk 
$$\underset{\hat{\Pi}\in \mathcal X _{\bar{\epsilon}}}{\inf} \underset{P_{l\in [L]}\in {\mathcal Y}(n, L, \rho, K, \sigma_{*})}{\sup} \mathbb E \|\hat{\Pi}-\Pi\|_2.$$

\begin{theorem}
\label{theo:opt}
Suppose the privacy parameter $\epsilon\in [0,\epsilon_0]$ for some constant $\epsilon_0>0$, $K>1$, and $K\sigma_{*}^2\leq n^2\rho^2/C_0$ for some constant $C_0>0$,
then we have 
\begin{equation*}
\underset{\hat{\Pi}\in \mathcal X _{\bar{\epsilon}}}{\inf} \underset{P_{l\in [L]}\in {\mathcal Y}(n, L, \rho, K, \sigma_{*})}{\sup} \mathbb E \|\hat{\Pi}-\Pi\|_2\gtrsim \sqrt{\frac{n {\rm e}^\epsilon}{\sigma_{*}^2L ({\rm e}^\epsilon-1)^2}\wedge 1}\asymp \sqrt{\frac{n(1-\nu)}{\sigma_{*}^2L \nu^2}\wedge 1}.
\end{equation*}
\end{theorem}

\begin{remark}
By the definition of parameter space $\mathcal Y$ in \eqref{eq: paramspace}, we naturally have that 
$$n^2\rho^2\geq {\rm tr} (P_lP_l^\intercal )=\sum_{i=1}^n\sigma_i^2(P_l)\geq \sigma_{*}^2K,$$ thus the condition $K\sigma_{*}^2\leq n^2\rho^2/C_0$ in Theorem is mild. In addition, Assumption \ref{assu:rank} used for Proposition \ref{theo: step1} requires that $\sigma_{*}\gtrsim n\rho$. Combining these two, when  $\sigma_{*}\asymp n\rho$, the lower bound in Theorem \ref{theo:opt} reduces to $\sqrt{\frac{1-\nu}{L n\rho^2\nu^2}\wedge 1},$ which matches the upper bound \eqref{eq:upp} up to log terms if $\rho$ is large such that $\frac{1-\nu}{L n\rho^2\nu^2}\lesssim 1$.
\end{remark}

\begin{remark}
The minimax results for directed networks can be adapted to the undirected setting through the standard \emph{symmetric dilation} technique. Specifically, for any matrix asymmetric $P_l\in\mathbb R^{\frac{n}{2}\times \frac{n}{2}}$ (for simplicity, we assume $n$ is even), define
\[
\mathscr D(P_l):=
\begin{pmatrix}
0 & P_l\\
P_l^\top & 0
\end{pmatrix}\in\mathbb R^{n\times n}.
\]
Then the left singular space of $P_l$ coincides with the eigenspace of $\mathscr D(P_l)$. One can thus follow the construction of $P_l$ (but modify the rank to $K/2$) to construct undirected population matrix $\mathscr D(P_l)$. The minimax bound is of the same order with that in Theorem \ref{theo:opt}.
\end{remark}

\begin{remark}
The minimax rate of subspace estimation under low rank models with Gaussian noise has been studied by \citet{cai2018rate,cai2021optimal}, among others. Compared with the Gaussian noise model, the network model requires that the population matrix lies in $[0,1]$, which makes the problem challenging. The minimax rate of subspace estimation under low rank network models without DP constraint has been studied by \citet{zhou2021rate,yan2023minimax}. We make use of the results in \citet{zhou2021rate} and the contraction of local DP mechanisms in \citet{asoodeh2024contraction} to establish Theorem~\ref{theo:opt}.
\end{remark}

In Theorem \ref{theo:opt}, we focus on the minimax optimality of the eigenspace estimation. 
In terms of clustering, the minimax rate of pure membership estimation in the non-private SBM setting has been studied by \citet{gao2017achieving,zhang2016minimax}, where exponential-type misclustering rates are established. Under local differential privacy, \citet{chakrabortty2024prime} derive minimax rates for mixed membership estimation in degree-corrected SBMs.
In our setting, which involves multiple networks and local privacy constraints, we expect the optimal rate of membership estimation to exhibit a similar exponential behavior. Establishing such a result would likely require either refined spectral clustering procedures \citep{gao2017achieving} or more delicate analysis techniques \citep{loffler2021optimality}. We leave this as an interesting direction for future work.
\iffalse
In terms of clustering, the minimax rate of pure membership estimation under SBMs has been studied by \citet{gao2017achieving,zhang2016minimax}, where the exponential-type misclustering rate is established. On the other hand, \citet{chakrabortty2024prime} established the minimax rate of mixed membership estimation under degree-corrected mixed membership SBMs with local DP constraints.
We expect that the minimax rate of membership estimation in our setting, which involves multiple networks and local DP constraints, may also be exponential. Achieving this lower bound requires either a refined spectral clustering method \citep{gao2017achieving} or a more delicate technical analysis \citep{loffler2021optimality}.
We leave this problem for future work. 
\fi

\subsubsection{Proof of Theorem \ref{theo:opt}}
Following \citet{zhou2021rate}, we establish a lower bound on the Bayes error rate. The minimax lower bound then follows from the fact that any minimax estimator is a Bayes estimator under a least favorable prior on the parameter space. For notational simplicity, we first consider the case $L=1$ and then extend the analysis to the multiple networks setting with $L>1$.

The construction of separable population matrices in parameter space $\mathcal Y$ is similar with \citet{zhou2021rate}. Define \(k_2 \ge 1\), \(\sigma>0\) and \(\mu \in (0,1)\) as
\begin{equation}\label{eq:k2-sigma-mu}
k_2=\left\lceil (10/\rho)^2 \sigma_*^2 / n \right\rceil,\quad
\sigma^2 = n k_2 (\rho/10)^2,\quad
\mu^2 = \min\{{\rm e}^\epsilon/(0.32k_2({\rm e}^\epsilon-1 )^2\rho^2),\,0.1\}/2.
\end{equation}
It is thus easy to see that \(\sigma_* \le \sigma \le \sqrt{2}\sigma_*\). The following construction holds as \(K\sigma_*^2 \le n^2\rho^2/C_0\),
$k_2 \le \frac{200n}{KC_0} \le \frac{n-1}{2K-2}$
for sufficiently large \(C_0\). Let
\[
H \in [-\sqrt{3},\sqrt{3}]^{n\times (K-1)}
\]
such that
\[
(H,1_n)^\intercal (H,1_n)/n = \mathbb I_K.
\]
Let \(R_i\), \(i=1,\ldots,N\), be distinct matrices in \(\{-1,1\}^{n\times (K-1)}\), with
$N=2^{n(K-1)}.$
Define
\[
W_i = \sqrt{1-\mu^2}\,H + \mu R_i, \qquad 0<\mu<1,
\]
and
\begin{equation}\label{eq:Pi-construction}
P_i = \frac{\rho}{2} 1_{n\times n}
+ \frac{\rho}{10}(W_i,-W_i,\ldots,W_i,-W_i,O),
\end{equation}
where \((W_i,-W_i)\) is repeated \(k_2\) times and $O$ represent the matrix with 0 entries. It is easy to see that
\[
\|W_i\|_\infty \le \sqrt{(1-\mu^2)3}+\mu \le 2,
\]
thus
\[
P_i \in [0.3\rho,0.7\rho]^{n\times n}.
\]
Let
\[
\Pi_i = \Pi_{P_i} = P_i P_i^\dagger \in \mathbb{R}^{n\times n}
\]
be the orthogonal projection onto the column space of \(P_i\). Let
\[
X_i = (W_i,1_n)=\bigl(\sqrt{1-\mu^2}\,H+\mu R_i,\,1_n\bigr)\in\mathbb{R}^{n\times K},
\]
so if \(\operatorname{rank}(X_i)=K\), then \(X_i\) has the same column space as \(P_i\) and we can write
\[
\Pi_i = X_i (X_i^\intercal X_i)^{-1} X_i^\intercal .
\]
By calculation, we have
\begin{equation}\label{eq:XtX}
n^{-1} X_i^\intercal X_j
=
\begin{cases}
\mathbb I_K + \Delta^{i,j} + \mu^2\bigl(V^{i,j}-\mathbb I_K\bigr), & i\neq j,\\[0.5ex]
\mathbb I_K + \Delta^{i,i}, & i=j,
\end{cases}
\end{equation}
where
\[
V^{i,j}
=
\left(
\begin{array}{c|c}
R_i^\intercal R_j / n & \mathbf{0} \\
\hline
\mathbf{0}^\intercal & 1
\end{array}
\right),
\qquad
\Delta_i
=
\frac{\mu}{n}
\left(
\begin{array}{c|c}
\sqrt{1-\mu^2}\,R_i^\intercal H & R_i^\intercal 1_n \\
\hline
\mathbf{0}^\intercal & 0
\end{array}
\right),
\]
and
\[
\Delta^{i,j} = \Delta_i + \Delta_j^\intercal + \mu^2\bigl(V^{i,j}-\mathbb I_K\bigr)\mathbf{1}_{\{i=j\}}.
\]
Therefore,  \(\operatorname{rank}(X_i)=K\) whenever \(\|\Delta^{i,i}\|_{2}<1\), that we will impose later. By Lemma 6 of \citet{zhou2021rate} (see also Lemma \label{lem:zhoulem6}), we have
\begin{equation}
\label{eq:sigP}
\sigma_K(P_i)
\ge
(\rho/10)\sqrt{
n\,2k_2\Bigl(
1-2\mu^2\Delta_i'
-\mu^2\Delta_i''
-(1+1/92)\mu^4(\Delta_i''')^2
\Bigr)_+
}
\end{equation}
 where
\[
\Delta_i' = \|R_i^\intercal H/(n\mu)\|_{2},\qquad
\Delta_i'' = \|R_i^\intercal R_i/n - \mathbb I_{K-1}\|_{2},
\qquad
\Delta_i''' = \|R_i^\intercal 1_n/(n\mu)\|_2.
\]
Let \(\delta_n\) satisfy \(0<\delta_n\le 1/(8\mu^2)\), to be determined later, and define
\[
\Omega^* = \{ i\le N:\ \Delta_i' \vee \Delta_i'' \vee \Delta_i''' \le \delta_n \}.
\]
As
\[
\|\Delta_i\|_{2} \le \mu^2\Delta_i' + \mu^2\Delta_i''',
\qquad
\|V^{i,i}-\mathbb I_K\|_{2}=\Delta_i'',
\]
we have
\[
\|\Delta^{i,i}\|_{2} \le 5\mu^2\delta_n<1
\]
for \(i\in\Omega^*\). When \(i\in\Omega^*\), we have \[
\sigma_K^2(P_i)
\ge
(\rho/10)^2 n\,2k_2 (1-4\mu^2\delta_n)_+
\ge
(\rho/10)^2 n\,k_2
=
\sigma^2
\ge
\sigma_*^2
\] by \eqref{eq:sigP}. Hence, $\{i\in\Omega^*\}$ implies that
\begin{equation}\label{eq:Omega-property}
\{P_i\in [0.3\rho,0.7\rho]^{n\times n},\ \sigma_K(P_i)\ge \sigma\ge \sigma_*,\ \operatorname{rank}(X_i)=K\}.
\end{equation}
Further, for \(\{i,j\}\subset \Omega^*\), we have
\[
\|\Delta^{i,j}\|_{2}
\le
(4+\mathbf{1}_{\{i=j\}})\mu^2\delta_n,
\]
so that by using \eqref{eq:XtX} we have
\begin{equation}\label{eq:trace-bound}
\begin{aligned}
\operatorname{tr}(\Pi_i\Pi_j)
&=\operatorname{tr}\!\Bigl(
(X_i^\intercal X_i)^{-1}X_i^\intercal X_j
(X_j^\intercal X_j)^{-1}X_j^\intercal X_i
\Bigr)\\
&\le
K + (C_1-1)\mu^2\delta_n K
+ \mu^2(1-\mu^2)\operatorname{tr}(V^{i,j}+V^{j,i}-2\mathbb I_K)
+ \mu^4\operatorname{tr}(V^{i,j}V^{j,i}-\mathbb I_K) \\
&\le
K + C_1\mu^2\delta_n K
+ \mu^2(1-\mu^2)\operatorname{tr}(V^{i,j}+V^{j,i}-V^{i,i}-V^{j,j}) \\
&=
K + C_1\mu^2\delta_n K
- \mu^2(1-\mu^2)\|R_i-R_j\|_F^2/n,
\qquad \forall\, i,j\in\Omega^*,
\end{aligned}
\end{equation}
where \(C_1\) is a numerical constant, and the details are provided in the proof of Lemma 7 of \citet{zhou2021rate}. The relationship \eqref{eq:trace-bound} will be used to transform the distance between subspace to that between parameter matrix $R_i$'s.

Let \(R\), \(P\), and \(\Pi_P\) be random matrices with the uniform prior distribution \(\pi(\cdot)\),
\[
\pi(i)=\mathbb P_{\pi}\bigl(R=R_i,\; P=P_i,\; \Pi_P=\Pi_i\bigr)=1/N=2^{-n(K-1)},
\]
so that the entries of \(R\) are i.i.d.\ Rademacher variables under \(\mathbb P_{\pi}\). Let
$\mathcal R^*=\{R_i:\ i\in\Omega^*\}$,
\(\pi^*\) be the uniform prior on \(\Omega^*\), and \(\mathbb P_{\pi^*}\) be the corresponding joint probability measure, so that \(\mathbb P_{\pi^*}\) is the conditional probability given \(R\in\mathcal R^*\) under \(\mathbb P_{\pi}\). By \eqref{eq:Omega-property},
$\mathbb P_{\pi^*}\bigl\{P\in\mathcal Y(n,1,\rho,K,\sigma_{*})\bigr\}=1.$

Now we proceed to establish the desired lower bound. By \eqref{eq:trace-bound} and Lemma~8 in \citet{zhou2021rate}, the Frobenius risk of the Bayes estimator under \(\mathbb P_{\pi^*}\) is bounded by
\begin{equation}\label{eq:bayes-risk-lower}
R_{\pi^*}^{\mathrm{Bayes}}
=
\mathbb E_{\pi^*}\bigl[\|\hat\Pi^*-\Pi_P\|_F^2\bigr]
\ge
\mu^2(1-\mu^2)n^{-1}\mathbb E_{\pi^*}\bigl[\|\hat R^*-R\|_F^2\bigr]
-
C_1\mu^2\delta_n K,
\end{equation}
where \(\hat\Pi^*\) and \(\hat R^*\) are respectively the posterior means of \(\Pi_P\) and \(R\) under \(\mathbb P_{\pi^*}\). Moreover, as
\[
\|\hat R^*\|_F^2 \vee \|R\|_F^2 \le Kn,
\]
so 
\begin{equation}\label{eq:bayes-risk-compare}
\mathbb E_{\pi^*}\bigl[\|\hat R^*-R\|_F^2\bigr]
+
\mathbb P_{\pi}(\Omega^{*c})\,4nK
\ge
\mathbb E_{\pi}\bigl[\|\hat R^*-R\|_F^2\bigr]
\ge
\mathbb E_{\pi}\bigl[\|\hat R-R\|_F^2\bigr],
\end{equation}
where \(\hat R\) is the Bayes estimator of \(R\) under \(\mathbb P_{\pi}\), by the optimality of \(\hat R\) under \(\mathbb P_{\pi}\). Now, we derive the lower bound of $\mathbb E_{\pi}\bigl[\|\hat R-R\|_F^2\bigr]$.
Under $\mathbb P_{\pi}$, the entries of the adjacency matrix $A$ are conditionally independent given $R$, and the privatized observations $Z=\{Z_{ij}\}$ are conditionally independent given $A$, where recall that
each $Z_{ij}$ is generated from $A_{ij}$ through the same non-interactive 
$\epsilon$-edge-local-DP mechanism. Moreover, the entries of $R$ are i.i.d.\
Rademacher variables under $\mathbb P_{\pi}$.
Since $(W_i,-W_i)$ is repeated $k_2$ times, conditionally on $R$, the $k_2$ i.i.d.\ privatized copies
of $(Z_{i,j},Z_{i,j+K-1})$ contain all the information for estimating the $(i,j)$-th entry $R_{i,j}$ of
$R$. Let
\[
\delta=\mu\rho/10.
\]
For fixed $(i,j)$, under $R_{i,j}=+1$ and $R_{i,j}=-1$, denote by $\nu_+$ and $\nu_-$ the laws of one
privatized pair $(Z_{i,j},Z_{i,j+K-1})$, respectively. Note that the unprivatized $A_{i,j}$ and $A_{i,j+K-1}$ are independent Bernoulli variables with probabilities $p_{i,j}+ \delta R_{i,j}\in [0.3\rho,0.7\rho]$ and $q_{i,j}- \delta R_{i,j}\in [0.3\rho,0.7\rho]$ for some $p_{i,j}$ and $q_{ij}$. Also note that the total variation distance satisfies
\[
{\rm TV}\!\bigl(\mathrm{Bern}(p_{i,j}+\delta)\otimes \mathrm{Bern}(q_{i,j}+\delta),\;\mathrm{Bern}(p_{i,j}-\delta)\otimes \mathrm{Bern}(q_{i,j}-\delta)\bigr)\leq 4 \delta.
\]
Hence, by the $\chi^2$-contraction inequality (Theorem 2 in \citet{asoodeh2024contraction}; see also Lemma \ref{lem:localcontraction}) for $\epsilon$-local DP mechanisms, we have the $\chi^2$-distance upper bounded as
\[
\chi^2(\nu_+\|\nu_-)
\le
64\Psi_\epsilon\,\delta^2,
\qquad
\Psi_\epsilon:=e^{-\epsilon}(e^\epsilon-1)^2.
\]
Now let $\hat R_{i,j}$ be the Bayes estimator of $R_{i,j}$ based on the $k_2$ i.i.d.\ privatized pairs.
By Lemma~9 of \citet{zhou2021rate} (see also Lemma \ref{lem:hell}), then
\[
\mathbb E_\pi\bigl[(\hat R_{i,j}-R_{i,j})^2\bigr]
\ge
\left(1-\frac12 \mathcal H^2(\nu_+,\nu_-)\right)^{k_2},
\]
where $\mathcal H$ denotes the Hellinger distance. 
Using $\mathcal H^2(\nu_+,\nu_-)\le \chi^2(\nu_+\|\nu_-)$, we obtain
\begin{equation}\label{eq:coordinate-risk-lower}
\begin{aligned}
\mathbb E_{\pi}\bigl[(\hat R_{i,j}-R_{i,j})^2\bigr]&\ge
\left(1-\frac12 \mathcal H^2(\nu_+,\nu_-)\right)^{k_2}\ge
1-\frac{k_2}{2} \mathcal H^2(\nu_+,\nu_-)\\
&\ge
1-\frac{k_2}{2} \chi^2(\nu_+\|\nu_-)\ge
1-32\,k_2\,\Psi_\epsilon\,\delta^2\ge
1-0.32\,k_2\,\Psi_\epsilon\,\mu^2\rho^2.
\end{aligned}
\end{equation}
By the definition of $\mu^2$ in \eqref{eq:k2-sigma-mu}, we have
$$1-0.32\,k_2\,\Psi_\epsilon\,\mu^2\rho^2\ge 0.5
\qquad\text{and}\qquad
1-\mu^2\ge 0.95.$$

Now, we are ready to derive the lower bound.
By \eqref{eq:bayes-risk-lower} and \eqref{eq:bayes-risk-compare}, we have
\[
R_{\pi^*}^{\mathrm{Bayes}}
\ge
\mu^2(1-\mu^2)\Bigl(
n^{-1}\mathbb E_{\pi}\bigl[\|\hat R-R\|_F^2\bigr]
-
\mathbb P_{\pi}(\Omega^{*c})\,4K
\Bigr)
-
C_1\mu^2K\delta_n
\]
and hence
\[
R_{\pi^*}^{\mathrm{Bayes}}
\ge
0.475\,\mu^2K
-
\bigl(4\mathbb P_{\pi}(\Omega^{*c})+C_1\delta_n\bigr)\mu^2K.
\]
We aim to show that $
4\mathbb P_{\pi}(\Omega^{*c})+C_1\delta_n \le 0.075$, and the lower bound under single network set-up follows. 
To this end, we take
\[
\delta_n
=
\max\left\{
\sqrt{\frac{32\pi K\sigma^2({\rm e}^\epsilon-1)^2}{n^2{\rm e}^\epsilon}}
+
\sqrt{\frac{128x_0\sigma^2({\rm e}^\epsilon-1)^2}{n^2{\rm e}^\epsilon}},
\;
4\sqrt{\frac{3K+x_0}{n}}
\right\},
\]
with \(x_0=\log(320)\) satisfying \(16e^{-x_0}=0.05\). As
$\sigma^2 \le 2\sigma_*^2 \le \frac{2n^2\rho^2}{KC_0}$, $C_0K \le n$ since we assume $K$ is fixed, and $\epsilon\in[0,\epsilon_0]$ for some small constant $\epsilon_0>0$, we have
\[
\delta_n
\le
\max\left\{
\sqrt{\frac{64\pi\rho^2}{C_2}}
+
\sqrt{\frac{256x_0\rho^2}{C_2}},
\;
4\sqrt{\frac{3+x_0}{C_2}}
\right\},
\]
for some constant $C_2$.
Thus, \(\mu^2\delta_n \le 1/8\) and \(C_1\delta_n \le 1/40\) for sufficiently large \(C_2\). Moreover, Lemma \ref{lem:delta-bound} implies that
\[
4\mathbb P_{\pi}\{\Omega^{*c}\} \le 16e^{-x_0} \le \frac{1}{20},
\]
so that $
4\mathbb P_{\pi}(\Omega^{*c}) + C_1\delta_n \le \frac{3}{40}$ holds. Consequently, by \eqref{eq:k2-sigma-mu},
\[
R_{\pi^*}^{\mathrm{Bayes}}
\ge
0.4K\mu^2\asymp \mu^2 \asymp \frac{n {\rm e}^\epsilon}{\sigma_{*}^2 ({\rm e}^\epsilon-1)^2}\wedge 1,
\]
and we obtain the minimax lower bound for $L=1$.

Finally, for $L>1$, since the $L$ networks are conditionally independent and share the same eigenspace $\Pi$ and privacy parameter $\epsilon$, the above argument extends directly to this setting. In particular, the information accumulates linearly over the $L$ networks, so the lower bound is improved by a factor of $L$. Therefore,
\[
R_{\pi^*}^{\mathrm{Bayes}} \gtrsim \mu^2 \asymp \frac{n e^\epsilon}{L\sigma^2 (e^\epsilon-1)^2}\wedge 1.
\]

\subsubsection{Useful lemmas for Theorem \ref{theo:opt}}

\begin{lemma}[Lemma 6 in \citet{zhou2021rate}]\label{lem:zhoulem6}
Let \(P_i\) be as in \eqref{eq:Pi-construction},
\[
\Delta_i' = \|R_i^\intercal H/(n\mu)\|_{2},\qquad
\Delta_i'' = \|R_i^\intercal R_i/n - \mathbb I_{K-1}\|_{2},
\qquad
\Delta_i''' = \|R_i^\intercal 1_n/(n\mu)\|_2.
\]
Then, the \(K\)-th singular value of \(P_i\) is lower bounded as 
\[
\sigma_K(P_i)
\ge
(\rho/10)\sqrt{
n\,2k_2\Bigl(
1-2\mu^2\Delta_i'
-\mu^2\Delta_i''
-(1+1/92)\mu^4(\Delta_i''')^2
\Bigr)_+
}.
\]
\end{lemma}

\begin{lemma}[Theorem 2 in \citet{asoodeh2024contraction}]
\label{lem:localcontraction}
If \( M\) is an \(\epsilon\)-local DP mechanism, then
\[
\chi^2(PM \,\|\, QM)
\le
\Psi_\epsilon \min\!\left\{4\,\mathrm{TV}^2(P,Q),\, \mathrm{TV}(P,Q)\right\},
\]
for any pair of distributions \((P,Q)\) and \(\epsilon \ge 0\), where
\[
\Psi_\epsilon := e^{-\epsilon}(e^\epsilon - 1)^2.
\]
\end{lemma}

\begin{lemma}[Lemma 9 in \citet{zhou2021rate}]
\label{lem:hell}
Let \(\mathbb{P}\) be a probability measure under which \(R\) is a Rademacher variable and given
\(R=\pm 1\), \(Y_1,\ldots,Y_k\) are i.i.d.\ random vectors with joint density \(f_{\pm}(y)\) with respect to a measure
\(\nu(dy)\). Let \(\mathcal H(f_{+},f_{-})\) be the Hellinger distance between \(f_{+}\) and \(f_{-}\) and \(\hat{R}\) be the Bayes estimator
of \(U\) based on \(Y_1,\ldots,Y_k\). Then,
\[
\mathbb{E}\bigl[(R-\hat{R})^2\bigr]
\ge
\left(1-\frac{\mathcal H^2(f_{+},f_{-})}{2}\right)^k.
\]
\end{lemma}

\begin{lemma}[Modification of Lemma 5 in \citet{zhou2021rate}]
\label{lem:delta-bound}
Let
$H \in \{-1,1\}^{n\times (K-1)}$
such that
$(H,1_n)^\intercal (H,1_n)/n = \mathbb I_K.$
Let \(K\ge 2\), and let
\[
R \in \{-1,1\}^{n\times (K-1)}
\]
have i.i.d.\ Rademacher entries. Then
\[
\mathbb P\left\{
\begin{aligned}
&\|R^\intercal H/n\|_{2} \vee \|R^\intercal 1_n/n\|_2
\le
\sqrt{{2\pi (K-1)}/{n}}+\sqrt{{8x}/{n}},\\
&\|R^\intercal R/n-\mathbb I_{K-1}\|_{2}
\le
4\sqrt{{(3(K-1)+x)}/{n}}
\end{aligned}
\right\}
\ge
1-4e^{-x}.
\]
Suppose \(n\rho \le \sigma^2\) and let
$\mu^2=\frac{n {\rm e}^\epsilon }{16 \sigma^2({\rm e}^\epsilon-1)^2}.$
Then, for
$$\delta_n=\max\left\{
\sqrt{\frac{32\pi K\sigma^2({\rm e}^\epsilon-1)^2}{n^2{\rm e}^\epsilon}}
+
\sqrt{\frac{128x_0\sigma^2({\rm e}^\epsilon-1)^2}{n^2{\rm e}^\epsilon}},
\;
4\sqrt{\frac{3K+x_0}{n}}
\right\},$$
we have
\[
\mathbb P\bigl\{\Delta_i' \vee \Delta_i'' \vee \Delta_i''' \le \delta_n\bigr\}
\ge
1-4e^{-x}.
\]
\end{lemma}

\subsection{Tuning parameter selection and additional numerical results}
\label{add:num}
\subsubsection{Tuning parameter selection}
The tuning parameter $\lambda$ in the regularization step of \texttt{TransNet} and \texttt{TransNetX} can be selected using the cross validation, following the idea in \citet{chen2018network}, which was initially proposed to select the number of communities in SBMs. Let $\Lambda$ be the set of candidate values for $\lambda$ and let the number of folds be $V\geq 2$ (in our experiments, $V=5$). The procedure is presented in the notation of \texttt{TransNet} (Algorithm \ref{alg:transnet}), with an analogous version for \texttt{TransNetX}.

\begin{enumerate}[leftmargin=*, itemsep=0.5ex]

\item {Randomly split the $n$ nodes into $V$ subsets such that the first $V-1$ subsets each contain $\lfloor n/V \rfloor$ nodes, and the remaining $V$-th subset contains $n - (V-1)\lfloor n/V \rfloor$ nodes.} Denote the subsets by
$\{\tilde{\mathcal N}_v: 1\le v\le V\}$. According to the resulting subsets, split the debiased adjacency matrix $\hat{A}_0$ for the target matrix into $V\times V$ equal-sized blocks
\[
\hat{A}_0=\big(\hat{A}^{(uv)}_0: 1\le u,v\le V\big),
\]
where $\hat {A}^{(uv)}_0$ is the submatrix of $\hat{A}_0$ with rows in $\tilde{\mathcal N}_u$
and columns in $\tilde{\mathcal N}_v$. 
% Define the submatrices:
% \[
% \hat {A}^{(-v)}_0=\big(\hat {A}^{(rs)}_0: r\ne v,\; 1\le r,s\le V\big)\quad{\rm and} \quad  \hat {A}^{(v)}_0=\big(\hat {A}^{(rs)}_0: r= v,\; 1\le r,s\le V\big).
% \]

\item For each $1\le v\le V$, and each $\lambda \in \Lambda$

\begin{enumerate}[label=(\alph*), itemsep=0.25ex]

\item \emph{Estimation of memberships:}
Define the rectangular
submatrix obtained by removing the rows of $\hat{A}_0$ in subset $\tilde{\mathcal N}_v$:
\[
\hat {A}^{(-v)}_0=\big(\hat {A}^{(rs)}_0: r\ne v,\; 1\le r,s\le V\big).
\]
Run the \texttt{TransNet} (Algorithm \ref{alg:transnet}) on $\{(\hat{A}_l)_{l\in[L]},\hat {A}^{(-v)}_0\}$ to obtain the membership $\hat g^{(v)}$ ($\hat g^{(v)}$ corresponds to the membership matrix $\hat\Theta^{(v)}$ via one-hot encoding), where
the right singular vectors $\hat{U}_0^{v}$ of $\hat {A}^{(-v)}_0$ and the eigenvectors $\hat{U}_l$ of $\hat{A}_l$ are used.   

\item \emph{Estimation of connectivity matrice $B_0$:}
Denote $\bar{N}_1:=\{\tilde{\mathcal N}_u: u\neq v\}$ and $\bar{N}_2:=\{\tilde{\mathcal N}_u: u=v\}$.
With $\hat g^{(v)}$, let $\mathcal N_{j,k}$ be the nodes in $\bar{N}_j$
with estimated membership $k$, and $n_{j,k} = |\mathcal N_{j,k}|$
($j=1,2,\; 1\le k \le K$). 
Estimate $B_0$ using the plug-in estimator:
\[
\hat B_{0,kk'}^{(v)} =
\begin{cases}
\displaystyle
\frac{\sum_{i\in \mathcal N_{1,k},\, j\in \mathcal N_{1,k'}\cup \mathcal N_{2,k'}} \hat{A}_{0,ij}}
      {n_{1,k}(n_{1,k'}+n_{2,k'})},
& k\ne k',
\\[3ex]
\displaystyle
\frac{\sum_{i,j\in \mathcal N_{1,k},\, i<j} \hat{A}_{0,ij} 
      + \sum_{i\in \mathcal N_{1,k},\, j\in \mathcal N_{2,k}} \hat{A}_{0,ij}}
     {(n_{1,k}-1)n_{1,k}/2 + n_{1,k}n_{2,k}},
& k=k'.
\end{cases}
\]

\item \emph{Validation:}

Calculate the predictive loss:
\[
\hat L^{(v)}(\hat{A}_0,\lambda)
= \sum_{i\in \tilde{\mathcal N}_v, j\notin \tilde{\mathcal N}_v, i\ne j}
(\hat{P}_{0,ij}-\hat{A}_{0,ij})^2, \quad {\rm with}\quad \hat{P}_{0,ij}:=\hat B^{(v)}_{0,\hat g^{(v)}_i\hat g^{(v)}_j}.\]
\end{enumerate}

\item Let $\hat L(\hat{A}_0,\lambda)=\sum_{v=1}^{V}\hat L^{(v)}(\hat{A}_0,\lambda)$ and output
\[
\hat \lambda=\arg\min_{\lambda\in\Lambda}\; \hat L(\hat{A}_0,\lambda)\, .
\]
\end{enumerate}

\subsubsection{Additional numerical results corresponds to Experiments I-III}
\begin{figure*}[!htbp]{}
%6.80 6.43
\centering

\subfigure[Case III]{\includegraphics[height=6cm,width=6cm,angle=0]{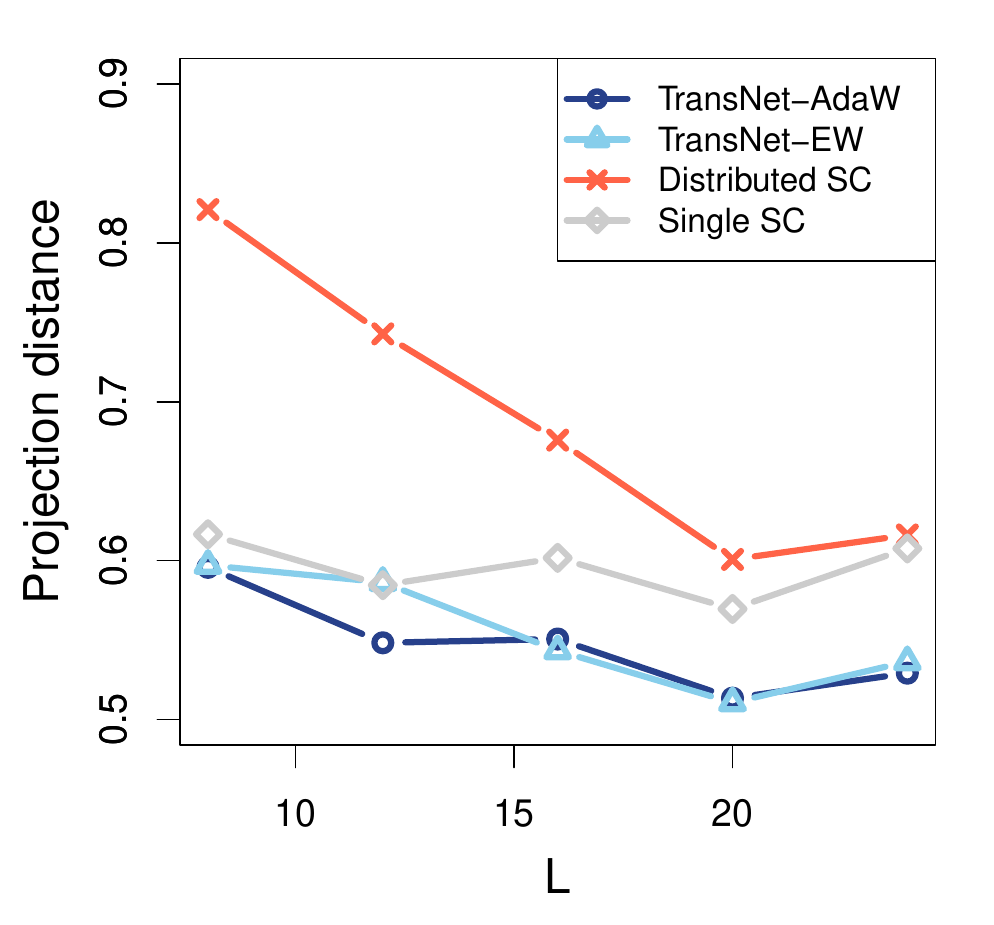}}
\subfigure[Case III]{\includegraphics[height=6cm,width=6cm,angle=0]{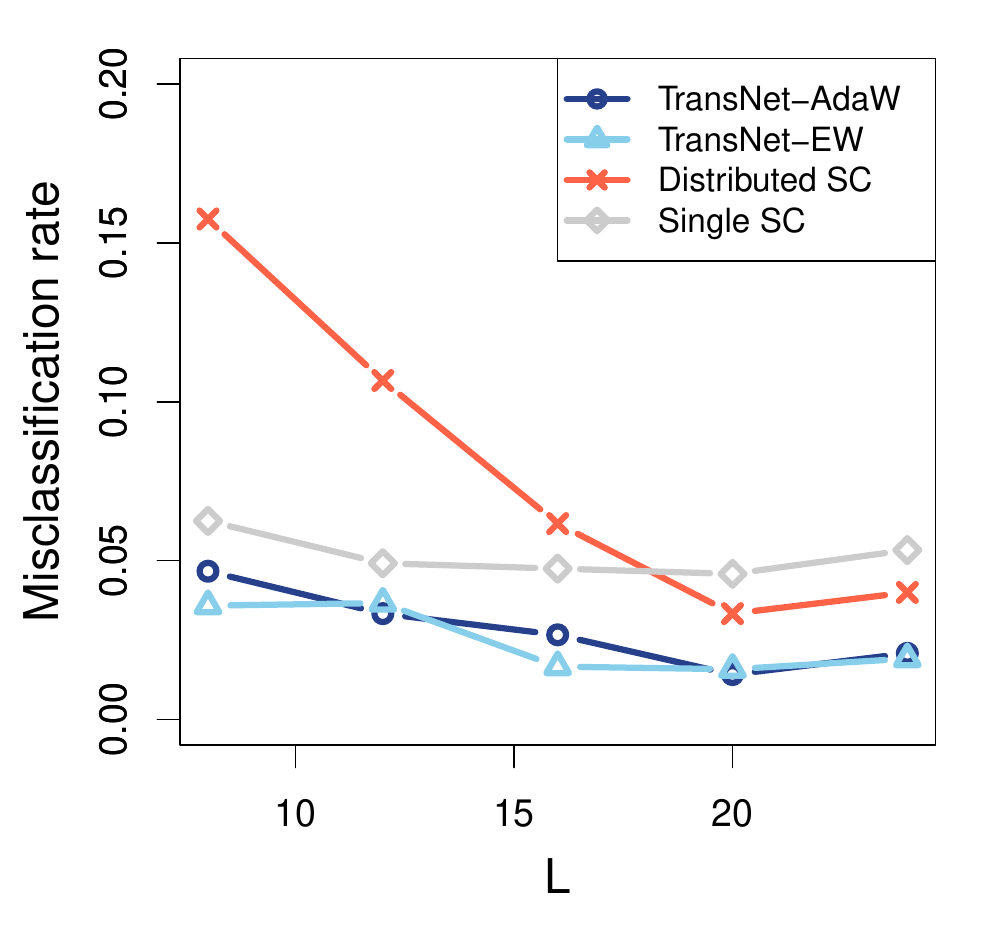}}
\caption{The projection distance (the first row) and misclassification rate (the second row) of each method under Case III of Experiment I (Private but non-heterogeneous). }\label{sim_pri_case3}
\end{figure*}

\begin{figure*}[!h]{}
\centering
\subfigure[Case III]{\includegraphics[height=6cm,width=6cm,angle=0]{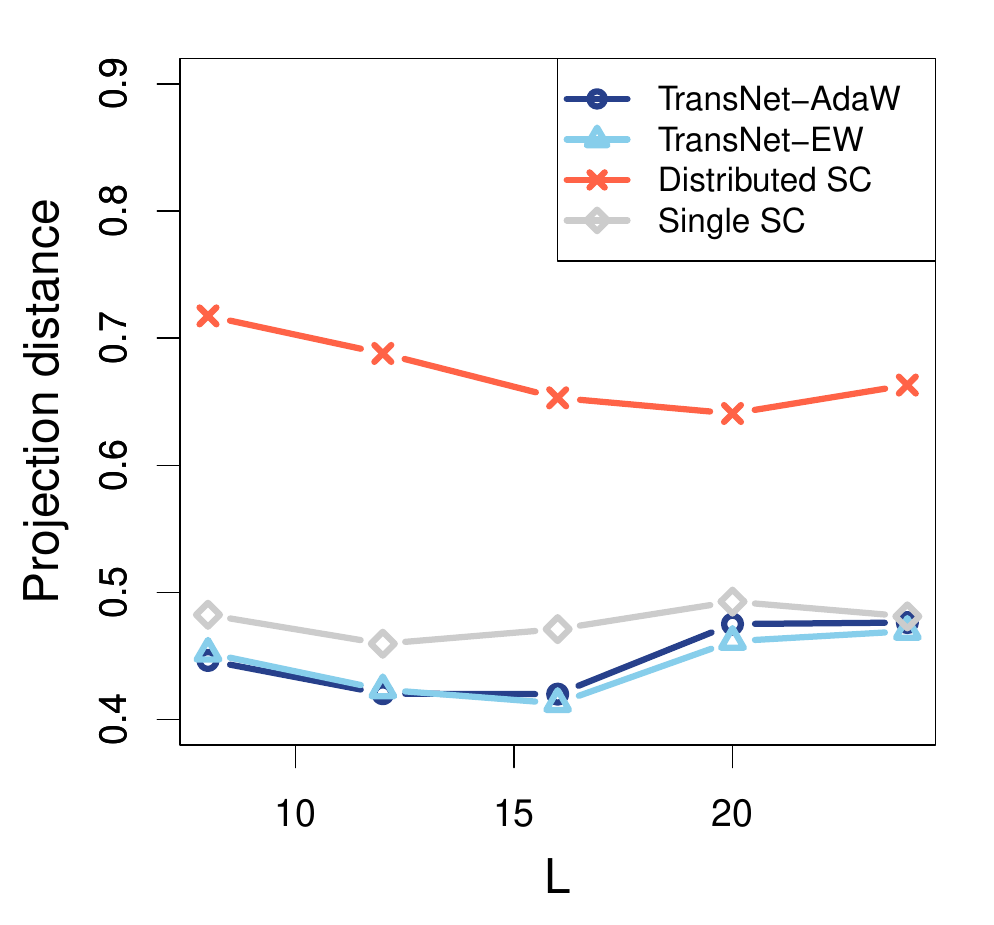}}
\subfigure[Case III]{\includegraphics[height=6cm,width=6cm,angle=0]{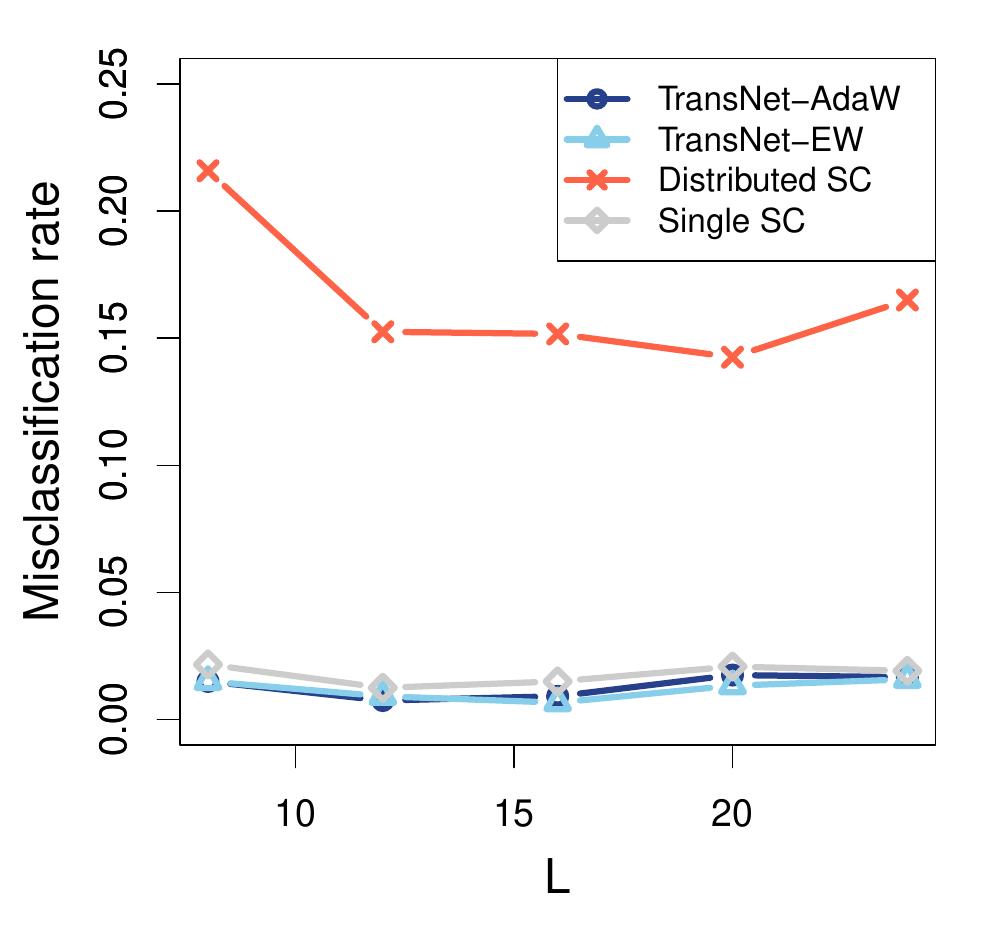}}
\caption{The projection distance (the first row) and misclassification rate (the second row) of each method under the Case III of Experiment II (Heterogeneous but non-private). }\label{sim_het_case3}
\end{figure*}

\begin{figure*}[!h]{}
\centering
\subfigure[Case III]{\includegraphics[height=6cm,width=6cm,angle=0]{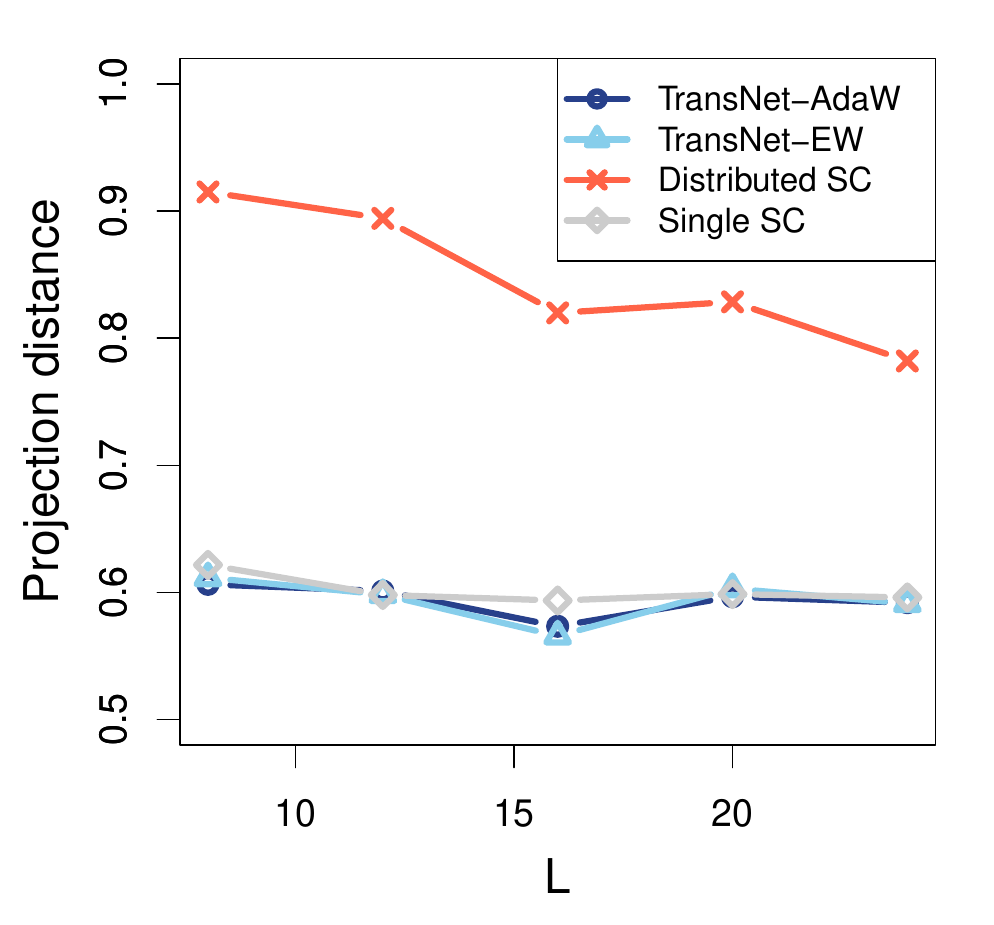}}
\subfigure[Case III]{\includegraphics[height=6cm,width=6cm,angle=0]{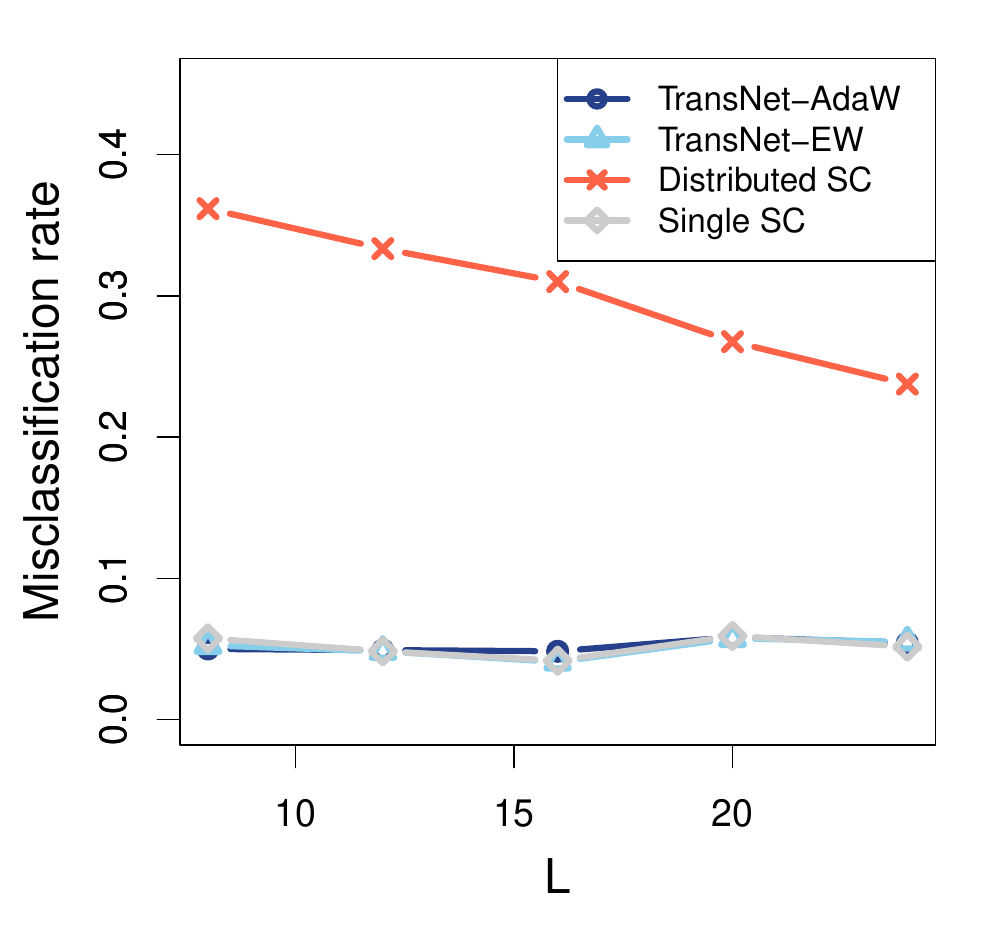}}
\caption{The projection distance (the first row) and misclassification rate (the second row) of each method under Case III of Experiment III (Heterogeneous and private). }\label{sim_het_pri_case3}
\end{figure*}

The numerical results of Case III (see the last column in Table \ref{table:experi}) in Experiments I-III are provided in Figures \ref{sim_pri_case3}-\ref{sim_het_pri_case3}. Please refer to Section \ref{sec:sim} for the method description and parameter settings.

\subsubsection{Sensitivity analysis}
We test the sensitivity of the proposed method. In addition to the aforementioned four methods, we also evaluate the oracle estimator, denoted by \texttt{TransNet-Oracle}, which serves as the counterpart of the proposed algorithm but enforces equal weights among the informative networks in the first step.

First, we evaluate the performance of the proposed method as the number of informative networks varies. The network generating process is similar to Section \ref{sec:sim}. Specifically, we fix $n=120$, $K=3$ and $L=20$, and let $m$ vary. The source networks are partitioned into $m$ informative networks and $L- m$ non-informative networks. 
The connectivity matrices for the target network, informative and non-informative networks are all fixed to be 
\begin{equation*}
\scriptsize
B=\begin{bmatrix}
0.33 & 0.1 & 0.1  \\
0.1 & 0.33 & 0.06\\
0.1 & 0.06 & 0.33
\end{bmatrix}.
\end{equation*}
The heterogeneity level and the privacy-preserving level are identical within the informative networks and within the non-informative networks. Specifically, we use $\mu = (\mu_S, \mu_{S^c})$ and $q = (q_S, q_{S^c})$ to denote the heterogeneity and privacy-preserving levels of the source networks, where $S$ and $S^c$ indicate the informative and non-informative networks, respectively. The parameters $\mu$ and $q$ are chosen to mimic the theoretical definition of informative networks. Specifically, under this setup, we have $\eta_n = \frac{\varsigma_n}{n\rho^2} \approx 0.296$. In particular, when $\mu_S = 0.1$ and $q_S = 0.85$, we obtain $\mathcal{P}_l^2 + \mathcal{E}_l^2 = 0.162 + 0.129 \approx \eta_n$. Thereby, in the following experiments, the networks with $\mu_S\leq 0.1$ and $q_S\geq 0.85$ are considered to be informative networks. We consider the following two experiments.  

\begin{figure*}[!h]{}
\centering
\subfigure [Case I]
{\includegraphics[height=4.3cm,width=4.5cm,angle=0]{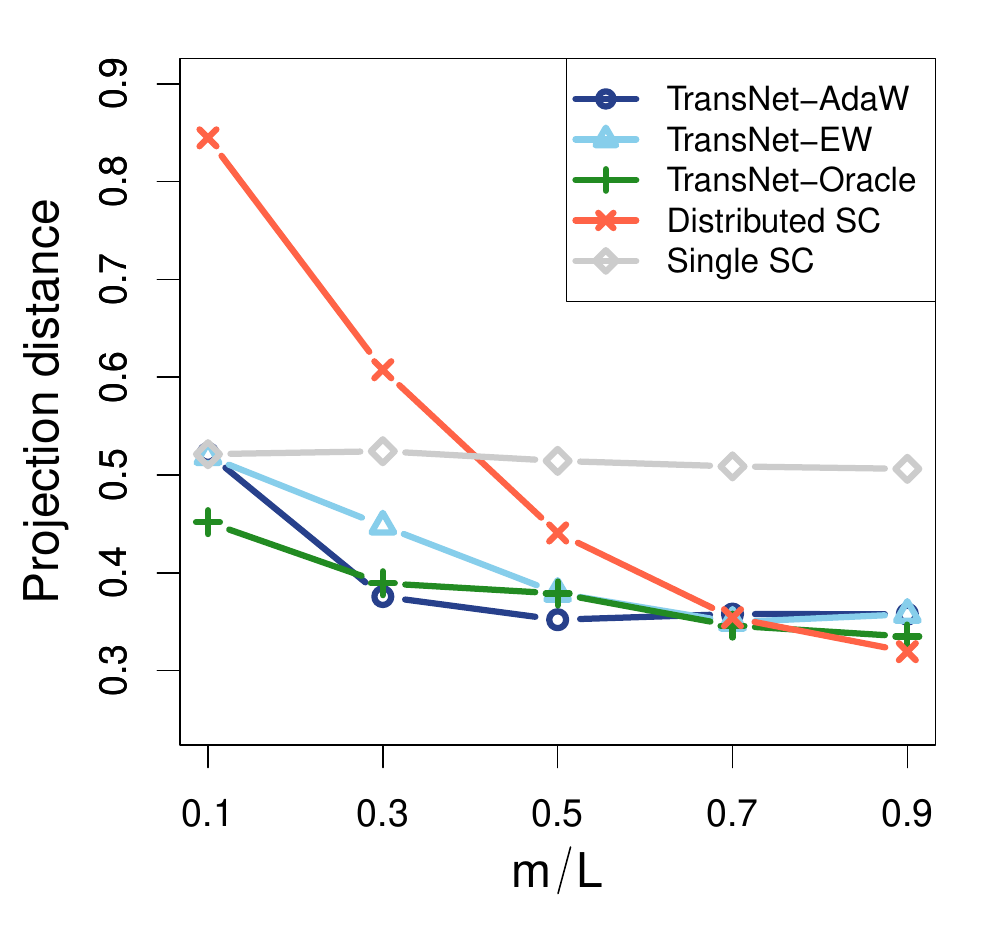}}
\subfigure[Case II]
{\includegraphics[height=4.3cm,width=4.5cm,angle=0]{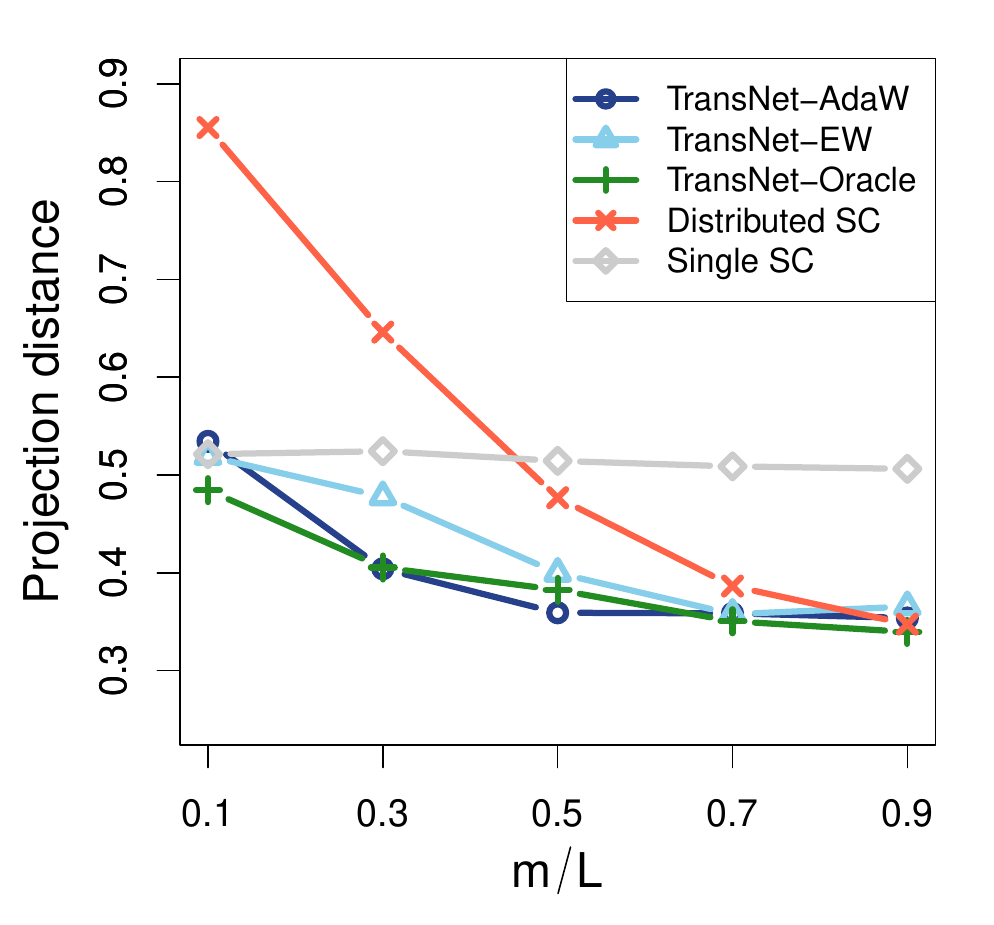}}
\subfigure[Case III]
{\includegraphics[height=4.3cm,width=4.5cm,angle=0]{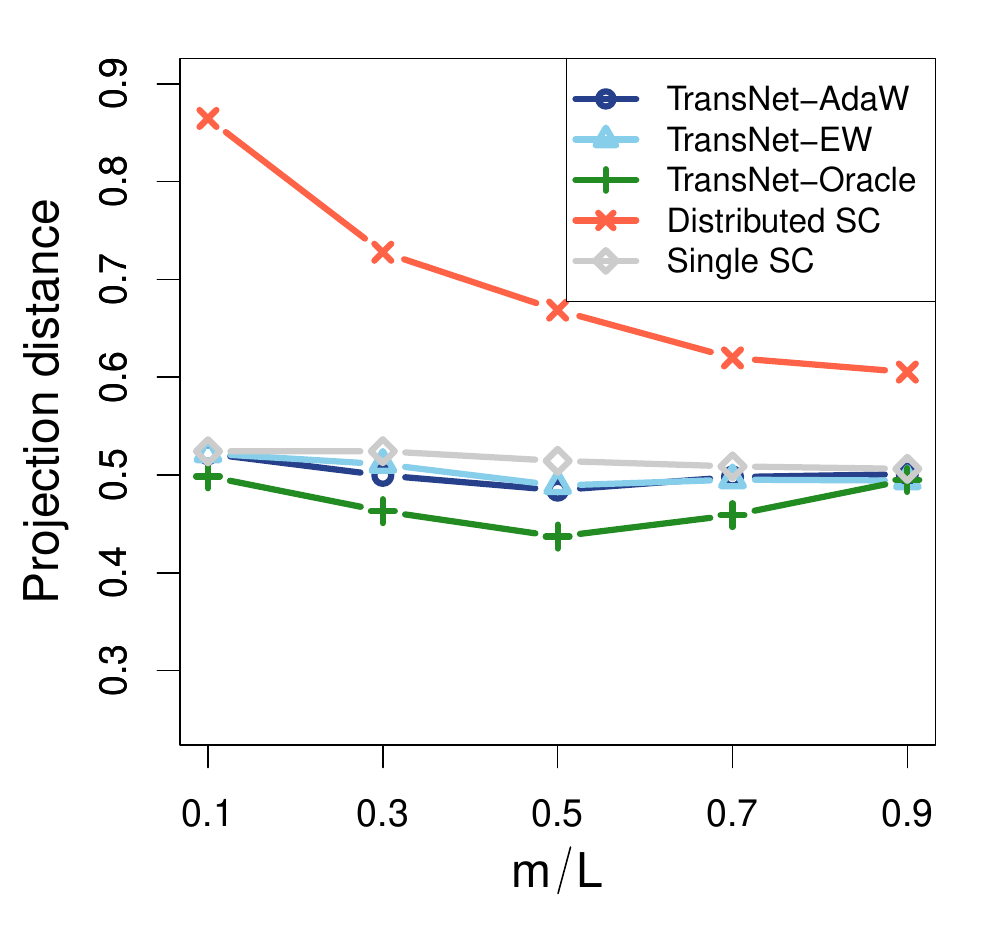}}
% \subfigure{\includegraphics[height=4.3cm,width=4.5cm,angle=0]{case11_mis.pdf}}
% \subfigure{\includegraphics[height=4.3cm,width=4.5cm,angle=0]{case12_mis.pdf}}
% \subfigure{\includegraphics[height=4.3cm,width=4.5cm,angle=0]{case13_mis.pdf}}
\caption{The projection distance of each method under Experiment IV.}\label{fig:expiv}
\end{figure*}

\paragraph{Experiment IV.} For the non-informative networks, the parameters $(\mu_{S^c},q_{S^c})$ are set as $(0.3,0.7)$. For the informative networks, we consider three cases; see the first column of Table \ref{table:experi2} for details. For reference, in Table \ref{table:experi2},  we also numerically calculate the lower bound for $m/L$ indicated by condition \eqref{eq:conditionfororacle} with multiplicative constant being $1$, i.e., $\frac{L-m}{m}\leq \frac{ \min_{l\in S^c} ({\mathcal E}_{\theta,l}^2+ {\mathcal P}_{l}^2) }{\eta_n} \cdot \min\{1,\frac{\log n}{L}\}$, which is a sufficient condition for \eqref{eq:conditionfororacle}. Figure \ref{fig:expiv} shows the average 
projection distance over 20 replications as the proportion of informative networks $m/L$ increases. The results show that the proposed method \texttt{TransNet-AdaW} performs comparably to the oracle method \texttt{TransNet-Oracle}, and outperforms other methods, across a wide range of $m/L$. In particular, in Cases I and II, the proportion of informative networks can be as small as 30$\%$. As the quality of the informative networks decreases, the performance of all methods deteriorates.

\begin{figure*}[!h]{}
\centering
\subfigure[Case I]
{\includegraphics[height=4.3cm,width=4.5cm,angle=0]{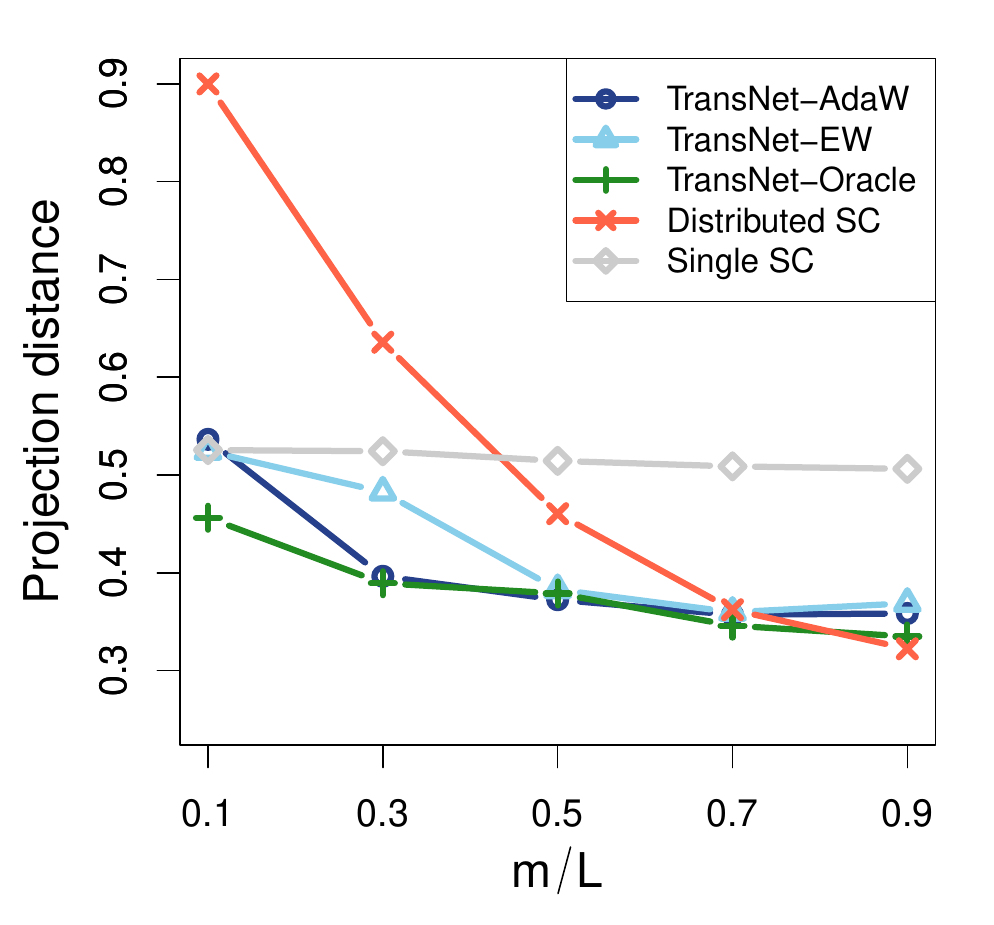}}
\subfigure[Case II]
{\includegraphics[height=4.3cm,width=4.5cm,angle=0]{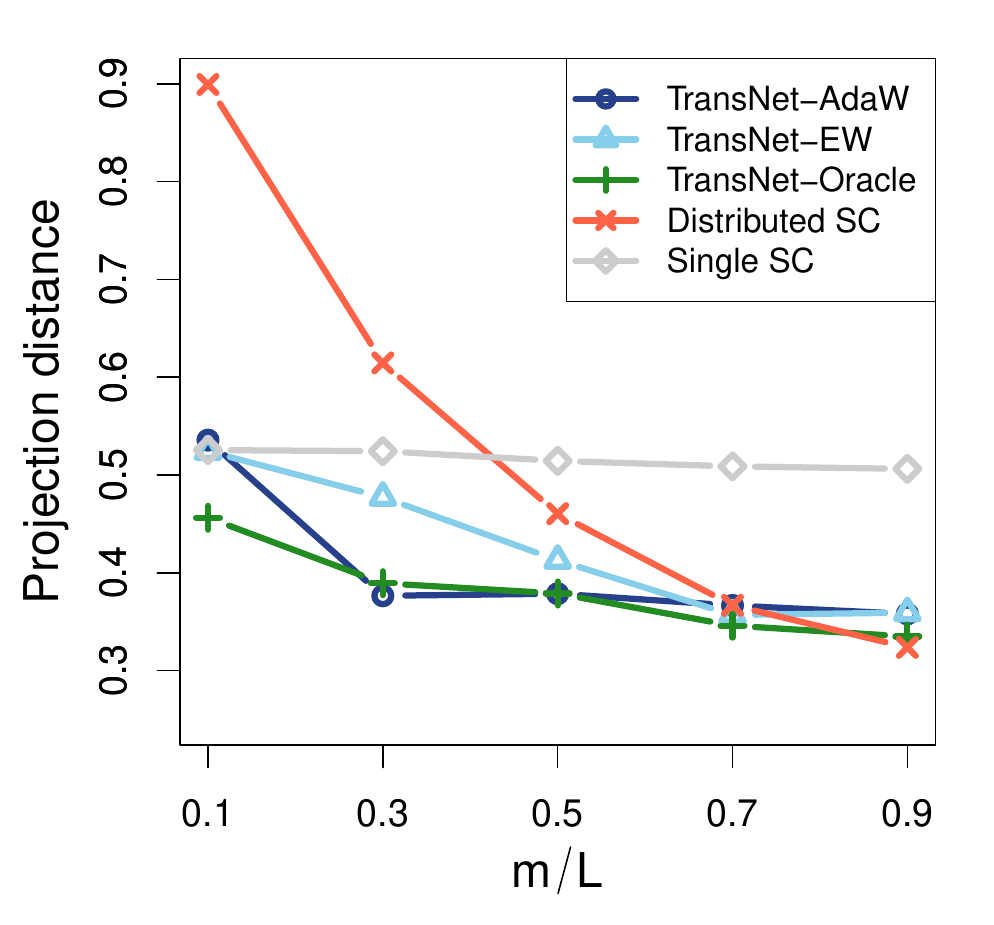}}
\subfigure[Case III]
{\includegraphics[height=4.3cm,width=4.5cm,angle=0]{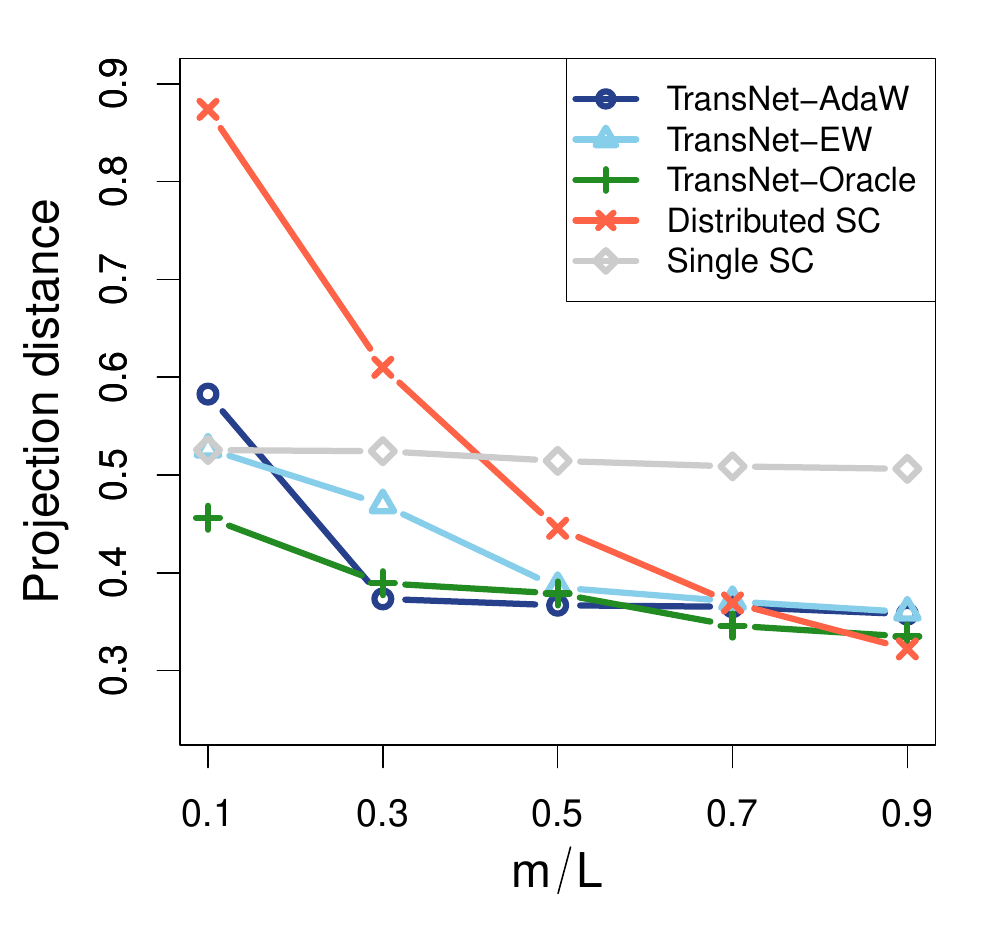}}
% \subfigure{\includegraphics[height=4.3cm,width=4.5cm,angle=0]{case21_mis.pdf}}
% \subfigure{\includegraphics[height=4.3cm,width=4.5cm,angle=0]{case22_mis.pdf}}
% \subfigure{\includegraphics[height=4.3cm,width=4.5cm,angle=0]{case23_mis.pdf}}
\caption{The projection distance of each method under Experiment V.}\label{fig:expv}
\end{figure*}

\paragraph{Experiment V.} We consider informative networks with strong signals, where the parameters $(\mu_{S},q_{S})$ are set as $(0.02,0.95)$. For the non-informative networks, we consider three cases; see the second column of Table \ref{table:experi2} for details. Figure \ref{fig:expv} shows the average 
projection distance over 20 replications as the proportion of informative networks $m/L$ increases. The results are similar to Experiment IV except that the proposed method is not sensitive to the quality of non-informative networks. In particular, under the considered parameter set-up, the privacy parameter $q_{S^c}$ of non-informative networks can be as small as $0.55$; the heterogeneity parameter $\mu_{S^c}$ can be as large as $0.5$. Note that in Cases I and II, the proposed method \texttt{TransNet-AdaW} performs comparably to the oracle method \texttt{TransNet-Oracle}, even when the lower bound for $m/L$ is not satisfied. In Case III, although the condition for $m/L$ is met across all considered values, \texttt{TransNet-AdaW} fails to match the performance of \texttt{TransNet-Oracle} when $m/L = 0.1$. This is because the theoretical requirement is derived up to unknown multiplicative constant, while in practice, too many non-informative networks deteriorates the effectiveness of the adaptive weights. 

\begin{table}[!htb]
\centering
\small
\caption{The parameter set-ups of Experiments IV-V. In all cases, the source networks are partitioned into two groups of unequal sizes. The first group consists of $m$ informative networks, whereas the second group consists of $L-m$ non-informative networks. Similar to Table \ref{table:experi}, we use $\mu = (\mu_S,\mu_{S^c})$ and $q=(q_{S},q_{S^c})$ to set the heterogeneity level and privacy-preserving level for each group.}\vspace{0.5cm}
\def\arraystretch{1.3}
\begin{tabular}{c|c|c|c|c|c|c}
\hline
\multirow{2}{*}{\textbf{Cases}} & \multicolumn{3}{c|}{\textbf{Experiment IV}} & \multicolumn{3}{c}{\textbf{Experiment V}} \\
\cline{2-7}
 & $(\mu_{S},q_{S})$ & $(\mu_{S^c},q_{S^c})$ & Condition \eqref{eq:conditionfororacle}&$(\mu_{S},q_{S})$ & $(\mu_{S^c},q_{S^c})$& Condition \eqref{eq:conditionfororacle} \\ 
\hline
Case I   & $(0.02, 0.95)$ & $(0.3, 0.7)$  & $\frac{m}{L}\geq 0.55$& $(0.02, 0.95)$ & $(0.5, 0.7)$& $\frac{m}{L}\geq 0.51$ \\ \hline
Case II  & $(0.02, 0.9)$ & $(0.3, 0.7)$  &$\frac{m}{L}\geq 0.55$&  $(0.02, 0.95)$ & $(0.5, 0.65)$& $\frac{m}{L}\geq 0.40$ \\ \hline
Case III & $(0.1, 0.85)$  & $(0.3, 0.7)$  &$\frac{m}{L}\geq 0.55$&  $(0.02, 0.95)$ & $(0.5, 0.55)$&$\frac{m}{L}\geq 0.08$ \\ \hline
\end{tabular}
\label{table:experi2}
\end{table}

% \begin{table}[!htb]
% \centering
% \small
% \caption{The parameter set-ups of Experiments IV-V. In all cases, the source networks are partitioned into four groups of unequal sizes. The first two groups are informative, each consisting of $m/2$ networks, whereas the last two groups are non-informative, each consisting of $(L - m)/2$ networks. Similar to Table \ref{table:experi}, we use $\mu = (\mu^{(1)},\mu^{(2)},\mu^{(3)},\mu^{(4)})$ and $q=(q^{(1)},q^{(2)},q^{(3)},q^{(4)})$ to set the heterogeneity level and privacy-preserving level for each group. Specifically, $\mu^{(1)}=\mu^{(2)}=\mu_S$ and $\mu^{(3)}=\mu^{(4)}=\mu_{S^c}$; $q^{(1)}=q^{(2)}=q_S$ and $q^{(3)}=q^{(4)}=q_{S^c}$. }\vspace{0.5cm}
% \def\arraystretch{1.3}
% \begin{tabular}{c|c|c}
% \hline
% \textbf{Cases} &\tabincell{c}{ \textbf{Experiment IV}} & \tabincell{c}{\textbf{Experiment V}} \\
% \hline
% {Case I}&\tabincell{c}{$\mu=c(0.02,0.02,0.5,0.5)$\\$q=c(0.95,0.95,0.7,0.7)$ }&\tabincell{c}{$\mu=c(0.02,0.02,0.3,0.3)$\\$q=c(0.95,0.95,0.75,0.75)$ }\\\hline
% {Case II}&\tabincell{c}{$\mu=c(0.02,0.02,0.5,0.5)$\\$q=c(0.9,0.9,0.7,0.7)$ }&\tabincell{c}{$\mu=c(0.02,0.02,0.5,0.5)$\\$q=c(0.95,0.95,0.75,0.75)$ }\\\hline
% {Case II}&\tabincell{c}{$\mu=c(0.1,0.1,0.5,0.5)$\\$q=c(0.9,0.9,0.7,0.7)$ }&\tabincell{c}{$\mu=c(0.02,0.02,0.5,0.5)$\\$q=c(0.95,0.95,0.65,0.65)$ }\\\hline

% \hline
% \end{tabular}
% \label{table:experi2}
% \end{table}

Second, we evaluate the performance of the proposed method as the privacy parameters vary. 
The network generating process is similar to Experiments IV and V. Specifically, we fix $n=120$, $K=3$, $L=20$, and $m=10$.  The connectivity matrices for the target network, informative and non-informative source networks are all set to be 
\begin{equation*}
\scriptsize
B=\rho\begin{bmatrix}
0.33 & 0.1 & 0.1  \\
0.1 & 0.33 & 0.06\\
0.1 & 0.06 & 0.33
\end{bmatrix}.
\end{equation*}
We fix $(\mu_S, \mu_{S^c})=(0.02,0.3)$ and consider the following experiment. 

\begin{figure*}[!h]{}
\centering
\subfigure[$\rho=1$]
{\includegraphics[height=4.3cm,width=4.5cm,angle=0]{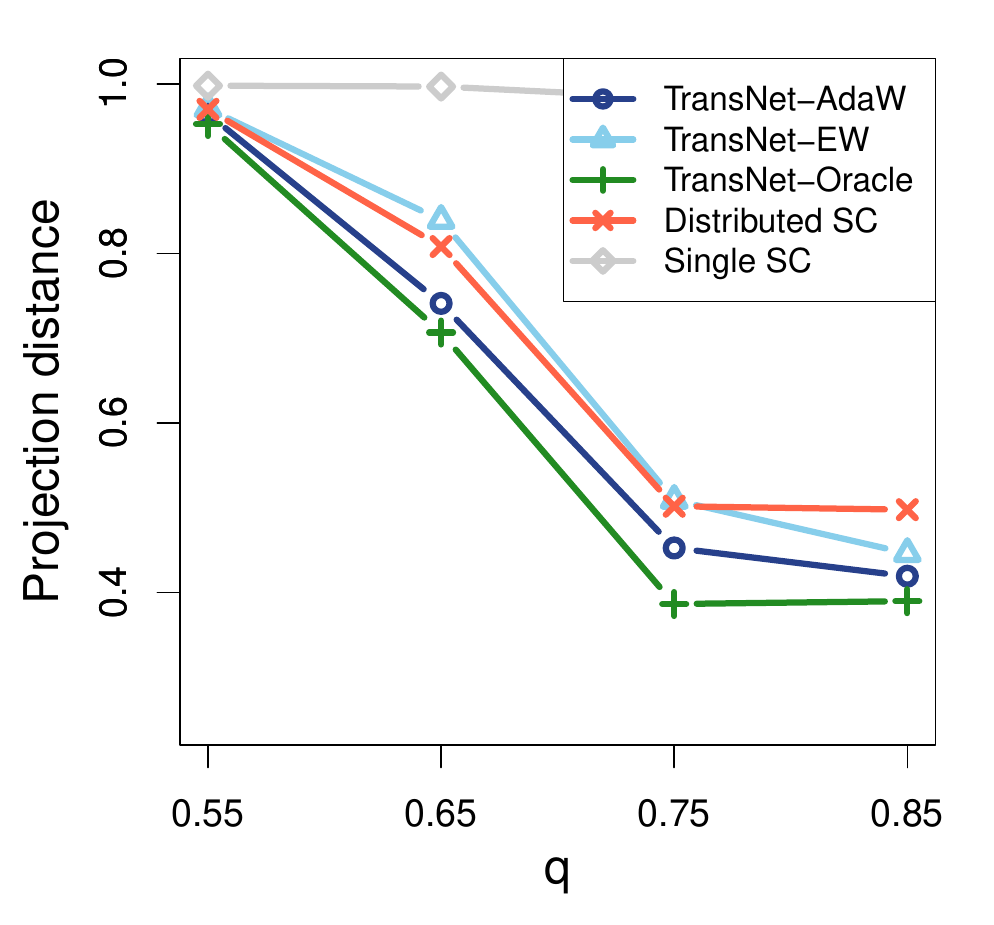}}
\subfigure[$\rho=1.5$]
{\includegraphics[height=4.3cm,width=4.5cm,angle=0]{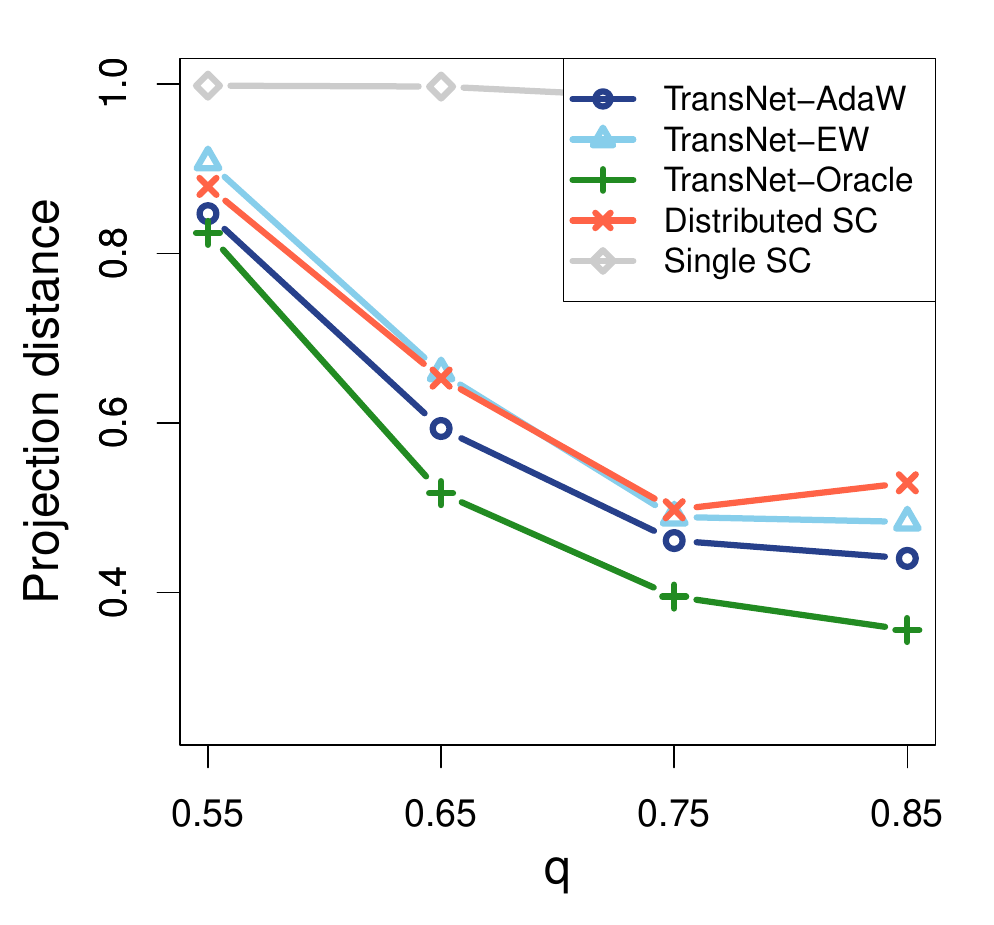}}
\subfigure[$\rho=2$]
{\includegraphics[height=4.3cm,width=4.5cm,angle=0]{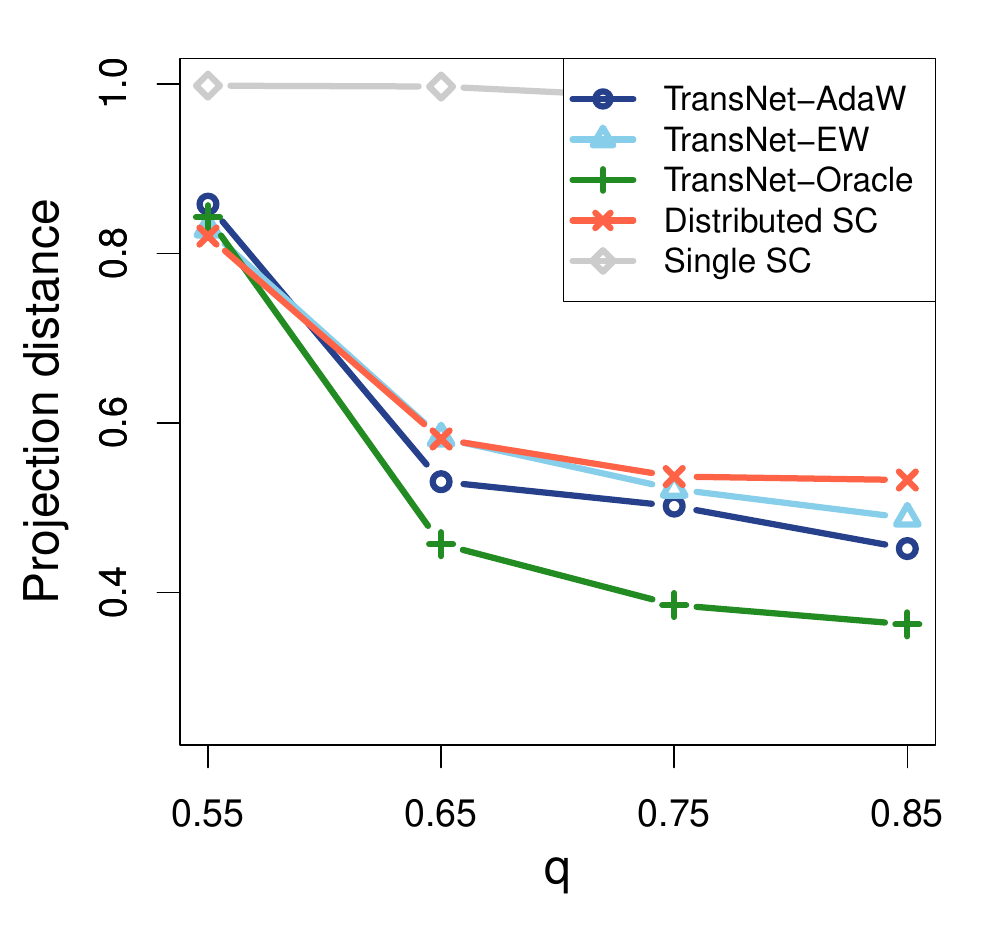}}
\caption{The projection distance of each method under Experiment VI.}\label{fig:expvi}
\end{figure*}

\paragraph{Experiment VI.} We let $q$ vary in $\{0.55, 0.65,0.75,0.85\}$ and set the privacy parameters $(q_S, q_{S^c})$ as $(q+0.2,q)$ when $q<0.85$. When $q=0.85$, $q_S$ is set to be 1. 
Therefore, by Lemma \ref{lem:dp} and equation \eqref{eq: qe}, the privacy budget is given by $\log\left(\frac{q}{1-q}\right)$, which takes values in $\{0.20, 0.62, 1.10, 1.73\}$ for the considered $q$. We consider three cases for the network sparsity parameter $\rho=1,1.5,2$. Figure \ref{fig:expvi} shows the average 
projection distance over 10 replications for different $\rho$ as $q$ increases. All four methods exhibit improved performance as $q$ increases, and the relative performance of all methods is consistent with that in Experiments IV and V. The proposed method exhibits satisfactory performance even when the privacy budget is as small as $0.62$. Consistent with the theory, as $\rho$ increases (i.e., the network becomes denser), the performance of all methods under small $q$ improves.

\subsubsection{Additional real data analysis}
\label{sec:politics}
\begin{figure*}[!h]{}
\centering
\subfigure[Work]{\includegraphics[height=4cm,width=4cm,angle=0]{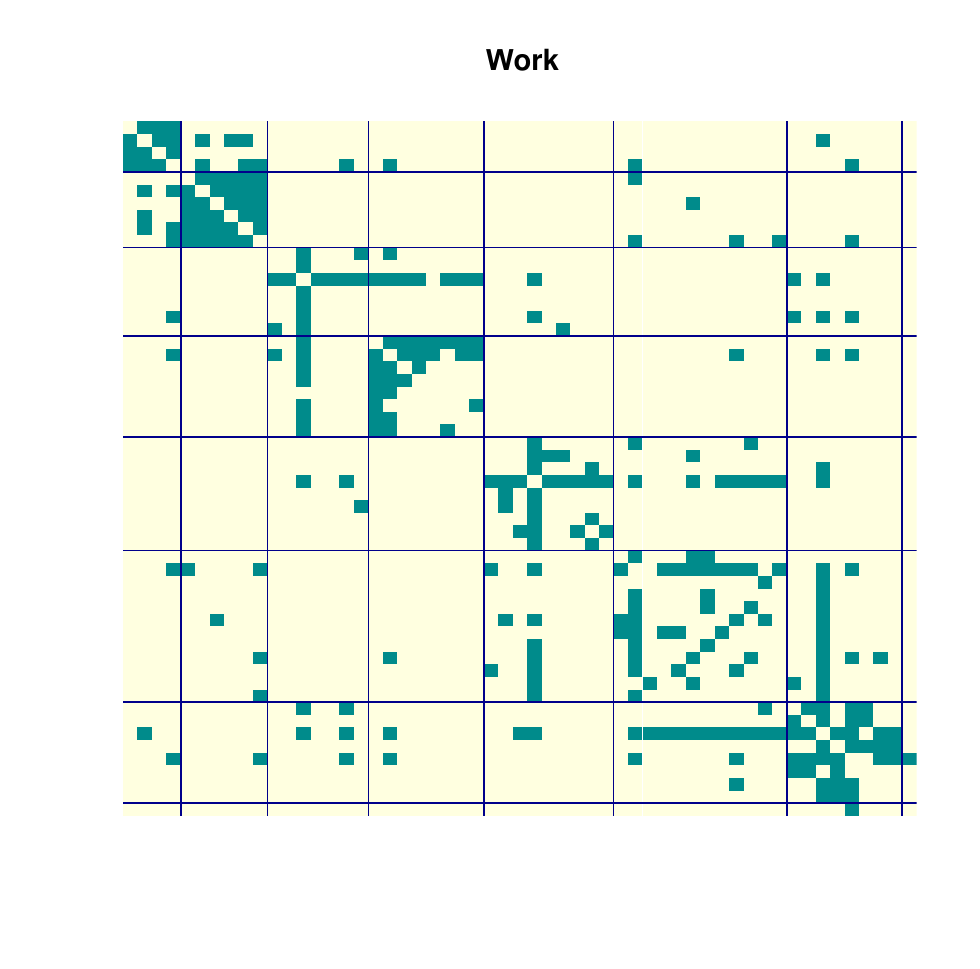}}
\subfigure[Facebook]{\includegraphics[height=4cm,width=4cm,angle=0]{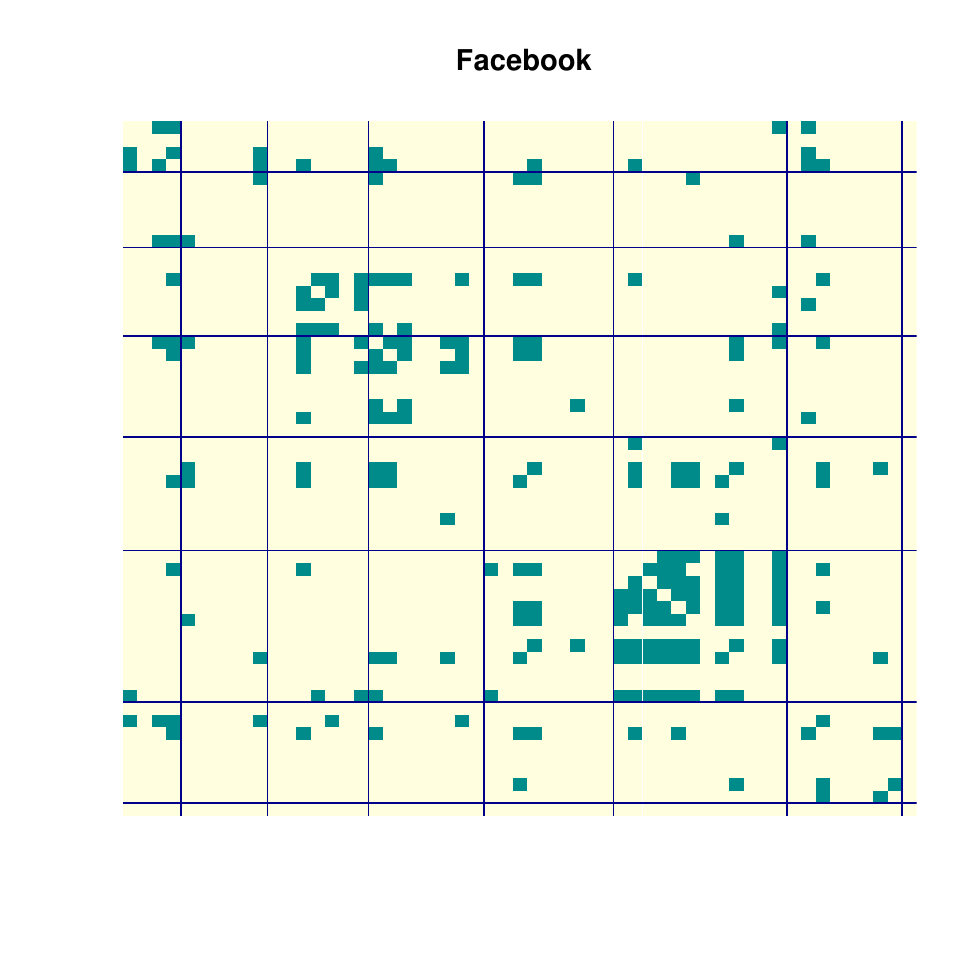}}
\subfigure[Leisure]{\includegraphics[height=4cm,width=4cm,angle=0]{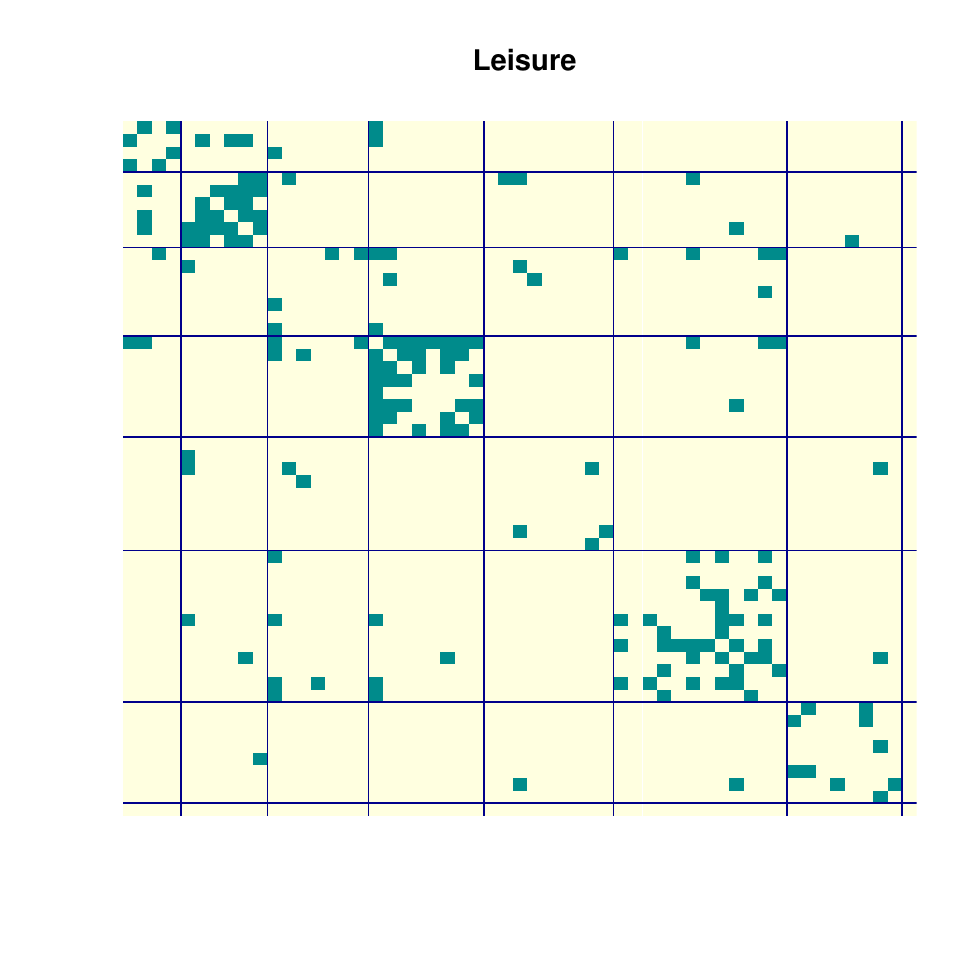}}
\subfigure[Lunch]{\includegraphics[height=4cm,width=4cm,angle=0]{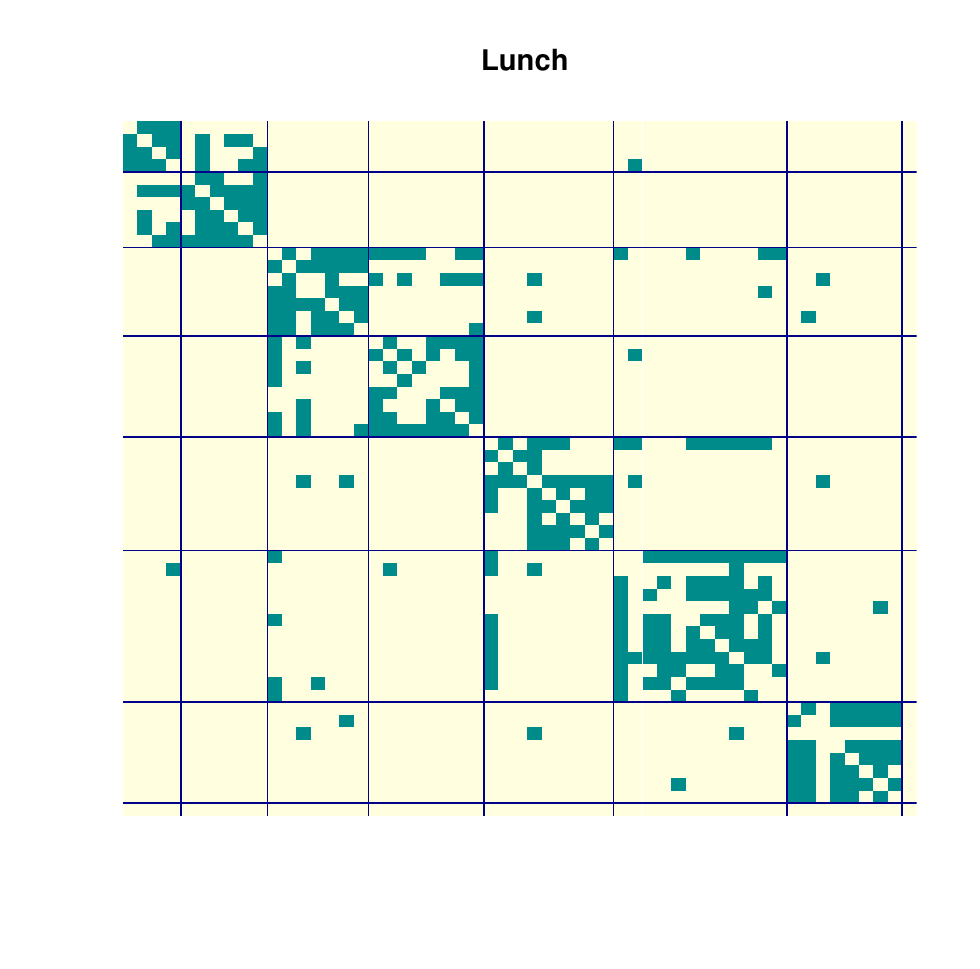}}
\subfigure[Coauthor]{\includegraphics[height=4cm,width=4cm,angle=0]{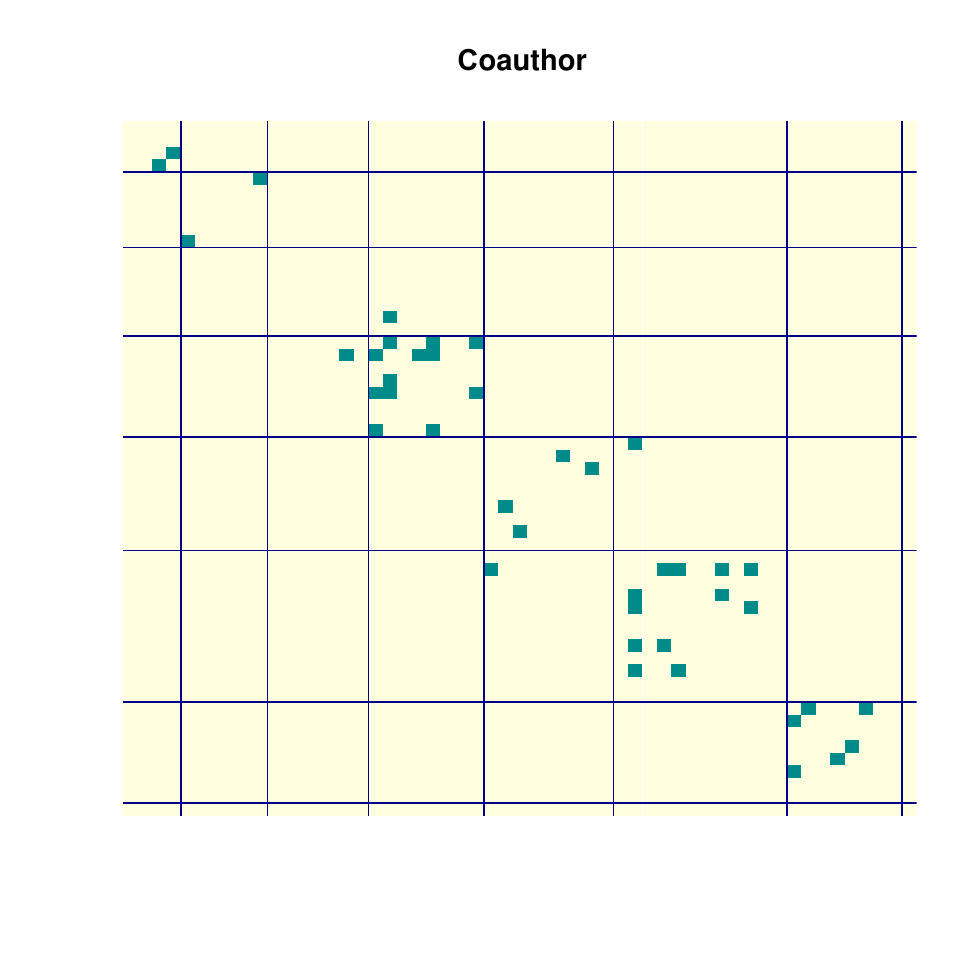}}
\caption{A visualization of the \texttt{AUCS} network data. Each sub-figure corresponds to one of the five relationships. The nodes 
are ordered according to underlying research groups.}\label{AUCS_vis}
\end{figure*}

First, we provide the visualization of the five networks of the \texttt{AUCS} data studied in the main paper in Figure \ref{AUCS_vis}.
The nodes in the same research group are ordered next to each other. 
It is obvious that the community structures of each network are heterogeneous. For example, for the lunch network, the nodes within the same research group are densely connected compared with those in distinct research groups. For the coauthor network, all the nodes are loosely connected, and in some research groups (say the third one), there are no connections among nodes.

\begin{figure*}[!h]{}
\centering
\subfigure[Follow]{\includegraphics[height=4.5cm,width=4.5cm,angle=0]{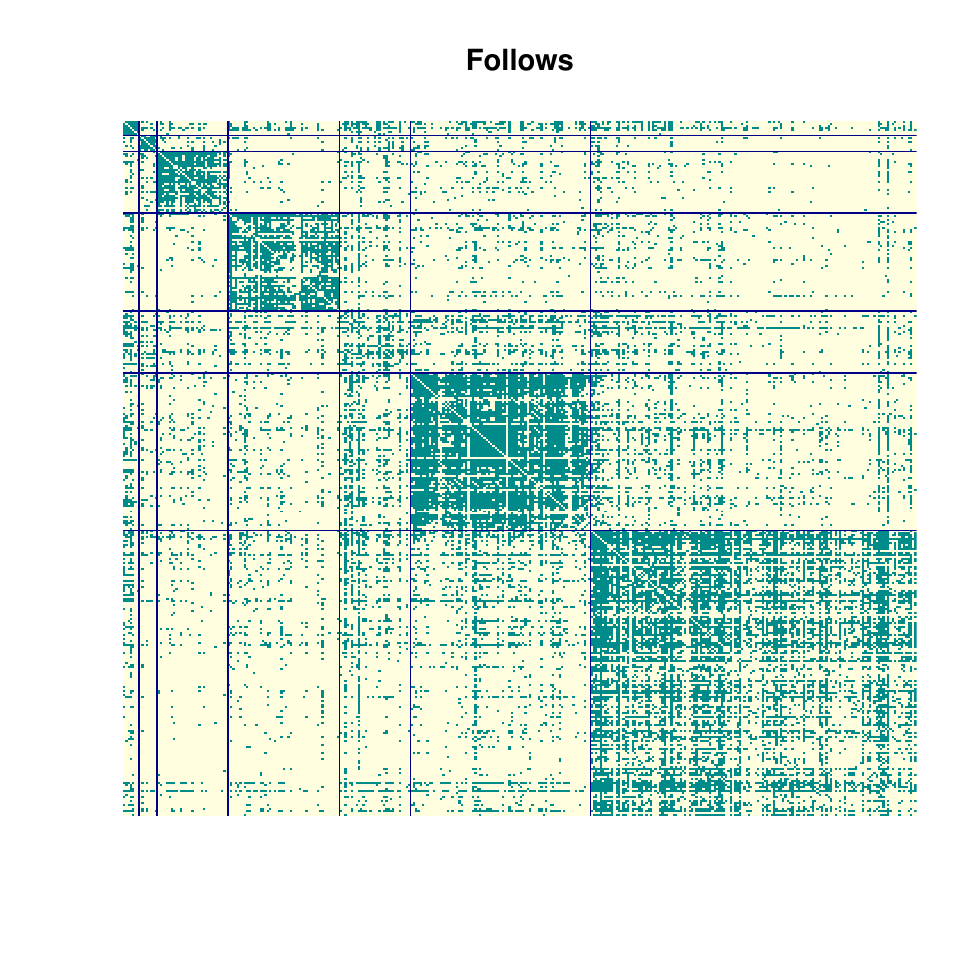}}
\subfigure[Mention]{\includegraphics[height=4.5cm,width=4.5cm,angle=0]{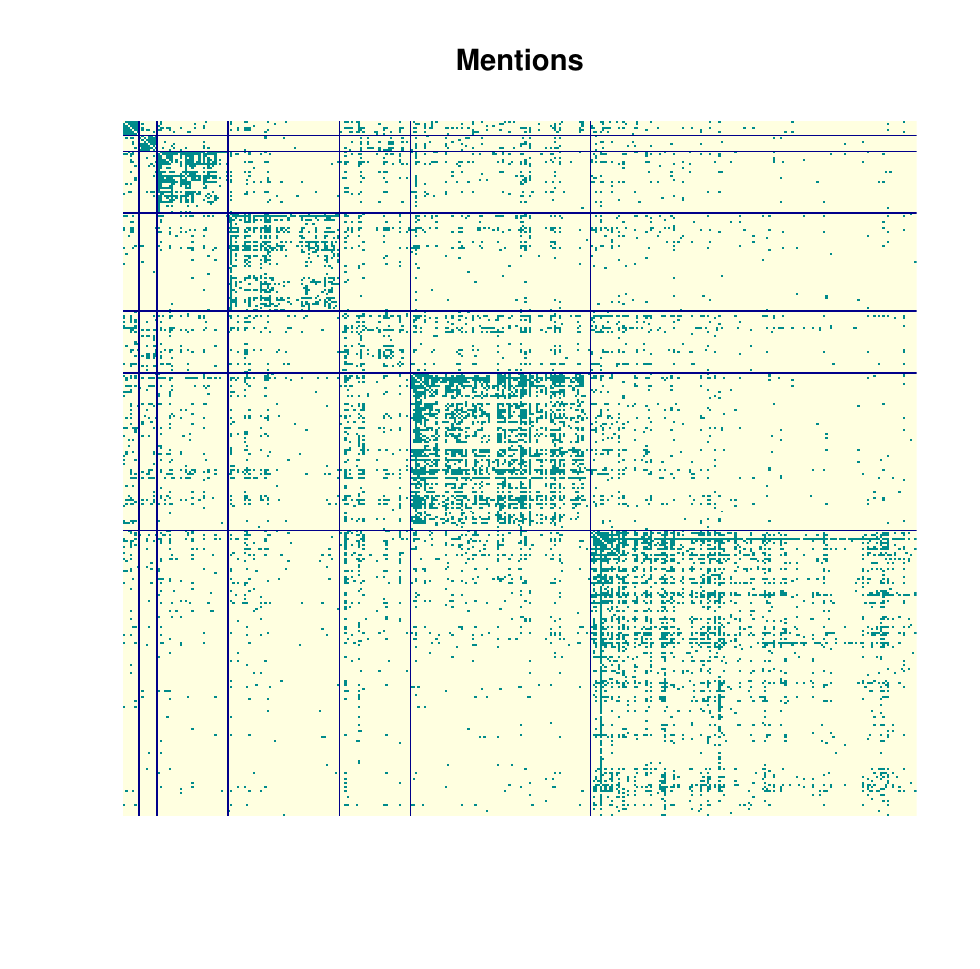}}
\subfigure[Retweet]{\includegraphics[height=4.5cm,width=4.5cm,angle=0]{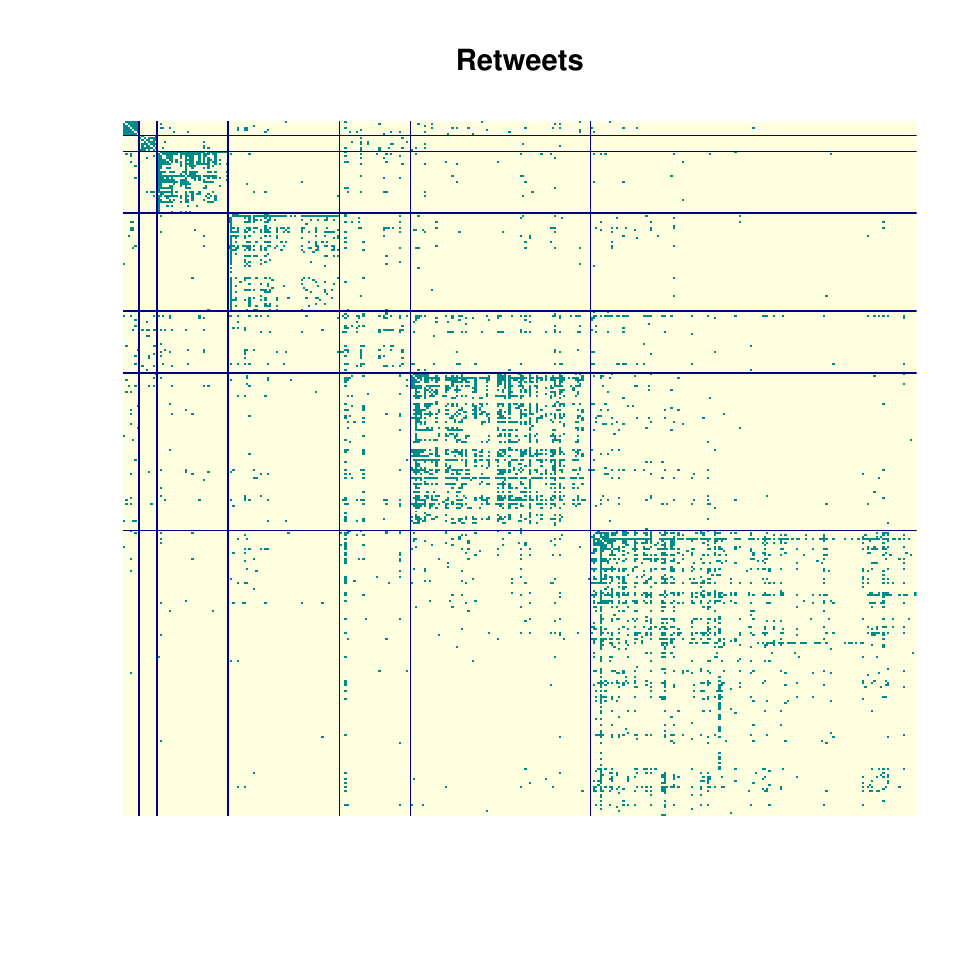}}
\caption{A visualization of the \texttt{Politics} network data. Each sub-figure corresponds to one of the three relationships. The nodes 
are ordered according to underlying affiliations.}\label{PB_vis}
\end{figure*}

Next, we analyze the \texttt{Politics} dataset \citep{greene2013producing} that includes the three relationships of 348 Twitter users including the Irish politicians and political organizations. The three relationships correspond to the users they Follow, Mention, and Retweet, respectively. The seven affiliations of these Twitter users are available. Figure \ref{PB_vis} provides a visualization of the three networks. The nodes in the same affiliation are ordered next to each other. These networks exhibit heterogeneous and varying strengths of community structures. For the Mention and Retweet networks, the signal strength of communities are weak compared with the Follow network.  

Similar to the \texttt{AUCS} data analyzed in the main paper, we regard one of three networks as the target network and the other two as source networks. Each network is alternately used as the target network. The affiliations of users are regarded as the true clustering of the target network. We vary the privacy parameters $q,q'(q'=q)$ of the target network and fix the privacy parameters of the source networks as: 0.95 (Follow), 0.8 (Mention), and 0.8 (Retweet). One can alternatively test other set-ups similarly.

\begin{figure*}[!h]{}
\centering
\subfigure[Follow]{\includegraphics[height=4.8cm,width=4.5cm,angle=0]{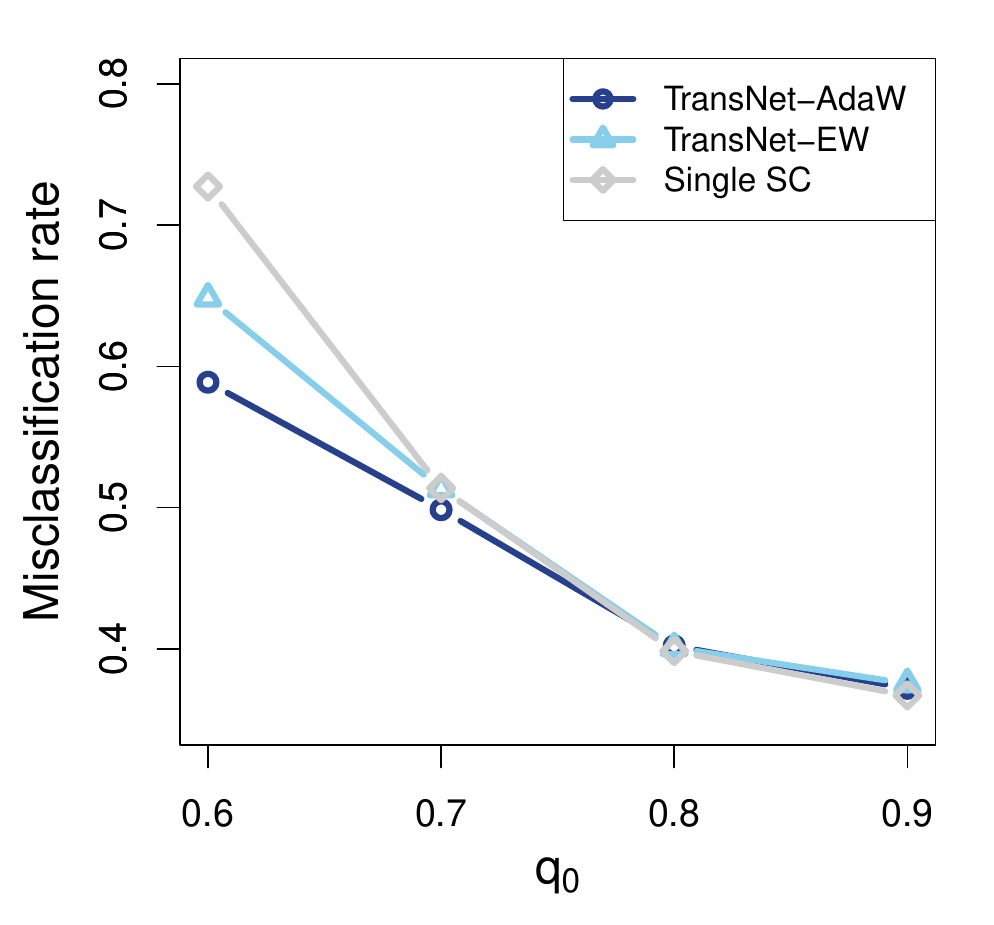}}
\subfigure[Mention]{\includegraphics[height=4.8cm,width=4.5cm,angle=0]{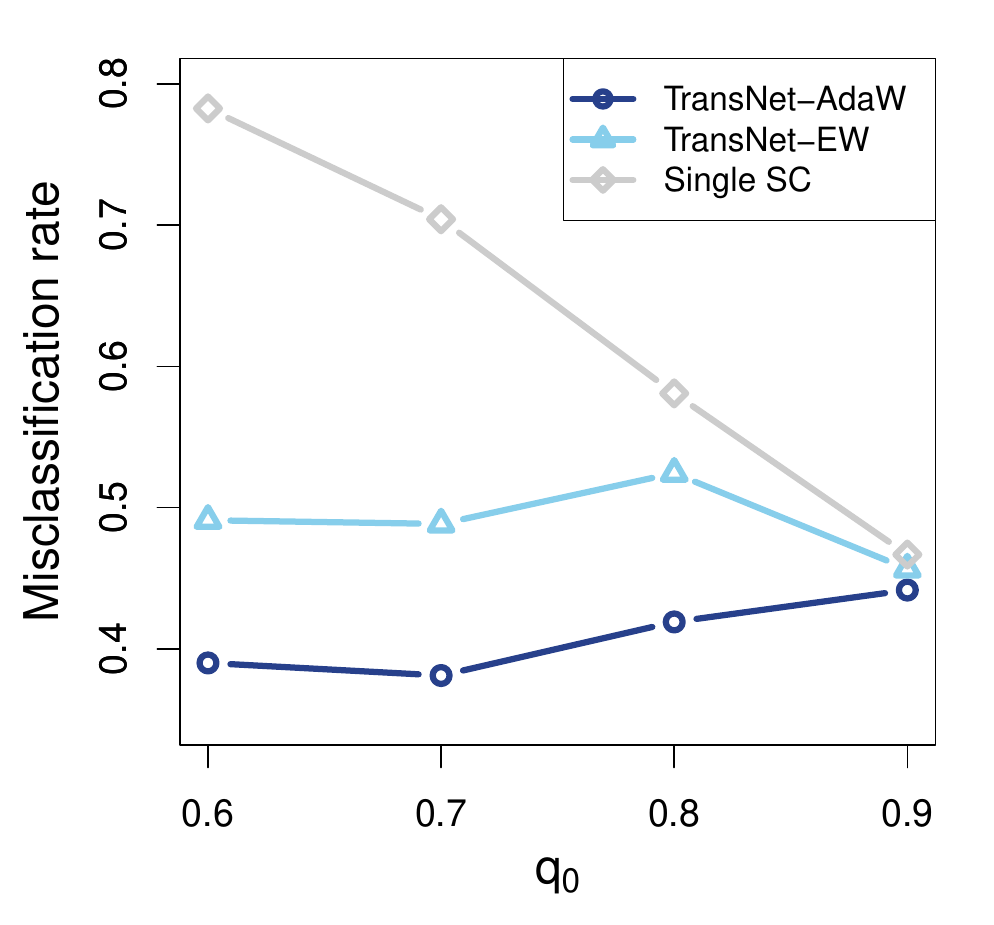}}
\subfigure[Retweet]{\includegraphics[height=4.8cm,width=4.5cm,angle=0]{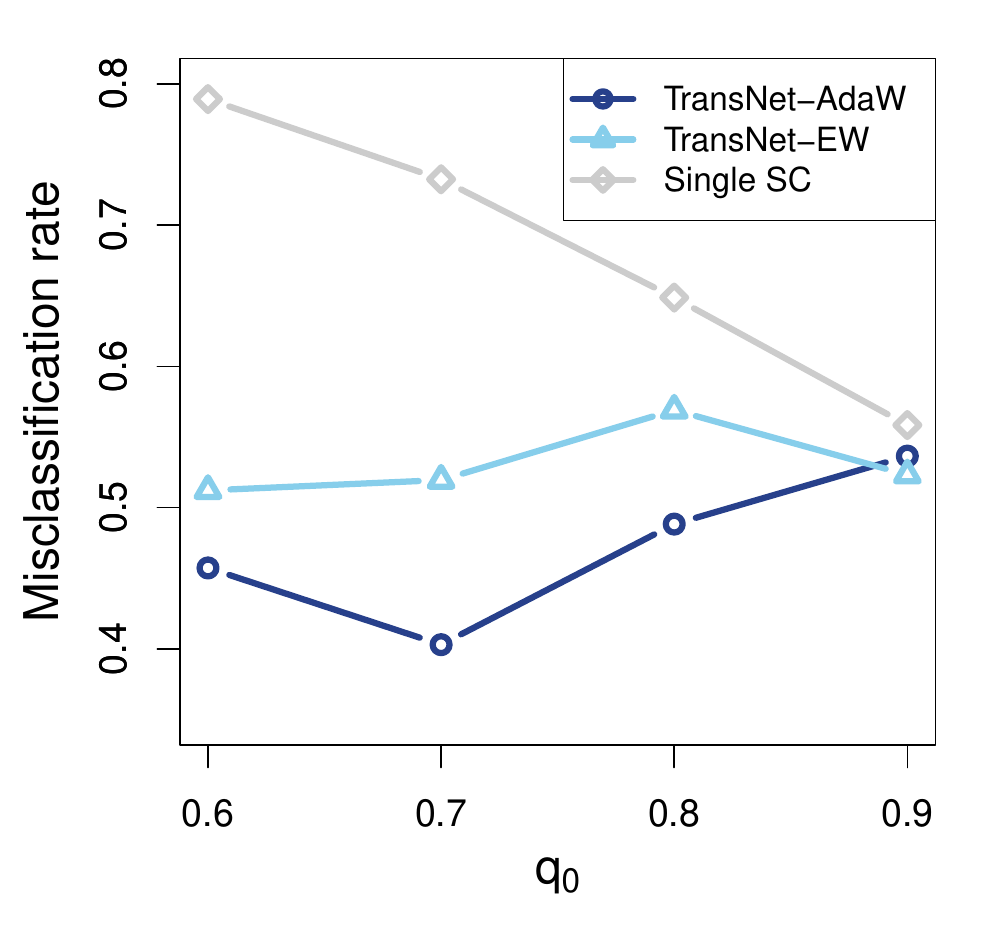}}
\caption{The misclassification rate of each method for the \texttt{Politics} network. (a)-(c) correspond to the scenario where Follow, Mention and Retweet are considered as the target network, with the remaining networks serving as source networks.}\label{PB_results}
\end{figure*}

The averaged results of the misclassification rates of three methods over 20 replications are given in Figure \ref{PB_vis}. Similar to the results of \texttt{AUCS} data, the transfer learning methods \texttt{TransNet-EW} and \texttt{TransNet-AdaW} yield accuracy gain in clustering compared to the \texttt{Single SC}, especially when the target network is less informative in clustering (see Figure \ref{PB_results} (b) and (c)). In addition,  \texttt{TransNet-AdaW} outperforms \texttt{TransNet-EW} when the source networks are heterogeneous and vary in quality for helping discover the underlying communities (see Figure \ref{PB_results} (b) and (c)).

%\vspace{-1cm}

\subsection{Extension}
\label{app:ext}
We have mainly considered the setup where each local data curator (local machine) is untrusted to the local data provider, thereby necessitating the input perturbation-based RR mechanism. In this section, we consider another typical set-up where the local data curators are trusted, and develop the output perturbation-based \emph{Gaussian mechanism} together with the transfer learning algorithm \texttt{TransNetX} and its theoretical guarantees.

The assumption of trusted data curators makes it theoretically preferable to perturb the summary statistics instead of raw data. For $l\in[L]$, denote the top-$K$ eigenvector of $A_l$ by $\tilde{U}_l$. We consider the following \emph{Gaussian mechanism} which adds noise on the projection matrix $\tilde{U}_l\tilde{U}_l^\intercal$,
\begin{equation}
\label{eq:gauour}
\mathcal Q_l({A}_l):= \tilde{U}_l \tilde{U}_l^\intercal+ N_l,\quad N_{l,ij}=N_{l,ji}\sim_{i.i.d.} \mathcal N(0, \sigma_l^2) \quad {\rm for}\; i\leq j,
\end{equation} 
where $\sigma_l$ depends on the privacy parameters $(\epsilon_l,\delta_l)$. 
Note that we use the projection matrix $\tilde{U}_l\tilde{U}_l^\intercal$ instead of the orthogonal transformed eigenspace in \eqref{eq:weiagg} of Algorithm \ref{alg:transnet} to better account for noise. We have the following results on the Gaussian mechanism.

{\color{black}
\begin{lemma}
\label{lem:cdp}
 Let
$\sigma_l=\frac{\Delta}{ \epsilon_l}\cdot\sqrt{2\log\left(\frac{1.25}{\delta_l}\right)},$
where $\epsilon_l>0$, $\delta_l\in [0,1)$ and $C>0$ are some constants, and $\Delta$ denotes the upper bound for the sensitivity and will be specified later. Then the following properties hold.
\begin{enumerate}[label=(\arabic*)]
\item  For each $l\in[L]$, $\mathcal Q_l(A_l)$ satisfies $(\epsilon_l,\delta_l)$-edge-DP whenever $$\max_{A_l,A'_l}\|\tilde{U}_l \tilde{U}_l^\intercal-\tilde{U}'_l (\tilde{U}'_l)^\intercal\|_F\leq \Delta,$$
where $\tilde{U}'_l$ denotes the top-$K$ eigenvector of the neighboring network $A'_l$.
In general, $$\Delta=\sqrt{2K}.$$
Moreover, suppose that $A_l$'s are generated from SBM, Assumption \ref{assu:sparse} holds, each population matrix $P_l$ satisfies $\lambda_{\min} (P_l)\gtrsim n\rho$ for all $i<j$ with $\rho\geq c\frac{\log n}{n}$ for some constant $c>0$,
each $B_l$ has full rank $K$, and the communities in both the source and target networks are balanced. Then, with high probability, 
$$\Delta=\frac{C\log^{1/2} n}{n^{3/2}\rho}$$ for some constant $C>0$, and thus $\mathcal Q_l(A_l)$ satisfies $(\epsilon_l,\delta_l)$-edge-DP with high probability.

\item (Parallel composition) Define 
$\overline{\mathcal{Q}}(\mathcal{A}):=(\mathcal Q_1(A_1),...,\mathcal Q_L(A_L))$, then
$\overline{\mathcal{Q}}(\mathcal{A})$ satisfies $(\max_{l\in[L]} \epsilon_l,\max_{l\in[L]} \delta_l)$-edge-DP {\color{black}with high probability.}
\end{enumerate}
\end{lemma}

}

\begin{remark}
Compared to Lemma \ref{lem:dp}, we here adopt a more general $(\epsilon_l,\delta_l)$-edge-DP \citep{dwork2014algorithmic} framework, which requires 
$$\mathbb P(\mathcal Q(A_l)=O_l|A_l)\leq {\rm e}^{\epsilon_l}\cdot\mathbb P(\mathcal Q(A'_l)=O_l|A'_l)+\delta_l$$
for any neighboring $A_l$ and $A'_l$. The parameter $\delta_l\in[0,1)$ is small and serves as a tolerance for the $\epsilon_l$-edge-DP in Definition \ref{edp}. See the formal definition of $(\epsilon_l,\delta_l)$-edge-DP in the Supplementary Materials.  
\end{remark}

% {\color{black}
% \begin{remark}
% The Gaussian mechanism is a standard approach to achieve
% $(\epsilon_l,\delta_l)$-edge-DP \citep{dwork2014algorithmic}. The utility of the Gaussian mechanism depends on the noise variance, which is determined by the sensitivity.
% Lemma \ref{lem:cdp} shows that, in general, the sensitivity $\max_{A_l,A'_l}\|\tilde{U}_l \tilde{U}_l^\intercal-\tilde{U}'_l (\tilde{U}'_l)^\intercal\|_F$ is upper bounded by $\sqrt{2K}$. Under the additional SBM assumptions on $A_l$ and some other additional assumptions, the sensitivity can be further reduced, yielding the sharper bound $O(\sqrt{\log n}/(n^{3/2}\rho))$. This sharper bound is obtained via using the tight form of Davis-Kahan theorem and the SBM assumption of $A_l$'s. In particular, the balanced community assumption is imposed to ensure eigenvector delocalization \citep{he2019local}.
% \end{remark}
% }

\begin{remark}
{\color{black}We note that a number of works in the existing literature on DP focus on worst-case privacy guarantees under minimal data assumptions. However, these worst-case bounds can be overly conservative in structured settings. In contrast, a growing body of work incorporates statistical models to obtain sharper utility guarantees. For example, \citet{cai2024optimal} study differentially private PCA under the spiked covariance model and derive high-probability sensitivity bounds by leveraging the underlying data-generating mechanism, {\color{black}which results in a high-probability privacy guarantee}. As discussed therein, such model-based analysis can substantially reduce the noise level compared to worst-case approaches. Similarly, \citet{hardt2013beyond} study DP singular vector estimation and show that worst-case bounds depending on the ambient dimension can be significantly improved by replacing the dimension with the matrix coherence, which is often much smaller both theoretically and empirically.
} 

{\color{black}In our setting, the utility of the Gaussian mechanism depends on the noise variance, which is determined by the sensitivity 
$\|\tilde{U}_l \tilde{U}_l^\intercal-\tilde{U}'_l (\tilde{U}'_l)^\intercal\|_F$, 
where $\tilde{U}'_l$ denotes the eigenvector corresponding to a neighboring network $A'_l$. As shown in Lemma \ref{lem:cdp}, the worst-case sensitivity is $\sqrt{2K}$, which leads to a non-vanishing noise level and may adversely affect the utility of the DP procedure.
Under additional assumptions (e.g., SBM and related regularity conditions), the sensitivity can be substantially reduced, yielding the sharper bound $O(\sqrt{\log n}/(n^{3/2}\rho))$. This improvement follows from a refined perturbation analysis (e.g., via the Davis--Kahan theorem) that exploits the underlying network structure. Such a refinement aligns naturally with our data-generating mechanism considered in our transfer learning framework.
}

\end{remark}

Based on the Gaussian mechanism, similar to Algorithm \ref{alg:transnet}, we develop the transfer learning algorithm \texttt{TransNetX} in this setting, summarized in Algorithm \ref{alg:transnet x}. {\color{black}To ensure tight error bound, we incorporate 
\begin{equation}
\label{eq:sigma}
\sigma_l=\frac{\Delta}{ \epsilon_l}\cdot\sqrt{2\log\left(\frac{1.25}{\delta_l}\right)}\quad {\rm with} \quad \Delta=\frac{C\log^{1/2} n}{n^{3/2}\rho}
\end{equation}
for the variance of Gaussian noise.}

\begin{remark}
Note that in Algorithm \ref{alg:transnet x}, we apply the RR mechanism with privacy budget $\epsilon_0$ to perturb the target network $A_0$ and then perform debiasing to obtain $\hat{A}_0$. The reason is as follows. The selection of the tuning parameter $\lambda$ in the regularization step requires estimating the connectivity matrix of the target network (see Section \ref{add:num}), for which using the debiased estimator $\hat{A}_0$ is more convenient than developing a separate privacy-preserving estimation procedure. Since $\hat{A}_0$ satisfies $(\epsilon_0,0)$-edge-DP, by the parallel composition property, $(\hat{A}_0, \overline{\mathcal{Q}}(\mathcal{A}))$ satisfies $(\max_{l\in[L]\cup\{0\}} \epsilon_l,\max_{l\in[L]} \delta_l)$-edge-DP.
\end{remark}
\begin{remark}
Also note that in addition to the privacy model and perturbation method, the alignment of the eigenspaces differs between Algorithm \ref{alg:transnet} and Algorithm \ref{alg:transnet x}. Computationally, the projection matrix used in Algorithm \ref{alg:transnet x} offers an alternative to the orthogonal transformation used in Algorithm \ref{alg:transnet}. However, theoretically, Algorithm \ref{alg:transnet x} requires each population matrix to be of rank $K$ to ensure the privacy analysis (see Lemma \ref{lem:cdp}) and to establish the statistical error bound for eigenspace under general weights (see Proposition \ref{theo: unifycdp}). This assumption is more stringent than the rank-$K$ assumption imposed on the weighted average of the population matrices (see Assumption \ref{assu:rank}) used in Proposition \ref{theo: step1}) for Algorithm \ref{alg:transnet}.
\end{remark}
\begin{algorithm}[!htbp]
\small

\renewcommand{\algorithmicrequire}{\textbf{Input:}}

\renewcommand\algorithmicensure {\textbf{Output:} }

\caption{\texttt{TransNetX}}

\label{alg:transnet x}

\begin{algorithmic}[1]

\STATE \textbf{Input:} Original source networks $\{{A}_l\}_{l=1}^L$, the bias-corrected target network $\hat{A}_0$, the number of clusters $K$, the Gaussian variance $\{\sigma_l^2\}_{l=0}^L$ defined in \eqref{eq:sigma}, and the tuning parameter $\lambda>0$.  \\
\STATE \,\underline{\textbf{Step 1 (Adaptive weighting):}}
\STATE \, \texttt{Eigen-decomposition:} Compute the $K$ leading eigenvectors $\tilde{U}_l$'s of ${A}_l$'s for $l\in [L]$, and Compute the $K$ leading eigenvectors $\hat{U}_0$'s of $\hat{A}_0$;  \\
\STATE\, \texttt{Weighted aggregation:} Using the adaptive weights defined later in \eqref{eq: adapweight2} to aggregate \\\;$\tilde{U}:=\sum_{l=1}^L \tilde{w}_l\left(\tilde{U}_l \tilde{U}_l^\intercal+ N_l\right)$ where $N_{l,ij}=N_{l,ji}\sim_{i.i.d.}~ \mathcal N(0, \sigma_l^2)$ for $i\leq j$, and obtain the \\\;top-$K$ eigenvectors $\bar{U}$ of $\tilde{U}$. \\

\STATE \underline{\textbf{Step 2 (Regularization):}}
\STATE \, Optimize and obtain ${\tilde{U}_0^{  RE}}(\lambda) \in \argmax_{V^\intercal V=I} {\rm tr}(V^\intercal {\left(\hat{U}_0\hat{U}_0^\intercal\right)}  V)-\frac{\lambda}{2} \|VV^\intercal- \bar{U}  \bar{U}^\intercal \|_F^2$.

\STATE \underline{\textbf{Step 3 (Clustering):}}
\STATE \, Conduct $k$-means clustering on the rows of ${\tilde{U}_0^{  RE}}(\lambda)$ to obtain $K$ clusters. \\
\STATE \textbf{Output:} The estimated eigenspace ${\tilde{U}_0^{  RE}}(\lambda)$ and the estimated $K$ clusters. 

\end{algorithmic}
\end{algorithm}

The following result provides the unified theory of the estimated eigenspace under general weights in Algorithm \ref{alg:transnet x}, which provides guidance for the selection of weights. 

\begin{proposition}  
\label{theo: unifycdp}
Suppose the assumptions in Lemma \ref{lem:cdp} all hold. Denote $\tau_n=\max\{1,\;(\rho\log n)^{1/2}\}$. If
\begin{equation}
\label{eq:conL2}
L \lesssim \min \{ n^{3/2},\,{n\rho}/{\tau_n^2} \},
\end{equation}
then we have with high probability that,
\begin{align}
\label{eq:bounduni2}
{\rm dist}(\bar{U},U_0) &\lesssim \sqrt{\frac{1}{n\rho}}\cdot\sqrt{\sum_{l=1}^L w_l^2} + \sqrt{\frac{1}{n\rho}}\cdot\sqrt{\sum_{l=1}^L w_l^2\cdot\frac{\log n\log(1.25/\delta_l)}{\epsilon_l^2n\rho}}
+ \sum_{l=1}^L w_l \mathcal{E}_{\theta,l}\\
&:=\tilde{\mathcal I}_1 + \tilde{\mathcal I}_2 + \tilde{\mathcal I}_3,\nonumber
\end{align}
provided that 
\vspace{-0.5cm}
\begin{equation}
\label{eq:conhet}
\tilde{\mathcal I}_2 + \tilde{\mathcal I}_3< 1/2-c_0\quad {\rm for\; some \;constant}\quad 0<c_0<1/2. 
\end{equation}
\end{proposition}

\begin{remark}
Condition \eqref{eq:conL2} is used to simplify the error bound. Condition \eqref{eq:conhet} is a technical condition for the von Neumann series expansion \citep{bhatia2013matrix} of $\bar{U}$;  it is mild as we notice that $\tilde{\mathcal I}_2$ is $o_p(1)$ when $\epsilon_l\gtrsim \sqrt{\log n\log (1.25/\delta_l)} /(n\rho)^{1-\alpha}$ for each $l\in[L]$ and some $\alpha>0$, and $\tilde{\mathcal I}_3\leq 2$ naturally.  
\end{remark}

\begin{remark}
Similar to Proposition \ref{theo: step1}, $\tilde{\mathcal I}_1$, $\tilde{\mathcal I}_2$ and $\tilde{\mathcal I}_3$ correspond to the statistical error without privacy and heterogeneity, the privacy cost and the heterogeneity cost, respectively. It is worth mentioning that the cost of privacy is proportional to $O_p(\frac{1}{\epsilon_l n\rho})$ (suppose $L$ is 1) compared to the statistical error $O_p(\frac{1}{\sqrt{ n\rho}})$ without privacy, which exhibits a similar pattern to the lower bounds derived for DP constrained high-dimensional linear regression~\citep{cai2021cost} and PCA \citep{cai2024optimal}. 
\end{remark}

Similar to Section \ref{sec:adative weighting}, we next define the informative networks and adaptive weighting strategy, followed by establishing the error-bound-oracle property. For simplicity, in what follows, we assume for $l\in [L]\cup \{0\}$ that
\begin{equation}
\label{con:epsiloncdp}
\frac{\epsilon_l}{\log ^{1/2}(1.25/\delta_l)}\geq \sqrt{\frac{\log n}{n\rho}}.
\end{equation}
The set of informative networks, $S$, is defined analogously to Definition~\ref{def:informative}. Specifically,  
\begin{align*}
S:=\left\{1\leq l\leq L:\;{\mathcal E}_{\theta,l}^2+{\mathcal{P}}_l^{'2}\leq {\eta}'_n\right\},
\;\;{\rm where}\; {\mathcal{P}}_l^{'}:=\sqrt{\frac{\log n\cdot \log(1.25/\delta_l)}{\epsilon_l^2n^2\rho^2L}} \; {\rm and}\;\eta'_n\asymp \frac{1}{n\rho},
\end{align*}
and $\mathcal E_{\theta,l}$ is defined in \eqref{eq:heterP}.
We also denote the cardinality of $S$ by $m$. The adaptive weights are defined as
\begin{equation}
\label{eq: adapweight2}
\tilde{w}_l\varpropto \left(\hat{\mathcal P}_{l}^{'2}+\hat{\mathcal E}_{\theta,l}^{'2}  + \frac{1}{n\hat{\rho}'_l}\right)^{-1} \varpropto \left(\hat{\mathcal E}_{\theta,l}^{'2} + \frac{1}{n\hat{\rho}'_l}\right)^{-1}\quad {\rm for}\quad l\in[L],
\end{equation}
where 
$\hat{\mathcal E}'_{\theta,l}:=\|\tilde{U}_l\tilde{U}_l^\intercal +N_l-(\tilde{U}_0\tilde{U}_0^\intercal +N_0)\|_2$,
$\hat{\mathcal P}'_{l}$ is the estimator of ${\mathcal P}'_{l}$ with $\rho$ replaced by $\hat{\rho}'_l$, and $\hat{\rho}'_l$ is a privacy-preserved consistent estimator of $\rho$ under privacy budget $(\epsilon_{\rho},\delta_{\rho})$.
For simplicity, we can apply the Gaussian mechanism (see Lemma \ref{lem:senandgau}) to  $\sum_{ij}A_{l,ij}/(n(n-1))$ to obtain $\hat{\rho}'_l$.
As a result, by the sequential composition theorem of DP \citep{dwork2014algorithmic} and Lemma \ref{lem:cdp}, Algorithm \ref{alg:transnet x} with adaptive weights defined in \eqref{eq: adapweight2} and the $\rho$ in the Gaussian variance replaced by $\hat{\rho}'_l$, satisfies $(\epsilon_{\rho}+\max_{l\in [L]\cup\{0\}} \epsilon_l ,\; \delta_{\rho}+\max_{l\in [L]} \delta_l )$. 

Similar to Theorem \ref{theo: effectofadaptive} and \ref{theo: effectofadaptivesecond}, we have the following the results.

\begin{theorem}
\label{theo:ebo2}
Suppose the assumptions in Proposition \ref{theo: unifycdp} all hold. If
\begin{equation}
\label{eq:con21}
\frac{L-m}{m}\lesssim \frac{ \min_{l\in S^c} {\mathcal E}_{\theta,l}^2}{\eta'_nL},
\end{equation}
then with high probability, the adaptive weights $\tilde{w}_l$'s defined in \eqref{eq: adapweight2} satisfies $\tilde{w}_l\asymp 1/m$ for $l\in S$ and 
\begin{equation}
\label{eq:bound2}
{\rm dist}(\bar{U},U_0) \lesssim \sqrt{\frac{1}{mn\rho}} + \sqrt{\frac{1}{n\rho}}\cdot\sqrt{\sum_{l\in S}\frac{\log n\log(1.25/\delta_l)}{m^2\epsilon_l^2n\rho}}
+ \sum_{l\in S} \frac{\mathcal{E}_{\theta,l}}{m}\lesssim \sqrt{\frac{1}{mn\rho}}
+ \sum_{l\in S} \frac{\mathcal{E}_{\theta,l}}{m}.
\end{equation}
If we additionally have
\begin{equation}
\label{eq:con22}
 \frac{ \min_{l\in S^c} {\mathcal E}_{\theta,l}^2 }{\eta'_n}\cdot (L-m)^2=w\left(\frac{L^3}{m}\right),
\end{equation}
then with high probability, the error bound of $\bar{U}$ under the adaptive weights $\tilde{w}_l$'s (i.e., \eqref{eq:bound2}) is of smaller order than that under the equal weights (i.e., \eqref{eq:bounduni2} with $w_l=1/L$ for all $l\in[L]$).
\end{theorem}

The results in Theorem \ref{theo:ebo2} can be similarly discussed as those in Theorem \ref{theo: effectofadaptive} and \ref{theo: effectofadaptivesecond}. In addition, based on the error bound for the aggregated eigenspace, it is easy to derive the error bound after regularization (Step 2) and clustering (Step 3). These are therefore omitted for brevity.

\paragraph{Comparison between \texttt{TransNet} and \texttt{TransNetX}.} The output perturbed Gaussian-based \texttt{TransNetX} and the input perturbed RR-based \texttt{TransNet} can be compared from the following aspects. 
\begin{itemize}
    \item \textbf{Privacy-utility trade-off:}
The \texttt{TransNetX} incurs smaller accuracy loss due to privacy since $\epsilon_l$ affects only the second-order term in ${\rm dist}(\bar U,U_0)$, whereas the \texttt{TransNet} adds noise to edges, {which affects both the first and second-order terms in ${\rm dist}(\bar U,U_0)$, though its effect can be alleviated by the adaptive weighting strategy.}
\item \textbf{Computation:}
The \texttt{TransNet} is computationally lighter, as it directly aligns eigenvectors (with computational complexity proportional to $n$) instead of constructing the projection matrices (with computational complexity proportional to $n^2$).
\item \textbf{Practicality:} RR in \texttt{TransNet} involves no unknown parameters, whereas the Gaussian mechanism in \texttt{TransNetX} depends on the sparsity parameter $\rho$ and the sensitivity constant $C$, both of which may require separate, privacy-consuming estimation and careful selection.
\end{itemize}
\subsubsection{Numerical experiments on \texttt{TransNetX}}
We evaluate the performance of the proposed output perturbed Gaussian-based transfer learning method, denoted by \texttt{TransNetX-AdaW}. Its equal weights counterpart is denoted by \texttt{TransNetX-EW}. For comparison, we also include the input perturbed RR-based methods \texttt{TransNet-AdaW} and \texttt{TransNet-EW}.

The network generation process follows that of Experiment V in the main paper, and we thus highlight only the differences. Specifically, we fix $L=12,K=3$ and let the sample size $n$ vary. The $L$ source networks are classified into two equal sized groups. The parameters in the same groups are identical. 
The heterogeneity parameters $(\mu^{(1)},\mu^{(2)})=(0.02,0.3)$. The privacy parameters $(\epsilon^{(1)},\epsilon^{(2)})=(1,0.6)$ for four methods. The parameters $(\delta^{(1)},\delta^{(2)})=(0.2,0.2)$ for methods \texttt{TransNetX-AdaW} and \texttt{TransNetX-EW}. For the target network, the privacy parameter $\epsilon$ is 1. 
The connectivity matrices are identical across the source and target networks and are the same as those in Experiment V.

We note that the Gaussian mechanism applied to the source networks in \texttt{TransNetX-AdaW} and \texttt{TransNetX-EW} involves two unknown parameters, i.e., the network sparsity $\rho$ and the multiplicative constant $C$ in the variance of Gaussian noise. For the network sparsity $\rho$, we consider two set-ups. In the first set-up, we use population $\rho$. For a fair comparison, the $\hat{\rho}_l$ used in the adaptive weights of \texttt{TransNet-AdaW} is replaced with the population $\rho$. In the second setup, we instead use the estimated network sparsity for each network. Note that in this DP regime, the estimation of $\rho$ cause additional privacy loss. By the sequential parallel composition property of DP \citep{dwork2014algorithmic}, we therefore decompose the privacy budget into two parts. The first part, $(\tfrac{1}{3}\epsilon^{(1)}, \tfrac{1}{3}\epsilon^{(2)})$, is used to estimate $\rho$. 
The second part, $(\tfrac{2}{3}\epsilon^{(1)}, \tfrac{2}{3}\epsilon^{(2)})$, is allocated to the remaining steps in \texttt{TransNetX}. For $\delta$, we allocate it evenly between the two parts, i.e., $(\tfrac{1}{2}\delta^{(1)}, \tfrac{1}{2}\delta^{(2)})$ for each source network. For the target network, recall that we applied the RR mechanism, so the allocation of privacy budget is not involved.
Using the specified privacy parameters, we apply the Gaussian mechanism (see Lemma \ref{lem:senandgau}) to the $\sum_{ij}A_{l,ij}/(n(n-1))$ to estimate the network sparsity for each source and target network. The multiplicative constant $C$ arises from the sharp analysis used to bound the sensitivity of the projection matrices $U_lU_l^\intercal$. Through numerical evaluation, we find that the sensitivity constant $C$ generally ranges between 2 and 10. Accordingly, we evaluate cases where $C \in \{4, 6, 8\}$.

\begin{figure*}[!h]{}
\centering
\subfigure[Population $\rho$, $C=4$]{\includegraphics[height=4.3cm,width=4.5cm,angle=0]{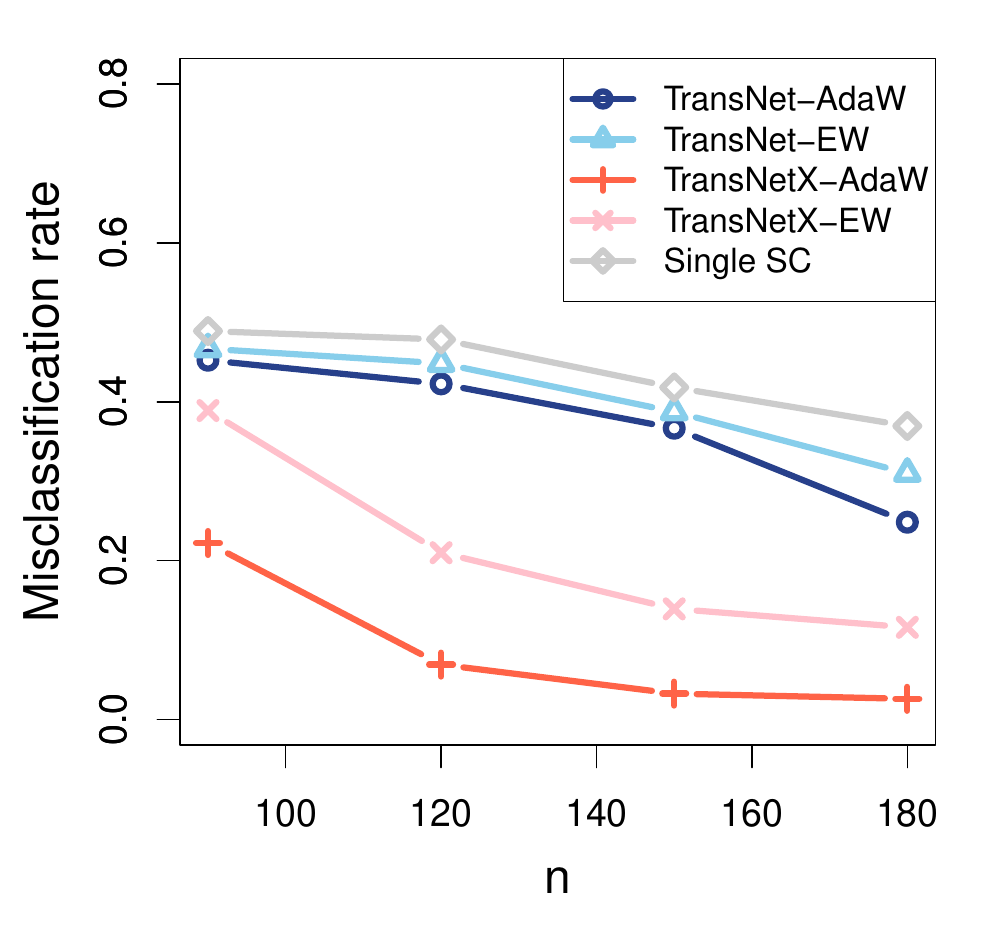}}
\subfigure[Population $\rho$, $C=6$]{\includegraphics[height=4.3cm,width=4.5cm,angle=0]{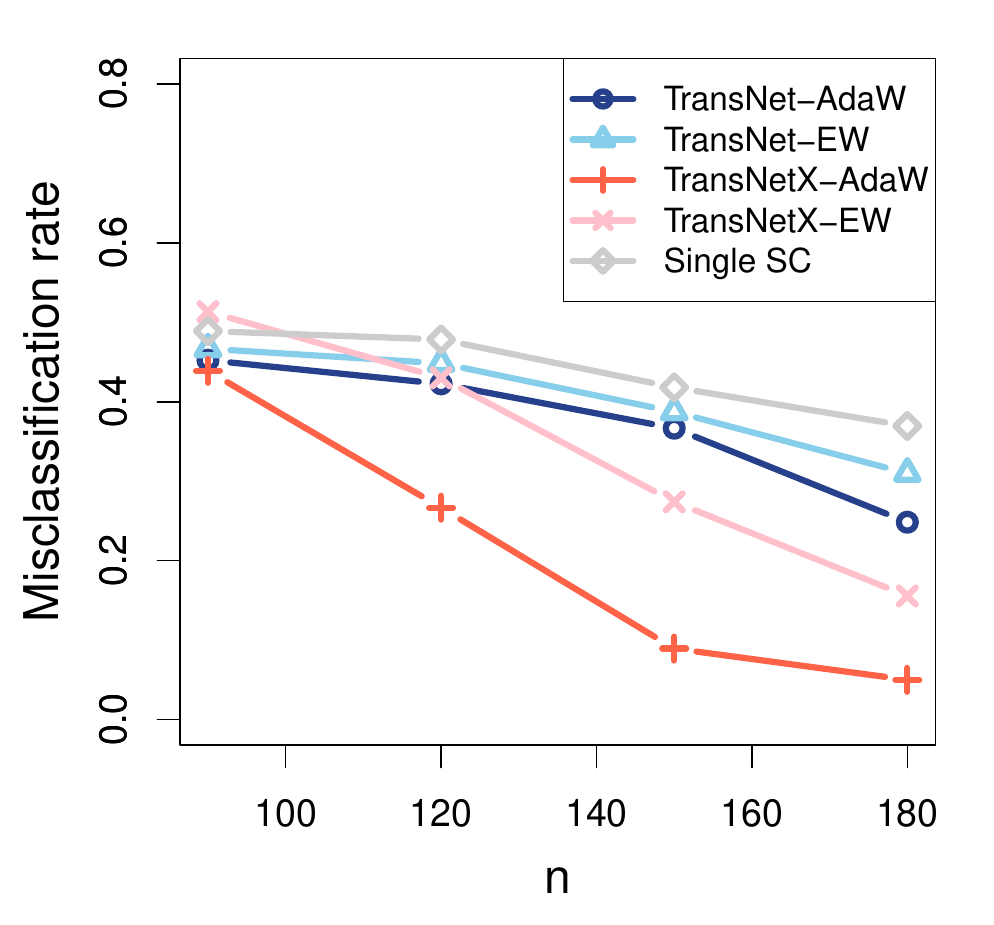}}
\subfigure[Population $\rho$, $C=8$]{\includegraphics[height=4.3cm,width=4.5cm,angle=0]{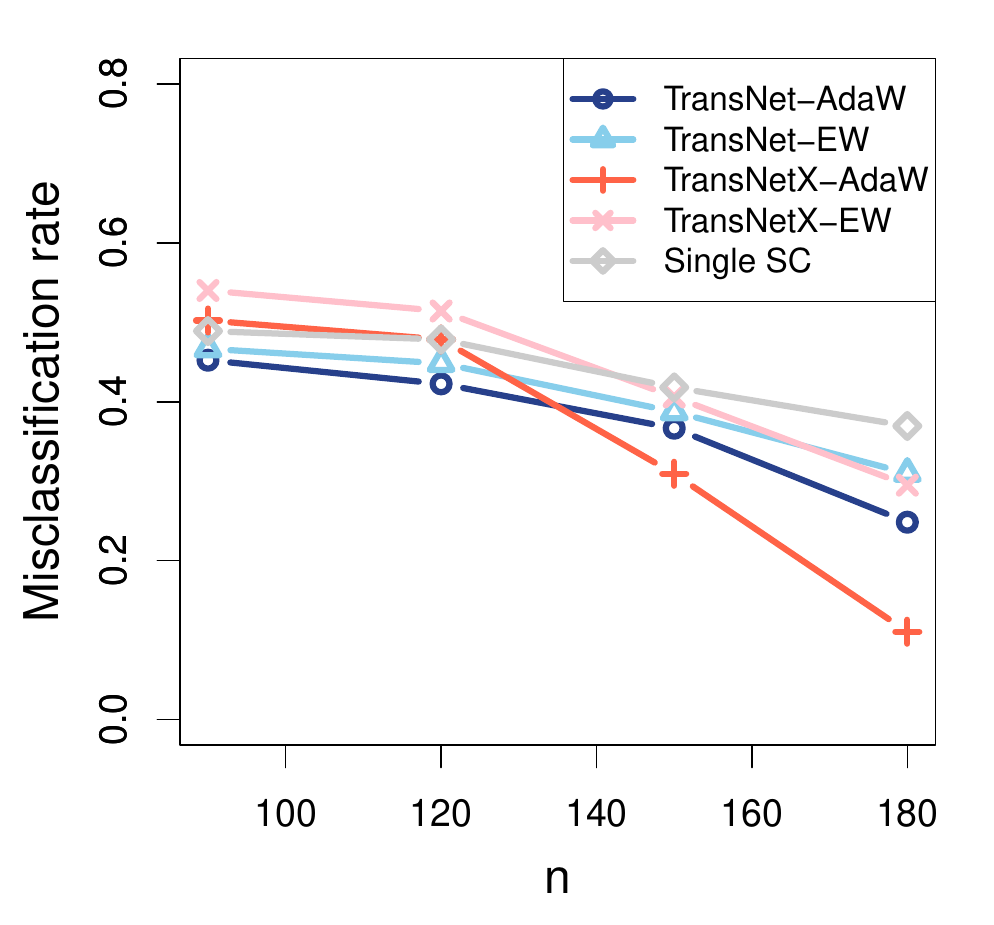}}
\subfigure[Estimated $\rho$, $C=4$]{\includegraphics[height=4.3cm,width=4.5cm,angle=0]{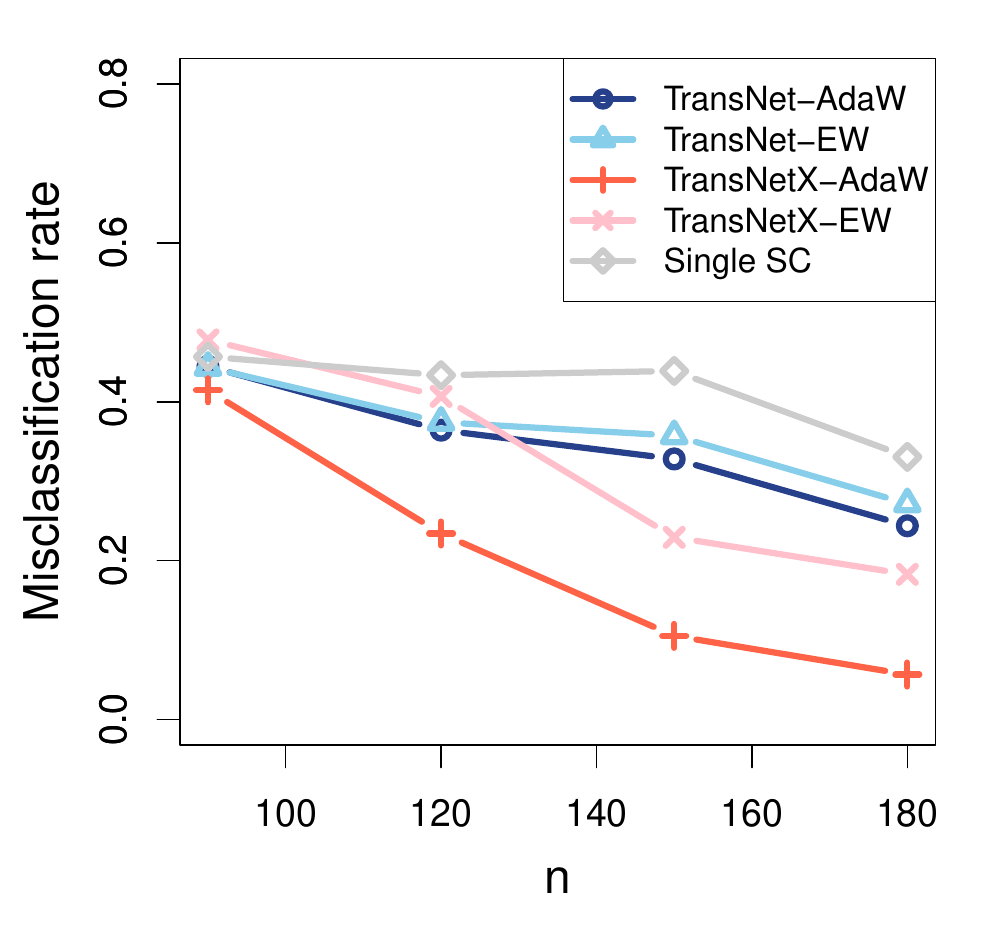}}
\subfigure[Estimated $\rho$, $C=6$]{\includegraphics[height=4.3cm,width=4.5cm,angle=0]{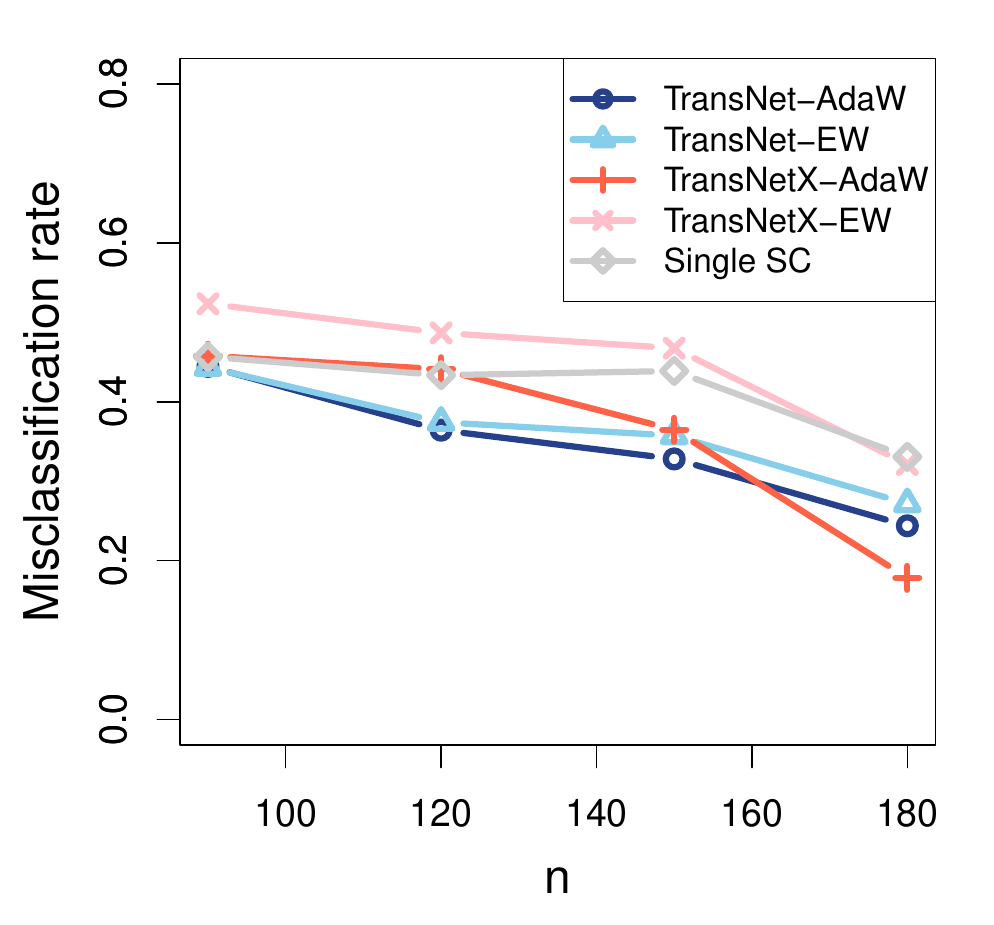}}
\subfigure[Estimated $\rho$, $C=8$]{\includegraphics[height=4.3cm,width=4.5cm,angle=0]{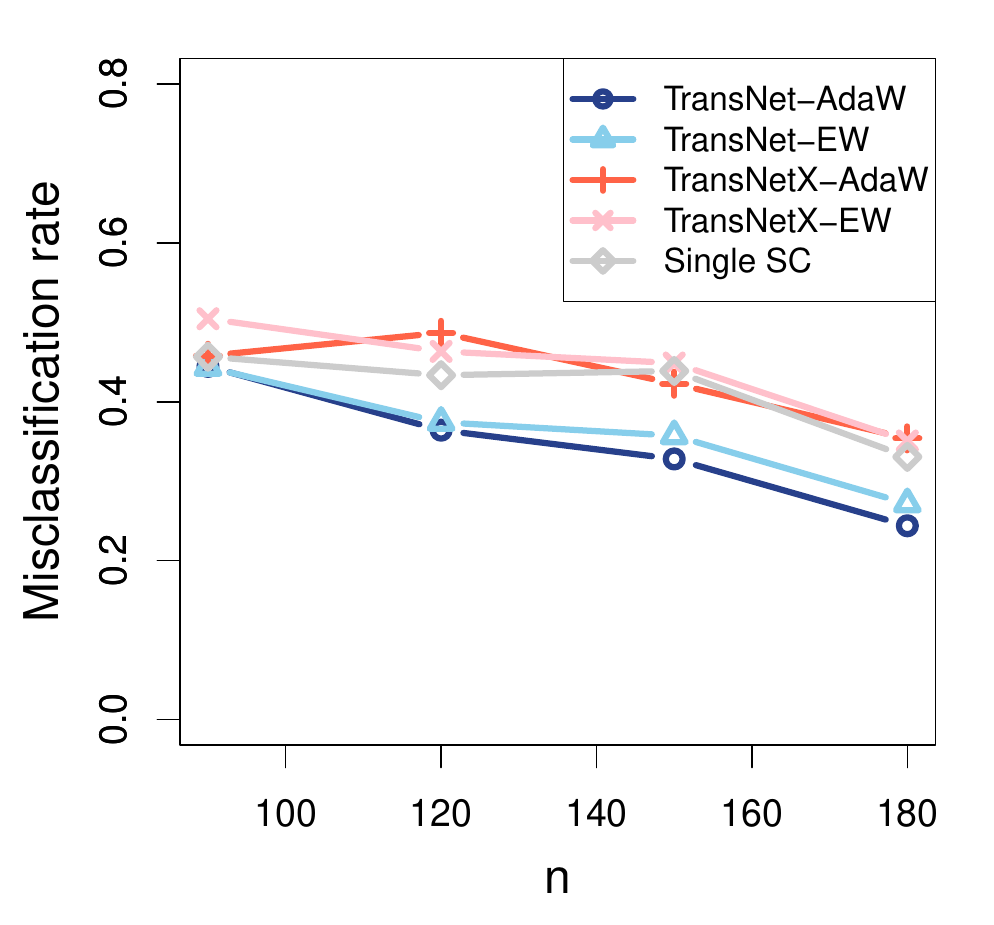}}
\caption{The misclassification rate of four methods using both the population and estimated $\rho$, under different values of $C$.}\label{fig:cdp}
\end{figure*}

Figure \ref{fig:cdp} presents the average misclassification rates of the four methods over 10 replications using both the population and estimated $\rho$, under different values of $C$. The results show that  \texttt{TransNetX-AdaW} and \texttt{TransNetX-EW} are highly sensitive to $C$. When $C$ is small, e.g., $C=4$, they perform better than  \texttt{TransNet-AdaW} and \texttt{TransNet-EW}. When $C$ is large, e.g., $C=8$, the performance of the output perturbed Gaussian-based methods deteriorates. In addition, the accuracy of the network sparsity is important. Using the population $\rho$, the output perturbed Gaussian-based methods outperform the input perturbed RR-based methods across a wider range of $C$ values than using the estimated $\rho$.

\subsection{{Proofs and supplementary materials  on \texttt{TransNetX}}}
\label{app:cdp}
This section is self-contained which includes all the preliminaries, lemmas and proofs involved in Section \ref{app:ext} of the Supplementary Materials.  We first introduce the preliminaries of $(\epsilon,\delta)$-edge-DP and Gaussian mechanism \citep{dwork2014algorithmic}, followed by the privacy analysis of \texttt{TransNetX}. We then present technical lemmas and proofs related to the statistical analysis of \texttt{TransNetX.} 

\subsubsection{Preliminaries of $(\epsilon,\delta)$-edge-DP and proof of Lemma \ref{lem:cdp}}

The next definition introduces the $(\epsilon,\delta)$-edge-DP \citep{dwork2014algorithmic} for a single network.

\begin{definition}[$(\epsilon,\delta)$-edge-DP]
\label{edeltadp}
Let $A\in \{0,1\}^{n\times n}$ be a network of interest. For any range $\mathcal R$, let $\mathcal Q(A)\in \mathcal R$ be a randomized mechanism that generates some synthetic information from $A$, indicated by a known parameter $\omega$. Given the privacy parameters $\epsilon>0$ and $\delta\in[0,1)$, $\mathcal Q(A)$ satisfies $(\epsilon,\delta)$-edge-DP if
\begin{equation}
\label{dpdeltadef}
\mathbb P_{\omega}(\mathcal Q(A)=O\,| \,A)\leq {\rm e}^\epsilon\cdot\mathbb P_{\omega}(\mathcal Q(A')=O \,|\,A')+\delta
\end{equation}
for any $O\in \mathcal R$ and any neighboring graphs $A$ and $A'$ which differ by only one edge.
\end{definition}

The $(\epsilon,\delta)$-edge-DP can be achieved by the following Gaussian mechanism \citep{dwork2014algorithmic}.

\begin{lemma}[Sensitivity and Gaussian mechanism]
\label{lem:senandgau}
The $l_2$-sensitivity of a function $f$ that maps $A$ into $\mathbb R^{n\times n}$ is defined as
\begin{equation*}
\Delta_f:= \underset{{\rm neighboring} (A,A')}{\rm sup} \|f(A)-f(A')\|_F,
\end{equation*}
where $A$ and $A'$ are called neighboring if they differ by only one edge. Then, for any $\epsilon>0$ and $\delta\in[0,1)$, the mechanism 
\begin{equation*}
\mathcal Q(A)=f(A)+N \quad {\rm with} \quad N\sim_{i.i.d.} \mathcal N(0,\sigma^2), 
\end{equation*} 
satisfies $(\epsilon,\delta)$-edge-DP if $\sigma\geq \frac{\Delta_f}{ \epsilon}\cdot\sqrt{2\log\left(\frac{1.25}{\delta}\right)}$.
\end{lemma}

We next use Lemma \ref{lem:senandgau} to prove Lemma \ref{lem:cdp}.

\subsubsection*{Proof of Lemma \ref{lem:cdp}}
We first prove the property (1).
The first part result follows directly from the fact that $\|\tilde{U}_l\tilde{U}^\intercal_l - \tilde{U}'_l(\tilde{U}'_l)^\intercal\|_F\leq \sqrt{2K}$. Now we show the second part result of (1).
Since this property holds for each $l$, we drop the subscript $l$ during the proof for notational simplicity. Let $A$ and $A'$ be two neighboring graphs and let $\tilde{U}\in\mathbb R^{n\times K}$ and $\tilde{U}'\in\mathbb R^{n\times K}$ be their corresponding eigenvectors, respectively. By Lemma \ref{lem:senandgau}, we only need to bound the sensitivity $\|\tilde{U}\tilde{U}^\intercal - \tilde{U}'(\tilde{U}')^\intercal\|_F$

By the Davis-Kahan theorem (see Lemma \ref{pro:DK}), we have 
\begin{equation}
\label{eq:dav}
\|\tilde{U}\tilde{U}^\intercal - \tilde{U}'(\tilde{U}')^\intercal\|_F\leq \frac{\sqrt{2} \|(A-A')\tilde{U}\|_F}{\delta},
\end{equation}
where $\delta=\min_{1\leq i\leq K,K+1\leq j\leq n}|\lambda_i(A)-\lambda_j(A')|$. Next, we derive the lower bound of $\delta$ and upper bound of $\|(A-A')\tilde{U}\|_F$, respectively. 

For $\delta$, we have
\begin{equation}
\label{eq:delta}
\delta\geq \min_{1\leq i\leq K,K+1\leq j\leq n}(\lambda_i(A)-\lambda_j(A')) \overset{(i)}{\geq} \min_{1\leq i\leq K,K+1\leq j\leq n}(\lambda_i(A)-\lambda_j(A)-1)
\overset{(ii)}{\gtrsim} n\rho.
\end{equation}
The inequality $(i)$ comes from the Weyl's inequality that 
\begin{equation*}
|\lambda_j(A')-\lambda_j(A)|\leq \|A-A'\|_2=1.
\end{equation*}
The inequality $(ii)$ comes from the following facts
\begin{align*}
&\min_{1\leq i\leq K,K+1\leq j\leq n}(\lambda_i(A)-\lambda_j(A))-1\\
=&\min_{1\leq i\leq K,K+1\leq j\leq n}\left(\lambda_i(P)-\lambda_j(P)-(\lambda_j(A)-\lambda_j(P)+\lambda_i(P)-\lambda_i(A))\right)-1\\
\overset{(a)}{\geq}& \min_{1\leq i\leq K,K+1\leq j\leq n}\left(\lambda_i(P)-\lambda_j(P)\right)-2\|A-P\|_2-1\overset{(b)}{\gtrsim} n\rho-\sqrt{n\rho}-1 \overset{(c)}\asymp n\rho, 
\end{align*}
where $P$ denotes the population of $A$, $(a)$ follows from the Weyl's inequality, $(b)$ follows from the assumption that $P$'s rank is $K$ and
$\lambda_{\min}(P)\gtrsim n\rho$, and $(c)$ is indicated by the assumption that $\rho\geq \frac{C\log n}{n}$.

For $\|(A-A')\tilde{U}\|_F$, denote the top-$K$ eigenvector of $P$ by $U$, we then observe that for any orthogonal matrix $O\in \mathbb R^{K\times K}$ determined later,  
\begin{align*}
\|(A-A')\tilde{U}\|_F&=\|(A-A')(\tilde{U}-UO+UO)\|_F\\
&\leq \|(A-A')UO\|_F + \|(A-A')(\tilde{U}-UO)\|_F:=I+II.
\end{align*}
For $I$, assuming that $A$ and $A'$ differs only by (1,2) and (2,1) entries, we have 
\begin{equation*}
I:=\|(A-A')UO\|_F\leq \|(UO)_{1*}\|_2+\|(UO)_{2*}\|_2,
\end{equation*}
where $X_{i*}$ denotes the $i$th row of the matrix $X$. By Lemma \ref{pro: eigendecom} and the balanced community assumption, we have
\begin{equation}
\label{eq:delo}
\max_i\|(UO)_{i*}\|_2=\|UO\|_{2,\infty}=\|\Theta D ^{-1}LO\|_{2,\infty}\leq \|\Theta D ^{-1}\|_{2,\infty}\|LO\|_2\lesssim \frac{1}{\sqrt{n}},
\end{equation}
where recall that $\Theta$ is the membership matrix, $D$ is a diagonal matrix with entries of order $\sqrt{n}$, and $L$ and $O$ are both orthogonal matrices. Hence, we have $I\lesssim 1/\sqrt{n}$.
For $II$, also noting that $A$ and $A'$ differs only by (1,2) and (2,1) entries, we have 
\begin{equation*}
II:=\|(A-A')(\tilde{U}-UO)\|_F\leq \|(\tilde{U}-UO)_{1*}\|_2+\|(\tilde{U}-UO)_{2*}\|_2.
\end{equation*}
By the entrywise bound of eigenvector from SBM (see Lemma 3 in \citet{zheng2022limit} for example), we have that under our condition $\rho\geq C(\frac{\log n}{n})$, there exists an orthogonal matrix $O\in \mathbb R^{K\times K}$ such that  
\begin{equation*}
\|\tilde{U}-UO\|_{2,\infty}\lesssim \frac{\log^{1/2} n}{n\sqrt{\rho}},
\end{equation*}
with large probability, where $c>0$ is some constant. Thus, $II\lesssim \frac{\log^{1/2} n}{n\sqrt{\rho}}$ with high probability. We then have
\begin{equation}
\label{eq:aa}
\|(A-A')\tilde{U}\|_F=I+II\lesssim \frac{1}{\sqrt{n}}+\frac{\log^{1/2} n}{n\sqrt{\rho}}\lesssim \frac{\log^{1/2} n}{\sqrt{n}}.
\end{equation}

As a result, combining \eqref{eq:delta} and \eqref{eq:aa} with \eqref{eq:dav}, we conclude that with high probability, the sensitivity is bounded as
\begin{equation}
\|\tilde{U}\tilde{U}^\intercal - \tilde{U}'(\tilde{U}')^\intercal\|_F\lesssim \frac{\log^{1/2} n}{n^{3/2}\rho}. 
\end{equation}
The property (1) is achieved. 

Next, we prove property (2). 
Suppose $\mathcal{A} = ({A}_1, \dots, {A}_L)$ and $\mathcal{A}' = ({A}_1', \dots, {A}_L')$ are neighboring datasets that differ in the $j$th network $A_j$. That is,
\[
{A}_l = {A}_l' \quad \text{for all } l \neq j, \quad {A}_j \sim {A}_j',
\]
where $\sim$ denotes the neighboring relationships of two networks that differ by only one edge. 
Let $O = (O_1, \dots, O_L)$ be any measurable set of outputs. Since the randomized mechanisms $\mathcal Q_l$'s are independently enforced to each data $A_l$'s, we then have
\[
\begin{aligned}
\mathbb{P}\big( \overline{\mathcal{Q}}(\mathcal{A}) = O |\mathcal A\big) 
&= \prod_{l=1}^L \mathbb{P}\big( \mathcal{Q}_l(A_l) = O_l|A_l \big) = \left( \prod_{l \neq j} \mathbb{P}\big( \mathcal{Q}_l(A_l) = O_l|A_l \big) \right) \cdot \mathbb{P}\big( \mathcal{Q}_j(A_j) = O_j |A_j\big),
\end{aligned}
\]
Similarly,
\[
\mathbb{P}\big( \overline{\mathcal{Q}}(\mathcal{A}') = O |\mathcal A'\big) 
= \left( \prod_{l \neq j} \mathbb{P}\big( \mathcal{Q}_l(A_l) = O_l|A_l \big) \right) \cdot \mathbb{P}\big( \mathcal{Q}_j(A_j') = O_j| A_j'\big).
\]
By the $(\epsilon_j, \delta_j)$-edge-DP of $\mathcal{Q}_j$, we have
\[
\mathbb{P}\big( \mathcal{Q}_j(A_j) = O_j|A_j \big) 
\leq {\rm e}^{\epsilon_j} \cdot \mathbb{P}\big( \mathcal{Q}_j(A_j') = O_j |A_j'\big) + \delta_j.
\]
Therefore,
\[
\begin{aligned}
\mathbb{P}\big( \overline{\mathcal{Q}}(\mathcal{A}) = O|\mathcal A\big) 
&\leq \left( \prod_{l \neq j} \mathbb{P}\big( \mathcal{Q}_l(A_l) = O_l|A_l\big) \right) 
\cdot \left( {\rm e}^{\epsilon_j} \cdot \mathbb{P}\big( \mathcal{Q}_j(A_j') = O_j|A_j' \big) + \delta_j \right) \\
&= {\rm e}^{\epsilon_j} \cdot \mathbb{P}\big( \overline{\mathcal{Q}}(\mathcal{A}') = O |\mathcal {A}'\big) 
+ \delta_j \cdot \prod_{l \neq j} \mathbb{P}\big( \mathcal{Q}_l(A_l) = O_l|A_l \big) \\
&\leq {\rm e}^{\epsilon^*} \cdot \mathbb{P}\big( \overline{\mathcal{Q}}(\mathcal{A}') = O|\mathcal{A}' \big) + \delta^*,
\end{aligned}
\]
where \( \epsilon^* = \max_l \epsilon_l \) and \( \delta^* = \max_l \delta_l \). The parallel composition property is thus achieved. 

\QEDA

\subsubsection{Technical theorems and lemmas for \texttt{TransNetX}}

We first recall and introduce some notations. Recall that for $l\in[L]\cup \{0\}$, $\tilde{U}_l$ and $U_l$ denote the top-$K$ eigenvector of $A_l$ and its population $P_l$, respectively. 
Denote the eigen-decomposition of $\sum_{l=1}^L w_l\left(\tilde{U}_l \tilde{U}_l^\intercal+ N_l\right)$ by
\begin{equation}
\label{eq:eigenuu}
\bar{U}\bar{\Sigma}\bar{U}^\intercal + \bar{U}_{\perp}\bar{\Sigma}_{\perp}\bar{U}^\intercal_{\perp},
\end{equation}
where $\bar{U}$ is the top-$K$ eigenvector and $\bar{U}_{\perp}$ is its orthogonal complement. For the source networks, i.e., $l\in[L]$, define 
\begin{equation}
\label{eq:wl}
W_l:=\argmin_{O\in \mathcal O_{K}} \|\tilde{U}_lO-U_l\|_F.
\end{equation}
For the target network, define
\begin{equation}
\label{eq:w}
W:=\argmin_{O\in \mathcal O_{K}} \|\bar{U}O-U_0\|_F.
\end{equation}
Define 
\begin{equation}
\label{eq:H}
H:= \sum_{l=1}^L w_l H_l\quad {\rm with} \quad H_l:=U_lU_l^\intercal -U_0U_0^\intercal,
\end{equation}
and 
\begin{equation}
\label{eq:N}
N:= \sum_{l=1}^L w_l N_l\quad {\rm with} \quad N_l\sim_{i.i.d.}\mathcal N(0,\sigma_l^2).
\end{equation}

The next theorem provides the basis for the statistical performance of 
\texttt{TransNetX}. 

\begin{theorem}
\label{thm:mainforx}
Suppose that for each $l\in[L]$, we have
\begin{equation}
\label{eq:singexp}
\tilde{U}_lW_l-U_l =T_l^0+T_l,
\end{equation}
where $W_l$ is defined in \eqref{eq:wl}. If $T_l^0$ and $T_l$ satisfies
\begin{equation}
\label{eq:cont}
\sum_{l=1}^L w_l \left( 2\|T_l^0\|_2+2\|T_l\|_2+ \|T_l^0+T_l\|_2^2 \|\right)\leq c-\|H\|_2-\|N\|_2
\end{equation}
for some constant $0<c<1/2$ with $H$ and $N$ defined in 
\eqref{eq:H} and \eqref{eq:N}, respectively, then we have
\begin{equation}
\label{eq:Q}
\bar{U}W-U_0= \sum_{l=1}^L w_l T_l^0U_l^\intercal U_0+ Q,
\end{equation}
where 
\begin{equation*}
Q\lesssim \sum_{l=1}^L w_l \|U_l^\intercal T_l^0\|_2+ \sum_{l=1}^L w_l \|T_l^0\|_2^2 + \sum_{l=1}^L w_l \|T_l\|_2 + \sum_{l=1}^L w_l\|H_l\|_2 + \|\sum_{l=1}^Lw_lN_l\|_2, 
\end{equation*}
and $W$ is defined in \eqref{eq:w}.
\end{theorem}

\subsubsection*{Proof of Theorem \ref{thm:mainforx}}
By the expansion \eqref{eq:singexp} for each single $\tilde{U}_l$, we have the following decomposition,
\begin{align*}
\sum_{l=1}^L w_l\left(\tilde{U}_l \tilde{U}_l^\intercal+ N_l\right) &= \sum_{l=1}^L w_l(\tilde{U}_lW_l) (\tilde{U}_lW_l)^\intercal +N\\
&=\sum_{l=1}^L w_l({U}_l+T_l^0+T_l) ({U}_l+T_l^0+T_l)^\intercal +N\\
&=\sum_{l=1}^L w_l{U}_lU_l^\intercal  + \tilde{E} + N\\
& = U_0U_0^\intercal + \tilde{E} + H+N,\\
\end{align*}
where the last equality follows from the definition of $H$ in \eqref{eq:H} and $\tilde{E}$ is defined as
\begin{align*}
\tilde{E}& = \sum_{l=1}^L w_l [T_l^0U_l^\intercal + U_l (T_l^{0})^\intercal] + \tilde{L},\\
\tilde{L}& = \sum_{l=1}^L w_l [T_lU_l^\intercal + U_l T_l^\intercal+ (T_l^0+T_l)(T_l^0+T_l)^\intercal].
\end{align*}
By the eigen-decomposition \eqref{eq:eigenuu}, we then have
\begin{equation}
\label{eq:decomu}
\bar{U}\bar{\Sigma}\bar{U}^\intercal + \bar{U}_{\perp}\bar{\Sigma}_{\perp}\bar{U}^\intercal_{\perp}=U_0U_0^\intercal + \tilde{E} + H+N.
\end{equation}
We aim to make use of the von Neumann series expansion \citep{bhatia2013matrix} to represent $\bar{U}$, which is also used in \citet{zheng2022limit,tang2022asymptotically}. To that end, we then show that the eigenvalues of $\bar{\Sigma}$ and $\tilde{E} + H+N$ are disjoint. Indeed, applying the Weyl's inequality to $\bar{U}\bar{\Sigma}\bar{U}^\intercal + \bar{U}_{\perp}\bar{\Sigma}_{\perp}\bar{U}^\intercal_{\perp}$ and $U_0U_0^\intercal$, we have by \eqref{eq:decomu} that
\begin{equation}
\label{eq:sigm}
\max_i|\bar{\Sigma}_{ii}-1| \leq \|\tilde{E}\| + \|H\|_2+\|N\|_2,
\end{equation}
which indicates that $\bar{\Sigma}_{ii}\geq 1-\|\tilde{E}\| -\|H\|_2-\|N\|_2$. On the other hand, we have $\|\tilde{E}+H+N\|_2\leq \|\tilde{E}\| +\|H\|_2+\|N\|_2$. Therefore, it suffices that 
\begin{equation*}
\|\tilde{E}\|_2 < \frac{1-2(\|H\|_2+\|N\|_2)}{2},
\end{equation*}
which is actually implied by our condition \eqref{eq:cont} and the definition of $\tilde{E}$. 
As a result, we have
\begin{equation*}
\bar{U}=\sum_{k=0}^\infty (\tilde{E}+ H+ N)^k U_0U_0^\intercal \bar{U} \bar{\Sigma}^{-(k+1)}. 
\end{equation*}
For any $W \in \mathcal O_K$, the following error decomposition then holds,
\begin{equation*}
\bar{U}W-U_0= \sum_{l=1}^L w_l T_l^0U_l^\intercal U_0+ Q,
\end{equation*}
where $Q=Q_1+Q_2+Q_3+Q_4+Q_5+Q_6$ defined respectively as follows,
\begin{align*}
Q_1& =U_0U_0^\intercal \bar{U}\bar{\Sigma}^{-1}W-U_0,\\
Q_2&= \sum_{l=1}^L w_lT_l^0 U_l^\intercal U_0(U_0^\intercal \bar{U}\bar{\Sigma}^{-2}-W^\intercal)W,\\
Q_3&=\sum_{l=1}^L w_l U_l(T_l^0)^\intercal U_0U_0^\intercal \bar{U}\bar{\Sigma}^{-2}W,\\
Q_4&= \tilde{L} U_0U_0^\intercal \bar{U}\bar{\Sigma}^{-2}W,\\
Q_5&= (H+N) U_0U_0^\intercal \bar{U}\bar{\Sigma}^{-2}W,\\
Q_6&=\sum_{k=2}^\infty (\tilde{E}+ H+ N)^k U_0U_0^\intercal \bar{U} \bar{\Sigma}^{-(k+1)}W.
\end{align*}

To facilitate the subsequent analysis, we provide some notation and analysis. Let
\begin{align*}
{\bm e}_{T^0}=\sum_l &w_l \|T_l^0\|_2, \; {\bm e}_{(T^0)^2}=\sum_l w_l \|T_l^0\|_2^2, \; {\bm e}_{T}=\sum_l w_l \|T_l\|_2, \\& {\bm e}_{*}=\sum_l w_l \|U_l^\intercal T_l^0\|_2,\,
{\bm e}_{H}=\|H\|_2,\; {\bm e}_{N}=\|N\|_2.
\end{align*}
By the definition of $\tilde{E}$ and $\tilde{L}$, it is easy to see that,
\begin{equation}
\label{eq:lande}
{\bm e}_{\tilde{L}}:=\|\tilde{L}\|_2\lesssim {\bm e}_{(T^0)^2}+{\bm e}_{T}, \; {\bm e}_{\tilde{E}}:=\|\tilde{E}\|_2\lesssim {\bm e}_{T^0}+{\bm e}_{T}, \; 
\end{equation}
where we used the fact that the error terms $\|T_l^0\|_2$'s and $\|T_l\|_2$'s are all smaller than 1, which is verified in Lemma \ref{lem:errorx}. 
The following bounds on $\|\bar{\Sigma}^{-1}\|_2$ will be used repeatedly. By \eqref{eq:sigm}, we have 
\begin{equation}
\label{eq:siginv}
\|\bar{\Sigma}^{-1}\|_2=\max_i(\Sigma_{ii})^{-1}\leq \frac{1}{1-{\bm e}_{\tilde{E}}-{\bm e}_{H}-{\bm e}_{N}}< \frac{1}{1-c},
\end{equation}
where $0<c<1/2$ is the constant in condition \ref{eq:cont} and the last inequality can be derived as follows. By the definition of $\tilde{E}$ and condition \eqref{eq:cont}, 
${\bm e}_{\tilde{E}}\leq c-{\bm e}_{H}-{\bm e}_{N}$, which indicates that
${\bm e}_{\tilde{E}}+{\bm e}_{H}+{\bm e}_{N}\leq c$.

Next, we bound $Q_1$-$Q_6$, respectively. For $Q_1$, we have the following decomposition,
\begin{align*}
Q_1&=U_0U_0^\intercal \bar{U}\bar{\Sigma}^{-1}W-U_0\\
&=U_0(U_0^\intercal \bar{U}\bar{\Sigma}^{-1}-W^\intercal)W\\
&=-U_0(U_0^\intercal \bar{U}\bar{\Sigma}-U_0^\intercal \bar{U})\bar{\Sigma}^{-1}W + U_0(U_0^\intercal \bar{U}-W^\intercal)W\\
&\overset{(i)}{=}-U_0U_0^\intercal(H+N+\tilde{E})\bar{U}\bar{\Sigma}^{-1}W+ U_0(U_0^\intercal \bar{U}-W^\intercal)W\\
&=-U_0U_0^\intercal(H+N)\bar{U}\bar{\Sigma}^{-1}W+U_0(U_0^\intercal \bar{U}-W^\intercal)W-U_0U_0^\intercal\tilde{E}\bar{U}\bar{\Sigma}^{-1}W\\
&:=Q_{11}+Q_{12}+Q_{13},
\end{align*}
where $(i)$ follows from \eqref{eq:decomu} that 
\begin{equation*}
\bar{U}\bar{\Sigma}=U_0U_0^\intercal\bar{U}+ (\tilde{E} + H+N)\bar{U}.
\end{equation*}
For $Q_{11}$, by \eqref{eq:siginv}, we have $$\|Q_{11}\|_2\leq ({\bm e}_{H}+{\bm e}_{N})\|\bar{\Sigma}^{-1}\|_2\lesssim {\bm e}_{H}+{\bm e}_{N}.$$
For $Q_{12}$, denote the minimum singular value of a matrix by $\sigma_{\min}(\cdot)$, we have
\begin{align}
\label{eq:Q12}
\|Q_{12}\|_2\leq \|U_0^\intercal \bar{U}-W^\intercal\|_2&\leq 1-\sigma_{\min}^2(U_0^\intercal \bar{U})\leq \|{\rm sin}\Theta(\bar{U},U_0)\|_2^2\lesssim {\bm e}_{\tilde{E}}^2+{\bm e}_{H}^2+{\bm e}_{N}^2,
\end{align}
where the last inequality follows from \eqref{eq:decomu} and the Davis-Kahan theorem (see Lemma \ref{pro:DK}) 
that 
\begin{equation}
\label{eq:sin}
\|{\rm sin}\Theta(\bar{U},U_0)\|_2^2\lesssim \frac{\|\tilde{E}\|_2+\|H\|_2+\|N\|_2}{\lambda_{\min}(U_0U_0^\intercal)}.
\end{equation}
For $Q_{13}$, we have 
\begin{align*}
\|Q_{13}\|_2 &=\|-U_0U_0^\intercal(\tilde{E}U_0U_0^\intercal\bar{U}+\tilde{E}(I-U_0U_0^\intercal)\bar{U})\bar{\Sigma}^{-1}W\|_2\\
& \leq (\|U_0^\intercal\tilde{E}U_0\|_2+\|\tilde{E}\|_2 \|(I-U_0U_0^\intercal)\bar{U}\|_2)\|\bar{\Sigma}^{-1}\|_2\\
&{\lesssim} {\bm e}_{*}+{\bm e}_{\tilde{L}}+{\bm e}_{\tilde{E}}({\bm e}_{H}+{\bm e}_{N}+{\bm e}_{\tilde{E}}),
\end{align*}
where the last inequality follows from the following facts: (i) by the definition of $\tilde{E}$, we have
\begin{align*}
U_0^\intercal\tilde{E}U_0 = \sum_{l=1}^L w_l U_0^\intercal[T_l^0U_l^\intercal + U_l (T_l^{0})^\intercal]U_0 + U_0^\intercal\tilde{L}U_0,
\end{align*}
which implies $\|U_0^\intercal\tilde{E}U_0\|_2\lesssim \sum_l w_l \|U_l^\intercal T_l^0\|_2+\|\tilde{L}\|_2={\bm e}_{*}+{\bm e}_{\tilde{L}}$; (ii) by \eqref{eq:sin} we have $\|(I-U_0U_0^\intercal)\bar{U}\|_2\lesssim {\bm e}_{H}+{\bm e}_{N}+{\bm e}_{\tilde{E}}$; (iii) by \eqref{eq:siginv}, we have $\|\bar{\Sigma}^{-1}\|_2$ upper bounded by some constant. 
As a result, we have
\begin{equation*}
\|Q_1\|_2 \lesssim  {\bm e}_{*}+{\bm e}_{\tilde{L}}+{\bm e}_{\tilde{E}}^2 + {\bm e}_{H} + {\bm e}_{N}\lesssim  {\bm e}_{*}+ {\bm e}_{(T^0)^2}+{\bm e}_{T}+ {\bm e}_{H} + {\bm e}_{N},
\end{equation*}
where the last inequality follows from \eqref{eq:lande}.

For $Q_2$, we first have the following decomposition of  $U_0^\intercal \bar{U}\bar{\Sigma}^{-2}-W^\intercal$, 
\begin{align*}
&U_0^\intercal \bar{U}\bar{\Sigma}^{-2}-W^\intercal\\
=&(U_0^\intercal \bar{U}-U_0^\intercal \bar{U}\bar{\Sigma}^{2})\bar{\Sigma}^{-2}+(U_0^\intercal \bar{U}-W^\intercal)\\
=&(U_0^\intercal \bar{U}-U_0^\intercal (U_0U_0^\intercal+H+N+\tilde{E})^2\bar{U})\bar{\Sigma}^{-2}+(U_0^\intercal \bar{U}-W^\intercal)\\
=& -U_0^\intercal\left((\tilde{E}+H+N)U_0U_0^\intercal+U_0U_0^\intercal(\tilde{E}+H+N)^\intercal+(\tilde{E}+H+N)^2\right)\bar{U}\bar{\Sigma}^{-2}+(U_0^\intercal \bar{U}-W^\intercal),
\end{align*}
where the second equality follows from \eqref{eq:decomu} by observing 
\begin{equation*}
\bar{U}\bar{\Sigma}^2\bar{U}^\intercal + \bar{U}_{\perp}\bar{\Sigma}_{\perp}^2\bar{U}^\intercal_{\perp}=(U_0U_0^\intercal + \tilde{E} + H+N)^2,
\end{equation*}
which implies  
\begin{equation*}
\bar{U}\bar{\Sigma}^2=(U_0U_0^\intercal + \tilde{E} + H+N)^2\bar{U}.
\end{equation*}
Then, noting \eqref{eq:siginv} and \eqref{eq:Q12}, we have
\begin{equation*}
\|U_0^\intercal \bar{U}\bar{\Sigma}^{-2}-W^\intercal\|_2\lesssim {\bm e}_{\tilde{E}}+{\bm e}_{H}+{\bm e}_{N}\lesssim {\bm e}_{T_0}+{\bm e}_{T}+{\bm e}_{H}+{\bm e}_{N}.
\end{equation*}
As a result,
\begin{align*}
\|Q_2\|_2\leq \sum_{l=1}^L w_l \|T_l^0\|_2\|U_0^\intercal \bar{U}\bar{\Sigma}^{-2}-W^\intercal\|_2\lesssim {\bm e}_{T_0}({\bm e}_{T_0}+{\bm e}_{T}+{\bm e}_{H}+{\bm e}_{N})\lesssim  {\bm e}_{(T^0)^2}+{\bm e}_{T}+ {\bm e}_{H} + {\bm e}_{N},
\end{align*}
where we used ${\bm e}_{T_0}^2\leq {\bm e}_{(T^0)^2}$ which holds by the Cauchy-Schwarz inequality. 

For $Q_3,Q_4$ and $Q_5$, by \eqref{eq:lande} and \eqref{eq:siginv}, it is easy to see
\begin{align*}
\|Q_3\|_2&=\|\sum_{l=1}^L w_l U_l(T_l^0)^\intercal U_0U_0^\intercal \bar{U}\bar{\Sigma}^{-2}W\|_2 \leq \sum_{l=1}^L w_l \|(T_l^0)^\intercal U_0\|_2\cdot\|\bar{\Sigma}^{-1}\|_2^2\lesssim {\bm e}_{*},\\
\|Q_4\|_2&=\|\tilde{L} U_0U_0^\intercal \bar{U}\bar{\Sigma}^{-2}W\|_2 \leq \|\tilde{L}\|_2\cdot\|\bar{\Sigma}^{-1}\|_2^2\lesssim{\bm e}_{(T^0)^2}+{\bm e}_{T};\\
\|Q_5\|_2&= \|(H+N) U_0U_0^\intercal \bar{U}\bar{\Sigma}^{-2}W\|_2\lesssim {\bm e}_{H}+{\bm e}_{N}.
\end{align*}

For $Q_6$, denote $\|\bar{\Sigma}^{-1}\|_2$ by $\bar{\lambda}^{-1}$, we then have
\begin{align*}
\|Q_6\|_2&=\|\sum_{k=2}^\infty (\tilde{E}+ H+ N)^k U_0U_0^\intercal \bar{U} \bar{\Sigma}^{-(k+1)}W\|_2\\
& \leq \sum_{k=2}^\infty \bar{\lambda}^{-(k+1)} ({\bm e}_{\tilde{E}}+{\bm e}_{H}+{\bm e}_{N})^k\\
& \lesssim ({\bm e}_{\tilde{E}}^2+{\bm e}_{H}^2+{\bm e}_{N}^2)\cdot \bar{\lambda}^{-3}\sum_{t=0}^\infty \bar{\lambda}^{-t}({\bm e}_{\tilde{E}}+{\bm e}_{H}+{\bm e}_{N})^t\\
&\lesssim ({\bm e}_{\tilde{E}}^2+{\bm e}_{H}^2+{\bm e}_{N}^2)\cdot \frac{1}{1-\bar{\lambda}^{-1}({\bm e}_{\tilde{E}}+{\bm e}_{H}+{\bm e}_{N})}\\
&\overset{(i)}{\lesssim} {\bm e}_{\tilde{E}}^2+{\bm e}_{H}^2+{\bm e}_{N}^2\lesssim {\bm e}_{(T^0)^2}+{\bm e}_{T}^2+ {\bm e}_{H}^2 + {\bm e}_{N}^2,
\end{align*}
where we used the fact that $\bar{\lambda}^{-1}$ is upper bounded by some constant and $$\bar{\lambda}^{-1}({\bm e}_{\tilde{E}}+{\bm e}_{H}+{\bm e}_{N})< \frac{c}{1-c}:=c'<1$$   by \eqref{eq:siginv} and its following analysis on ${\bm e}_{\tilde{E}}+{\bm e}_{H}+{\bm e}_{N}$.

Finally, combining the bound of $Q_1$-$Q_6$, we obtain that 
\begin{equation*}
Q\lesssim {\bm e}_{*}+{\bm e}_{\tilde{L}}+{\bm e}_{\tilde{E}}^2 + {\bm e}_{H} + {\bm e}_{N}\lesssim  {\bm e}_{*}+ {\bm e}_{(T^0)^2}+{\bm e}_{T}+ {\bm e}_{H} + {\bm e}_{N},
\end{equation*}
and the conclusion of Theorem \ref{thm:mainforx} is  arrived. 

\QEDA

The next lemma provides the expansion for single $\tilde{U}_l$ and the corresponding error bounds.

\begin{lemma}
\label{lem:errorx}
Suppose Assumption \ref{assu:sparse} hold, 
$B_l$'s are all of full rank $K$ and $\lambda_{\min}(P_l)\gtrsim n\rho$ for all $l\in [L]\cup\{0\}$, and
$\rho\geq C \log n/n$ for some constant $C>0$. Let $E_l=A_l-P_l$ and recall definition of $D_l$ in \eqref{eq:UandD},
then we have
\begin{equation*}
\tilde{U}_l W_l-U_l=T_l^0 +T_l \quad{\rm with}\quad T_l^0:=E_l \Theta_lD_l^{-2} B_l^{-1}D_l^{-1},
\end{equation*}
where with high probability,
\begin{align*}
\|T_l^0\|_2\lesssim \frac{1}{\sqrt{n\rho}}, \; \|U_l^\intercal T_l^0\|_2\lesssim \frac{(\rho\log n)^{1/2}}{n\rho}\quad {\rm and}\quad \|T_l\|_2\lesssim \frac{1}{n\rho}\cdot \max\{1,(\rho\log n)^{1/2}\}.
\end{align*}
\end{lemma}
The proof of Lemma \ref{lem:errorx} follows from Lemma 2 and Lemma 3 of \citet{zheng2022limit} if we note that SBM is a special case of the COSIE model studied therein.

\subsubsection{Main proofs for \texttt{TransNetX}}

\subsubsection*{Proof of Proposition \ref{theo: unifycdp}}
We make use of Theorem \ref{thm:mainforx} and Lemma \ref{lem:errorx} to prove. First, we bound the error term $Q$ in \eqref{eq:Q}. Then, we show that condition \eqref{eq:cont} holds. After that, we derive the bound for the main term $\sum_{l=1}^L w_l T_l^0U_l^\intercal U_0$. 

For $Q$, by Lemma \ref{lem:errorx}, we have with high probability that
\begin{equation*}
\sum_{l=1}^L w_l \|U_l^\intercal T_l^0\|_2+ \sum_{l=1}^L w_l \|T_l^0\|_2^2 + \sum_{l=1}^L w_l \|T_l\|_2 \lesssim \frac{1}{n\rho}\cdot \max\{1,(\rho\log n)^{1/2}\}.
\end{equation*}
It remains to bound $\sum_{l=1}^L w_l\|H_l\|_2$ and $\|\sum_{l=1}^L w_l N_l\|_2$. By Lemma \ref{lem: accuracyofe} and the definition \eqref{eq:H}, we have $$\sum_{l=1}^L w_l\|H_l\|_2\lesssim \sum_{l=1}^L w_l\mathcal E_{\theta,l}.$$
For $\|\sum_{l=1}^L w_l N_l\|_2$, we note that $\sum_{l=1}^L w_l N_l$ has i.i.d. $\mathcal N(0,\sigma^2)$ entries with
$$\sigma\asymp \sqrt{\sum_{l=1}^L w_l^2\cdot\frac{\log n\log(1.25/\delta_i)}{\epsilon_l^2n^3\rho^2}}.$$
By the spectral-norm bound for Gaussian matrix with i.i.d. entries (see Theorem 4.4.5 in \citet{vershynin2018high} for example), we have with high probability that
\begin{equation}
\label{eq:gau}
\|\sum_{l=1}^L w_l N_l\|_2\lesssim \sqrt{n}\sigma \asymp \sqrt{\sum_{l=1}^L w_l^2\cdot\frac{\log n\log(1.25/\delta_i)}{\epsilon_l^2n^2\rho^2}}.
\end{equation}
As a result, we obtain the following high probability bound for the error term $Q$,
\begin{equation}
\label{eq:QB}
\|Q\|_2\lesssim \frac{1}{n\rho}\cdot \max\{1,(\rho\log n)^{1/2}\}+\sum_{l=1}^L w_l\mathcal E_{\theta,l}+ \sqrt{\sum_{l=1}^L w_l^2\cdot\frac{\log n\log(1.25/\delta_i)}{\epsilon_l^2n^2\rho^2}}.
\end{equation}

We next show that condition \eqref{eq:cont} holds. By Lemma \ref{lem:errorx} and the condition for $\rho$, we have the LHS of \eqref{eq:cont} bounded as
\begin{equation*}
\sum_{l=1}^L w_l \left( 2\|T_l^0\|_2+2\|T_l\|_2+ \|T_l^0+T_l\|_2^2 \|\right)\lesssim \frac{1}{\sqrt{n\rho}}+ \frac{\max\{1,(\rho \log n)^{1/2}\}}{n\rho}\lesssim \frac{1}{\sqrt{n\rho}}=o_p(1).
\end{equation*}
On the other hand, by our condition \eqref{eq:conhet}, the RHS of \eqref{eq:cont} is upper bounded by constant. Hence, \eqref{eq:cont} is satisfied. 

Define $V_l:=\Theta_l D_l$ and $\Lambda_l^{-1}:= D_l^{-1}B_l^{-1}D_l^{-1}$, where $V_l$ actually serves as the eigenspace of $P_l$.
then we have by Lemma \ref{lem:errorx} that
\begin{equation*}
\sum_{l=1}^L w_l T_l^0U_l^\intercal U_0=\sum_{l=1}^L w_l E_l V_l \Lambda_l^{-1} U_l^\intercal U_0.
\end{equation*}
Before bounding the main term $\sum_{l=1}^L w_l T_l^0U_l^\intercal U_0$, we introduce the following results that will be used repeatedly. By the concentration bound of the adjacency matrix (see \citet{lei2015consistency} for example), we have with high probability that
\begin{equation}
\label{eq:e}
\|E_l\|_2=\|A_l-P_l\|_2 \lesssim \sqrt{n\rho}\quad {\rm if} \quad \rho\geq C \log n/n. 
\end{equation}
By $\lambda_{\min}(P_l)\gtrsim n\rho$ and the balanced community conditions, 
\begin{equation}
\label{eq:lam}
\|\Lambda_l^{-1}\|_2\lesssim \frac{1}{n\rho}.
\end{equation}
Similar to the derivation of \eqref{eq:delo}, we have by the balanced community condition that
\begin{equation}
\label{eq:delo2}
\max_{s,t}|V_{l,st}|\leq \max_i\|(V_l)_{i*}\|_2\lesssim \frac{1}{\sqrt{n}},
\end{equation}
where $V_{l,st}$ denotes the $(s,t)$th entry of $V_l$ and $(V_l)_{i*}$ denotes the $i$th row of $V_l$. For the eigenvector $U_l$, we also have 
\begin{equation}
\label{eq:delou}
\max_{s,t}|U_{l,st}|\leq \max_i\|(U_l)_{i*}\|_2\lesssim \frac{1}{\sqrt{n}}.
\end{equation}

Now we bound the main term $\|\sum_{l=1}^L w_l T_l^0U_l^\intercal U_0\|_F$. 
We have the following observation,
\begin{align*}
&\|\sum_{l=1}^L w_l E_l V_l \Lambda_l^{-1} U_l^\intercal U_0\|_F^2\\
=&{\rm tr}\left(\sum_{i=1}^L\sum_{j=1}^L w_iw_j U_0^\intercal U_i\Lambda_i^{-1}V_i^\intercal E_iE_jV_j \Lambda_j^{-1} U_j^\intercal U_0  \right)\\
=&\sum_{i=1}^Lw_i^2\|  E_i V_i \Lambda_i^{-1} U_i^\intercal U_0\|_F^2 + \sum_{i\neq j}w_iw_j {\rm tr}\left( U_0^\intercal U_i\Lambda_i^{-1}V_i^\intercal E_iE_jV_j \Lambda_j^{-1} U_j^\intercal U_0  \right):=I+II.
\end{align*}
For $I$, we have with high probability that
\begin{align}
\label{eq:I}
\sum_{i=1}^Lw_i^2\|  E_i V_i \Lambda_i^{-1} U_i^\intercal U_0\|_F^2 \leq \sum_{i=1}^Lw_i^2\|  E_i \|_2^2 \|   \Lambda_i^{-1} \|_F^2\lesssim \sum_{i=1}^Lw_i^2\frac{1}{n\rho},
\end{align}
where we have used \eqref{eq:e} and \eqref{eq:lam}, and $ \Lambda_i^{-1}$ is of rank $K$. For $II$, noting that the trace operator is the sum of the diagonal entries of a given matrix, we here bound each entry first. 
Define
\begin{equation*}
 \Upsilon_{ij}:=U_0^\intercal U_i\Lambda_i^{-1}V_i^\intercal E_iE_jV_j \Lambda_j^{-1} U_j^\intercal U_0
\end{equation*}
and denote its $(s,t)$th entry by $\Upsilon_{ij,st}$.   
It is easy to see
{\scriptsize
$${\Upsilon_{ij,st} = \sum_{k_1=1}^n \sum_{k_2=1}^K\sum_{k_3=1}^K\sum_{k_4=1}^n\sum_{k_5=1}^n\sum_{k_6=1}^n\sum_{k_7=1}^K\sum_{k_8=1}^K\sum_{k_9=1}^n U_{0,k_1s} U_{i,k_1k_2} (\Lambda_i^{-1})_{k_2k_3} V_{i,k_4k_3} E_{i,k_4k_5}E_{j,k_5k_6}V_{j,k_6k_7}  (\Lambda_j^{-1})_{k_7k_8}U_{j,k_9k_8}U_{0,k_9t}}.$$
}We will use the Chebyshev inequality to bound $\Upsilon_{ij,st}$. Note that the randomness of $\Upsilon_{ij,st} $ only comes from $E_{i,k_4k_5}E_{j,k_5k_6}$. It is easy to see that $\mathbb E(\Upsilon_{ij,st})=0$. For the variance, we note that though some of $\{E_{i,k_4k_5}E_{j,k_5k_6}\}_{k_4,k_5,k_6\in[n]}$
 are dependent, say $E_{i,12}E_{j,32}$ and $E_{i,12}E_{j,42}$, their covariances are 0.  Indeed, we have 
{\small
\begin{align*}
\mathrm{Cov} \left( E_{i,12} E_{j,32},\, E_{i,12} E_{j,42} \right) 
&= \mathbb{E} \left[ \left( E_{i,12} E_{j,32} - \mathbb{E}[E_{i,12} E_{j,32}] \right)
                     \left( E_{i,12} E_{j,42} - \mathbb{E}[E_{i,12} E_{j,42}] \right) \right] \\
&= \mathbb{E} \left[ 
    \mathbb{E} \left[ 
        \left( E_{i,12} E_{j,32} - \mathbb{E}[E_{i,12} E_{j,32}] \right)
        \left( E_{i,12} E_{j,42} - \mathbb{E}[E_{i,12} E_{j,42}] \right)
        \bigg| E_{i,12} 
    \right] 
\right] \\
&= \mathbb{E} \left[ E_{i,12}^2 \cdot 
    \mathbb{E} \left[ 
        \left( E_{j,32} - \mathbb{E}[E_{j,32}] \right)
        \left( E_{j,42} - \mathbb{E}[E_{j,42}] \right)
    \right] 
\right] \\
&= \mathbb{E} \left[ E_{i,12}^2 \right] \cdot 
   \mathbb{E} \left[ 
        \left( E_{j,32} - \mathbb{E}[E_{j,32}] \right)
        \left( E_{j,42} - \mathbb{E}[E_{j,42}] \right)
   \right] = 0.
\end{align*}
}Hence, the variance of $\Upsilon_{ij,st}$ can be written as the sum of variances for each individual term. Noting \eqref{eq:lam}-\eqref{eq:delou} and the fact that ${\rm Var}(E_{i,k_4k_5}E_{j,k_5k_6})\lesssim \rho^2$, we have
\begin{equation*}
{\rm Var}(\Upsilon_{ij,st})\lesssim n^5K^4\cdot \frac{\rho^2}{n^{10}\rho^4}\asymp \frac{1}{n^5\rho^2}.
\end{equation*}
By the Chebyshev inequality, we then have with high probability that
\begin{equation*}
\Upsilon_{ij,st}\asymp \sqrt{{\rm Var}(\Upsilon_{ij,st})}\asymp \frac{1}{n^{5/2}\rho}. 
\end{equation*}
Therefore, 
\begin{equation}
\label{eq:II}
II\lesssim \sum_{i\neq j}w_iw_j (K\cdot\max_s \Upsilon_{ij,ss} )\lesssim  \sum_{i\neq j}w_iw_j \frac{1}{n^{5/2}\rho}.
\end{equation}

Combining \eqref{eq:I} and \eqref{eq:II} with \eqref{eq:QB}, we obtain the bound 
\begin{align*}
{\rm dist}(\bar{U},U_0) &\lesssim \sqrt{\sum_{i=1}^Lw_i^2\frac{1}{n\rho}}+ \sqrt{\sum_{l=1}^L w_l^2\cdot\frac{\log n\log(1.25/\delta_l)}{\epsilon_l^2n^2\rho^2}}
\\&\quad\quad\quad+\sum_{l=1}^L w_l \mathcal{E}_{\theta,l}+\frac{1}{n\rho}\cdot \max\{1,(\rho\log n)^{1/2}\}+\sqrt{\sum_{i\neq j}w_iw_j \frac{1}{n^{5/2}\rho}}\\
&:=\tilde{\mathcal I}_1+\tilde{\mathcal I}_2+\tilde{\mathcal I}_3+\tilde{\mathcal I}_4+\tilde{\mathcal I}_5.
\end{align*}
By our condition, $\tilde{\mathcal I}_4$ and $\tilde{\mathcal I}_5$ are both dominated by $\tilde{\mathcal I}_1$. Indeed,
\begin{equation*}
\tilde{\mathcal I}_4:=\frac{1}{n\rho}\cdot \max\{1,(\rho\log n)^{1/2}\}\overset{(i)}{\lesssim} \sqrt{\frac{1}{Ln\rho}}\lesssim \sqrt{\sum_{i=1}^Lw_i^2\frac{1}{n\rho}}:=\tilde{\mathcal I}_1,
\end{equation*}
where $(i)$ comes from assumption \eqref{eq:conL2}. By $\sum_iw_i=1$, we have
\begin{equation*}
\tilde{\mathcal I}_5:=\sqrt{\sum_{i\neq j}w_iw_j \frac{1}{n^{5/2}\rho}}\lesssim \sqrt{\frac{1}{n^{5/2}\rho}}
\overset{(ii)}{\lesssim} \sqrt{\frac{1}{Ln\rho}}\lesssim \sqrt{\sum_{i=1}^Lw_i^2\frac{1}{n\rho}}:=\tilde{\mathcal I}_1,
\end{equation*}
where $(ii)$ comes from assumption \eqref{eq:conL2}. The conclusion of Proposition \ref{theo: unifycdp} is thus arrived. \QEDA

\subsubsection*{Proof of Theorem \ref{theo:ebo2}}
We first observe that for each $l\in[L]\cup\{0\}$,
\begin{equation*}
\|\tilde{U}_l\tilde{U}_l^\intercal + N_l -U_lU_l^\intercal \|_2\leq \|\tilde{U}_l\tilde{U}_l^\intercal  -U_lU_l^\intercal \|_2 + \|N_l\|_2\lesssim \frac{1}{\sqrt{n\rho}}+ \frac{\log^{1/2} n \sqrt{\log (1.25/\delta_l)}}{\epsilon_l n\rho}
\end{equation*}
with high probability, where the last inequality follows from Davis-Kahan theorem and the concentration inequality for random matrix with Gaussian entries; see similar derivations in \eqref{eq:gau}. Thus, following the proof logic of Lemma \ref{lem: accuracyofe}, we can show that 
$$\hat{\mathcal E}'_{\theta,l}\asymp {\mathcal E}_{\theta,l}+ O_p\left(\frac{1}{\sqrt{n\rho}}+ \frac{\log^{1/2} n\sqrt{\log (1.25/\delta_l)}}{\epsilon_l n\rho}\right)\asymp O_p\left(\frac{1}{\sqrt{n\rho}}\right),$$
where the last equality follows from condition \eqref{con:epsiloncdp}.
By this fact and Lemma \ref{lem: accuracyofrho}, we then have for $l\in[L]$ that
\begin{align*}
\tilde{w}_l&\varpropto \left({\mathcal P}_{l}^{'2} L+O_p({\mathcal P}_{l}^{'2})+ {\mathcal E}_{\theta,l}^2 +O_p(\frac{1}{n\rho}) + \frac{1}{n\rho}\right)^{-1} \varpropto \left( {\mathcal E}_{\theta,l}^2 + \frac{1}{n\rho}\right)^{-1}.
\end{align*}
By the condition ${\eta}'_n\asymp \frac{1}{n\rho}$, we have $\tilde{w}_l\varpropto n\rho$ for $l\in S$ and $\tilde{w}_l\varpropto \frac{1}{\mathcal E_{\theta,l}^2}$ for $l\in S^c$. In particular, we have
\begin{equation}
\label{eq:w1}
\tilde{w}_l\asymp \frac{ n\rho}{m\cdot n\rho+\sum_{l\in S^c}1/\mathcal E_{\theta,l}^2}\asymp\frac{1}{m}\quad {\rm for}\quad l\in S,
\end{equation}
where we used the fact that 
\begin{equation*}
\sum_{l\in S^c}1/\mathcal E_{\theta,l}^2\lesssim \frac{L-m}{\min _{l\in S^c} \mathcal  E_{\theta,l}^2} \lesssim m\cdot n\rho,
\end{equation*}
which holds by condition \eqref{eq:con21}. 

To establish the error-bound-oracle property, we need to show
{\small
\begin{equation*}
\left(\sum_{l\in S^c} \tilde{w}_l^2 \frac{1}{n\rho}\right)^{1/2}+\left(\sum_{l\in S^c} \tilde{w}_l^2 \mathcal P_{l}^{'2}L \right)^{1/2}+\sum_{l\in S^c}\tilde{w}_l\mathcal E_{\theta,l} \lesssim \left(\sum_{l\in S} \tilde{w}_l^2 \frac{1}{n\rho}\right)^{1/2}+\left(\sum_{l\in S} \tilde{w}_l^2 \mathcal P_{l}^{'2}L \right)^{1/2}+\sum_{l\in S}\tilde{w}_l\mathcal E_{\theta,l}.
\end{equation*}
}By condition \eqref{con:epsiloncdp}, we have $\mathcal P_l^{'2}L\lesssim \frac{1}{n\rho}$, so we only need to show,
{\small
\begin{equation}
\label{eq:inequa2}
\left(\sum_{l\in S^c} \tilde{w}_l^2 \frac{1}{n\rho}\right)^{1/2}+\sum_{l\in S^c}\tilde{w}_l\mathcal E_{\theta,l} \lesssim \left(\sum_{l\in S} \tilde{w}_l^2 \frac{1}{n\rho}\right)^{1/2}+\sum_{l\in S}\tilde{w}_l\mathcal E_{\theta,l}.
\end{equation}
}Note that 
\begin{equation*}
\sum_{l\in S^c}\tilde{w}_l\mathcal E_{\theta,l} \lesssim \left(\sum_{l\in S^c} \tilde{w}_l^2 \mathcal E_{\theta,l}^2L\right)^{1/2}.
\end{equation*}
To show \eqref{eq:inequa2}, it suffices to show 
\begin{equation}
\label{eq:suff2}
\sum_{l\in S^c} \tilde{w}_l^2 \left(\frac{1}{n\rho}+\mathcal E_{\theta,l}^2L\right)\lesssim \sum_{l\in S} \tilde{w}_l^2 \frac{1}{n\rho}\asymp \frac{1}{mn\rho},
\end{equation}
where we used \eqref{eq:w1} in the last equality. 
Note that for $l\in S^c$,
\begin{equation*}
\tilde{w}_l\asymp \frac{1/{\mathcal E_{\theta,l}^2}}{m \cdot n\rho+\sum_{l\in S^c}1/\mathcal E_{\theta,l}^2 }\asymp \frac{1}{\mathcal E_{\theta,l}^2mn\rho}.
\end{equation*}
Therefore, by the condition for ${\eta}'_n$, we have 
\begin{equation*}
\sum_{l\in S^c} \tilde{w}_l^2 L\left(\frac{1}{n\rho L}+\mathcal E_{\theta,l}^2\right)\asymp \sum_{l\in S^c} \frac{L}{m^2n^2\rho^2 \mathcal E_{\theta,l}^2},
\end{equation*}
which combining with \eqref{eq:con21} implies that \eqref{eq:suff2} holds. As a result, by \eqref{eq:w1}, we have established \eqref{eq:bound2}. 

Next we show that under \eqref{eq:con22}, the error bound of $\bar{U}$ under $\tilde{w}_l$'s (i.e., \eqref{eq:bound2}) is of smaller order than the error bound under equal weights. By Proposition \ref{theo: unifycdp}, the error bound under the equal weights is
\begin{equation*}
\left(\frac{1}{n\rho L}\right)^{1/2}+\frac{1}{L}\sum_{l=1}^L \mathcal E_{\theta,l}.
\end{equation*}
By \eqref{eq:bound2}, it suffices to show
\begin{equation}
\label{eq:sufficient2}
\frac{1}{m^2}\sum_{l\in S}\left(\frac{1}{n\rho}+ \mathcal E_{\theta,l}^2L\right) = o\left( \frac{1}{L^2} \left(\sum_{l\in S^c} \mathcal E_{\theta,l}\right)^2\right).
\end{equation}
By the condition of ${\eta}'_n$,
\begin{equation*}
\frac{1}{m^2}\sum_{l\in S}\left(\frac{1}{n\rho}+ \mathcal E_{\theta,l}^2L\right) \lesssim \frac{L}{mn\rho}.
\end{equation*}
On the other hand, we have
\begin{equation*}
 \frac{1}{L^2} \left(\sum_{l\in S^c} \mathcal E_{\theta,l}\right)^2 \gtrsim\frac{(L-m)^2}{L^2} \min_{l\in S^c} \mathcal E_{\theta,l}^2.
\end{equation*}
By our condition \eqref{eq:con22},  \eqref{eq:sufficient2} holds and the result follows.

\QEDA

\spacingset{1.5}
\bibliographystyle{plainnat}
\bibliography{transnet}
\end{document}